%% file: ijcai21.tex
\documentclass{article}
\usepackage{ijcai21}
\usepackage{color}
\usepackage{times}
\usepackage{soul}
\usepackage{url}
\usepackage[utf8]{inputenc}
\usepackage[font=small]{caption}
\usepackage{graphicx}
\usepackage{amsmath}
\usepackage{amsthm}
\usepackage{booktabs}
\usepackage{algorithm}
\usepackage{algorithmic}
\usepackage{multicol}
\usepackage{cleveref}
\usepackage{subfig}
\usepackage{tikz}
\usepackage{rotating}
\usepackage{pgfplots} 
\usepackage[skip=0pt]{caption}
\usepackage{enumitem}
\setlist{nosep}
\usepackage{xcolor}

\title{A Microscopic Pandemic Simulator for Pandemic Prediction \\Using Scalable Million-Agent Reinforcement Learning}

\author{
Zhenggang Tang$^1$
\and
Kai Yan$^1$\footnote{Equal contribution for the first two authors.}\and
Liting Sun$^2$\and
Wei Zhan$^2$\And
Changliu Liu$^3$
\affiliations
$^1$Peking University\\
$^2$University of California, Berkeley\\
$^3$Carnegie Mellon University\\
\emails
\{tangzhenggang, kaiyan\}@pku.edu.cn,
\{litingsun, wzhan\}@berkeley.edu,
cliu6@andrew.cmu.edu
}

\begin{document}

\maketitle

\begin{abstract}
  Microscopic epidemic models are powerful tools for government policy makers to predict and simulate epidemic outbreaks, which can capture the impact of individual behaviors on the macroscopic phenomenon. However, existing models only consider simple rule-based individual behaviors, limiting their applicability. This paper proposes a deep-reinforcement-learning-powered microscopic model named Microscopic Pandemic Simulator (MPS). By replacing rule-based agents with rational agents whose behaviors are driven to maximize rewards, the MPS provides a better approximation of real world dynamics. To efficiently simulate with massive amounts of agents in MPS, we propose Scalable Million-Agent DQN (SMADQN). The MPS allows us to efficiently evaluate the impact of different government strategies. This paper first calibrates the MPS against real-world data in Allegheny, US, then demonstratively evaluates two government strategies: information disclosure and quarantine. The results validate the effectiveness of the proposed method. As a broad impact, this paper provides novel insights for the application of DRL in large scale agent-based networks such as economic and social networks. 
\end{abstract}

\vspace{-1mm}
\section{Introduction}\label{sec:intro}
\vspace{-1mm}
A good epidemic prediction model is indispensable to mitigate pandemics such as COVID-19, which could help governments derive optimal policies that balance public health and economic resiliency. Various epidemic models have been proposed in literature, among which microscopic models~\cite{2ndwave} are particularly useful since they are fine-grained enough to encode individual behaviors and local contact traces. 
However, in the existing works, individuals are either modeled with fixed behaviors ~\cite{costbenefit} or are adhere to a given script or rule ~\cite{meloni2011modeling}. These rules may be too simple to fully capture people's diverse behaviors under environmental changes. 
On the other hand, reinforcement learning (RL) has been proved empirically good at generating complex behaviors, where the agent aims to optimize a relatively simple reward function without being guided by handcrafted scripts. RL models are more explainable and natural, because real-world individuals are driven by different motivations, which is coherent with the reward optimization process. 
Many recent works apply deep RL on microscopic epidemic models~\cite{kompella2020reinforcement}. However, these works mainly focus on policy optimization for governments to balance public health and economy without detailed modeling of individual behaviors. 

To address the individual behavior modeling problem in pandemic modelling, this paper proposes Microscopic Pandemic Simulator (MPS), a novel microscopic epidemic model where individuals are controlled by multi-agent (MA) RL policies and thus able to change their behaviours according to information gained from their surroundings and the government, as inspired by and expanded upon Liu's previous work\cite{Liu2020AME}. Since most practical epidemic models contain millions of agents~\cite{France}, the main challenge of applying MARL is its scalability. Without special design, it is impossible to make standard RL algorithms applicable to such a massive model. While there are previous works on MARL with million-level number of agents~\cite{zheng2017magent}, the pandemic environment imposes a much harder challenge for two reasons: 1) rewards of actions are significantly delayed up to several days due to the presence of incubation period, causing 1-step TD learning improper; 2) Our problem is non-ergodic: agents' joint-state significantly changes as the epidemic spreads, aggravating oscillation during training. To solve these problems with such a large scale, we proposed Scalable Million-Agent DQN (SMADQN), a novel DQN-based algorithm with specially designed replay-buffer and processes for calculating TD($\lambda$), which solves the difficulties above well with such scalability. 

To validate authenticity and adaptability of our model, we apply our model on a large scale COVID-19 simulation. Unlike existing works that consider simple contact networks with limited numbers of facilities or complex networks within a city~\cite{2ndwave}, this paper provides a large-scale county-level simulation with all population in the Allegheny county. It includes more than $10^6$ residents and a comprehensive contact network including $14$ types of facilities. 
By building a novel demographic dataset of Allegheny County with contact network, we show that the model fit real-world data well. 
We further tested the performance of two progressive government strategies on our model, namely, \textit{Information Disclosure} (ID) and \textit{Quarantine} (QT). ID means that the government will disclose information of the pandemic to residents, such as the number of infections in each facility. QT refers to the extra government mitigation strategy that requires symptomatic individuals and part of close contacts of them to be quarantined. We conclude that ID can help mitigate the epidemic and reduce people's activity levels. However, it is not strong enough to completely control the epidemic. Meanwhile, QT, although posting higher requirements for governments' administrative capability, can control the epidemic with minimum negative impacts on economic and social activities.

Our contributions are two-fold.
The \textbf{contributions to epidemic modeling} are the following. 
1) We proposed the Microscopic Pandemic Simulator (MPS) where individuals as modeled as rational agents instead of following simple rules. 
2) We built a comprehensive demographic dataset with a contact network of Allegheny County. To the authors' best knowledge, this is the first work in such details in this area.
3) With the MPS and the dataset, we have provided a flexible tool to test the impacts of different governmental strategies. 
The \textbf{contribution to MARL} is the novel soft-DQN algorithm, i.e., SMADQN. It improves the efficiency of learning by utilizing $\lambda$-return for million-level multi-agent problem with delayed reward signal and ever-changing joint-state.

\vspace{-1mm}
\section{Related Work}\label{sec:related}
\vspace{-1mm}
\textbf{Microscopic Epidemic models.} 
Microscopic epidemic models contain three parts~\cite{France}: personal status, contact network between individuals and a reasonable synthetic population. Generally, personal status includes health status, vulnerability data (e.g., age, basic medical condition~\cite{australia}) and social contacts; contact network is usually divided into different layers, each of which consists of a graph with clusters, representing different places (e.g., workplace and household); synthetic population is retrieved by either mobility data such as GPS~\cite{2ndwave} or generated from a certain distribution~\cite{costbenefit,covidabs} coherent to census data. To balance the granularity and accuracy against computational resources and data available, our work adopts basic personal status settings, fine contact network modeling with 14 layers, and generated synthetic population merged from ArcGIS~\cite{arcgis} and US synthetic population dataset~\cite{synpop}.


\textbf{Large-Scale Multi-Agent RL.} Our problem is framed as a large-scale MA problem. Qu and Li~\cite{qu2019exploiting}~\cite{qu2020scalable} considered cooperative games and exploited problem-specific structures, such as a tree or a network. In our work, agents also interact in a network. Instead of focusing on policy optimization, we exploit the explicitness of reward parameters and inherent ability of adaption of RL to produce a more explainable and flexible model in a non-cooperative setting. The work that proposed the most closely related algorithm to ours is~\cite{yang2018study}, where they simulated a million-level prey-predator world and proposed a DQN algorithm with redesigned replay buffers. However, their environment has an infinite horizon, and is ergodic: all agent's joint-states in different steps are similar, which makes the RL algorithm easier to converge. Furthermore, their reward signals have no delay, while our SMADQN algorithm is designed to solve the non-ergodic environment with long episodic length and delayed reward signal.

\vspace{-1mm}
\section{Problem Formulation}\label{sec:pre}
\vspace{-1mm}
In this work, we model epidemics using Multi-Agent Partial Observation Markov Decision Process (MA-POMDP) where every susceptible individual is viewed as one agent. We are interested in systems that contain more than $n{>}10^6$ agents. The MA-POMDP is defined by  $(\mathcal{S},\mathcal{O},\mathcal{A},R,P)$. $\mathcal{S}$ represents the joint-state space, and $\mathcal{O}=\otimes_{i \in \{1, 2,{\cdots}, n\}}\mathcal{O}_i$ is the union of all $n$ agents' observation spaces. 
$\mathcal{A}$ is the action space for each agent. Agent $i$'s action is denoted $a_i$. $R(s,a_1,a_2,...,a_n;i)$ is the reward function for agent $i$. $P(s'|s,a_1,a_2,...,a_n)$ is transition probability from state $s$ to state $s'$ when agent $i$ takes action $a_i$ for all $i$. Agent $i$ has a stochastic policy $\pi_i(o_i,a_i;\theta_i)$ parameterized by $\theta_i$ which outputs the probability to take action $a_i$ when observing $o_i$. Every agent $i$ optimizes its expected accumulative reward $U_i(\theta_i)=E_{a_1\sim\pi_1,a_2\sim\pi_2,...,a_n\sim\pi_n,}\left[\sum_t \gamma^t R(s^t,a^t_1,a^t_2,...,a^t_n;i)\right]$ with the discounted factor $\gamma$, where $t$ indicates discrete time steps. 

\textbf{Actions}: Agent $i$'s action is modeled into three parts: the activity level, whether to wear a mask, and shopping decision, i.e., $a_{i} = a_{i,act} \times a_{i,mask} \times a_{i,shop}$. To avoid being infected, the most effective way is staying away from risky (with high probability of infection) and non-compulsory facilities. However, it is cumbersome and requires a large action space to allow the agent to choose which facility to visit. So, we assign each facility with a risk level: $MinAct$, and only enable the agent to decide a discrete activity level, then let it visit facilities whose risk level, $MinAct$, are under this level. Note that $MinAct$ of workplaces, schools and households are all set to $0$ as they are compulsory although visiting these facilities may also be risky. That's how we model $a_{i,act}$ and it's still consistent with reality since it's natural for individual to stop visiting the riskiest facility first for health. Besides activity level, wearing reduces infection probability during contact while also brings with a little inconvenience, and we build its decision as $a_{i,mask}$. Furthermore, we separate the decision of shopping $a_{i,shop}$, which indicates whether to go shopping in retail stores, or shopping online, or not shopping at all, from $a_{act}$ because $a_{i,shop}$ is influenced by both the epidemic and supply level of home.

\textbf{Observations}: Agent $i$'s default observation is $o_{i,no\ info} = o_{i,hea} \times o_{i,hou} \times o_{i,sup} \times o_{city} \times o_{t}$. Like any real-life individual, an agent gathers information from both its surroundings and the government. For the former part, we assume that they always know the current health condition of itself ($o_{i,hea}$) and other people living in the same household ($o_{i,hou}$), as well as the amount of necessary stocks for living ($o_{i,sup}$), such as food; for the latter, we choose the total number of infection ($o_{city}$) and the number of days since the first cases are discovered locally ($o_t$), which are accessible from almost every government in real life. With information disclosure, the observation space will be increased by one dimension, i.e., $o_{i,sur}$, which indicates the severity of infections in each facility that agent $i$ may visit calculated by the government.

\textbf{States}: Similar to the observation, $S=\{S_i, i \in \{1, 2,{\cdots}, n\}\} \times s_t$ and $S_i = s_{i,hea} \times s_{i,sup}$. The only difference between state and observation is health condition, as people cannot tell whether they are infected under the presence of incubation period.

 
\textbf{Rewards}: Reward of an agent reflects the incentive of balancing the two goals: ``avoiding infection" and ``maintaining normal life". The former incentive results in a huge one-time negative reward $R_{ill}$ when getting infected (smoothed in practice; see appendix for details) and a $R_{shop}=-1$ for offline shopping. The latter leads to 3 components of reward: a positive reward $R_{act}(a_i)$ increased with higher activity level $a_{i,act}$; a negative reward $R_{mask}$ for wearing masks, as many people are reluctant to wear masks; and a negative reward $R_{sup}$ as the stocked supplies (e.g., food) decreased with lower supply level $o_{i,sup}$. To further adapt to real-world situation, we impose a more negative reward $R_{eth}$, the ethical penalty, for higher activity levels or not wearing mask when having symptoms. This is because real-world individuals are not fully selfish: they will avoid infecting others and comply with the restrictions once fallen ill.


More details of the observation space, action space, reward and other RL parameters can be found in the appendix. With these definitions, we will derive the transition function of the simulator in the following sections.


\vspace{-1mm}
\section{Microscopic Pandemic Simulator}\label{sec:simu} 
\vspace{-1mm}







\begin{figure}
    \centering
    \includegraphics[width=\linewidth]{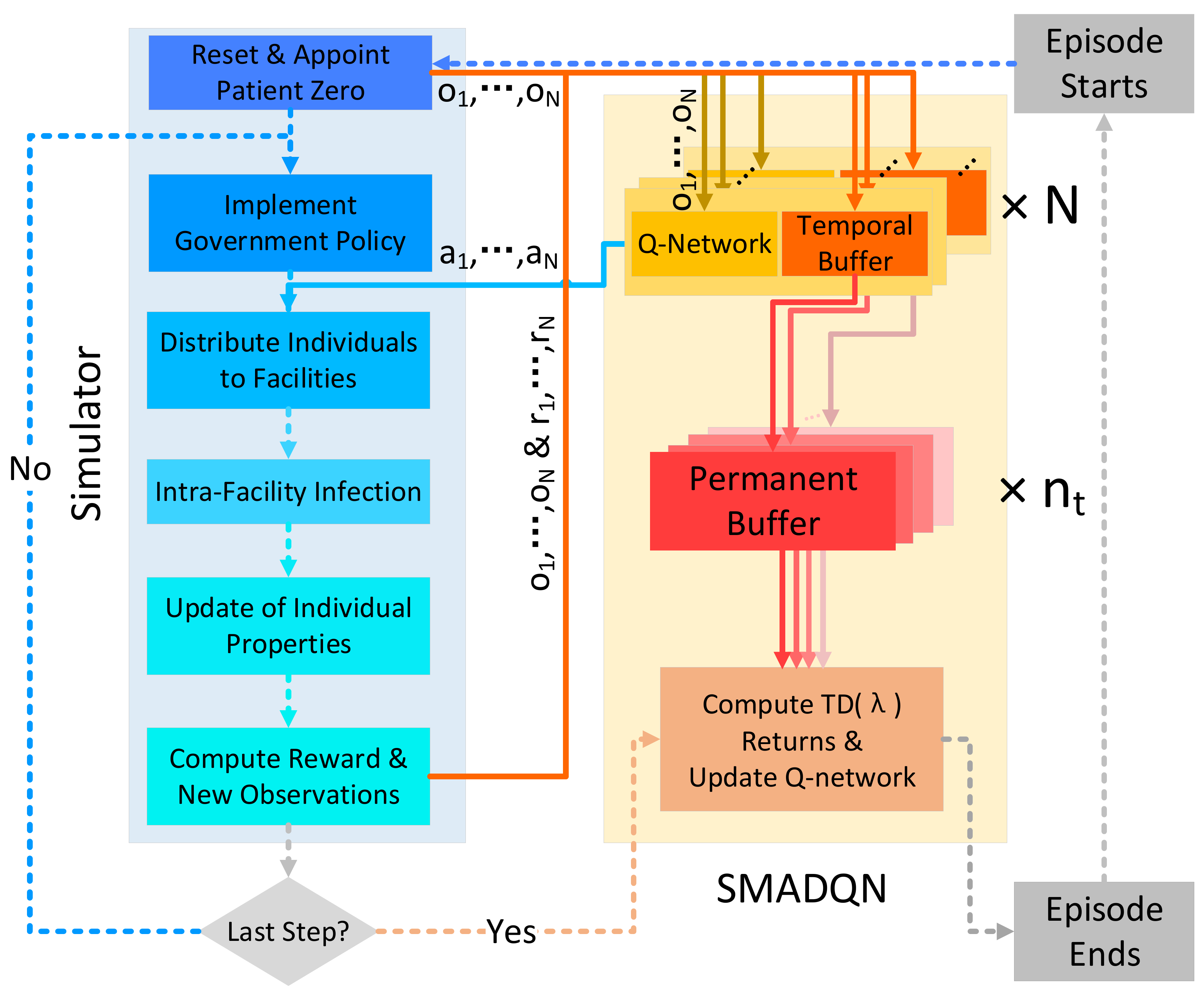}
    \caption{The flow chart of MPS and its interactions with SMADQN-controlled agents. The dotted lines are logical processes and solid lines are data flows.}
    \label{fig:1}
\end{figure}

This section introduces the basic settings of Microscopic Pandemic Simulator (MPS). MPS takes one day as a discrete round, which is the finest practical grain of time since hour-level simulation requires prohibitively expensive computational resources and GPS data for commute patterns are absent. We show in our experiment that this grain of time is fine enough for authentic simulation.  
\Cref{fig:1} provides an overview of the pipeline of the simulator and its interaction with the SMADQN-controlled agents. The pipeline is organized in a logical order: At the beginning of each day, the government implements some mitigation strategy (e.g., adjust maximum capacity ratio for restaurants); then, agents will observe the current situation and decide whether to visit different facilities, to wear masks and to shop for this day. With determined visit list for each facility, the disease will spread from the infected agents to healthy ones within each facility. Finally, modified by the spread, the properties of each agent, including health states $s_{hea}$ and supply level $s_{sup}$ will be updated. This section explains the three major steps of the simulator above in detail; the rest of our MPS model are explained in RL-related parts (for computing reward) and experiment (for government policy).  

\subsection{Distribution of Individuals to Facilities}

Every day after agents make their decisions, they will be distributed into different facilities for calculation of contact.

\textbf{Deciding the agent affiliation to facilities}. We collect facility data from ArcGIS~\cite{arcgis}, and we use synthesized US population dataset~\cite{synpop} for individual agent data. Similar to the reality that individuals prefer to visit near facilities, we attach each available agent to the nearest facility of each type on the map within a distance upper bound. For simplicity, one individual is only attached to one building for each facility type. Another important thing we should mention is that age is crucial for modeling agents in epidemic outbreak since it is related with the type of facilities where one is active. In our model, the population is divided into four age groups, which are preschool children (0-9 years old), students (10-18), adults (19-64) and seniors (65+). We don't consider preschool children and students' after-school activities so they do not have $a_{act}$ while adults and seniors have. Students are affiliated with schools and adults with workplaces while preschool children and seniors with neither.
 
\textbf{Facility Attributions}. 
How to distribute agents to facilities depends on agents' actions and attributions of facilities. 
The two most important attributions about a facility are its capacity and type. For places other than hospitals, if there are more people going to a facility than its capacity, extra individuals will be uniformly randomly picked and kicked out from the facility that day. The type of facility decides the basic infection coefficient: $I_F$, which represents the relative infection probability in this type of facilities, and the minimum daily $a_{act}$ for agents to join: $MinAct$. An agent will visit a facility with $MinAct = x$ when its $a_{act} \ge x$. For all non-compulsory facilities, $MinAct$ is assigned based on $I_F$: The higher $I_F$ is, the higher the $MinAct$ is. 

\subsection{Intra-Facility Infection}

The contact pattern of people inside a facility could be very complicated. In reality, some people may wander around while others may stay isolated. However, with the absence of GPS data, it would be too arbitrary to make any assumption on agents' behavior patterns inside a facility. Even such assumptions can be made, the computational cost would be too high to keep track of agents' trajectories. Therefore, we simply assume that all infection happen in facilities and each person makes a constant number of contacts in a particular facility irrelevant of its capacity, and the contacts between agents are independently uniformly drawn.  The probability $p$ of an individual $x$ infected in facility $F$ with $n$ people this day is: $p = \min(\beta I_F f_F p_{x}\sum\nolimits_{y\in  F}p_{y}/C_F, 1)$ where $p_{x} = g(age_x, a_x)$ and $p_{y} = h(s_{y,hea}, a_y)$. 
In this formula, $\beta$ is the overall hyper-parameter for infection rate. $p_{x}$ is the factor related to the victim $x$, and $p_{y}$ is the factor related to the contagious patient $y$. $I_F$ is the basic coefficient of infection in facility $F$, $f_F$ is the normal frequency of people going to facility $F$ and $C_F$ is the maximum capacity of facility $F$. $p_x$ and $p_y$ are calculated respectively by function $g$ and $h$ which are determined by the agent's specific health state, action and age. For example, young and mask-worn people have less probability to get infected and asymptomatic, presymptomatic or mask-worn patients are less infectious.

There are two issues worth noting about this formula. First, it \textit{smooths out} the infection probability by multiplying the frequency in normal life $f_{F}$ onto the infectious probability. In another word, when an agent decides $a_{act} = x$, it will visit all facilities with $MinAct \leq x$ on that day being less infectious and susceptible to infection, and the reduction is inversely proportional to $f_F$, as if it will visit each facility with probability equal to $1/f_{F}$. Second, the formula above is an approximation to the real probability
$
p= \beta I_Ff_Fp_{x}(1-\prod_{y}(1-p_{y}))/C_F.
$
This simplification accelerates the calculation process.

Detailed values of parameters, settings of functions and the algorithm deciding the agents' affiliation are listed in the appendix. Moreover, community has a slightly different formula for infection, which is also stated in the appendix.

\subsection{Update of Individual Properties}
After the intra-facility disease spread, the two major properties of each agent $i$ will be updated: supply level $s_{i,sup}$ and health state $s_{hea}$.

\textbf{Supply Level}. Supply level models the amount of stock one remains. It resembles a balance between sheltering-at-home and going out less and going shopping for essential goods to survive. Supply level drops monotonically from $1$ to $0$ with decreasing speed as people get more thrift with scarce supplies, and can only be reset to $1$ when one of the agents in the household selects $a_{shop}\in\{\text{offline, online}\}$ and succeeds in shopping. Shopping offline is risky but will always succeed. Shopping online avoids infection, but the number of people it successfully serves every day is very limited, and failed online shopping brings nothing. Agents will receive increasing negative reward with decreasing supply level.

\textbf{Health State}.
\begin{figure}
    \centering
    \includegraphics[width=\linewidth]{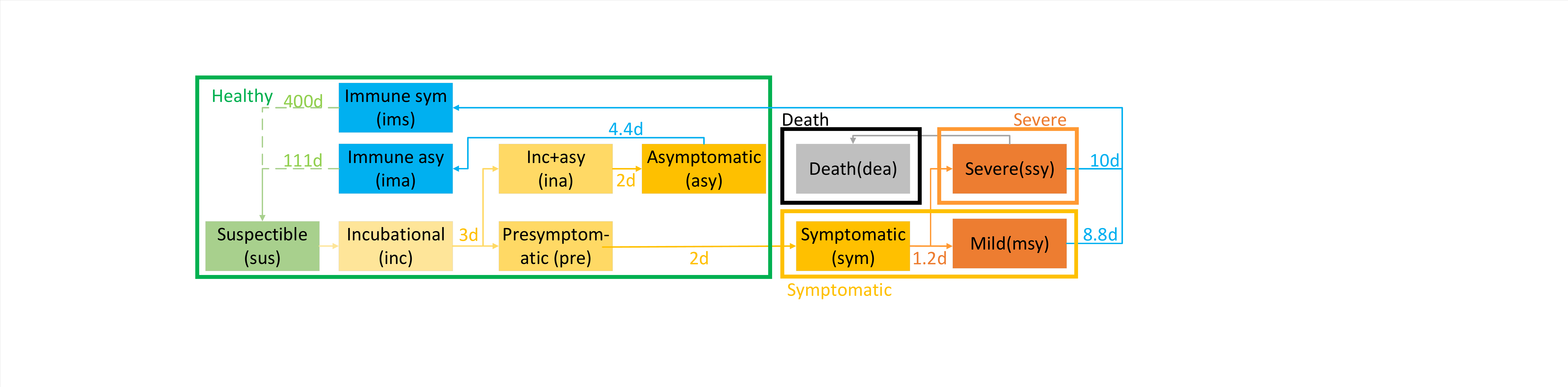}
    \caption{The natural history of disease based on the SEIR model. Each rectangle represents a health state ($s_{hea}$) and they are 4 boxes representing health observations ($o_{hea}$). The numbers on the edges are the expectation time of transition. The transition is also significantly infected by age; older people have much higher probability of being severe (from ``sym" to ``ssy") or dead (from ``ssy" to ``dea").}
    \label{fig:2}
\end{figure}
Each individual's health state $s_{hea}$ is updated at every simulation step, and the observation $o_{hea}$ is decided by $s_{hea}$; the transitions and correspondence are illustrated in \cref{fig:2}. Agents cannot directly access their health state $s_{hea}$, but can observe the observations $o_{hea}$ instead.

We model the natural history of disease as a variant of the SEIR model~\cite{SEIR} with some changes: (1) presymptomatic, asymptomatic and immune states are added to better adapt to the real-world situation; (2) incubation and immune states adjacent to the symptomatic and asymptomatic state are separated to two states respectively for their infectivity and length differs; (3) the patients are divided into two sets: ``mild" and ``severe" to better model the hospitalization process. The mild patients do not need to be treated at hospital, and will recover automatically; the severe patients must be treated at hospital, otherwise it will die soon. Due to the lack of effective cure, treatment in hospital will not accelerate recovery of mild patients in our model.
 

\vspace{-1mm}
\section{SMADQN}\label{sec:marl}
\vspace{-1mm}

In our settings, although sufficient data are generated after each episode, the value-policy iterations are very slow for simulating a whole episode costs long time. So, we choose a value-based algorithm like DQN as our base algorithm rather than other policy-based algorithms, whose policy only changes a little in the local area after each training step. But vanilla DQN is still problematic for such a massive MARL problem. 
First, for the property of partial observation, the consequence of actions may only be observed after several steps, so one-step TD learning is not sufficient.
Second, it requires parameter sharing in policies to keep computation efficient with millions of agents. However, under deterministic policy, although the behaviour change for a single agent is incremental in training, the multi-agent system may experience dramatic change due to parameter sharing, which may lead to severe oscillation during training.
Third, ``Off-policy" methods may store some expired transitions in the replay buffer and disturb the training process while ``on-policy" method may cause huge oscillation.

To alleviate these problems, we make changes below to form our algorithm. 
First, we use TD($\lambda$) to train the Q-network. Like a standard DQN, we also have two Q-networks: the source network $Q$ and the target network $Q'$, while $Q$ is trained using MSE-loss to fit $\lambda$-returns. But to reduce computational cost, $\lambda$-returns are calculated using $Q'$ only after an episode finished and before training $Q$, and $Q'$ is only updated after training Q. So $\lambda$-returns do not have to be re-updated when $Q$ is being trained.
Second, like \cite{haarnoja2017reinforcement}, we also use Soft Q-learning where the probability of the individual $i$ choosing $a_i$ when observing $o_i$: $P(a_i|o_i)$ is directly proportional to $Q(a_i|o_i) / \alpha$, where $\alpha$ is the soft rate of Q-learning. Besides exploration provided by ``Softness", $\epsilon$-greedy method is added to encourage extra exploration. And $\epsilon$ is set to decrease gradually.
Third, we keep a balance between fully ``off-policy" and fully ``on-policy": We build 2 types of replay buffers: the temporary one stores trajectories generated from the current episode, and the permanent one stores trajectories used for DQN's training. When an episode finished, $\beta$ (in our implementation, $\beta = 1/3$) of trajectories in the permanent buffer are randomly chosen and replaced by random $\beta$ of trajectories in the temporal buffer, and the other trajectories in the temporal buffer are thrown away.

In our epidemic model, individuals in the same age stage share parameters with each other. We use $n_t=4$ sets of DQN network to model all individuals, one for each age group. 
More details of hyperparameters and pseudocodes are in the appendix.

\vspace{-1mm}
\section{Experiments}\label{sec:exp}
\vspace{-1mm}


    

    
 
This section aims to validate that the MPS is detailed, authentic, explainable and adaptive to external changes, hence can be used for decision makers to derive public strategies upon. We first show that our model has the ability to present comprehensive microscopic details in \cref{fig:corner}. Then, we show that our model can describe and predict the real-world dynamics by fitting the real world data on the spread of covid. Furthermore, in order to exploit our model and provide new candidate strategies for decision makers, we evaluate \textit{information disclosure}, a novel mitigation strategy which gives agents information on the infection status of each facility except household with a 2-day delay to account for the time for information collection. Finally, we further explore our model by running \textit{quarantine (QT)}, a widely-used government mitigation strategy, to show that our model is compatible with the main toolbox of real-world government and provide detailed information of the pandemic when executing this mitigation strategy for the decision makers.

\textbf{Basic Settings of Dataset}. We chose to evaluate and explore our model and SMADQN based on real-world data at Allegheny county in US, Pittsburgh. For simplicity, our model is closed in the sense that there is no incoming or outgoing agent (e.g., tourists). Based on ArcGIS~\cite{arcgis} and US population synthetic dataset~\cite{synpop}, there are $N=1188112$ residents (agents), among whom are $130451$ preschool children (0-9 years old), $114867$ students (10-18), $739183$ adults (19-64) and $200011$ seniors (65+). For the beginning of each episode, $2$ agents are chosen to be symptomatic (health state is ``sym") and $8$ agents to be exposed (health state is ``inc") near West Penn hospital, according to news~\cite{firstcovid}. For all scenarios, the action of first $10$ days is locked to be a fixed risky behaviour due to the latency of both the government and people; we chose $10$ days because it is the interval between the first discovered cases in Allegheny and the day when stay-at-home order is put into effect (and this is why there is a sudden infectivity drop in most experiments). To generate a model that catches the local epidemic contact structure better and is prepared for fine-grained government strategy, we modeled $14$ types of different facility and divided them into $4$ categories by the amount of infection risk. Their distributions are illustrated in the last subfigure of \cref{fig:corner}. 


    
    

\textbf{Calibration}. The first and most important requirement for any model applied in real-world decision making is authenticity to ground truth. Thus, before testing different government strategies with our model, we first calibrated our model to fit the real-world data in Allegheny between March 14th, 2020 (when the first two cases in Allegheny are discovered) and the end of May, after which mass unpredictable protest took place. If not specified, all experiments in this section run for $80$ days. And just as the real-world situation, the government only discloses the number of infected people on each day. 
There are three major criteria for our calibration: infected cases, hospitality and fatality. However, infected cases might be under-reported at the beginning of an outbreak due to insufficient testing. Therefore, we did not fit official reported cases, but infected cases inducted by hospitalized cases instead. The induction is based on data collected from US nationwide~\cite{covidabs}~\cite{CDCchart}.

Most hyper-parameters about Covid-19 can be obtained by previous works on the disease. But for the lack of data about individuals' behaviours, reward parameters are hard to summarize, and we could only hypothesize them. We left validation of reward parameters as future work. Among all reward parameters, $R_{ill}$ is the most important factor for agent behavior control, which is the immediate penalty of agents fallen ill: Higher $R_{ill}$ leads to more discreet agent behavior, such as wearing masks more often, stricter social distancing practice and less shopping frequency. Therefore, we mainly tuned $R_{ill}$ in the calibration. We also sampled different $R_{ill}$ in some experiments while fixing other reward parameters. Other hyper-parameters are left for sensitivity analysis in the appendix.
\Cref{fig:ca} show our calibration result. All curves in our paper are averaged from 3 $\times$ last 10 episodes of 3 random seeds after training for 100 episodes if not stated otherwise. For total infected cases and total hospitalization cases, our model yields a result with the error between predicted summed infected cases and the actual one being 2.72\% and the error for summed severe cases being 3.41\%. The error is calculated and averaged only using 20th to 80th days because data were not accurate enough at the very beginning of the epidemic. The death cases are less coherent but still reasonable. 
This result proves that our model is authentic and can be explored for more government policies. 

We modeled $R_{ill}$ as $4500(\frac{80-d}{80})^4+21000(\frac{d}{80})^4$ where $d$ is the day that first infection occurs. The reason is two-fold: first, it is clearly shown from the real-world news and data that people are (monotonically) increasingly aware of the virus, otherwise the number of infections will not drop; second, we tried to fit the data using $R_{ill}$ functions with the relatively simple form of power function for better explainability and generalizability. Another thing worth noting is that there is a steep drop of infectivity from day 10, which resembles the effect of stay-at-home order. See appendix for details about the timeline of government strategy in our model.

 \begin{figure}[b]
 \vspace{-10pt}
{
\begingroup
\setlength{\abovecaptionskip}{3pt plus 3pt minus 2pt}
\setlength{\belowcaptionskip}{3pt plus 3pt minus 2pt}
\renewcommand{\arraystretch}{0}
\pgfplotsset{width=0.2\linewidth, every axis legend/.append style={
at={(1.02,1)},
anchor=north west}}
\pgfplotsset{
legend image code/.code={
\draw[mark repeat=2,mark phase=2]
plot coordinates {
(0cm,0cm)
(0.15cm,0cm)        
(0.3cm,0cm)         
};%
}
}
\begin{center}
\begin{tabular}{@{\hspace{-1.2\tabcolsep}} c  @{\hspace{-1.3\tabcolsep}} c  @{\hspace{-0.5\tabcolsep}} c  @{\hspace{-1.3\tabcolsep}} c}
  \input{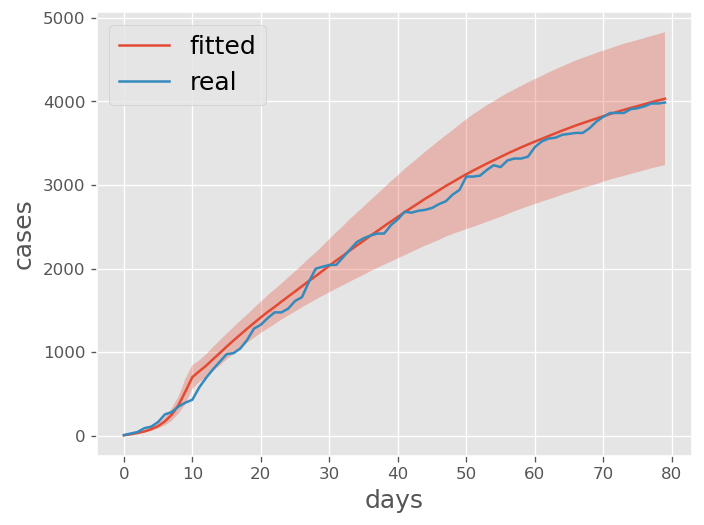}&
  \input{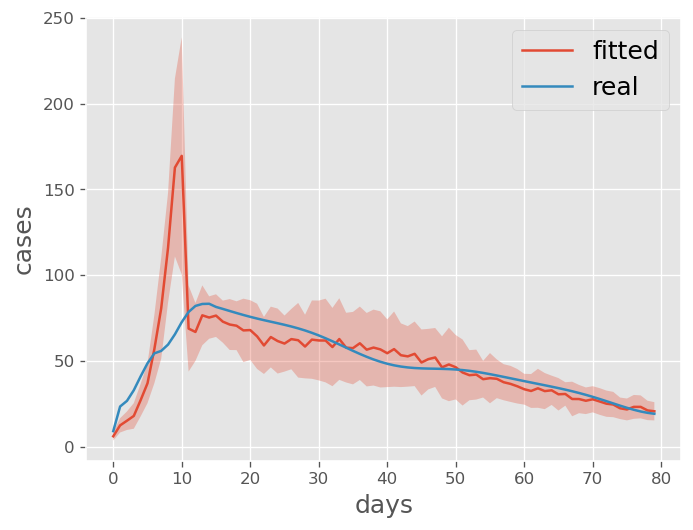}
  &\input{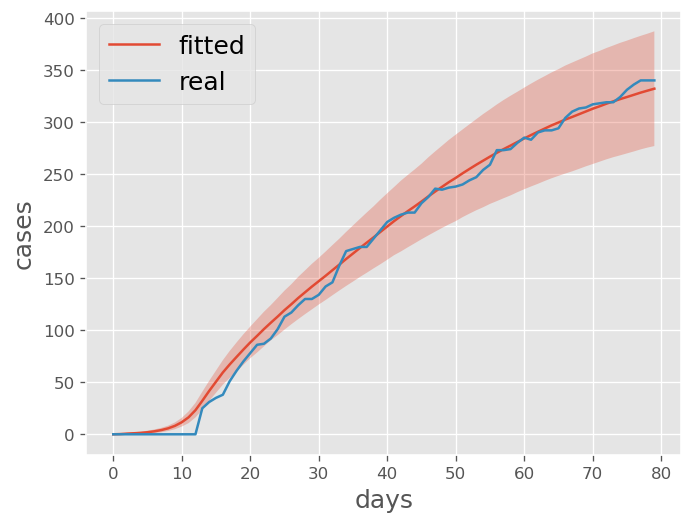}&\input{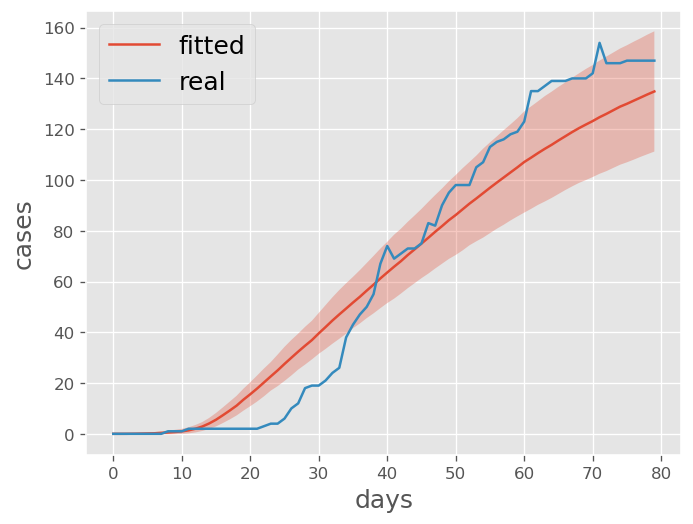}\\
  \small a) total cases&\small b) daily cases&\small c) hospitalization &\small d) fatality
\end{tabular}
\end{center}
\endgroup
}
\caption{The result of calibration. The 4 sub-figures present total cases, daily cases, total hospitalization, and total deaths respectively. The real daily cases curve is smoothed for better readability.}
\label{fig:ca}
\end{figure}


{
\begingroup
\begin{figure}[h]
\begin{minipage}[c]{0.24\linewidth}
\includegraphics[width=\linewidth]{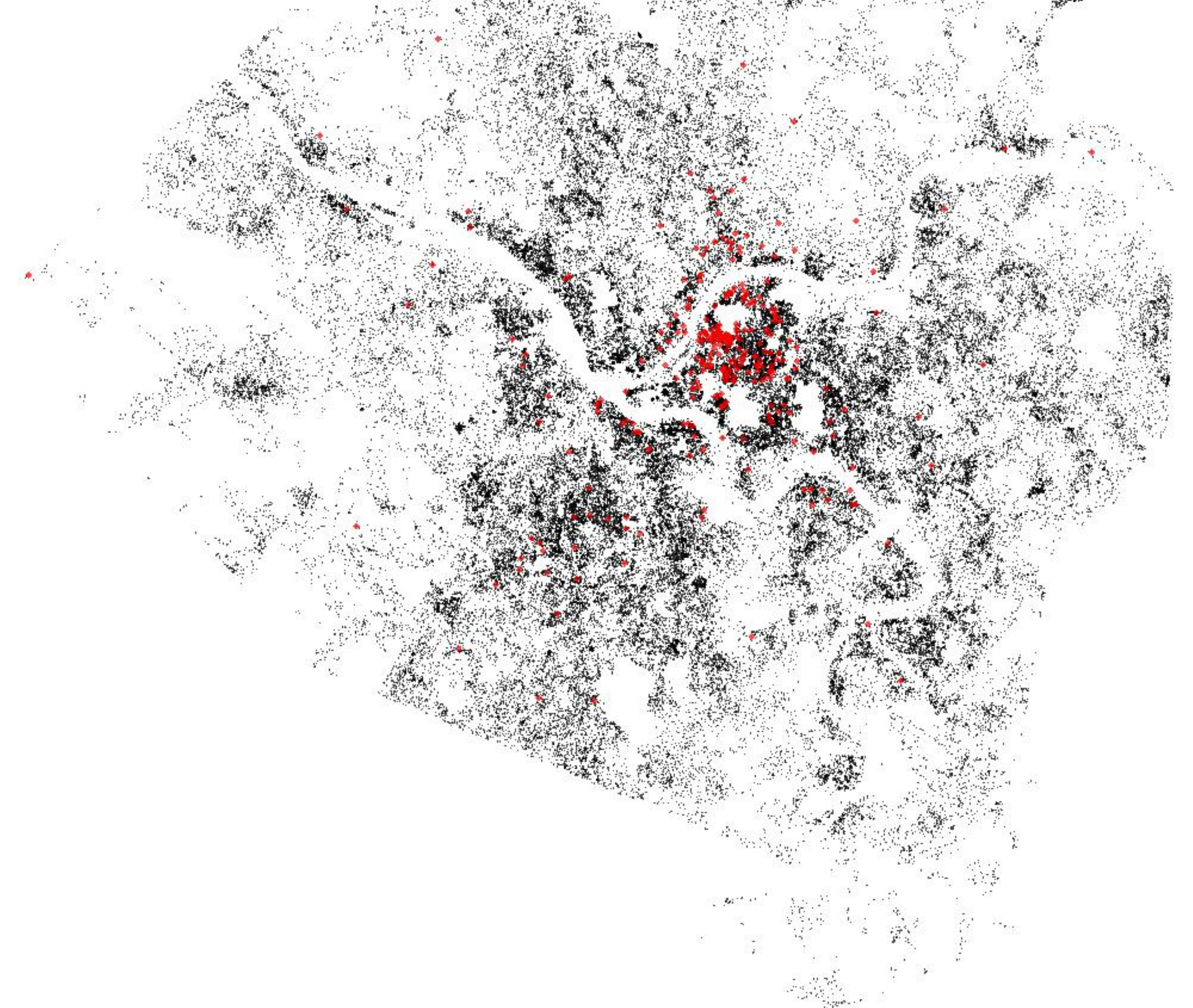}
\caption*{\scalebox{0.85}{a) 10d from center}}
\includegraphics[width=\linewidth]{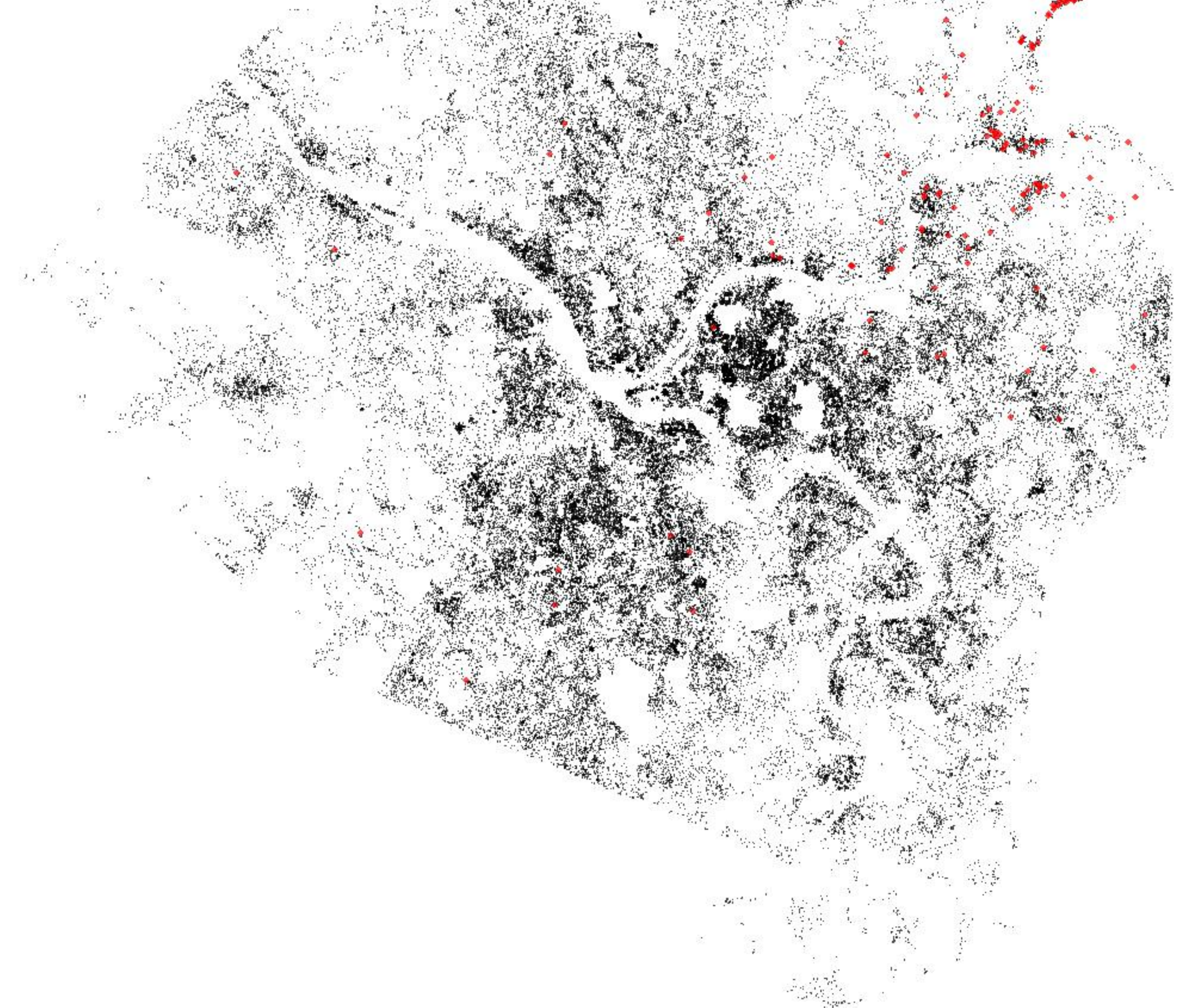}
\caption*{\scalebox{0.85}{d) 10d from corner}}
\end{minipage}
\begin{minipage}[c]{0.24\linewidth}
\includegraphics[width=\linewidth]{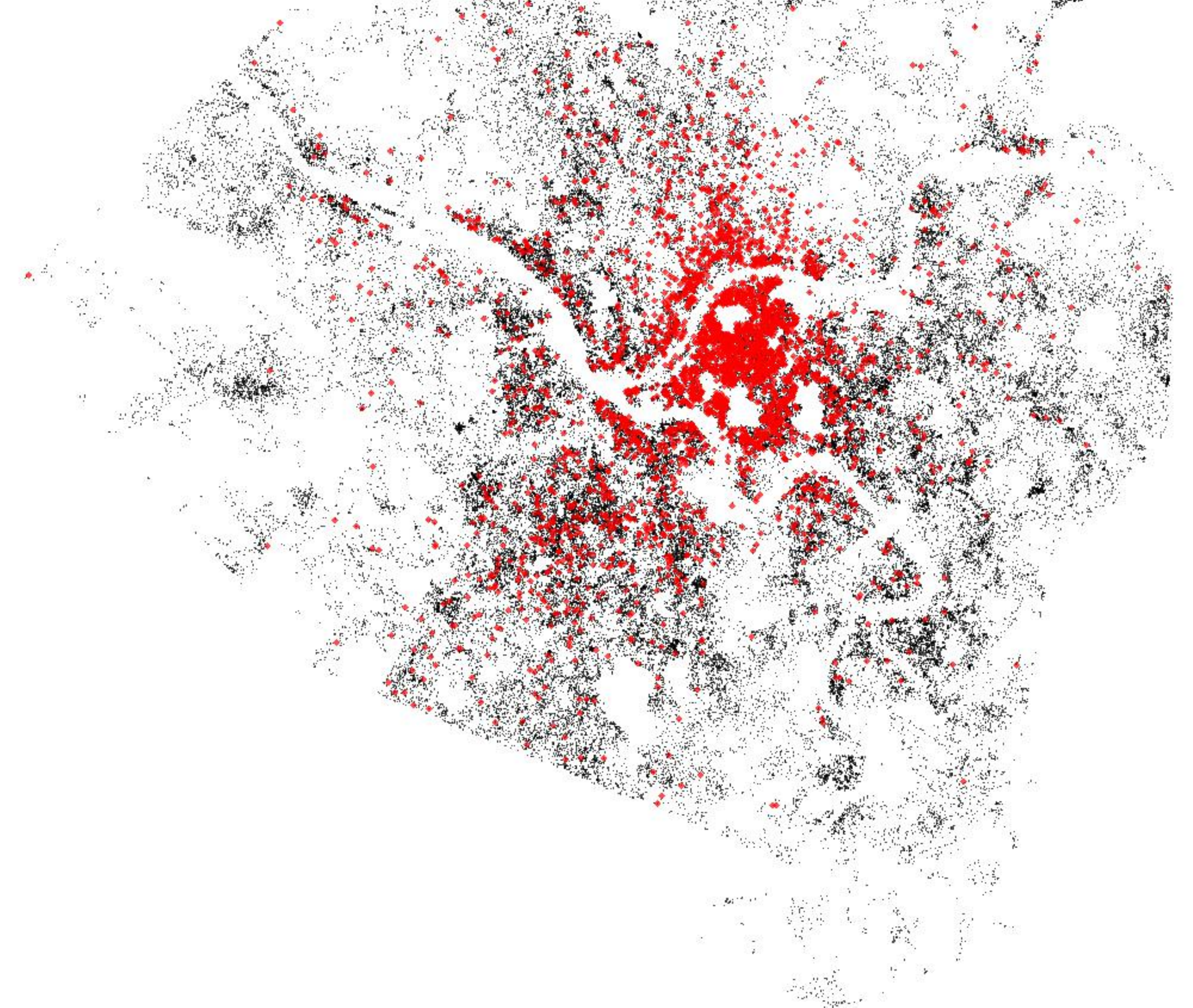}
\caption*{\scalebox{0.85}{b) 25d from center}}
\includegraphics[width=\linewidth]{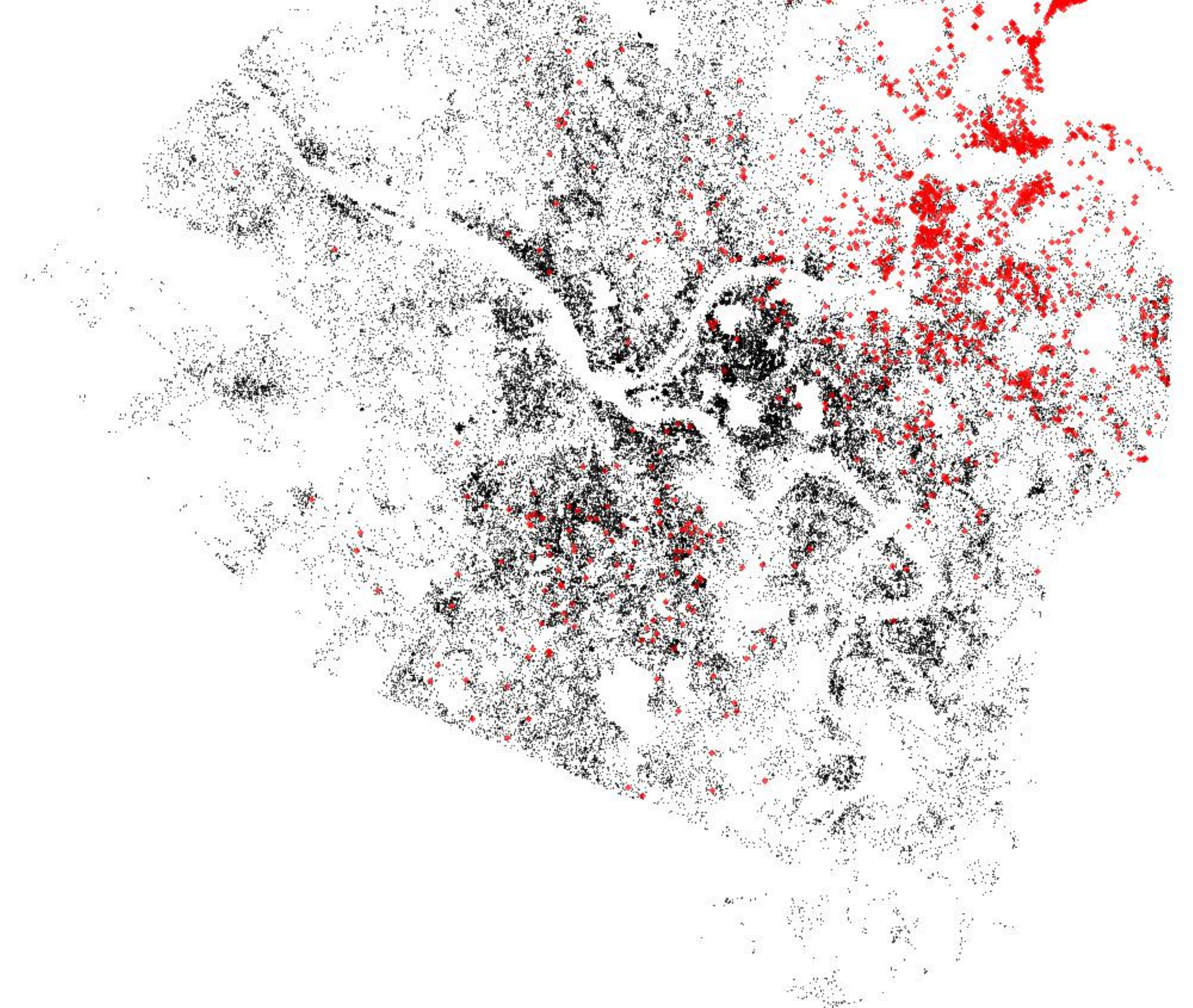}
\caption*{\scalebox{0.85}{e) 25d from corner}}
\end{minipage}
\begin{minipage}[c]{0.24\linewidth}
\includegraphics[width=\linewidth]{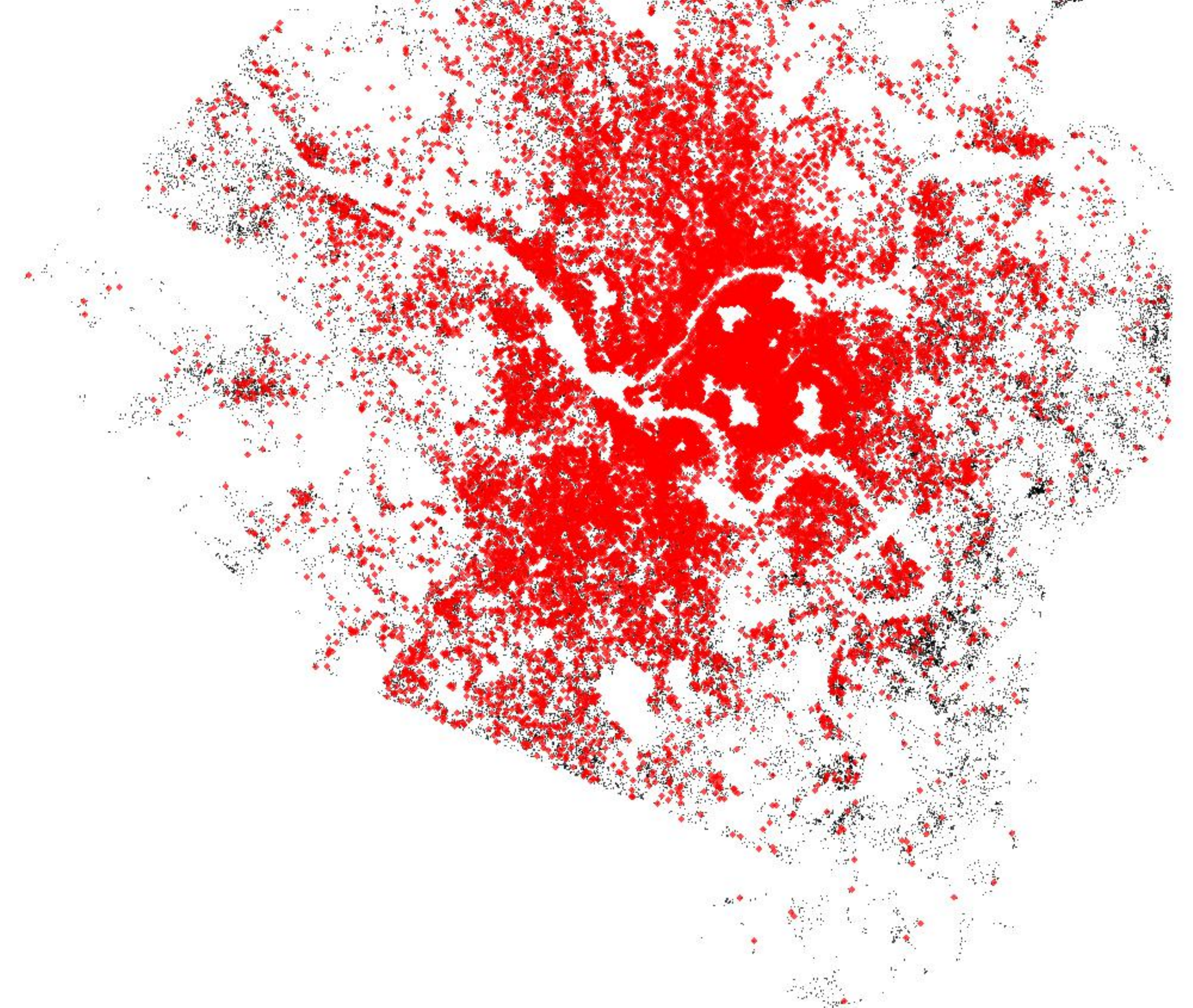}
\caption*{\scalebox{0.85}{c) 40d from center}}
\centering
\includegraphics[width=\linewidth]{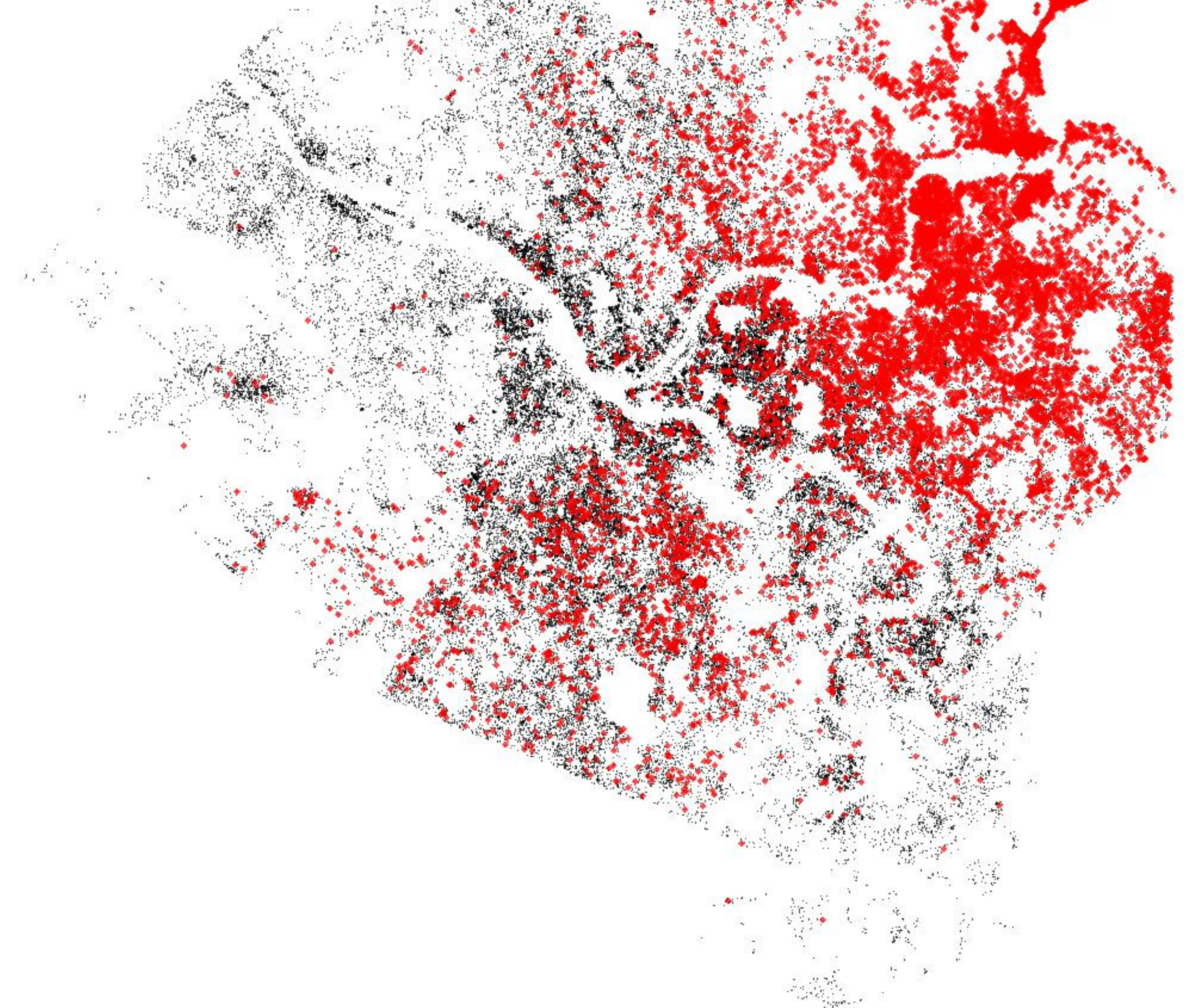}
\caption*{\scalebox{0.85}{f) 40d from corner}}
\end{minipage}
\begin{minipage}[c]{0.24\linewidth}
\centering
\includegraphics[width=0.88\linewidth]{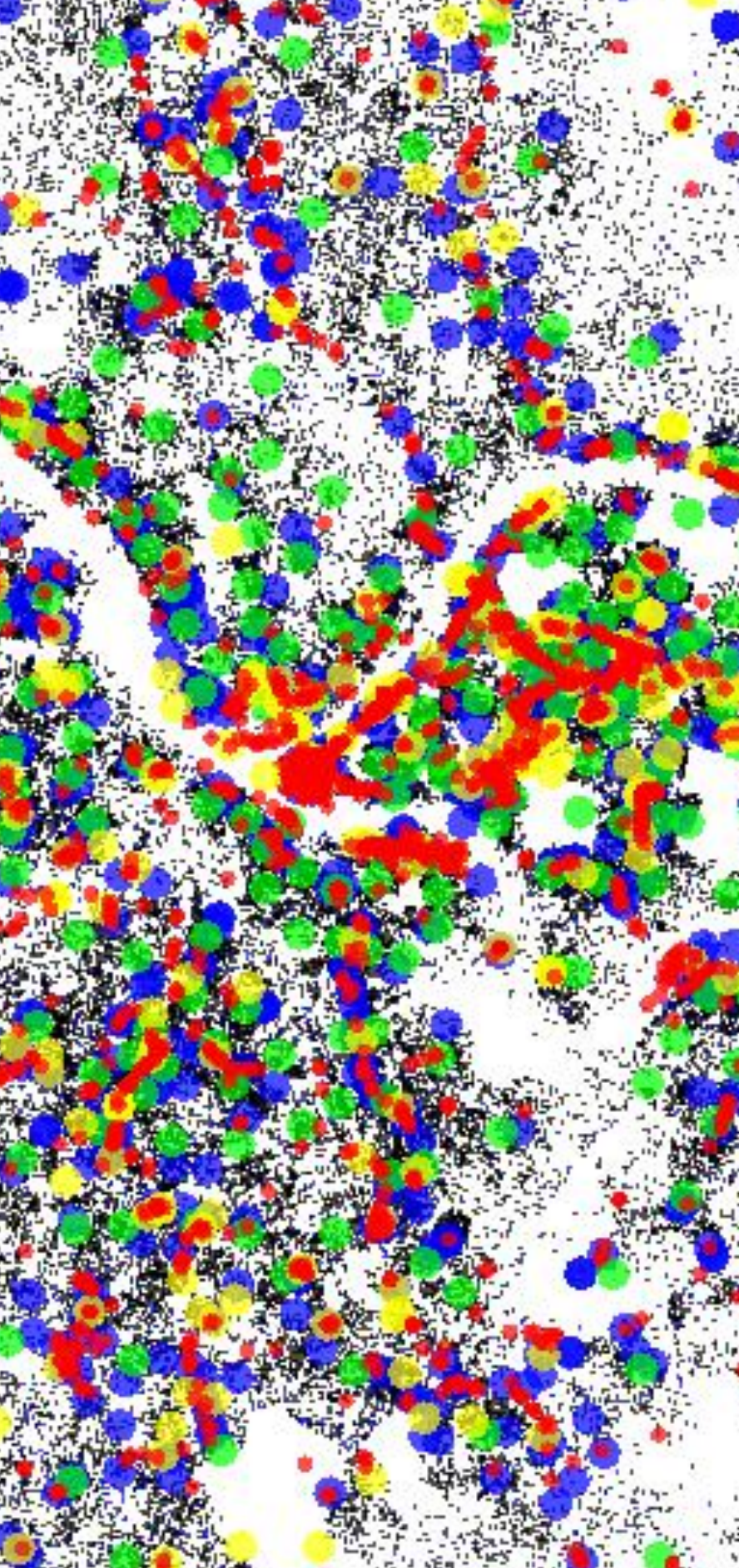}
\caption*{\scalebox{0.85}{g) city overview}}

\end{minipage}
\caption{An illustration of how our microscopic model captures the local structure of the spread of disease and some other details. 
 a) to c) are the global infection status of day 10, 25 and 40 starting from downtown, while d) to f) are the status of day 10, 25 and 40 starting from the edge of the map. The point is red if the household has at least 1 victim, and black otherwise. g) is an illustration of facility distribution in downtown area. The black points are households, and the red, yellow, green and blue points are the facilities with MinAct as $3$, $2$, $1$ and $0$.}
\end{figure}
\label{fig:corner}
\endgroup
}
    
\begin{figure}[h]
{
\begingroup
\small
\setlength{\abovecaptionskip}{3pt plus 3pt minus 2pt}
\setlength{\belowcaptionskip}{3pt plus 3pt minus 2pt}
\renewcommand{\arraystretch}{0}
\pgfplotsset{width=0.2\linewidth, every axis legend/.append style={
at={(1.02,1)},
anchor=north west}}
\pgfplotsset{
legend image code/.code={
\draw[mark repeat=2,mark phase=2]
plot coordinates {
(0cm,0cm)
(0.15cm,0cm)        
(0.3cm,0cm)         
};%
}
}
\begin{center}
\begin{tabular}{@{\hspace{-1\tabcolsep}}c  @{\hspace{-1.5\tabcolsep}} c  @{\hspace{-1.5\tabcolsep}} c  @{\hspace{-1.5\tabcolsep}} c}
  \input{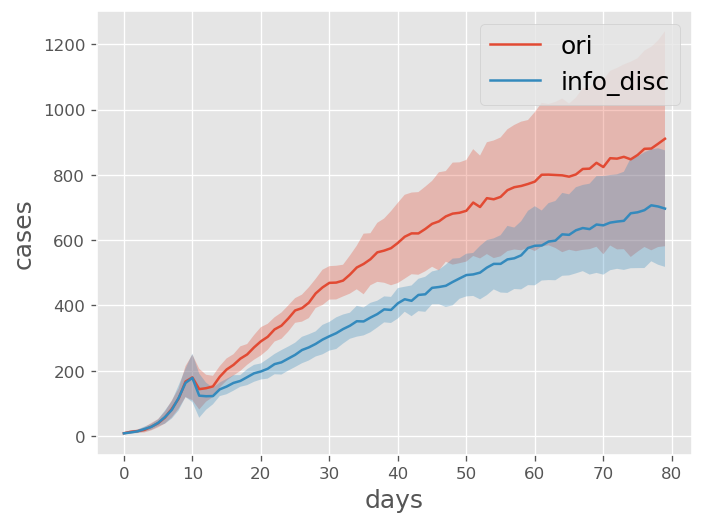}&\input{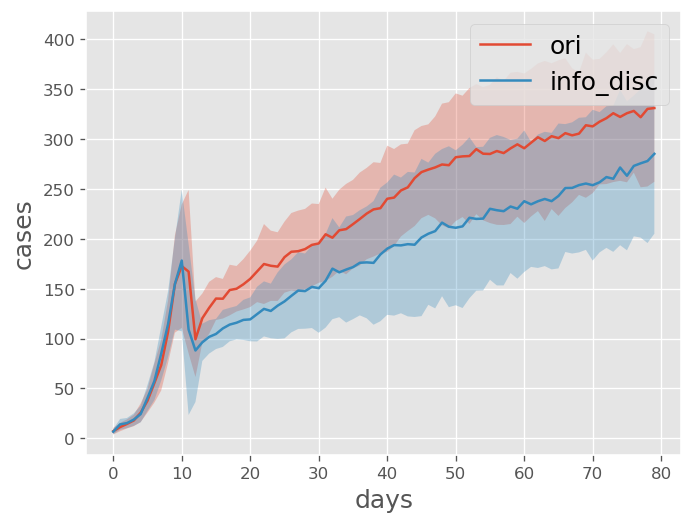}&\input{pic/new217/217dic10k_2}&\input{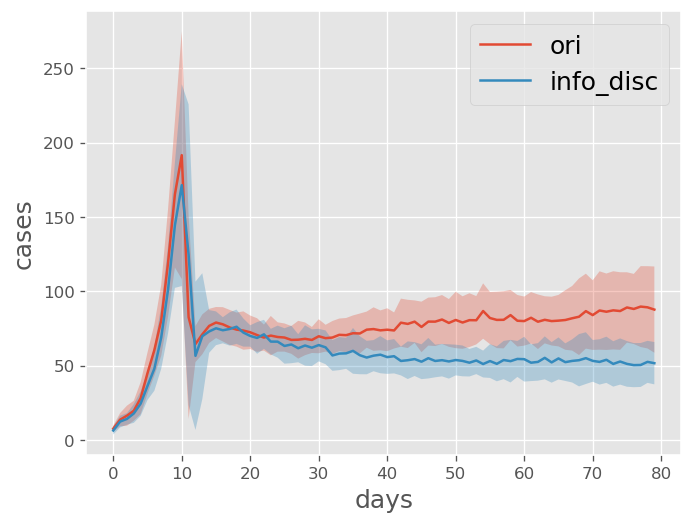}\\
  \input{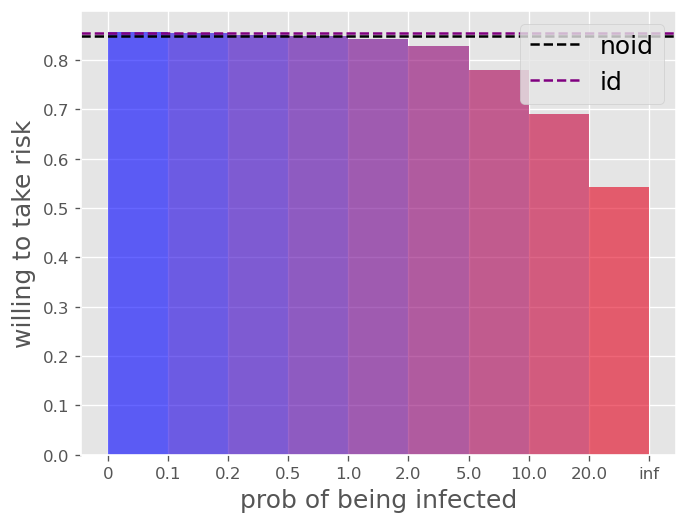}&\input{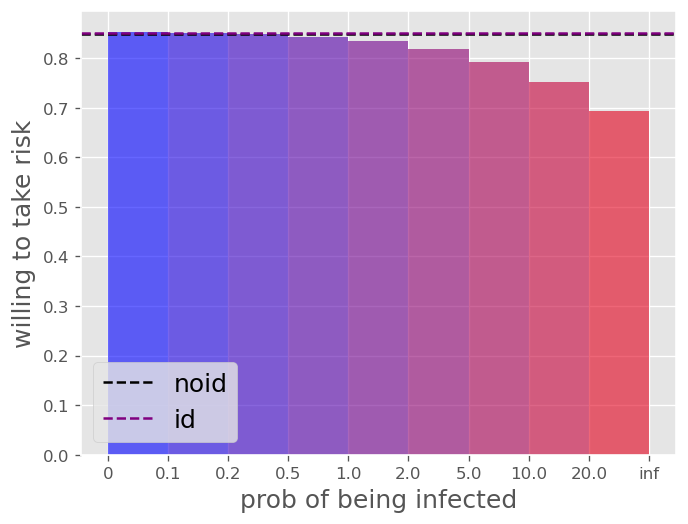}&\input{pic/new217/217act10k_2}&\input{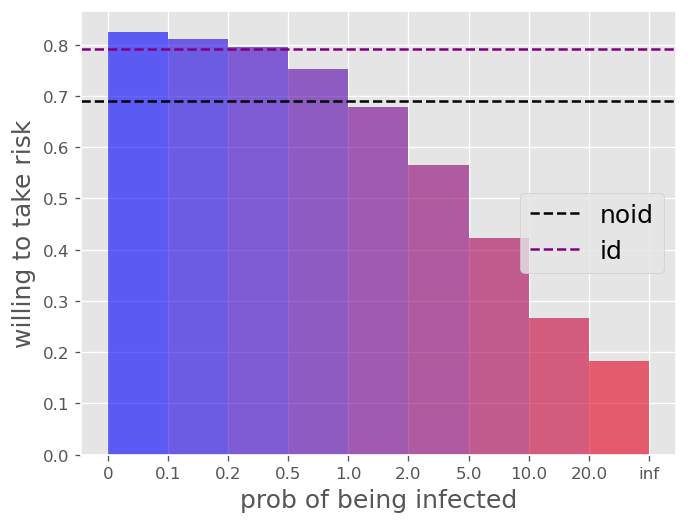}\\
  \input{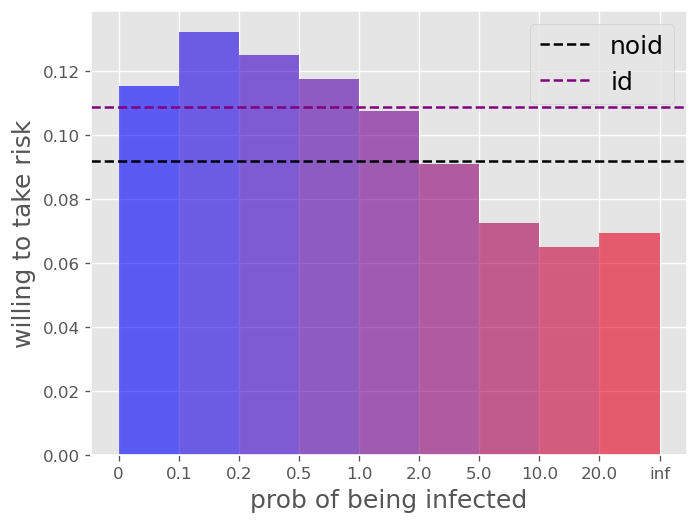}&\input{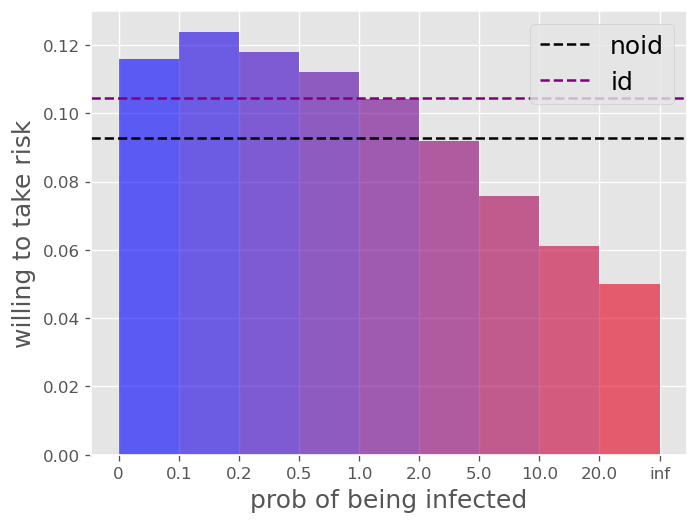}&\input{pic/new217/217mar10k_2}&\input{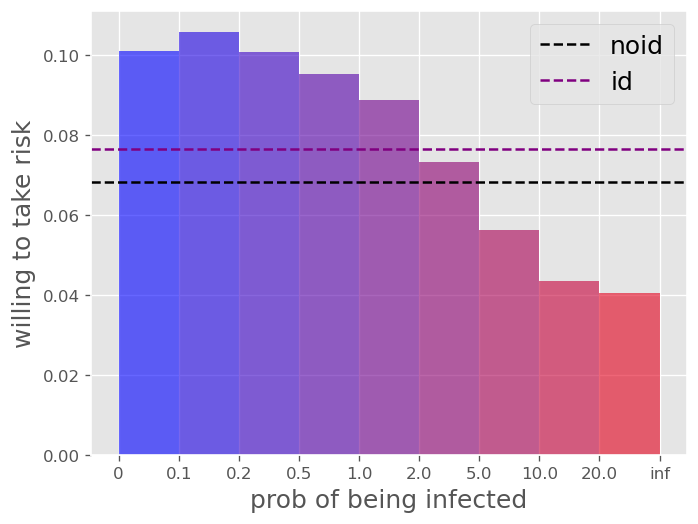}\\
  \input{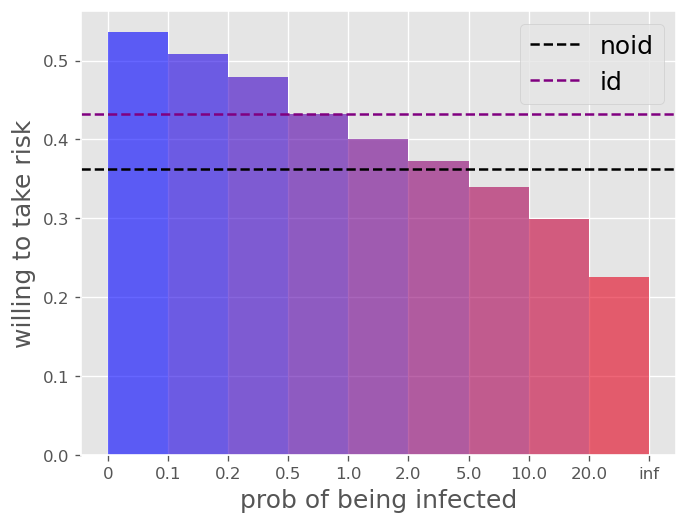}&\input{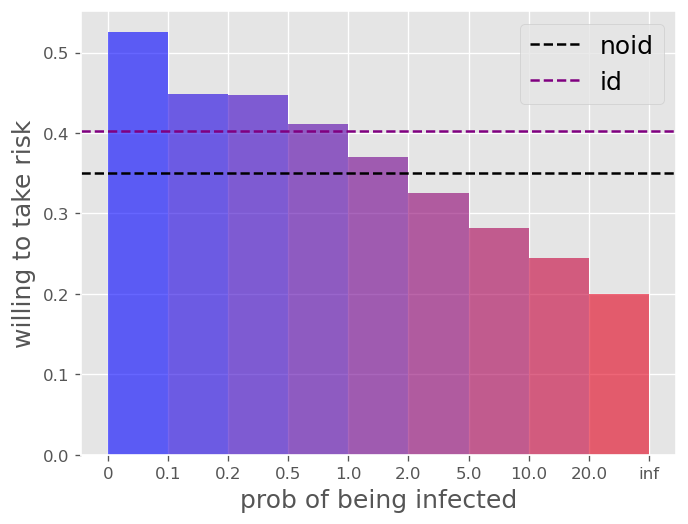}&\input{pic/new217/217mas10k_2}&\input{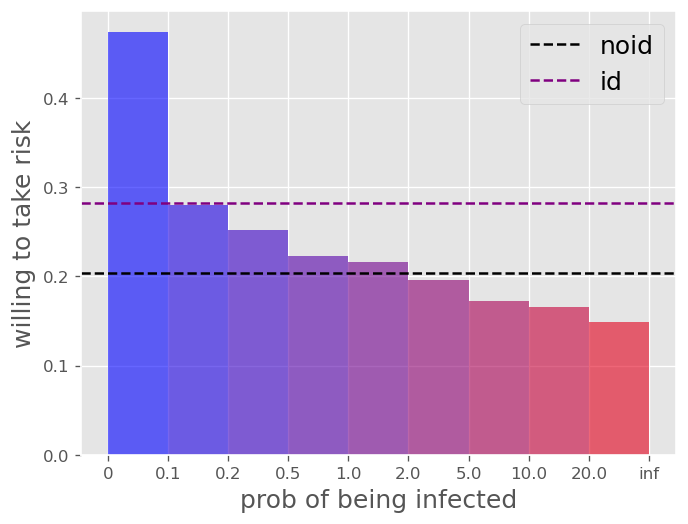}\\
  \quad
  \text{ R\textsubscript{ill}=1k}&\quad\text{ R\textsubscript{ill}=3k}&\ \text{ R\textsubscript{ill}=10k}&\text{R\textsubscript{ill}=25k}
\end{tabular}
\end{center}
\endgroup
}
\caption{The experiment result of the \textit{information disclosure} strategy. Every column is a different setting of $R_{ill}$. 
}
\label{fig:id}
\end{figure}

\textbf{Information Disclosure}. Transparency is crucial for epidemic outbreak handling; not only will timely information about the infected cases alleviate panic, but ideally, when information is clear enough, agents will avoid infection sources for themselves and return to normal when the source disappears, thus minimizing the impact to economy and government costs. This proposes a new candidate strategy, \textit{information disclosure}, to test the effect of pandemic control under ideal information coverage without any mandatory measures.
 In \textit{information disclosure}, governments will keep monitoring new cases, and announce the probability of being infected in every non-household facility 2 days ago on a daily basis (we assumed that the government needs 2 days to collect information). \Cref{fig:id} shows the result of our experiments; the higher the $R_{ill}$ is, which indicates that people are more serious about the disease, the lower the total number of cases is. The first row of \cref{fig:id} clearly indicates that \textit{information disclosure} is an effective strategy since it results in less cases than that without \textit{information disclosure} in all four scenarios. It indeed reduces the number of daily cases, but the strategy may not be strong enough when individuals are less discreet since it still cannot flatten the curve when $R_{ill} \leq 3k$. The next rows illustrate how our RL model learns an adaptive response to different levels of risks. Higher observed risk leads to a drop in choosing $A_{act}=3$ and offline shopping, as well as a rise in mask rate. Agents generally learned to avoid danger when the risk of infection increases, without explicitly setting a script for exact policy. More importantly, the learned policy has a higher average probability of taking risk with better control of epidemic, which illustrates that the behavior learned by RL is non-trivial in the sense that agents adopts a better policy to mitigate the pandemic while keeping a more normal life, which is possible because people in more dangerous areas are much more discreet. This effect slows down the spread dramatically.

\begin{figure}[h]
{
\begingroup
\small
\renewcommand{\arraystretch}{0}
\setlength{\abovecaptionskip}{2pt plus 1pt minus 1pt}
\setlength{\belowcaptionskip}{2pt plus 1pt minus 1pt}
\pgfplotsset{width=0.2\linewidth,every axis legend/.append style={
at={(1.02,1)},
anchor=north west}}
\pgfplotsset{
legend image code/.code={
\draw[mark repeat=2,mark phase=2]
plot coordinates {
(0cm,0cm)
(0.15cm,0cm)        
(0.3cm,0cm)         
};%
}
}
\begin{center}
\begin{tabular}{@{\hspace{-1\tabcolsep}}c  @{\hspace{-1.3\tabcolsep}} c  @{\hspace{-1.3\tabcolsep}} c  @{\hspace{-1.3\tabcolsep}} c}
  \input{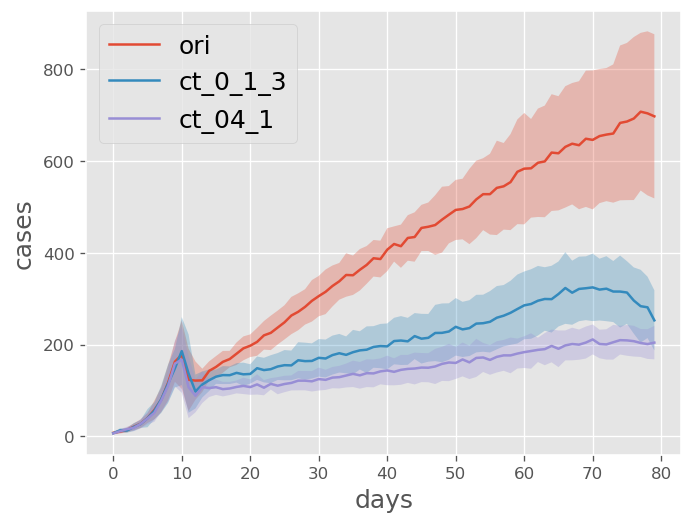}&\input{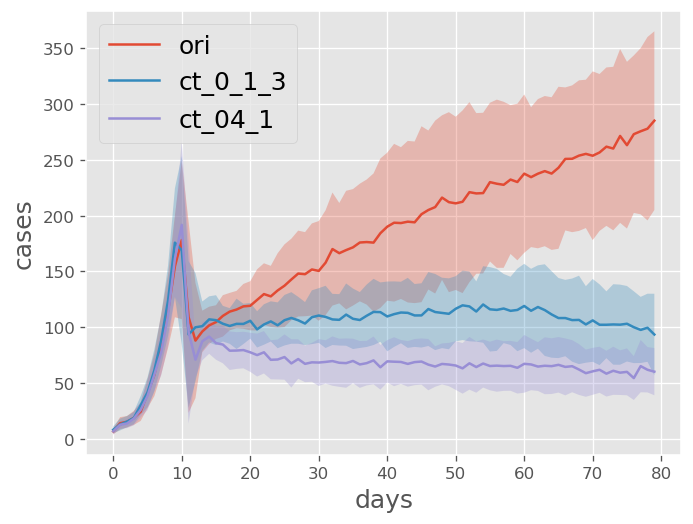}&\input{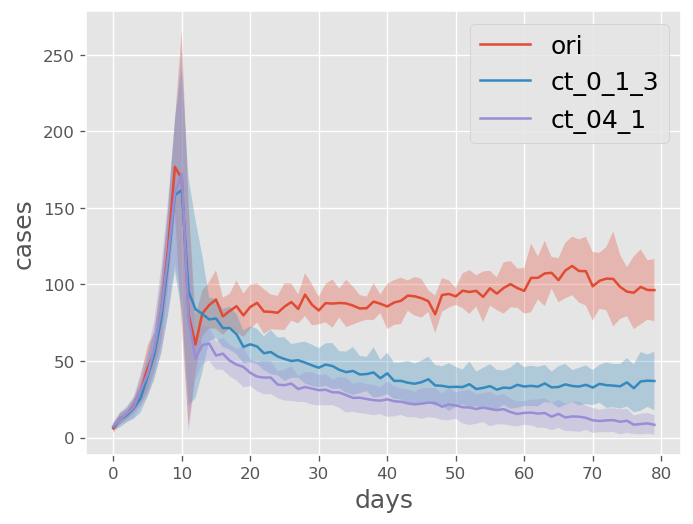}&\input{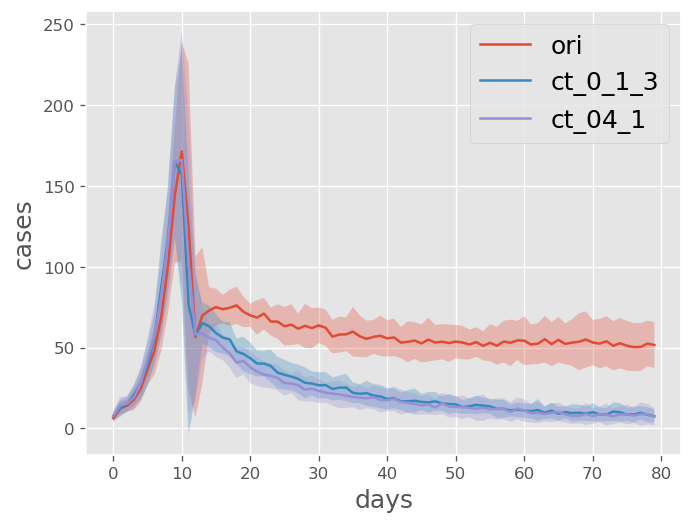}\\
  \input{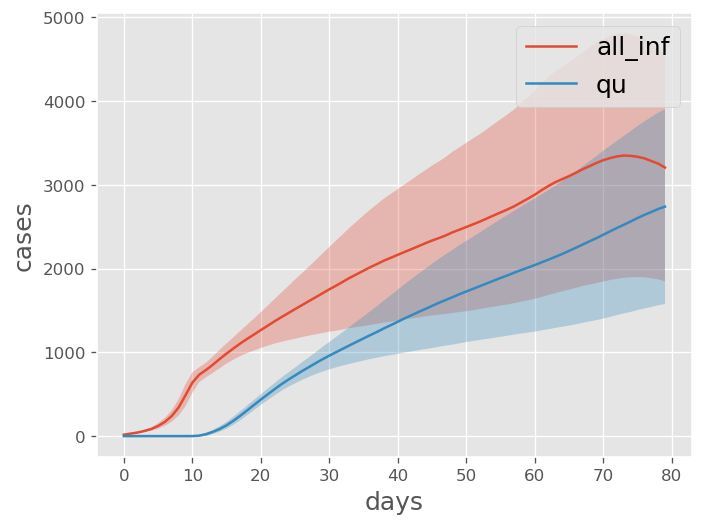}&\input{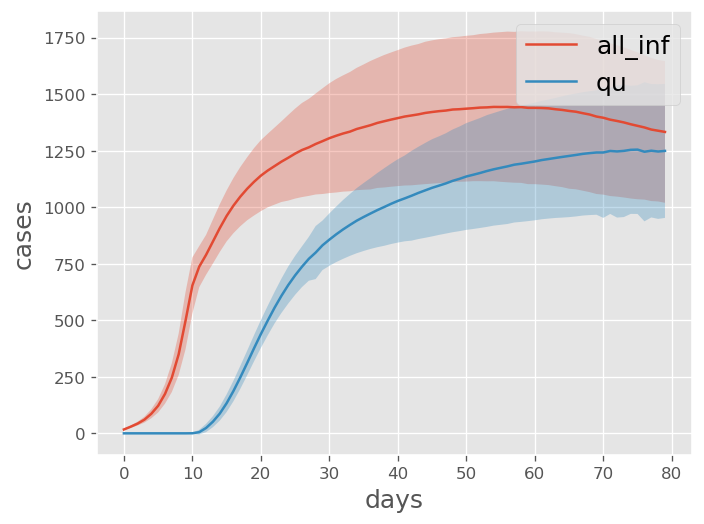}&\input{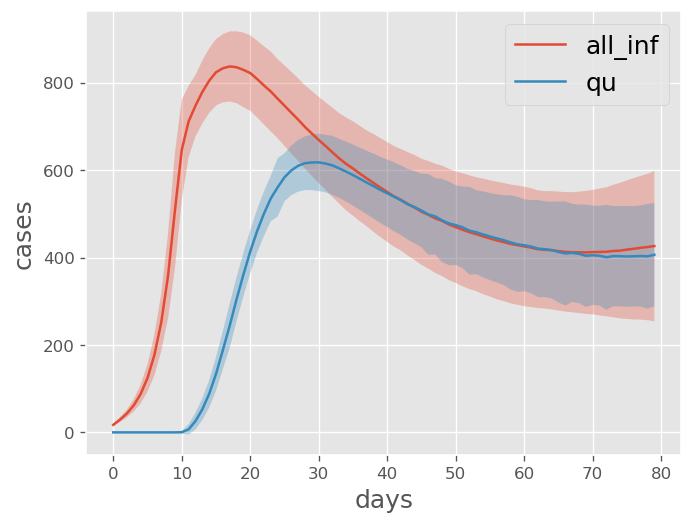}&\input{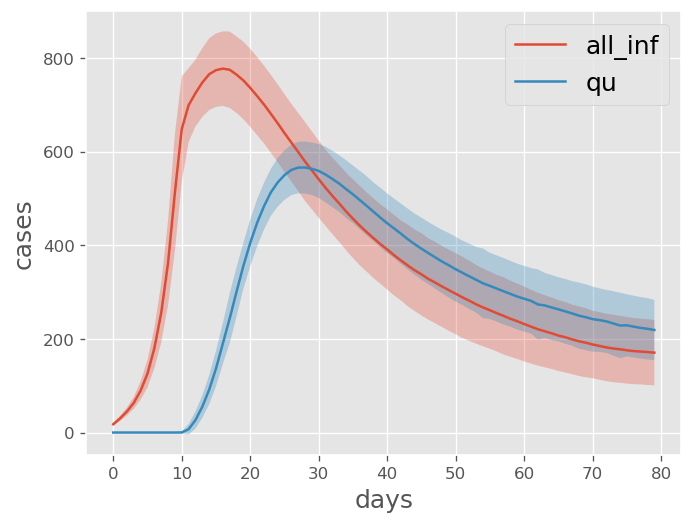}\\
  \input{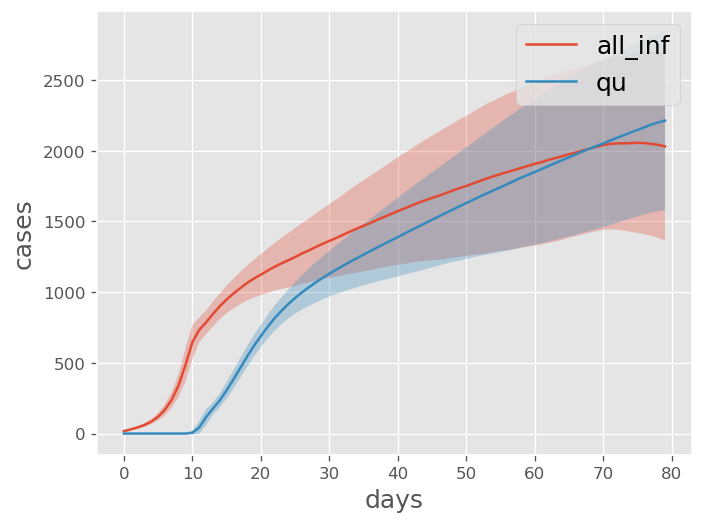}&\input{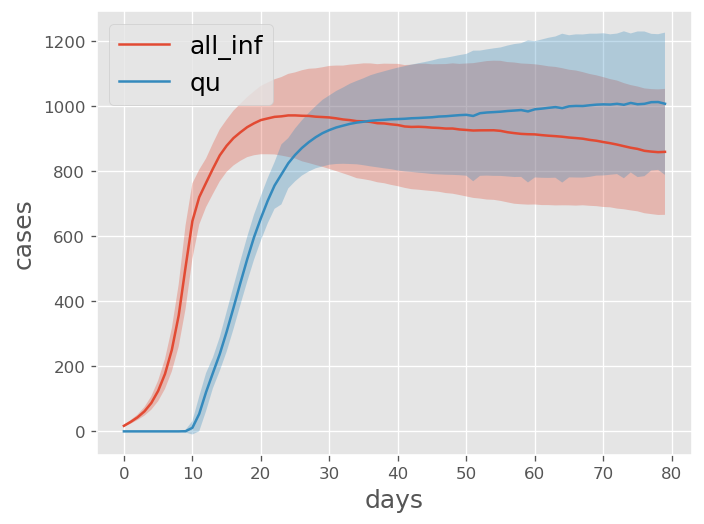}&\input{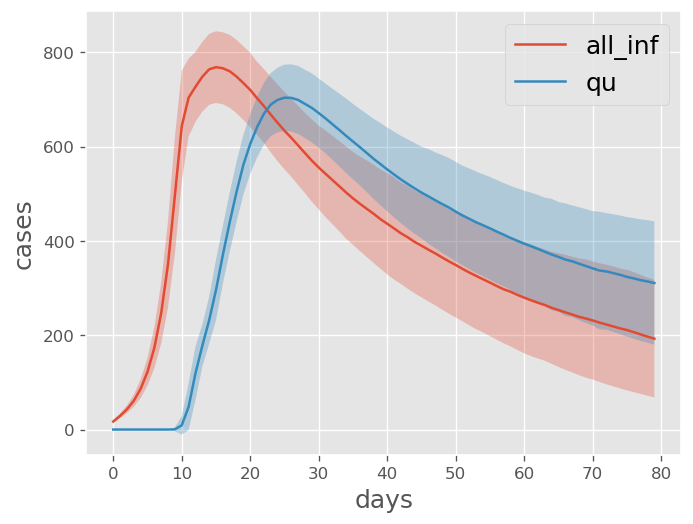}&\input{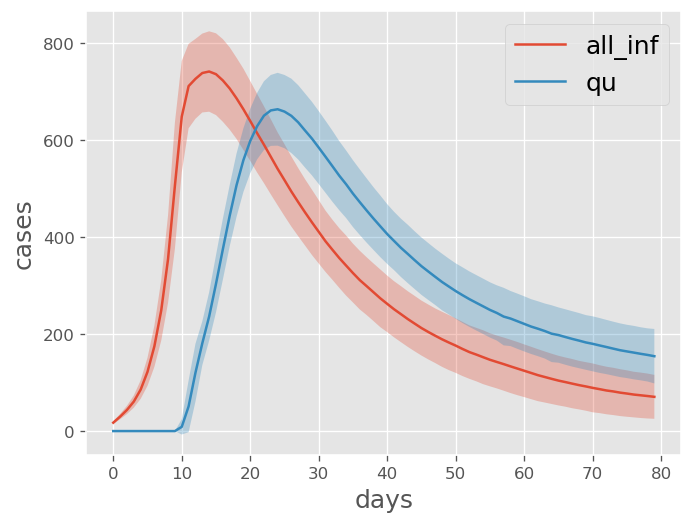}\\
   \text{\qquad\quad R\textsubscript{ill}=1k}&\quad\text{ R\textsubscript{ill}=3k}&\ \text{ R\textsubscript{ill}=10k}&\text{\quad R\textsubscript{ill}=25k}
\end{tabular}
\end{center}
\endgroup
}
\caption{The experiment result of the \textit{quarantine} (QT) strategy. Every column is a different setting of $R_{ill}$. The first row is the total cases of infection under three cases (red for no measure, blue for weak quarantine, and purple for strong quarantine. The second and third rows are the number of people isolated (blue) versus current number of cases (red) in weak QT and strong QT.}
\label{fig:qt}
\end{figure}

\textbf{Quarantine}. 
In this scenario, besides information disclosure, the government will isolate every agent having been symptomatic for above 2 days (we assumed that it's inevitable for governments to have delay). We tested two strategies with different strength: The weaker strategy can only discover and quarantine a 3-day-infected people with probability = 1/3 each day. While The stronger one can discover it with probability = 1, and the strategy can also quarantine $40\%$ agents that has directly infected by it as well, which needs contact tracing in reality. The isolated agent is excluded from the contact network. It cannot infect anybody upon isolation, including those in the same household, and will be released $9$ days after recovery (in ``ima" or ``ims" state). \Cref{fig:qt} shows the result of the three scenarios: no measure, weak quarantine, and strong quarantine. The first row shows that quarantine is very effective with most patients isolated from the contact network; even the weak quarantine can effectively control the pandemic (flatten the curve), and the strong quarantine eliminates pandemic within 80 days in half scenarios. The following two rows respectively shows the number of people isolated versus current number of cases under weak and strong quarantine strategy.


\vspace{-1mm}
\section{Conclusion}\label{sec:conc}
\vspace{-1mm}



This paper introduced a microscopic epidemic simulator and the SMADQN algorithm to deal with millions of agents. We synthesized a comprehensive dataset for Allegheny county, US for evaluation, and explored two possible government policy candidates on our model: \textit{information disclosure} and \textit{quarantine}. Both proved useful for epidemic control.

\vspace{-1mm}
\appendix
\section{The Dataset based on Demographic Data of Pittsburgh}
\subsection{Data Source}

Our dataset of Pittsburgh includes the following two types of information:
\begin{itemize}
    \item Agents' characteristics including their ages and trajectories/positions.
    \item Facility characteristics including locations, capacities and members of different facilities. For households, workplace and schools, we adopted the data from ~\cite{synpop}. For recreational places, we utilized the data from \cite{arcgis}. Regarding hospitals in Allegheny county, data from \cite{hospital} was adopted.
\end{itemize}

\subsection{Age Characteristics}

\begin{figure}[h!]
    \centering
    \includegraphics[width=\linewidth]{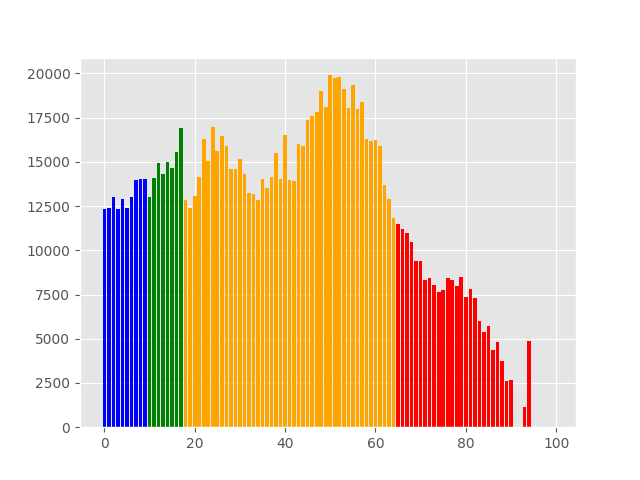}
    \caption{The distribution of residents' ages in the US synthetic population dataset; the total population is $1,188,112$. Blue, green, orange and red data stands for ``chd" (children, 0-9 years old),``sch" (school students, 10-17 years old),``adu" (adults, 18-64 years old) and ``rtr" (retired people, 65+ years old) respectively.}
    \label{fig:agesyn}
\end{figure}

\Cref{fig:agesyn} shows the distribution of residents' ages in our dataset. The distribution is synthesized based on the US population dataset \cite{population}.

\subsection{Facility Characteristics}
\Cref{tab1} lists the numbers of different types of facilities and their corresponding capacities (on average and maximally) in our synthesized dataset. In total, we have 14 types of facilities.
\begin{table}[h]
    \centering
    \begin{tabular}{p{50pt}|p{30pt}|p{60pt}|p{50pt}}
    \hline
        Facility & Number & Capacity (Average) & Capacity (Max)\\
        \hline
        Hospital & 14 & 1577.43+384 & 8613+1542\\
        Household & 533919 & 2.22527 & 13\\
        Workplace & 38333 & 13.5461 & 7741\\
        School & 338 & 517.583 & 2239\\
        Retail & 601 & 247.977 & 1533\\
        Supermarket & 87 & 1900.97 & 5380 \\ Community & 358 & 3318.75 & 15889\\
        Library & 88 & 302.17 & 3069\\
        Museum & 78 & 77.8333 & 2347 \\
        Gym & 193 & 110.782 & 1797 \\
        Restaurant & 2691 & 201.458 & 3462\\
        Stadium & 3 & 3963.33 & 3976\\
        Theatre & 59 & 159.576 & 758\\
        Cinema & 36 & 367.917 & 3811 \\
        \hline
    \end{tabular}
    \vspace{2mm}
    \caption{The number, average capacity and maximum capacity of each type of facility (for hospitals, the capacity is in the ``doctor + beds" format). We only model general hospitals in our dataset; specialized medical facilities such as rehabilitation centers and women's hospitals are excluded.}
    \label{tab1}
\end{table}

\subsection{Construction of the Contact Network}
We followed the paradigm in \cite{2ndwave} and \cite{kompella2020reinforcement} to construct the contact network among agents. The principles are: 1) all agents are affiliated with certain facilities; 2) at each simulation step, agents will be assigned into the affiliated facilities as long as they are not dead or hospitalized; 3) within the same facility, each pair of agents have equal probability to meet each other, i.e., equal probability to spread virus potentially; and 4) once an agent enters a facility, a minimum level of activity is required (see \cref{tab1} in the main manuscript for details about the requirements).

Following such principles, we first assigned each agent to a household based on the dataset in \cite{synpop}. Similarly, agents above a certain age are also attached to schools and work places. However, the work in \cite{synpop} didn't provide other types of facilities as in our dataset (\Cref{tab1}). To solve such a problem, we proposed to assign agents to facilities based on their distances to the facilities. The process is as follows:
\begin{itemize}
    \item Step I: extract facility characteristics such as locations and capacities from \cite{arcgis};
    \item Step II: determine the connection between agents and facilities based on their characteristics. For instance, children go to schools, adults go to workplaces, but the retired people go neither.
    \item Step III: establish the distribution of agents for each connected agent-facility group. We obtained such distributions by approximating the ratios of people visiting facilities in data from \cite{freq}, \cite{freqres} and \cite{statista}. Two major factors are considered here: capacities of the facilities and distances between agents and facilities. We prioritized agents closer to facilities, and facilities with larger capacities. For example, the priority radii for each facility type are $2$km for conventional or retail stores, $5$km for restaurants, gyms and supermarkets, $10$km for theaters/cinemas, libraries and museums, and $35$km for stadiums. \Cref{fig:BG}, \cref{fig:CVS}, \cref{fig:GYM} and \cref{fig:MUS} illustrate agent allocation for different types of facilities. Note that hospitals are treated differently. We assume agents will go to the nearest hospital with remaining capacity $>0$ considering the fact that life threaten is of highest priority. 
\end{itemize}

{
\begingroup
\begin{figure*}[h]
\centering
\begin{minipage}[c]{0.23\linewidth}
\includegraphics[width=\linewidth]{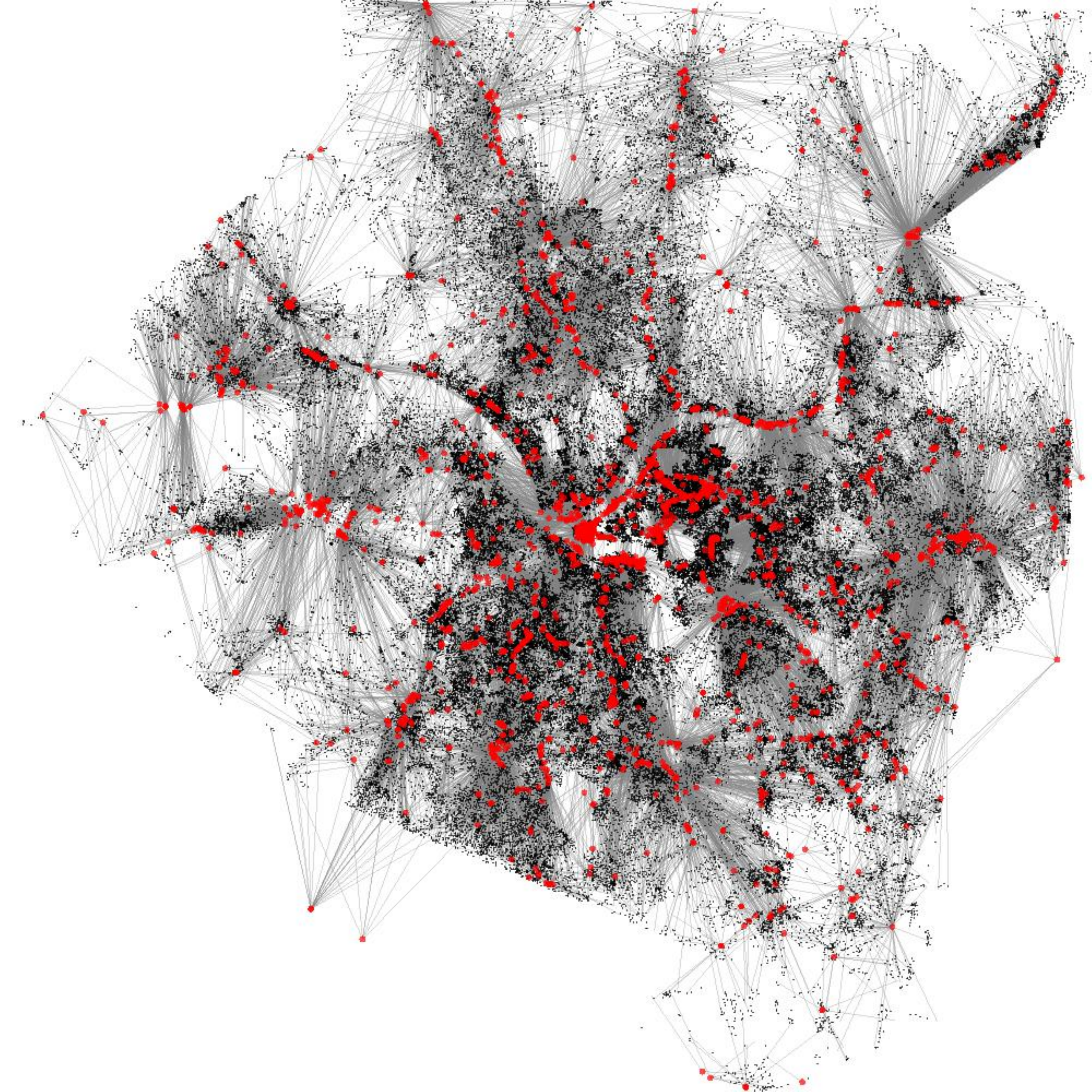}
\caption*{\scalebox{1}{restaurants (red)}}
\label{fig:BG}
\end{minipage}
\begin{minipage}[c]{0.23\linewidth}
\includegraphics[width=\linewidth]{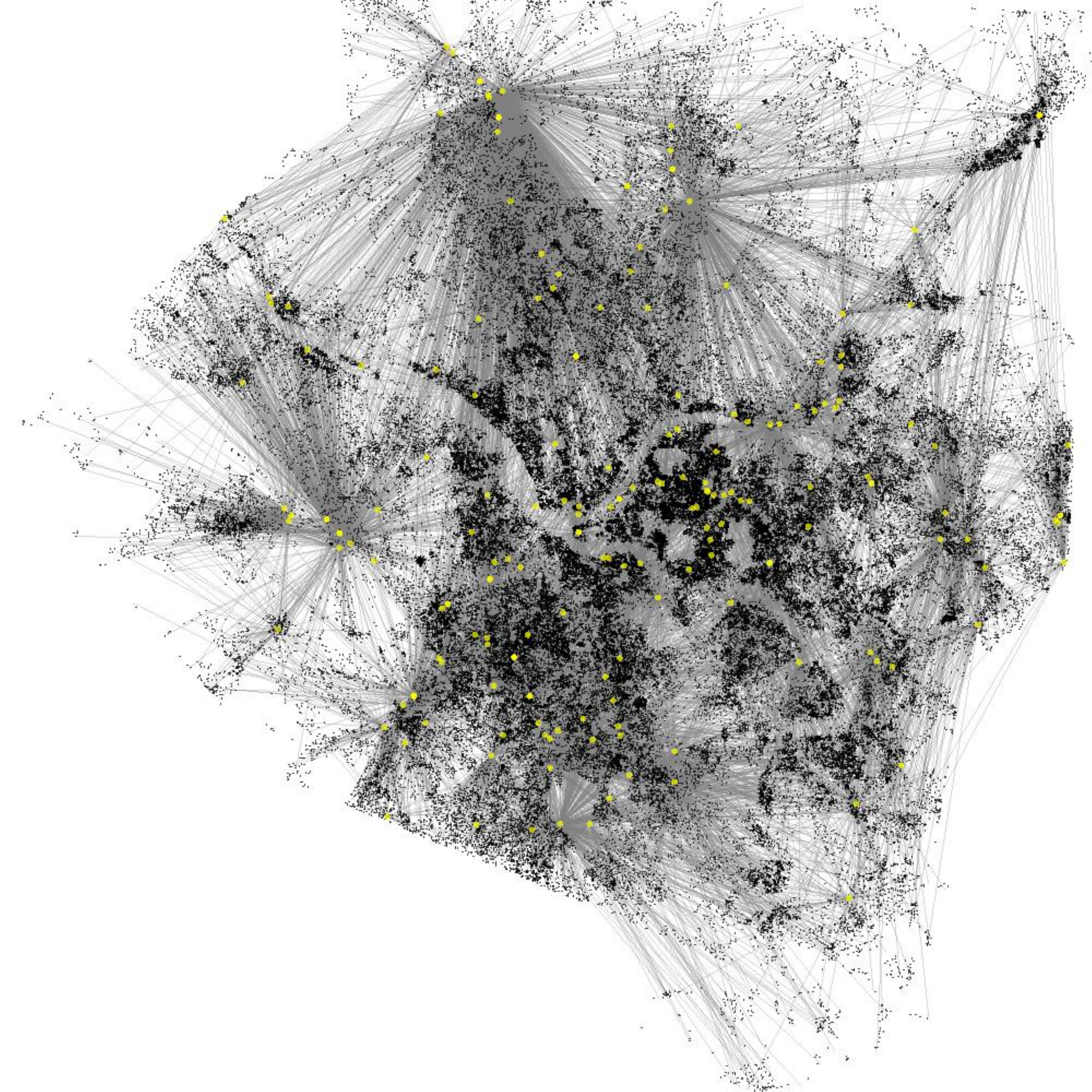}
\caption*{\scalebox{1}{gyms (yellow)}}
\label{fig:GYM}
\end{minipage}
\begin{minipage}[c]{0.23\linewidth}
\includegraphics[width=\linewidth]{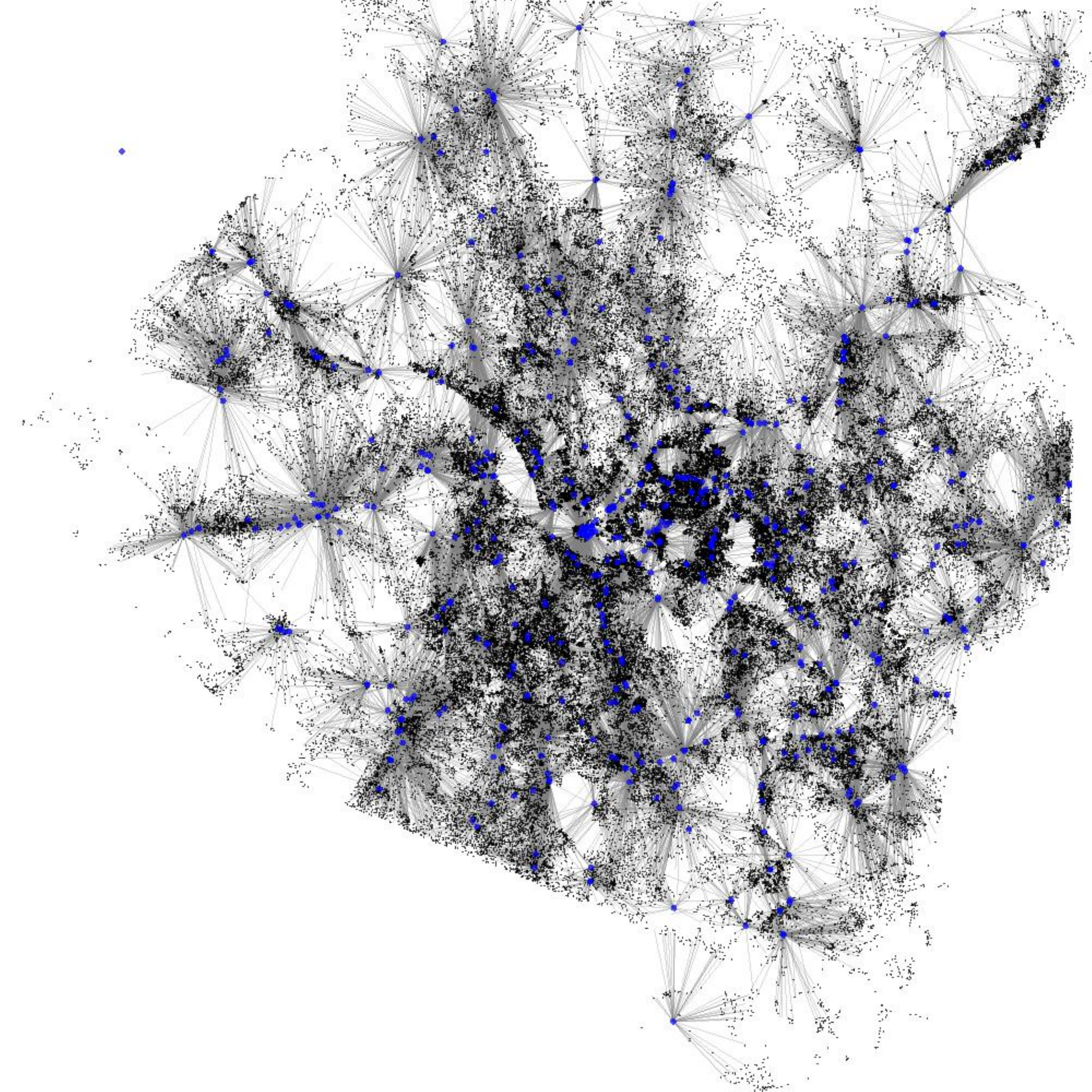}
\caption*{\scalebox{1}{convenient stores (blue)}}
\label{fig:CVS}
\end{minipage}
\begin{minipage}[c]{0.23\linewidth}
\includegraphics[width=\linewidth]{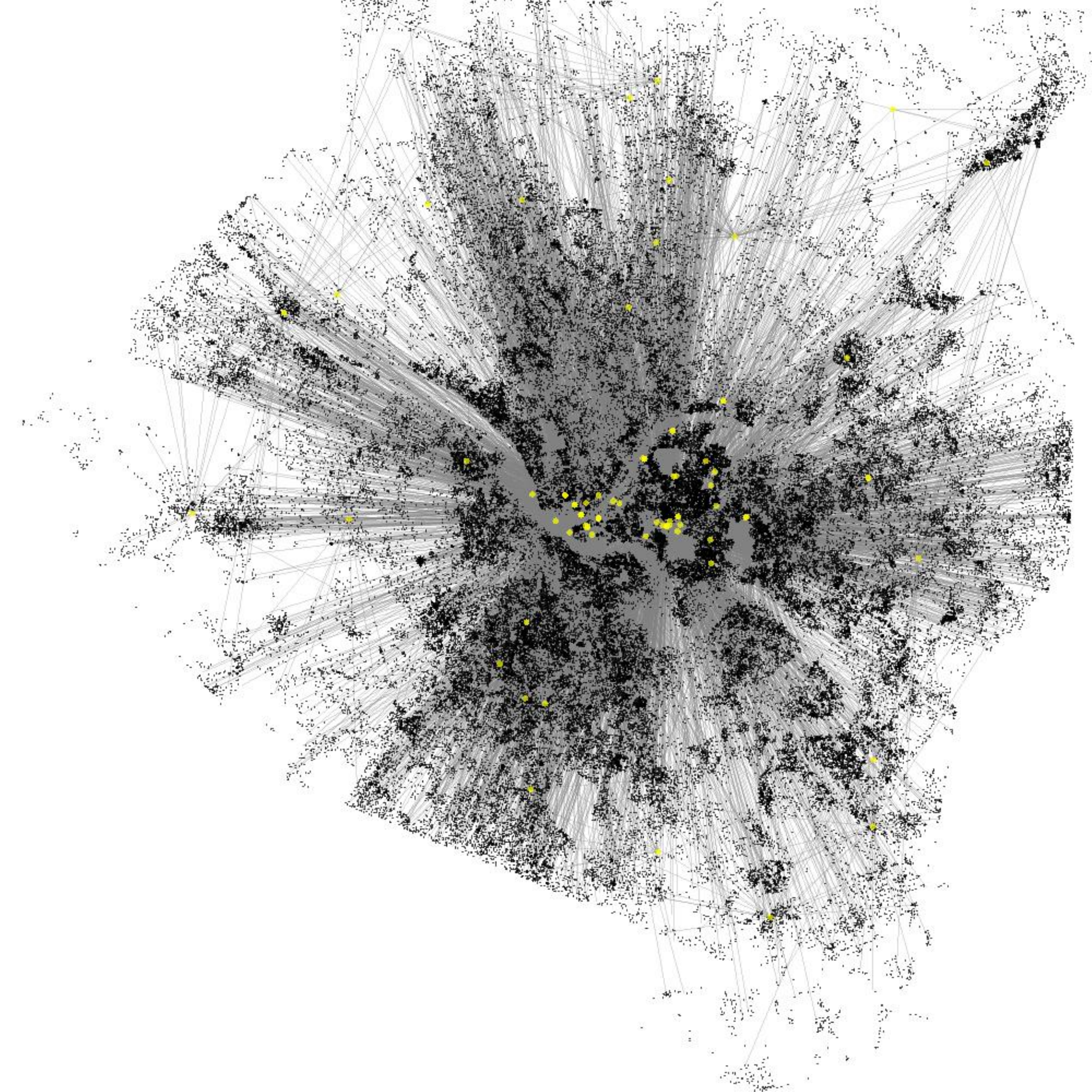}
\caption*{\scalebox{1}{museums (yellow)}}
\label{fig:MUS}
\end{minipage}
\caption{Agent allocation of some specific facilities.}
\end{figure*}
\endgroup
}


\section{Parameters for the Agent Model}
\subsection{Hospitality and Fatality Rates}
\par The death rate and hospitality rate used in our work are estimated based on the US population data \cite{population}, and the CDC report on fatality cases and age-wise hospitalization rate per 100k people \cite{CDCinfectdeath} (\cref{tab2} and \cref{tab3}).
\begin{table}[h]
    \centering
    \begin{tabular}{p{25pt}|p{90pt}|p{85pt}}
    \toprule
      Age & proportion in infection & proportion in fatality 
    \\\midrule
    0-17 & 8.5\% & 0.06\% \\
    18-64 & 76.3\% & 20.71\% \\
    65+ & 15.2\% & 79.23\% \\
    \bottomrule
    \end{tabular}
    \vspace{2mm}
    \caption{The proportion of each age group in overall infected and death cases\protect\cite{CDCinfectdeath} as of September 19th, 2020.}
    \label{tab2}
\end{table}
\begin{table}[h]
    \centering
    \begin{tabular}{p{40pt}|p{80pt}|p{90pt}}
    \toprule
      Age & Cumulative Hospitalization per 100k & Estimated US population in 2019
    \\\midrule
    0-4 & 17.9 & 19756683  \\
    5-17 & 10.3 & 53462467  \\
    18-49 & 119.2 & 138216422 \\
    50-64 & 261.5 & 62925688 \\
    65+ & 472.3 & 54058263 \\
    \bottomrule
    \end{tabular}
    \vspace{2mm}
    \caption{The estimated population, hospitalization rate and hospitalization cases for each age group.}
    \label{tab3}
\end{table}
\subsection{Other Parameters}
\begin{table*}[h]
    \centering
    \begin{tabular}{p{25pt}|p{80pt}|p{90pt}|p{50pt}|p{50pt}|p{60pt}}
    \toprule
      Age & Estimated Infection & Estimated Cumulative Hospitalization Cases & Estimated Fatality & Severity Rate & Unrevised Death Rate
    \\\midrule 
    0-17 & 574796 & 9043 & 120 & 1.57\% & 0.021\% \\
    18-64 & 5159636 & 329304 & 41256 & 6.38\% & 0.8\% \\
    65+ & 1027869 & 255317 & 157831 & 24.84\% & 15.36\%\\
    \bottomrule
    \end{tabular}
    \vspace{2mm}
    \caption{The calculated death rate and hospitalization rate for each age group in the US by Sept. 19th, 2020. We timed 1.25 to the final fatality rate for there exists on-going cases that will be dead in the future.}
    \label{tab4}
\end{table*}
\par \Cref{tab5} shows the list of other parameters about the individual epidemiology.
\begin{table*}
    \centering
    \begin{tabular}{p{50pt}p{310pt}p{30pt}p{80pt}}
    \toprule
    Notation & Meaning  & Value & Source\\
    \midrule
    p\_sev2cri\_nhos & The daily probability of death for severe patients out of hospital(0-18 yrs old/18-65/65+) & 0.6,0.8,1 & estimated\\
    p\_sev2rec & The daily probability of recovery for severe patients in hospital & 1/10 & \cite{2ndwave} \\
    p\_inc2pre & The daily probability of developing from "inc" state to "pre" or "ina" state & 1/3 & \cite{2ndwave}\\
    p\_rec\_sym & The daily probability of recovery from mild symptom & 1/8.8 & \cite{howlong} \\
    p\_hos & The daily probability of developing from "sym" state to "msy"/"ssy" state &  1/1.2 & \cite{incubation}\\
    p\_deimm\_asy & The daily probability of losing immunity after being asymptomatic & 0.009 & \cite{immuneloss}\\
    p\_deimm\_sym & The daily probability of losing immunity after being symptomatic & 0.0025 &\cite{immuneloss} \\
    asy\_infect\_rate & The infectivity of asymptomatic patients (symptomatic is 1) & 0.31 & \cite{2ndwave}\\
    pre\_infect\_rate & The infectivity of pre-symptomatic patients & 0.12 & \cite{2ndwave}\\
    asy\_pop\_rate & The proportion of patients that turn out to be asymptomatic.  & 0.25  & calibrated \\
    \bottomrule
    \end{tabular}
    \vspace{2mm}
    \caption{The individual epidemiology parameter list. All daily probabilities are geometrical distribution and independent for each day.}
    \label{tab5}
\end{table*}

\section{Details of Intra-Facility Infection}
\subsection{The calculation of $p_x$ and $p_y$}
As we mentioned in the main manuscript, $p_x=g(s_x,a_x)$ for a victim $x$ and $p_y=h(s_y,a_y)$ for an infectious agent $y$ are determined by multiple factors, such as whether the agent is wearing mask and the agent's current health state. We considered the fact that people in different phases of the disease have different power of virus spreading. 

More specifically, for all facilities except communities, $p_x=g(s_x, a_x)$ can be written as 
$$p_x=p_{age,x}p_{mask,x},$$ and $p_y=h(s_y,a_y)$ as
$$p_y=p_{hs,y}p_{mask,y}.$$
In the formulae above, $p_{age, x}$ is the age factor of the victim $x$; older people are more vulnerable to the infection and have bigger $p_{age, x}$. $p_{hs,y}$ is the factor for current health state of $y$; people in pre-symptomatic phase and asymptomatic patients have limited power to spread virus. For any agent $z$, $p_{mask, z}=0.4$ if the agent is wearing a mask ($a_{z,mask}=\text{mask}$), and $1$ otherwise. For example, if both $x$ and $y$ are wearing masks, then the probability of infection is reduced to 16\% compared to that without masks on both sides. For $p_{age,x}$ and $p_{hs,y}$, we have 
$$p_{age,x}=\begin{cases} 0.4&\mbox{x is ``chd"}\\ 0.38&\mbox{x is ``sch"}\\0.8175&\mbox{x is ``adu"}\\0.81&\mbox{x is ``rtr"}\end{cases}$$
and 
$$p_{hs,y}=\begin{cases} 0.12&\mbox{Health state is pre-symptomatic}\\ 0.31&\mbox{Health state is asymptomatic} \\1&\mbox{Health state is in \{``sym",``msy" and ``ssy"\}}\\0&\mbox{otherwise}\end{cases}$$

As for other facility-related parameters for intra-facility infection, \cref{tab6} shows the frequency and basic coefficient of each type of facilities as well as their $MinAct$.   

\begin{table}[h]
\centering
\vspace{2mm}\footnotesize
\begin{tabular}{p{45pt}|p{30pt}|p{70pt}|p{30pt}}
\toprule
Facility & $I_F$ & $f_F$ & MinAct\\
\midrule
    Hospital & 0 & N/A & 0\\
    Household & 0.23 & 1 & 0\\
    Workplace & 0.14 & 5/7 & 0\\
    School & 0.21 & 5/7 & 0\\
    Retail & 0.09 & 1 & 0\\
    Supermarket & 0.09 & 1&0\\
    Community & 0.0075 & 1&1\\
    Library & 0.12 & 10.5/365&2\\
    Museum & 0.12 & 2.5/365 * 1/0.54&2\\
    Gym & 0.15 & 0.47 &2\\
    
    Restaurant & 0.21 & 4.2 / 7 & 3\\
    Stadium & 0.42 & 4.7/365 * 1/0.17& 3\\
    Theatre & 0.42 & 3.8/365 * 1/0.35& 3\\
    Cinema & 0.42 & 5.3/365 * 1/0.59 & 3\\
\bottomrule

\end{tabular}
\vspace{2mm} 
\caption{The basic coefficient($I_F$), normal frequency($f_F$) and minimal $a_{act}$(MinAct) for joining the calculation of each facility(Fac). Some data (fractions) are corrected by the proportion of active person in our model of Allegheny. Data are based on \protect\cite{2ndwave}\protect\cite{freqres}\protect\cite{freq}\protect\cite{statista}}
\label{tab6}
\vspace{-3mm}
\end{table}
\subsection{Special Rules for Community in Intra-Facility Infection}
In our model, community infection stands for the probability of getting infected by walking by or chatting with the infected agents in open space. Infections on public transportation and other places are not considered here. The reason for us to distinguish community infection from other facilities is that the risk of community infection and agents' activity levels are co-related. It is not accurate enough to model the risk of community infection by a single threshold as in other facilities.

Therefore, we let the activity level $A_{act}$ as a factor of the spreading power in community infection. For example, if the infected individual chooses $A_{act}=2$ and the victim chooses $A_{act}=3$, then the probability of infection calculated from the formula in the paper should multiply $\frac{2}{2}*\frac{3}{2}=1.5$. A cautious person will avoid community infection if and only if he/she chooses $A_{act}=0$, which stands for staying at home as much as possible.
\subsection{The Calculation of Contact Trace}
Strong quarantine in our experiment requires contact tracing, namely, if a case is discovered by the government, ``other people he/she has physically contacted" must be quarantined as well. To calculate the contact trace in the intra-facility infection process, we assume the patient-victim pair (\textit{i.e.} contact trace) is distributed proportionally to the virus-spreading power of all patients in the same facility every day. For example, if two patients $A$ and $B$ infect $3$ victims in a workplace on a particular day, and $A$'s virus-spreading power is twice of $B$'s, then $2$ people (randomly drawn) will be modeled as infected by $A$.

\section{Reinforcement Learning Agent Setup}

\subsection{detailed settings of the MA-POMDP}

\textbf{Observations}: in experiments without information disclosure, an agent $i$'s observation is $o_{i,no info} = o_{i,hea} \times o_{i,rel} \times o_{i,sup} \times o_{city}$. $o_{i,hea}$ indicates the agent's health states, which is a one-hot vector with a dimension of $5$\footnote[1]{Note that although there are $11$ types of health states, the agent cannot distinguish them all before medical check. For a comprehensive relationship between $o_{i,hea}$ and the health states, one can refer to figure 2 in our main manuscript.}. In our experiment, for simplicity, we assumed that all people without medical checks will believe themselves as infected by COVID-19 only when they have symptoms, and before that, people do not take medical tests. $o_{i,rel}$ indicates agent $i$'s relatives' health states (a relative is defined as another person in the same household), which is a real number and equals to the probability of agent $i$ being infected by the relatives. $o_{i,sup}\in[0, 1]$ represents agent $i$'s supply level\footnote[2]{If the supply is not replenished, we let the supply level $o_{i,sup}$ drop at a decreasing rate. More specifically, denote the supply level as $L$, then $L=\max(0, 1-(\frac{d}{21})^2)$, where $d$ is the number of days since last replenishment of supply.}. Finally, $o_{city}$ is the number of infected cases in the city, normalized by dividing $1000$. 

In experiments with information disclosure\footnote[3]{We assumed that the government can only know and disclose information of people infected $2$ days ago.} the observation of agent $i$ is $o_{i,info} = o_{i,no info} \times o_{i,sur}$. We separate all types of facilities into $4$ groups and each group of facilities has the same level $\text{MinAct}\in\{0,1,2,3\}$. The component of each group can be seen in \cref{tab6} in the appendix. 
$o_{i,sur}\in\mathrm{R}^4$ indicates severity of infections in all four groups of facilities that agent $i$ may visit. We assumed that severity approximately equals to the probability of agent $i$ being infected in those facilities. 

All observations are concatenated together and feed into the Q-network. We use different sets of hyper-parameters to learn different policies.

\textbf{Actions}: Agent $i$'s action space is discrete, represented by $a_{i} = a_{i,mask} \times a_{i,act} \times a_{i,shop}$. $a_{i,mask} = \{\text{mask, no\_mask}\}$ indicates wearing a mask or not. $a_{i,act} = \{0, 1, 2, 3\}$ indicates the activity level for other public facilities except retail stores. $a_{i,shop} = \{\text{no\_shopping, shopping\_online, shopping\_offline\}}$ indicates the ways for shopping to replenish an agent's supply. In each simulation step (a day), there will be only $17,000$ people, or roughly $\frac{1}{70}$ of the total population randomly chosen from the pool of agents with $a_{i,act}=\text{``shopping online"}$. The number $\frac{1}{70}$ is estimated from \cite{shopnum}.

\textbf{Rewards}: An agent rewards include rewards on activity levels, rewards on health states, rewards for wearing masks, rewards for offline shopping, and rewards related to the supply level. 

We let the activity rewards $R_{act}(a_i)$ be positively proportional to the activity level if no symptom shows, i.e., $R_{act}(a_i) = \alpha_{act}a_i$. Once infected, an one-time negatively high reward $R_{ill}$ will be posted on the agent. $R_{mask} = r_{mask}<0$ is assigned for wearing a mask, $R_{shop}=-1$ is assigned for selecting offline shopping, and $R_{eth} = -(\alpha_{act}a_i + r_{mask})$ (the ethical penalty) is given for high activity level or not wearing mask with symptoms. Also, an agent gets penalty for low supply level. More specifically, it gets a negative reward of $r_{sup}=-\frac{1}{0.58}(1-L)$ with $L$ as the supply level. The constant $\frac{1}{0.58}$ is calculated to make a rational agent's shopping frequency at around $7-8$ days, beyond which the incentive to avoid infection in offline shopping by $R_{shop}$ is overwhelmed.

In our settings, we assumed $\alpha_{act} = 1$ and $r_{mask} = 0.1$. We tried different values for $R_{ill}$ to generate different settings where people care about their health in different degrees. We left finding a realistic set of more hyper-parameters such as $\alpha_{act}$ and $r_{mask}$ via inverse RL as future work. 

Note that by the one-time penalty $R_{ill}$, the RL training in our work is naturally smoothing. The smoothing process is two-fold: 1) one-time penalty is smoothed into each simulation step; for example, if a healthy person goes to places where the probability of infection is 10\%, then he would receive 0.1$R_{ill}$; 2) people don't go to every facility at each step, but we smoothed it by correcting the factor with frequency. For example, if people goes to workplaces 5 times per week and libraries 10 times per year, we will assume that they go to such places every day, with the virus-spreading power multiplied by $\frac{5}{7}$ and $\frac{11}{365}$.

%

\subsection{details of SMADQN}
\Cref{algo:DQN} and \cref{tab:DQN} are, respectively, the pseudocode and hyper-parameters of the SMADQN algorithm.

\begin{table}[H]
\centering
\vspace{2mm}
\begin{tabular} {p{110pt}|p{100pt}}
\toprule
Hyper-parameters       & Value \\
\midrule 
minimum of $\epsilon$ ($\epsilon_m$) & 0.9 \\
maximum of $\epsilon$ ($\epsilon_M$) & 1.2 \\
step-size of $\epsilon$ ($\epsilon_s$) & 0.1 \\
Soft rate ($\alpha$) & 1 / 3 \\
training optimizer & Adam \\
Learning rate  & $0.01x+0.001(1-x)$, $x = n_{episode}/20-1$ clipped to $[0,1]$ \\
\# mini-batch ($n_b$) & 100\\
Adam step-size & 1e-5  \\
Discount rate ($\gamma$)  & 0.9  \\
GAE parameter ($\lambda$) & 0.9  \\
episode length ($T$)        & 80    \\ 
\bottomrule
\end{tabular}
\vspace{2mm}
\caption{The hyper-parameters in the SMADQN algorithm}
\label{tab:DQN}
\vspace{-3mm}
\end{table}

\section{Major Covid-Related Events of Allegheny County and Corresponding Government Policies}
To better simulate the real-world infection, we collected major news related to the spread of Covid-19 in Allegheny County, US. \Cref{tab8} listed the major events since Covid-19 began to spread in Allegheny. The collected events helped us calibrate the hyper-parameters of our model from two major aspects: 1) we fixed all actions of individuals and implemented no government policy for the first 10 days in all experiments to simulate the delay of the government and regular people; 2) we chose $80$ days as the length of one episode to match the fact that Covid-19 fighting in Allegheny achieved a stage of success in the first $80$ days since the first case was discovered. 
\Cref{tab8} shows the major events and the corresponding government policies in our model.

\begin{algorithm}[H]
\begin{algorithmic}[1]
   \caption{SMADQN\label{algo:DQN}}
   \STATE Randomly initialize source Q network for each agent type $t$: $Q_t(o,a|\theta_t)$ with weights $\theta_t$.
   
   \STATE Initialize target Q network for each agent type $t$: $Q'_t$ with weights $\mu_t \leftarrow \theta_t$
   
   \STATE Initialize permanent buffer for each agent type $t$: $R_{perm, t}$.
   
   \STATE Initialize a standard normal distribution generator $G$
   
   \STATE Initialize the greedy threshold $\epsilon = \epsilon_m$
   
   \FOR{episode $i = 1,2,...,+\infty$} {
   
   \STATE Initialize temporal buffer for each agent type $t$: $R_{temp, t}$
   
   \STATE Receive initial observation for each individual $i$ of each type $t$ at step 1: $o_{t,i,1}$
   
   \FOR{$j=1,...,T$}
   {
   \FOR{each individual $i$ of each type $t$}
   {
   \STATE Select action of this individual according to its source Q network: $a_{t,i,j} \propto Q_{t,i}(a_{t,i,j}, o_{t,i,j}|\theta_t) / \alpha$
   
   \STATE use $G$ to generate a random number $g$
   
   \IF{$g \ge \epsilon$}
   {
   \STATE Resample $a_{t,i,j}$ uniformly 
   }
   
   \ENDIF
   
   }
   \ENDFOR
   \STATE Execute actions and observe reward $r_{t,i,j}$ nad observe new observation $o_{t,i,j+1}$
   }
   \ENDFOR
   \FOR{each type $t$}
   {
        \IF{$episode = 1$}
        {
        \STATE Store all individuals' trajectories: ($o_{t,i,1}, a_{t,i,1}, r_{t,i,1}, o_{t,i,2}, a_{t,i,2}, r_{t,i,2}, ..., o_{t,i,T}, $ $a_{t,i,T}, r_{t,i,T}$) in $R_{temp, t}$
        }
        \ELSE{\STATE Replace $\beta$ of trajectories in $R_{perm, t}$ with trajectories in $R_{temp, t}$}
        \ENDIF
        \STATE Update $Q'_t$: $\mu_t \gets \theta_t$
        \STATE Use $Q'_t$ to update GAE values for each $o_{t,i,j}$ in $R_{perm, t}$ as $y_{t,i,j}$
        \STATE Separate data in $R_{perm, t}$ to $n_b$ mini-batches, and train $Q_t$ with MSE using $y$ as target.
   }
   \ENDFOR 
   \STATE $\epsilon \gets min(\epsilon + \epsilon_s, \epsilon_M)$
   }
   \ENDFOR
\end{algorithmic}
\end{algorithm}

\begin{table}[h]
    \centering
    \begin{tabular}{p{20pt}|p{70pt}|p{120pt}}
    \toprule
        Day & Event & Government Policy (Capacity Restraint) \\
        \midrule
        0  & First two cases are discovered in Allegheny& N/A\\
        \midrule
        10 & The stay-at-home order is in effect&workplace 25\%, supermarket,  community \& retail 100\%, others 0\% (capacity)\\
        \midrule
        62 & Allegheny move to "yellow" phase for reopening&workplace 50\%  community 100\% supermarket \& retail 100\% restaurant 25\%\\
        \midrule
        80-100 & Massive protest&N/A\\
        \midrule
        82 & Allegheny move to "green" phase for reopening&workplace 75\% community 100\% supermarket \& retail 100\% school 0\% others 50\%\\
        \midrule
        110 & Temporal ban on restaurants & workplace 75\% community 100\% supermarket \& retail 100\% restaurant 0\% others 50\%\\
        \midrule
        116 & Soften regulations for restaurants & workplace 75\% community 100\% supermarket \& retail 100\% restaurant 10\% others 50\%\\
        \midrule
        123 & Soften regulations for restaurants & workplace 75\% community 100\% supermarket \& retail 100\% restaurant 35\%
others 50\%\\
\bottomrule
    \end{tabular}
    \vspace{2mm}
    \caption{Major events related to Covid-19 outbreak in Allegheny County \protect\cite{Alleghenynews} and the corresponding government policies. Day $0$ is March 14th, 2020. Although we did not simulate the situation after 80 days, we still assigned government policies for them.}
    \label{tab8}
\end{table}
\section{Computational Feasibility}

Our experiment code was developed with multi-threaded C++ for simulation and Python for MARL training and data processing. We connected the two parts with Cython. We run our experiments on a server with a CPU: Intel(R) Core(TM) i9-9940X CPU @ 3.30GHz (14 cores) and a GPU: RTX 2080 Ti. A typical step usually took around 15-20 seconds in our experiments, and about half an hour for a complete epoch. The whole program needs 40G RAM and 3G VRAM. The whole algorithm has $O(n)$ time complexity and $O(n)$ space complexity, where $n$ is the number of agents. We use multi-thread to reduce the constant of $O(n)$. Most experiments were trained for 100 episodes to guarantee convergence, which usually has much redundancy in practice. Hence, the training process of our experiments could be even much faster with less episodes.

\section{Boosting Training with Policy Transfer}

A useful property of SMADQN is that it does not need to be trained from scratch for each experiment. Instead, we show in this section that the policy can be transferred between different experiments to boost training: 1) train a policy on the settings without any government control, and 2) transfer the policy on a setting with much strong control and fine-tune it for a few episodes. 

\begin{figure}[H]
    \centering
    \includegraphics[width=6cm]{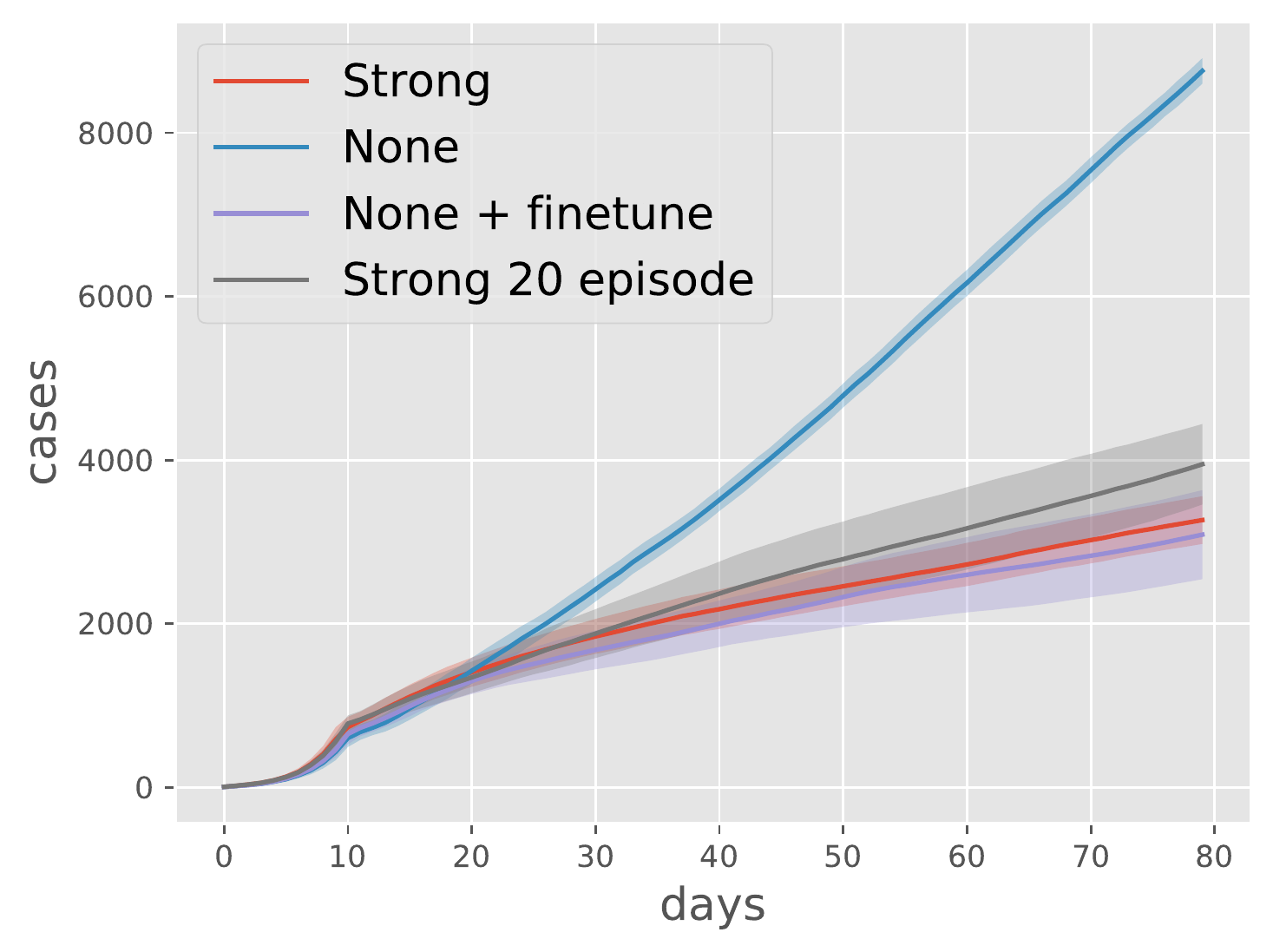}
    \caption{The averaged number of summed cases with strong government control (the same with the calibration). ``Strong" stands for training with strong control for 100 episodes"; ``None" stands for training with no governmental control for 100 episodes; ``None+finetune" stands for training with no control for 80 episodes and with strong control for 20 episodes; ``Strong 20 episode" stands for training with strong control for 20 episodes from scratch.}
    \label{fig:6}
\end{figure}

\Cref{fig:6} illustrates the performance of SMADQN with different settings under strong government control, i.e., the real-life government strategy used in calibration. Via the results, we can see that the policy transfer worked well. We use the similarity of average daily cases as the performance metric. 

We can see that with policy transfer (``None+finetune"), SMADQN quickly yielded a policy which behaved very similarly to the policy trained under strong control from scratch for 100 episodes (``Strong"). The outcome is more similar than the policy trained from scratch for 20 episodes (``Strong 20 episode"). Such observation indicates that policy transfer with SMADQN is more efficient than training from scratch. Direct application of policy trained on non-control settings yielded a very different outcome (``None"), which shows the necessity of policy transfer. Moreover, the difference is coherent with the common knowledge that being less serious about the virus will lead to much poorer control effect of the pandemic. 

\section{Sensitivity Analysis}
\begin{figure}[H]
    \centering
    \includegraphics[width=6cm]{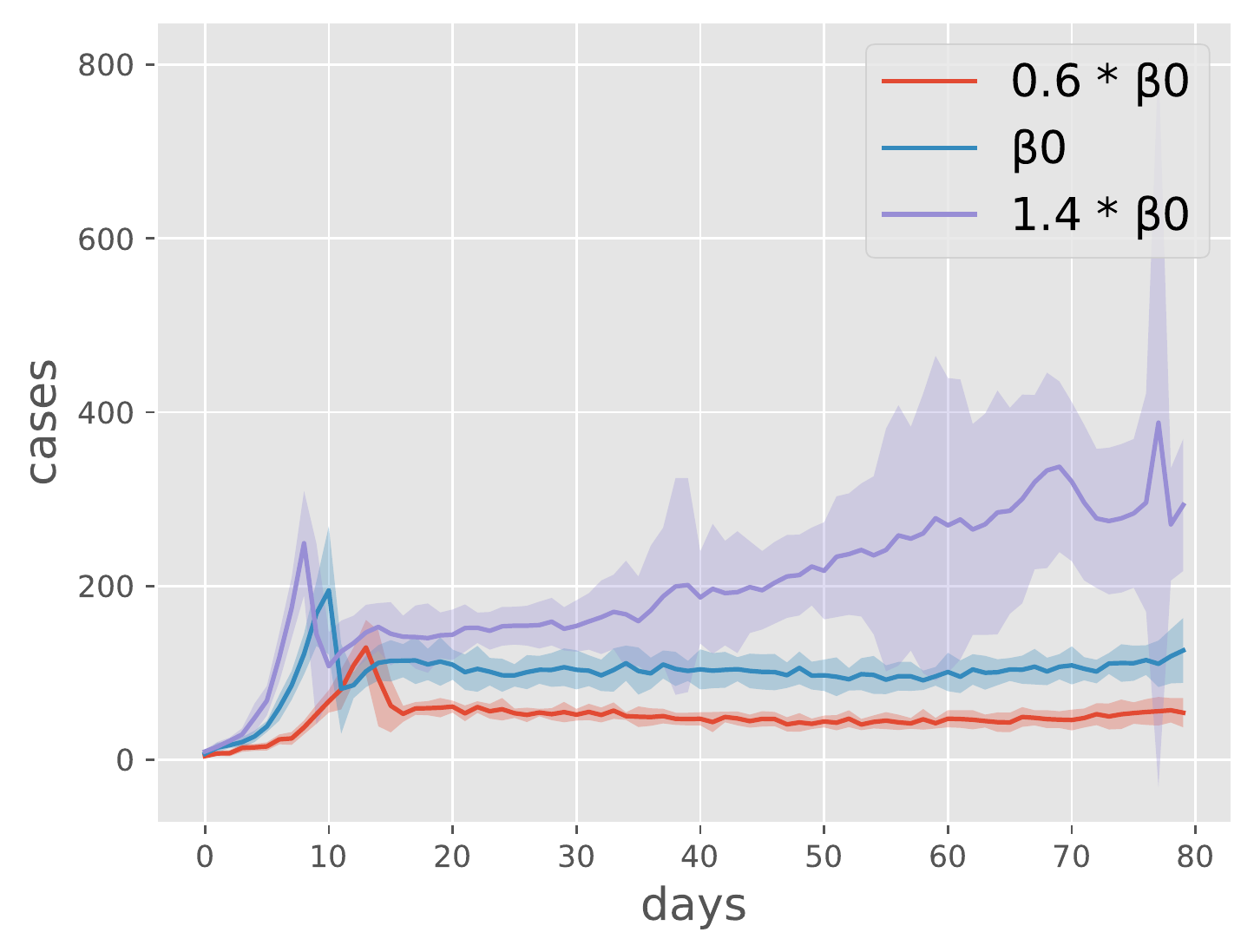}
    \caption{Daily cases with different $\beta$ and no government control strategy.}
    \label{fig:7}
\end{figure}

\begin{figure}[H]
    \centering
    \includegraphics[width=6cm]{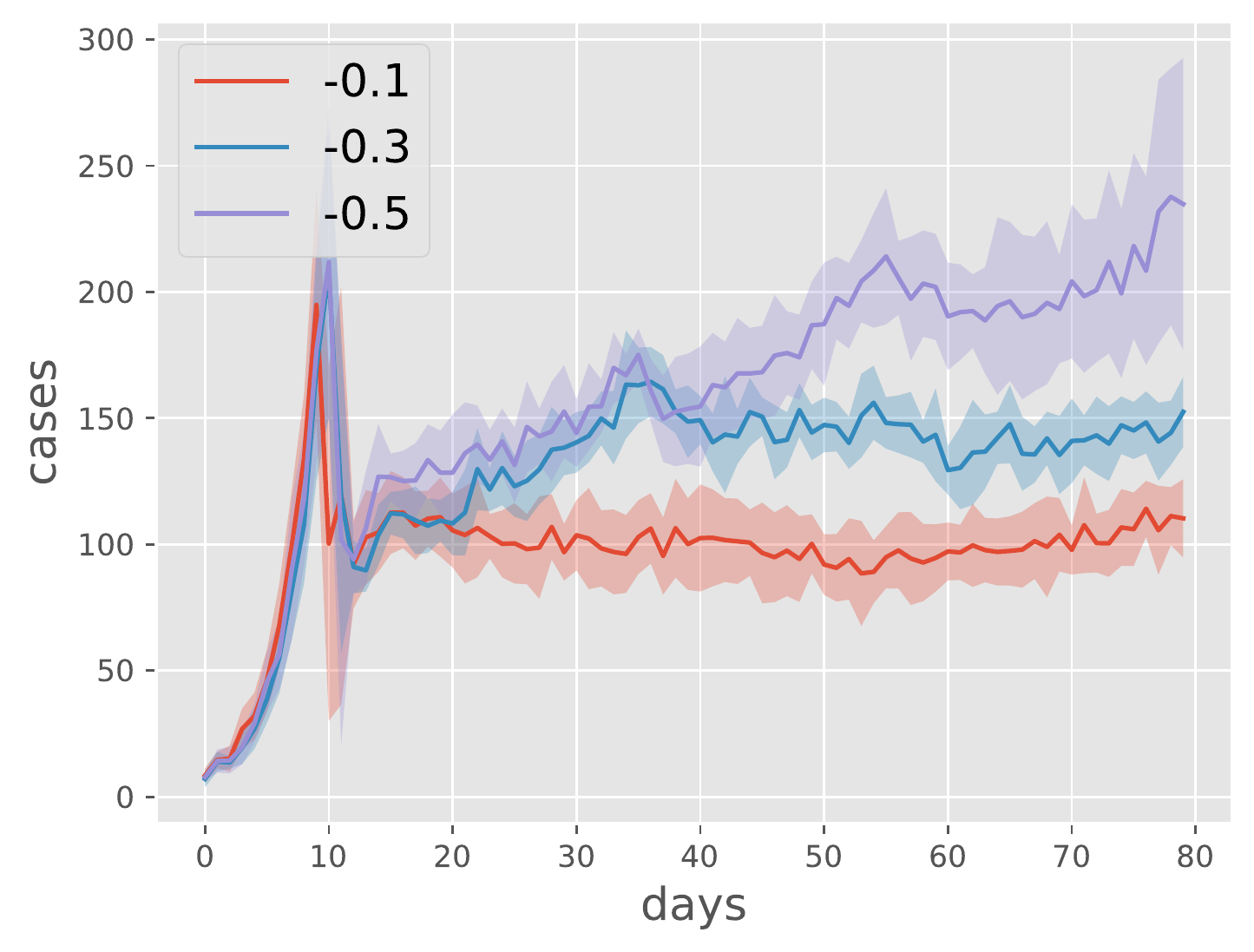}
    \caption{Daily cases with different penalty $R_{mask}\in\{-0.1,-0.3,-0.5\}$ for wearing masks.}
    \label{fig:8}
\end{figure}

We show that our MPS model and the SMADQN algorithm behave reasonably in the sensitivity analysis by varying two hyper-parameters: the infection rate $\beta$ and the penalty of wearing mask $R_{mask}$. The results are shown in \cref{fig:7} and \cref{fig:8}. All data in the sensitivity analysis was the average of last $10$ episodes. 

  
\Cref{fig:7} shows the daily number of cases under different settings of $\beta$.  $\beta_0$ = 15.8 is the default $beta$ in our experiments. All RL agents were trained without government control strategies. We can see that the daily cases increased monotonically with $\beta$, which shows that the MPS is stable and can produce reasonable results for different $\beta$. Furthermore, for all three $\beta$s, the epidemic was controlled to some extent instead of growing exponentially, which means that our SMADQN can handle a wide range of hyper-parameters related to the real-world environment.

\Cref{fig:8} depicts the daily number of cases under different settings of $R_{mask}$ ($R_{mask} = -0.1$ in a default experiment setting). RL agents were all trained without government control strategies. The daily cases increased monotonically with more harsh $R_{mask}$'s penalties, which shows that our SMADQN can handle a wide range of reward parameters and yield reasonable policies in coherent with common knowledge. Hence, it indicates that our reward parameters can successfully capture the influence of individual values towards the macroscopic development of pandemic. Moreover, the SMADQN algorithm can also be utilized as a feasible tool for real-world decision makers.

\section{Code and Data Availability}

All codes and data are accessible in \url{https://github.com/recordmp3/Microscopic-epidemic-model/settings}, and the format of the data can be seen in the code.

\vspace{-1mm}

\bibliographystyle{named}
\bibliography{ijcai21}
\end{document}

%% file: pic/new198/198_daily_infected_cases_summed.tex
\begin{tikzpicture}

\definecolor{color0}{rgb}{0.886274509803922,0.290196078431373,0.2}
\definecolor{color1}{rgb}{0.203921568627451,0.541176470588235,0.741176470588235}
\pgfmathsetlengthmacro\MajorTickLength{
      \pgfkeysvalueof{/pgfplots/major tick length} * 0.5
    }
\begin{axis}[
axis line style={white},
major tick length=\MajorTickLength,
legend cell align={left},
legend style={fill opacity=0.8, draw opacity=1, text opacity=1, at={(0.03,0.97)}, anchor=north west, draw=white!80!black, fill=white!89.8039215686275!black},
tick align=inside,
tick pos=left,
x grid style={white},
xlabel={days},
height=1in,width=0.38\linewidth,
font = \fontsize{7}{7.5}\selectfont,
xmajorgrids,
xmin=0, xmax=80,
xtick style={color=white!33.3333333333333!black},
y grid style={white},
x label style={at={(0.5, 0.3)}},
y label style={at={(0,0.5)}},
ymajorgrids,
ymin=0, ymax=5000,
ytick={0,2500,5000},
yticklabels={0,2.5k,5k},
yticklabel style={rotate=90},
xtick={0,20,40,60,80},
xticklabels={0,20,40,60,80},
ytick style={color=white!33.3333333333333!black}
]
\path [fill=color0, fill opacity=0.3, very thin]
(axis cs:0,8.63085461043683)
--(axis cs:0,3.7024787228965)
--(axis cs:1,13.7728642532738)
--(axis cs:2,26.4681549102219)
--(axis cs:3,39.0632218208556)
--(axis cs:4,58.9130916462317)
--(axis cs:5,86.2687385102346)
--(axis cs:6,125.272963375969)
--(axis cs:7,178.893249342016)
--(axis cs:8,267.399783690472)
--(axis cs:9,381.419231143576)
--(axis cs:10,558.716887871582)
--(axis cs:11,641.857954156235)
--(axis cs:12,703.319135329981)
--(axis cs:13,769.750933787456)
--(axis cs:14,839.810253354791)
--(axis cs:15,912.501269880752)
--(axis cs:16,980.587445566898)
--(axis cs:17,1045.47645510268)
--(axis cs:18,1112.08899073228)
--(axis cs:19,1171.26032647041)
--(axis cs:20,1230.55268381113)
--(axis cs:21,1283.57154893809)
--(axis cs:22,1333.09304677427)
--(axis cs:23,1389.69065645575)
--(axis cs:24,1438.62964128765)
--(axis cs:25,1485.88206948902)
--(axis cs:26,1536.71069664511)
--(axis cs:27,1583.36557278965)
--(axis cs:28,1628.81983267069)
--(axis cs:29,1674.13804747565)
--(axis cs:30,1716.7991200515)
--(axis cs:31,1760.25111081046)
--(axis cs:32,1801.06095529617)
--(axis cs:33,1843.4929485561)
--(axis cs:34,1885.3643117405)
--(axis cs:35,1924.95913129691)
--(axis cs:36,1968.6667346352)
--(axis cs:37,2007.63913238243)
--(axis cs:38,2046.78006111569)
--(axis cs:39,2084.2094704205)
--(axis cs:40,2123.15415787092)
--(axis cs:41,2160.76841703048)
--(axis cs:42,2198.49478666129)
--(axis cs:43,2237.62114359339)
--(axis cs:44,2275.93099610287)
--(axis cs:45,2307.809518947)
--(axis cs:46,2344.0481138865)
--(axis cs:47,2382.48184217455)
--(axis cs:48,2414.03177373998)
--(axis cs:49,2443.39238516259)
--(axis cs:50,2473.62962051863)
--(axis cs:51,2499.97024122779)
--(axis cs:52,2529.99286470917)
--(axis cs:53,2559.79630108508)
--(axis cs:54,2591.27471497062)
--(axis cs:55,2619.85329750071)
--(axis cs:56,2652.45061331241)
--(axis cs:57,2685.11068366311)
--(axis cs:58,2715.6921893128)
--(axis cs:59,2744.2791613424)
--(axis cs:60,2772.24827803619)
--(axis cs:61,2800.63200659422)
--(axis cs:62,2827.26309535104)
--(axis cs:63,2854.13880197803)
--(axis cs:64,2884.24185669184)
--(axis cs:65,2909.53998594149)
--(axis cs:66,2935.60933201781)
--(axis cs:67,2961.85124118882)
--(axis cs:68,2988.0652440512)
--(axis cs:69,3011.43863515841)
--(axis cs:70,3039.94943806506)
--(axis cs:71,3065.60185806964)
--(axis cs:72,3088.17132826562)
--(axis cs:73,3109.43176554245)
--(axis cs:74,3132.84199914699)
--(axis cs:75,3154.26350492897)
--(axis cs:76,3176.87057320077)
--(axis cs:77,3199.23140834234)
--(axis cs:78,3219.30295367183)
--(axis cs:79,3238.31070586728)
--(axis cs:79,4827.35596079938)
--(axis cs:79,4827.35596079938)
--(axis cs:78,4804.89704632817)
--(axis cs:77,4782.50192499099)
--(axis cs:76,4758.26276013256)
--(axis cs:75,4734.26982840436)
--(axis cs:74,4712.09133418634)
--(axis cs:73,4690.56823445755)
--(axis cs:72,4662.49533840105)
--(axis cs:71,4634.73147526369)
--(axis cs:70,4607.45056193494)
--(axis cs:69,4580.49469817492)
--(axis cs:68,4550.1347559488)
--(axis cs:67,4520.61542547784)
--(axis cs:66,4491.05733464885)
--(axis cs:65,4455.46001405851)
--(axis cs:64,4419.49147664149)
--(axis cs:63,4383.66119802197)
--(axis cs:62,4345.53690464896)
--(axis cs:61,4303.90132673911)
--(axis cs:60,4267.15172196381)
--(axis cs:59,4227.9208386576)
--(axis cs:58,4185.9078106872)
--(axis cs:57,4143.08931633689)
--(axis cs:56,4100.41605335425)
--(axis cs:55,4053.68003583263)
--(axis cs:54,4002.19195169604)
--(axis cs:53,3954.80369891492)
--(axis cs:52,3900.34046862416)
--(axis cs:51,3846.82975877221)
--(axis cs:50,3786.5037128147)
--(axis cs:49,3723.94094817074)
--(axis cs:48,3657.23489292669)
--(axis cs:47,3596.11815782545)
--(axis cs:46,3530.3518861135)
--(axis cs:45,3464.32381438634)
--(axis cs:44,3397.86900389713)
--(axis cs:43,3327.71218973994)
--(axis cs:42,3261.37188000537)
--(axis cs:41,3192.29824963619)
--(axis cs:40,3115.97917546241)
--(axis cs:39,3045.7905295795)
--(axis cs:38,2969.61993888431)
--(axis cs:37,2893.02753428423)
--(axis cs:36,2818.7332653648)
--(axis cs:35,2741.70753536976)
--(axis cs:34,2666.2356882595)
--(axis cs:33,2592.44038477724)
--(axis cs:32,2509.20571137049)
--(axis cs:31,2433.68222252288)
--(axis cs:30,2353.2008799485)
--(axis cs:29,2271.86195252435)
--(axis cs:28,2192.18016732931)
--(axis cs:27,2120.70109387701)
--(axis cs:26,2043.08930335489)
--(axis cs:25,1968.31793051098)
--(axis cs:24,1895.17035871235)
--(axis cs:23,1820.77601021092)
--(axis cs:22,1749.3736198924)
--(axis cs:21,1680.69511772857)
--(axis cs:20,1604.91398285554)
--(axis cs:19,1528.07300686292)
--(axis cs:18,1451.51100926772)
--(axis cs:17,1376.92354489732)
--(axis cs:16,1299.2125544331)
--(axis cs:15,1221.36539678591)
--(axis cs:14,1141.05641331188)
--(axis cs:13,1060.38239954588)
--(axis cs:12,973.480864670019)
--(axis cs:11,901.008712510432)
--(axis cs:10,846.283112128418)
--(axis cs:9,684.380768856424)
--(axis cs:8,472.866882976194)
--(axis cs:7,329.173417324651)
--(axis cs:6,221.260369957365)
--(axis cs:5,146.331261489765)
--(axis cs:4,99.5535750204349)
--(axis cs:3,64.8701115124778)
--(axis cs:2,41.3985117564448)
--(axis cs:1,23.6271357467262)
--(axis cs:0,8.63085461043683)
--cycle;

\path [fill=color1, fill opacity=0.3, very thin]
(axis cs:0,9.16)
--(axis cs:0,9.16)
--(axis cs:1,27.48)
--(axis cs:2,45.8)
--(axis cs:3,91.6)
--(axis cs:4,109.92)
--(axis cs:5,164.88)
--(axis cs:6,256.48)
--(axis cs:7,283.9969964)
--(axis cs:8,352.1562756)
--(axis cs:9,397.595795)
--(axis cs:10,431.6754346)
--(axis cs:11,579.3538728)
--(axis cs:12,692.9526713)
--(axis cs:13,795.1915901)
--(axis cs:14,886.0706289)
--(axis cs:15,976.9496678)
--(axis cs:16,988.3095476)
--(axis cs:17,1045.108947)
--(axis cs:18,1147.347866)
--(axis cs:19,1283.666424)
--(axis cs:20,1329.105943)
--(axis cs:21,1408.625102)
--(axis cs:22,1476.784382)
--(axis cs:23,1476.784382)
--(axis cs:24,1522.223901)
--(axis cs:25,1613.10294)
--(axis cs:26,1658.542459)
--(axis cs:27,1840.300537)
--(axis cs:28,1999.338855)
--(axis cs:29,2022.058615)
--(axis cs:30,2044.778374)
--(axis cs:31,2044.778374)
--(axis cs:32,2135.657413)
--(axis cs:33,2226.536452)
--(axis cs:34,2317.415491)
--(axis cs:35,2362.85501)
--(axis cs:36,2396.93465)
--(axis cs:37,2419.65441)
--(axis cs:38,2419.65441)
--(axis cs:39,2521.893328)
--(axis cs:40,2590.052608)
--(axis cs:41,2680.931646)
--(axis cs:42,2669.571767)
--(axis cs:43,2692.291526)
--(axis cs:44,2703.651406)
--(axis cs:45,2726.371166)
--(axis cs:46,2771.810685)
--(axis cs:47,2805.890325)
--(axis cs:48,2885.409484)
--(axis cs:49,2942.208883)
--(axis cs:50,3101.247201)
--(axis cs:51,3101.247201)
--(axis cs:52,3112.607081)
--(axis cs:53,3180.76636)
--(axis cs:54,3237.56576)
--(axis cs:55,3214.846)
--(axis cs:56,3294.365159)
--(axis cs:57,3317.084919)
--(axis cs:58,3317.084919)
--(axis cs:59,3339.804678)
--(axis cs:60,3453.403477)
--(axis cs:61,3521.562756)
--(axis cs:62,3555.642396)
--(axis cs:63,3567.002275)
--(axis cs:64,3601.081915)
--(axis cs:65,3612.441795)
--(axis cs:66,3623.801675)
--(axis cs:67,3623.801675)
--(axis cs:68,3680.601074)
--(axis cs:69,3760.120233)
--(axis cs:70,3816.919632)
--(axis cs:71,3862.359152)
--(axis cs:72,3862.359152)
--(axis cs:73,3862.359152)
--(axis cs:74,3907.798671)
--(axis cs:75,3919.158551)
--(axis cs:76,3941.878311)
--(axis cs:77,3975.95795)
--(axis cs:78,3975.95795)
--(axis cs:79,3987.31783)
--(axis cs:79,3987.31783)
--(axis cs:79,3987.31783)
--(axis cs:78,3975.95795)
--(axis cs:77,3975.95795)
--(axis cs:76,3941.878311)
--(axis cs:75,3919.158551)
--(axis cs:74,3907.798671)
--(axis cs:73,3862.359152)
--(axis cs:72,3862.359152)
--(axis cs:71,3862.359152)
--(axis cs:70,3816.919632)
--(axis cs:69,3760.120233)
--(axis cs:68,3680.601074)
--(axis cs:67,3623.801675)
--(axis cs:66,3623.801675)
--(axis cs:65,3612.441795)
--(axis cs:64,3601.081915)
--(axis cs:63,3567.002275)
--(axis cs:62,3555.642396)
--(axis cs:61,3521.562756)
--(axis cs:60,3453.403477)
--(axis cs:59,3339.804678)
--(axis cs:58,3317.084919)
--(axis cs:57,3317.084919)
--(axis cs:56,3294.365159)
--(axis cs:55,3214.846)
--(axis cs:54,3237.56576)
--(axis cs:53,3180.76636)
--(axis cs:52,3112.607081)
--(axis cs:51,3101.247201)
--(axis cs:50,3101.247201)
--(axis cs:49,2942.208883)
--(axis cs:48,2885.409484)
--(axis cs:47,2805.890325)
--(axis cs:46,2771.810685)
--(axis cs:45,2726.371166)
--(axis cs:44,2703.651406)
--(axis cs:43,2692.291526)
--(axis cs:42,2669.571767)
--(axis cs:41,2680.931646)
--(axis cs:40,2590.052608)
--(axis cs:39,2521.893328)
--(axis cs:38,2419.65441)
--(axis cs:37,2419.65441)
--(axis cs:36,2396.93465)
--(axis cs:35,2362.85501)
--(axis cs:34,2317.415491)
--(axis cs:33,2226.536452)
--(axis cs:32,2135.657413)
--(axis cs:31,2044.778374)
--(axis cs:30,2044.778374)
--(axis cs:29,2022.058615)
--(axis cs:28,1999.338855)
--(axis cs:27,1840.300537)
--(axis cs:26,1658.542459)
--(axis cs:25,1613.10294)
--(axis cs:24,1522.223901)
--(axis cs:23,1476.784382)
--(axis cs:22,1476.784382)
--(axis cs:21,1408.625102)
--(axis cs:20,1329.105943)
--(axis cs:19,1283.666424)
--(axis cs:18,1147.347866)
--(axis cs:17,1045.108947)
--(axis cs:16,988.3095476)
--(axis cs:15,976.9496678)
--(axis cs:14,886.0706289)
--(axis cs:13,795.1915901)
--(axis cs:12,692.9526713)
--(axis cs:11,579.3538728)
--(axis cs:10,431.6754346)
--(axis cs:9,397.595795)
--(axis cs:8,352.1562756)
--(axis cs:7,283.9969964)
--(axis cs:6,256.48)
--(axis cs:5,164.88)
--(axis cs:4,109.92)
--(axis cs:3,91.6)
--(axis cs:2,45.8)
--(axis cs:1,27.48)
--(axis cs:0,9.16)
--cycle;

\addplot [semithick, color0]
table {%
0 6.16666666666667
1 18.7
2 33.9333333333333
3 51.9666666666667
4 79.2333333333333
5 116.3
6 173.266666666667
7 254.033333333333
8 370.133333333333
9 532.9
10 702.5
11 771.433333333333
12 838.4
13 915.066666666667
14 990.433333333333
15 1066.93333333333
16 1139.9
17 1211.2
18 1281.8
19 1349.66666666667
20 1417.73333333333
21 1482.13333333333
22 1541.23333333333
23 1605.23333333333
24 1666.9
25 1727.1
26 1789.9
27 1852.03333333333
28 1910.5
29 1973
30 2035
31 2096.96666666667
32 2155.13333333333
33 2217.96666666667
34 2275.8
35 2333.33333333333
36 2393.7
37 2450.33333333333
38 2508.2
39 2565
40 2619.56666666667
41 2676.53333333333
42 2729.93333333333
43 2782.66666666667
44 2836.9
45 2886.06666666667
46 2937.2
47 2989.3
48 3035.63333333333
49 3083.66666666667
50 3130.06666666667
51 3173.4
52 3215.16666666667
53 3257.3
54 3296.73333333333
55 3336.76666666667
56 3376.43333333333
57 3414.1
58 3450.8
59 3486.1
60 3519.7
61 3552.26666666667
62 3586.4
63 3618.9
64 3651.86666666667
65 3682.5
66 3713.33333333333
67 3741.23333333333
68 3769.1
69 3795.96666666667
70 3823.7
71 3850.16666666667
72 3875.33333333333
73 3900
74 3922.46666666667
75 3944.26666666667
76 3967.56666666667
77 3990.86666666667
78 4012.1
79 4032.83333333333
};
\addplot [semithick, color1]
table {%
0 9.16
1 27.48
2 45.8
3 91.6
4 109.92
5 164.88
6 256.48
7 283.9969964
8 352.1562756
9 397.595795
10 431.6754346
11 579.3538728
12 692.9526713
13 795.1915901
14 886.0706289
15 976.9496678
16 988.3095476
17 1045.108947
18 1147.347866
19 1283.666424
20 1329.105943
21 1408.625102
22 1476.784382
23 1476.784382
24 1522.223901
25 1613.10294
26 1658.542459
27 1840.300537
28 1999.338855
29 2022.058615
30 2044.778374
31 2044.778374
32 2135.657413
33 2226.536452
34 2317.415491
35 2362.85501
36 2396.93465
37 2419.65441
38 2419.65441
39 2521.893328
40 2590.052608
41 2680.931646
42 2669.571767
43 2692.291526
44 2703.651406
45 2726.371166
46 2771.810685
47 2805.890325
48 2885.409484
49 2942.208883
50 3101.247201
51 3101.247201
52 3112.607081
53 3180.76636
54 3237.56576
55 3214.846
56 3294.365159
57 3317.084919
58 3317.084919
59 3339.804678
60 3453.403477
61 3521.562756
62 3555.642396
63 3567.002275
64 3601.081915
65 3612.441795
66 3623.801675
67 3623.801675
68 3680.601074
69 3760.120233
70 3816.919632
71 3862.359152
72 3862.359152
73 3862.359152
74 3907.798671
75 3919.158551
76 3941.878311
77 3975.95795
78 3975.95795
79 3987.31783
};
\end{axis}

\end{tikzpicture}

%% file: pic/new198/198_daily_infected_cases.tex
\begin{tikzpicture}

\definecolor{color0}{rgb}{0.886274509803922,0.290196078431373,0.2}
\definecolor{color1}{rgb}{0.203921568627451,0.541176470588235,0.741176470588235}
\pgfmathsetlengthmacro\MajorTickLength{
      \pgfkeysvalueof{/pgfplots/major tick length} * 0.5
    }
\begin{axis}[
axis line style={white},
legend cell align={left},
major tick length=\MajorTickLength,
legend style={fill opacity=0.8, draw opacity=1, text opacity=1, draw=white!80!black,at={(0.5,0.97)},inner xsep=1pt, inner ysep=-1pt, row sep=-3pt, fill=white!89.8039215686275!black},
tick align=inside,
tick pos=left,
x grid style={white},
xlabel={days},
xmajorgrids,
font = \fontsize{7}{7.5}\selectfont,
height=1in,width=0.38\linewidth,
xmin=-3.95, xmax=82.95,
xtick style={color=white!33.3333333333333!black},
x label style={at={(0.5, 0.3)}},
y grid style={white},
ymajorgrids,
ytick={0,120,240},
yticklabels={0,120,240},
yticklabel style={rotate=90},
xtick={0,20,40,60,80},
xticklabels={0,20,40,60,80},
ymin=-8.04673278114588, ymax=250.435920307786,
ytick style={color=white!33.3333333333333!black}
]
\path [fill=color0, fill opacity=0.3, very thin]
(axis cs:0,8.63085461043683)
--(axis cs:0,3.7024787228965)
--(axis cs:1,8.3721369241794)
--(axis cs:2,9.84394045884779)
--(axis cs:3,10.570164509778)
--(axis cs:4,17.7695325960349)
--(axis cs:5,25.4737569561774)
--(axis cs:6,37.1046705713874)
--(axis cs:7,51.3606771029363)
--(axis cs:8,84.0996354187435)
--(axis cs:9,110.939593089669)
--(axis cs:10,100.513291196256)
--(axis cs:11,43.8734938375274)
--(axis cs:12,50.1134111288715)
--(axis cs:13,59.2189670411509)
--(axis cs:14,63.0095274795858)
--(axis cs:15,64.0722085630632)
--(axis cs:16,60.7097411866688)
--(axis cs:17,56.4582795247092)
--(axis cs:18,56.3742018384439)
--(axis cs:19,49.3775106856899)
--(axis cs:20,50.6637294943896)
--(axis cs:21,45.4007017673459)
--(axis cs:22,42.4840438132499)
--(axis cs:23,46.2950101195548)
--(axis cs:24,42.7275830328451)
--(axis cs:25,43.8570096575525)
--(axis cs:26,45.2132625727984)
--(axis cs:27,40.3775167420041)
--(axis cs:28,39.923503805685)
--(axis cs:29,39.6749406718551)
--(axis cs:30,38.7391602329867)
--(axis cs:31,37.5884204440835)
--(axis cs:32,35.3528063571482)
--(axis cs:33,39.0480671189541)
--(axis cs:34,37.5206388898196)
--(axis cs:35,36.4197779192819)
--(axis cs:36,38.9774430040629)
--(axis cs:37,35.2269739459405)
--(axis cs:38,35.7476954227573)
--(axis cs:39,34.609311262003)
--(axis cs:40,34.8221596949655)
--(axis cs:41,35.0479025084797)
--(axis cs:42,34.8409051944875)
--(axis cs:43,35.0888189198141)
--(axis cs:44,35.4172350248222)
--(axis cs:45,29.9207435774627)
--(axis cs:46,33.5005356098989)
--(axis cs:47,34.9256974911158)
--(axis cs:48,28.2329957060342)
--(axis cs:49,26.6207461732331)
--(axis cs:50,27.639047643221)
--(axis cs:51,24.0817821692785)
--(axis cs:52,27.1705865231639)
--(axis cs:53,27.5737240589667)
--(axis cs:54,28.7541219600364)
--(axis cs:55,25.3943822476548)
--(axis cs:56,28.3873837895409)
--(axis cs:57,27.1648149780886)
--(axis cs:58,26.1923837146573)
--(axis cs:59,25.2562623159171)
--(axis cs:60,24.6200222717425)
--(axis cs:61,22.6985924427419)
--(axis cs:62,22.7821708867427)
--(axis cs:63,21.9714673387029)
--(axis cs:64,24.4285820519015)
--(axis cs:65,21.1764722144302)
--(axis cs:66,24.0016262097644)
--(axis cs:67,17.8224672331633)
--(axis cs:68,19.6377204670883)
--(axis cs:69,19.1046508918251)
--(axis cs:70,20.1178520251325)
--(axis cs:71,18.6960668042222)
--(axis cs:72,17.5243139166586)
--(axis cs:73,17.2895247931345)
--(axis cs:74,16.2017733005439)
--(axis cs:75,15.4260948652599)
--(axis cs:76,16.3470629707056)
--(axis cs:77,16.6059728115282)
--(axis cs:78,15.5842429444103)
--(axis cs:79,15.4108443583434)
--(axis cs:79,26.0558223083232)
--(axis cs:79,26.0558223083232)
--(axis cs:78,26.8824237222563)
--(axis cs:77,29.9940271884718)
--(axis cs:76,30.2529370292944)
--(axis cs:75,28.1739051347401)
--(axis cs:74,28.7315600327894)
--(axis cs:73,32.0438085401989)
--(axis cs:72,32.8090194166747)
--(axis cs:71,34.2372665291111)
--(axis cs:70,35.3488146415341)
--(axis cs:69,34.6286824415082)
--(axis cs:68,36.0956128662451)
--(axis cs:67,37.9775327668367)
--(axis cs:66,37.6650404569022)
--(axis cs:65,40.0901944522365)
--(axis cs:64,41.5047512814318)
--(axis cs:63,43.0285326612971)
--(axis cs:62,45.4844957799239)
--(axis cs:61,42.4347408905915)
--(axis cs:60,42.5799777282575)
--(axis cs:59,45.3437376840829)
--(axis cs:58,47.2076162853427)
--(axis cs:57,48.1685183552447)
--(axis cs:56,50.9459495437924)
--(axis cs:55,54.6722844190119)
--(axis cs:54,50.1125447066303)
--(axis cs:53,56.6929426076999)
--(axis cs:52,56.3627468101694)
--(axis cs:51,62.5848844973882)
--(axis cs:50,65.160952356779)
--(axis cs:49,69.4459204934336)
--(axis cs:48,64.4336709606325)
--(axis cs:47,69.2743025088842)
--(axis cs:46,68.7661310567678)
--(axis cs:45,68.4125897558707)
--(axis cs:44,73.0494316418445)
--(axis cs:43,70.3778477468526)
--(axis cs:42,71.9590948055125)
--(axis cs:41,78.8854308248536)
--(axis cs:40,74.3111736383678)
--(axis cs:39,78.990688737997)
--(axis cs:38,79.985637910576)
--(axis cs:37,78.0396927207262)
--(axis cs:36,81.7558903292704)
--(axis cs:35,78.6468887473848)
--(axis cs:34,78.1460277768471)
--(axis cs:33,86.6185995477125)
--(axis cs:32,80.9805269761851)
--(axis cs:31,86.3449128892498)
--(axis cs:30,85.2608397670133)
--(axis cs:29,85.3250593281449)
--(axis cs:28,77.0098295276483)
--(axis cs:27,83.8891499246625)
--(axis cs:26,80.3867374272016)
--(axis cs:25,76.5429903424475)
--(axis cs:24,80.6057503004883)
--(axis cs:23,81.7049898804452)
--(axis cs:22,75.7159561867501)
--(axis cs:21,83.3992982326541)
--(axis cs:20,85.4696038389437)
--(axis cs:19,86.3558226476434)
--(axis cs:18,84.8257981615561)
--(axis cs:17,86.1417204752908)
--(axis cs:16,85.2235921466646)
--(axis cs:15,88.9277914369368)
--(axis cs:14,87.7238058537475)
--(axis cs:13,94.1143662921824)
--(axis cs:12,83.8199222044618)
--(axis cs:11,93.9931728291393)
--(axis cs:10,238.686708803744)
--(axis cs:9,214.593740243665)
--(axis cs:8,148.100364581256)
--(axis cs:7,110.172656230397)
--(axis cs:6,76.8286627619459)
--(axis cs:5,48.6595763771559)
--(axis cs:4,36.7638007372985)
--(axis cs:3,25.4965021568887)
--(axis cs:2,20.6227262078189)
--(axis cs:1,16.6945297424873)
--(axis cs:0,8.63085461043683)
--cycle;

\path [fill=color1, fill opacity=0.3, very thin]
(axis cs:0,9.16)
--(axis cs:0,9.16)
--(axis cs:1,23.521975308642)
--(axis cs:2,26.8014814814815)
--(axis cs:3,33.0216913135803)
--(axis cs:4,41.1020649975309)
--(axis cs:5,48.7136617308642)
--(axis cs:6,54.4914404101995)
--(axis cs:7,55.9298596538907)
--(axis cs:8,59.6141543672163)
--(axis cs:9,65.58784379075)
--(axis cs:10,72.7167803932891)
--(axis cs:11,78.6311407671078)
--(axis cs:12,82.1357829466809)
--(axis cs:13,83.2802393950438)
--(axis cs:14,83.3933966024927)
--(axis cs:15,81.5521334029727)
--(axis cs:16,80.388381202014)
--(axis cs:17,79.1898995897244)
--(axis cs:18,78.0015189685032)
--(axis cs:19,76.8579926331872)
--(axis cs:20,75.7808397469851)
--(axis cs:21,74.7765632328228)
--(axis cs:22,73.8365078663398)
--(axis cs:23,72.9384409279464)
--(axis cs:24,72.0497173610795)
--(axis cs:25,71.1316671851744)
--(axis cs:26,70.1446573969508)
--(axis cs:27,69.05317489905)
--(axis cs:28,67.8302790162129)
--(axis cs:29,66.4608867771788)
--(axis cs:30,64.943559319918)
--(axis cs:31,63.2907085943432)
--(axis cs:32,61.527383562971)
--(axis cs:33,59.6889719557307)
--(axis cs:34,57.818233881484)
--(axis cs:35,55.962062164328)
--(axis cs:36,54.1682652538143)
--(axis cs:37,52.482536978478)
--(axis cs:38,50.9456644262814)
--(axis cs:39,49.590971376394)
--(axis cs:40,48.4420164568155)
--(axis cs:41,47.510649219907)
--(axis cs:42,46.7956343690603)
--(axis cs:43,46.2821319803297)
--(axis cs:44,45.9423218256447)
--(axis cs:45,45.7373567515824)
--(axis cs:46,45.6206305041314)
--(axis cs:47,45.5420905794904)
--(axis cs:48,45.4530818739208)
--(axis cs:49,45.3110427502606)
--(axis cs:50,45.0833459119863)
--(axis cs:51,44.7497021253307)
--(axis cs:52,44.3028028460858)
--(axis cs:53,43.7472080128318)
--(axis cs:54,43.0968064701292)
--(axis cs:55,42.3714100008006)
--(axis cs:56,41.5931345993054)
--(axis cs:57,40.7831618581708)
--(axis cs:58,39.9592896522249)
--(axis cs:59,39.1344372577241)
--(axis cs:60,38.3160397171064)
--(axis cs:61,37.5061119448837)
--(axis cs:62,36.701716850966)
--(axis cs:63,35.8956277921352)
--(axis cs:64,35.0770951849322)
--(axis cs:65,34.2327538643844)
--(axis cs:66,33.3477881569106)
--(axis cs:67,32.407472616901)
--(axis cs:68,31.399125300894)
--(axis cs:69,30.314374092975)
--(axis cs:70,29.1514905570371)
--(axis cs:71,27.9174386007458)
--(axis cs:72,26.6292528089243)
--(axis cs:73,25.3144168376352)
--(axis cs:74,24.0100441934633)
--(axis cs:75,22.7608414257793)
--(axis cs:76,21.6160180933471)
--(axis cs:77,20.6254625679026)
--(axis cs:78,19.8356029711031)
--(axis cs:79,19.2854085087038)
--(axis cs:79,19.2854085087038)
--(axis cs:79,19.2854085087038)
--(axis cs:78,19.8356029711031)
--(axis cs:77,20.6254625679026)
--(axis cs:76,21.6160180933471)
--(axis cs:75,22.7608414257793)
--(axis cs:74,24.0100441934633)
--(axis cs:73,25.3144168376352)
--(axis cs:72,26.6292528089243)
--(axis cs:71,27.9174386007458)
--(axis cs:70,29.1514905570371)
--(axis cs:69,30.314374092975)
--(axis cs:68,31.399125300894)
--(axis cs:67,32.407472616901)
--(axis cs:66,33.3477881569106)
--(axis cs:65,34.2327538643844)
--(axis cs:64,35.0770951849322)
--(axis cs:63,35.8956277921352)
--(axis cs:62,36.701716850966)
--(axis cs:61,37.5061119448837)
--(axis cs:60,38.3160397171064)
--(axis cs:59,39.1344372577241)
--(axis cs:58,39.9592896522249)
--(axis cs:57,40.7831618581708)
--(axis cs:56,41.5931345993054)
--(axis cs:55,42.3714100008006)
--(axis cs:54,43.0968064701292)
--(axis cs:53,43.7472080128318)
--(axis cs:52,44.3028028460858)
--(axis cs:51,44.7497021253307)
--(axis cs:50,45.0833459119863)
--(axis cs:49,45.3110427502606)
--(axis cs:48,45.4530818739208)
--(axis cs:47,45.5420905794904)
--(axis cs:46,45.6206305041314)
--(axis cs:45,45.7373567515824)
--(axis cs:44,45.9423218256447)
--(axis cs:43,46.2821319803297)
--(axis cs:42,46.7956343690603)
--(axis cs:41,47.510649219907)
--(axis cs:40,48.4420164568155)
--(axis cs:39,49.590971376394)
--(axis cs:38,50.9456644262814)
--(axis cs:37,52.482536978478)
--(axis cs:36,54.1682652538143)
--(axis cs:35,55.962062164328)
--(axis cs:34,57.818233881484)
--(axis cs:33,59.6889719557307)
--(axis cs:32,61.527383562971)
--(axis cs:31,63.2907085943432)
--(axis cs:30,64.943559319918)
--(axis cs:29,66.4608867771788)
--(axis cs:28,67.8302790162129)
--(axis cs:27,69.05317489905)
--(axis cs:26,70.1446573969508)
--(axis cs:25,71.1316671851744)
--(axis cs:24,72.0497173610795)
--(axis cs:23,72.9384409279464)
--(axis cs:22,73.8365078663398)
--(axis cs:21,74.7765632328228)
--(axis cs:20,75.7808397469851)
--(axis cs:19,76.8579926331872)
--(axis cs:18,78.0015189685032)
--(axis cs:17,79.1898995897244)
--(axis cs:16,80.388381202014)
--(axis cs:15,81.5521334029727)
--(axis cs:14,83.3933966024927)
--(axis cs:13,83.2802393950438)
--(axis cs:12,82.1357829466809)
--(axis cs:11,78.6311407671078)
--(axis cs:10,72.7167803932891)
--(axis cs:9,65.58784379075)
--(axis cs:8,59.6141543672163)
--(axis cs:7,55.9298596538907)
--(axis cs:6,54.4914404101995)
--(axis cs:5,48.7136617308642)
--(axis cs:4,41.1020649975309)
--(axis cs:3,33.0216913135803)
--(axis cs:2,26.8014814814815)
--(axis cs:1,23.521975308642)
--(axis cs:0,9.16)
--cycle;

\addplot [semithick, color0]
table {%
0 6.16666666666667
1 12.5333333333333
2 15.2333333333333
3 18.0333333333333
4 27.2666666666667
5 37.0666666666667
6 56.9666666666667
7 80.7666666666667
8 116.1
9 162.766666666667
10 169.6
11 68.9333333333333
12 66.9666666666667
13 76.6666666666667
14 75.3666666666667
15 76.5
16 72.9666666666667
17 71.3
18 70.6
19 67.8666666666667
20 68.0666666666667
21 64.4
22 59.1
23 64
24 61.6666666666667
25 60.2
26 62.8
27 62.1333333333333
28 58.4666666666667
29 62.5
30 62
31 61.9666666666667
32 58.1666666666667
33 62.8333333333333
34 57.8333333333333
35 57.5333333333333
36 60.3666666666667
37 56.6333333333333
38 57.8666666666667
39 56.8
40 54.5666666666667
41 56.9666666666667
42 53.4
43 52.7333333333333
44 54.2333333333333
45 49.1666666666667
46 51.1333333333333
47 52.1
48 46.3333333333333
49 48.0333333333333
50 46.4
51 43.3333333333333
52 41.7666666666667
53 42.1333333333333
54 39.4333333333333
55 40.0333333333333
56 39.6666666666667
57 37.6666666666667
58 36.7
59 35.3
60 33.6
61 32.5666666666667
62 34.1333333333333
63 32.5
64 32.9666666666667
65 30.6333333333333
66 30.8333333333333
67 27.9
68 27.8666666666667
69 26.8666666666667
70 27.7333333333333
71 26.4666666666667
72 25.1666666666667
73 24.6666666666667
74 22.4666666666667
75 21.8
76 23.3
77 23.3
78 21.2333333333333
79 20.7333333333333
};
\addlegendentry{fitted}
\addplot [semithick, color1]
table {%
0 9.16
1 23.521975308642
2 26.8014814814815
3 33.0216913135803
4 41.1020649975309
5 48.7136617308642
6 54.4914404101995
7 55.9298596538907
8 59.6141543672163
9 65.58784379075
10 72.7167803932891
11 78.6311407671078
12 82.1357829466809
13 83.2802393950438
14 83.3933966024927
15 81.5521334029727
16 80.388381202014
17 79.1898995897244
18 78.0015189685032
19 76.8579926331872
20 75.7808397469851
21 74.7765632328228
22 73.8365078663398
23 72.9384409279464
24 72.0497173610795
25 71.1316671851744
26 70.1446573969508
27 69.05317489905
28 67.8302790162129
29 66.4608867771788
30 64.943559319918
31 63.2907085943432
32 61.527383562971
33 59.6889719557307
34 57.818233881484
35 55.962062164328
36 54.1682652538143
37 52.482536978478
38 50.9456644262814
39 49.590971376394
40 48.4420164568155
41 47.510649219907
42 46.7956343690603
43 46.2821319803297
44 45.9423218256447
45 45.7373567515824
46 45.6206305041314
47 45.5420905794904
48 45.4530818739208
49 45.3110427502606
50 45.0833459119863
51 44.7497021253307
52 44.3028028460858
53 43.7472080128318
54 43.0968064701292
55 42.3714100008006
56 41.5931345993054
57 40.7831618581708
58 39.9592896522249
59 39.1344372577241
60 38.3160397171064
61 37.5061119448837
62 36.701716850966
63 35.8956277921352
64 35.0770951849322
65 34.2327538643844
66 33.3477881569106
67 32.407472616901
68 31.399125300894
69 30.314374092975
70 29.1514905570371
71 27.9174386007458
72 26.6292528089243
73 25.3144168376352
74 24.0100441934633
75 22.7608414257793
76 21.6160180933471
77 20.6254625679026
78 19.8356029711031
79 19.2854085087038
};
\addlegendentry{real}
\end{axis}

\end{tikzpicture}

%% file: pic/new198/198_daily_sev_cases.tex
\begin{tikzpicture}

\definecolor{color0}{rgb}{0.886274509803922,0.290196078431373,0.2}
\definecolor{color1}{rgb}{0.203921568627451,0.541176470588235,0.741176470588235}
\pgfmathsetlengthmacro\MajorTickLength{
      \pgfkeysvalueof{/pgfplots/major tick length} * 0.5
    }
\begin{axis}[
axis line style={white},
legend cell align={left},
legend style={fill opacity=0.8, draw opacity=1, text opacity=1, at={(0.03,0.97)}, anchor=north west, draw=white!80!black, fill=white!89.8039215686275!black},
tick align=inside,
major tick length=\MajorTickLength,
tick pos=left,
x grid style={white},
font = \fontsize{7}{7.5}\selectfont,
height=1in,width=0.38\linewidth,
xlabel={days},
xmajorgrids,
xmin=0, xmax=80,
xtick style={color=white!33.3333333333333!black},
y grid style={white},
x label style={at={(0.5, 0.3)}},
ymajorgrids,
y label style={at={(0.55,0.5)}},
ytick={0,200,400},
yticklabels={0,200,400},
yticklabel style={rotate=90},
xtick={0,20,40,60,80},
xticklabels={0,20,40,60,80},
ymin=-19.576427790496, ymax=406.540297403553,
ytick style={color=white!33.3333333333333!black}
]
\path [fill=color0, fill opacity=0.3, very thin]
(axis cs:0,0)
--(axis cs:0,0)
--(axis cs:1,-0.207485736221056)
--(axis cs:2,-0.126743392203945)
--(axis cs:3,-0.00447430817126304)
--(axis cs:4,0.157133333371425)
--(axis cs:5,0.5109081981993)
--(axis cs:6,1.0843799160693)
--(axis cs:7,1.89960054882631)
--(axis cs:8,3.14806857548518)
--(axis cs:9,4.85676528315149)
--(axis cs:10,7.40156585162102)
--(axis cs:11,10.7103598235887)
--(axis cs:12,16.1679124739245)
--(axis cs:13,23.511047177559)
--(axis cs:14,31.9445594801994)
--(axis cs:15,39.8594507499765)
--(axis cs:16,47.6278779677215)
--(axis cs:17,54.1749706793095)
--(axis cs:18,60.4074351254861)
--(axis cs:19,66.7411374521865)
--(axis cs:20,73.0525025365884)
--(axis cs:21,78.544377966101)
--(axis cs:22,84.4341313190517)
--(axis cs:23,90.2220556864806)
--(axis cs:24,95.1717069735508)
--(axis cs:25,100.55963478773)
--(axis cs:26,105.706413680357)
--(axis cs:27,110.623793315455)
--(axis cs:28,115.398295535913)
--(axis cs:29,120.015019084789)
--(axis cs:30,124.649508113832)
--(axis cs:31,129.006127154164)
--(axis cs:32,133.573132000736)
--(axis cs:33,138.091443798454)
--(axis cs:34,142.502610441466)
--(axis cs:35,146.729706821828)
--(axis cs:36,151.049870072916)
--(axis cs:37,155.148549734395)
--(axis cs:38,159.358533584813)
--(axis cs:39,163.399603847137)
--(axis cs:40,167.678075703829)
--(axis cs:41,172.087294004161)
--(axis cs:42,175.654249822963)
--(axis cs:43,179.645556661046)
--(axis cs:44,183.464160489549)
--(axis cs:45,187.35904110702)
--(axis cs:46,191.058038671705)
--(axis cs:47,194.667945233123)
--(axis cs:48,198.236100304616)
--(axis cs:49,201.638730908482)
--(axis cs:50,204.807170228343)
--(axis cs:51,208.62513558681)
--(axis cs:52,211.99100036251)
--(axis cs:53,215.235640759603)
--(axis cs:54,218.121154529044)
--(axis cs:55,221.342232946125)
--(axis cs:56,224.043872430765)
--(axis cs:57,226.824711881671)
--(axis cs:58,229.51822901818)
--(axis cs:59,232.583810266506)
--(axis cs:60,235.43249577872)
--(axis cs:61,238.107264320403)
--(axis cs:62,240.676703145777)
--(axis cs:63,243.409654420909)
--(axis cs:64,245.984904433796)
--(axis cs:65,248.301275328557)
--(axis cs:66,250.524786132217)
--(axis cs:67,252.622645742167)
--(axis cs:68,255.010250301971)
--(axis cs:69,257.438813809991)
--(axis cs:70,259.641644279073)
--(axis cs:71,261.904744859915)
--(axis cs:72,264.159300559256)
--(axis cs:73,266.211940945801)
--(axis cs:74,267.997018540625)
--(axis cs:75,269.869978087114)
--(axis cs:76,271.736647766651)
--(axis cs:77,273.729869710646)
--(axis cs:78,275.451804313658)
--(axis cs:79,276.92388274596)
--(axis cs:79,387.171355349278)
--(axis cs:79,387.171355349278)
--(axis cs:78,384.891052829199)
--(axis cs:77,382.803463622687)
--(axis cs:76,380.568114138111)
--(axis cs:75,378.225260008124)
--(axis cs:74,376.060124316518)
--(axis cs:73,373.530916197056)
--(axis cs:72,371.088318488363)
--(axis cs:71,368.33335037818)
--(axis cs:70,365.929784292355)
--(axis cs:69,362.8373766662)
--(axis cs:68,360.008797317077)
--(axis cs:67,357.158306638786)
--(axis cs:66,354.265690058259)
--(axis cs:65,350.803486576205)
--(axis cs:64,347.653190804299)
--(axis cs:63,344.104631293377)
--(axis cs:62,340.637582568509)
--(axis cs:61,336.911783298644)
--(axis cs:60,333.034170887947)
--(axis cs:59,329.063808781113)
--(axis cs:58,325.224628124677)
--(axis cs:57,321.184811927853)
--(axis cs:56,316.841841854949)
--(axis cs:55,312.162528958637)
--(axis cs:54,307.631226423337)
--(axis cs:53,302.745311621349)
--(axis cs:52,297.818523447013)
--(axis cs:51,292.841531079857)
--(axis cs:50,287.859496438323)
--(axis cs:49,282.827935758185)
--(axis cs:48,277.259137790622)
--(axis cs:47,271.779673814496)
--(axis cs:46,266.084818471152)
--(axis cs:45,259.926673178694)
--(axis cs:44,254.402506177117)
--(axis cs:43,249.135395719906)
--(axis cs:42,243.660035891323)
--(axis cs:41,237.54127742441)
--(axis cs:40,231.56954334379)
--(axis cs:39,225.448015200482)
--(axis cs:38,218.984323558045)
--(axis cs:37,213.013355027509)
--(axis cs:36,206.854891831846)
--(axis cs:35,200.632197940077)
--(axis cs:34,194.402151463296)
--(axis cs:33,187.889508582498)
--(axis cs:32,181.760201332598)
--(axis cs:31,175.527206179169)
--(axis cs:30,169.493349029025)
--(axis cs:29,163.765933296164)
--(axis cs:28,157.439799702182)
--(axis cs:27,151.195254303593)
--(axis cs:26,144.198348224405)
--(axis cs:25,138.002269974175)
--(axis cs:24,131.009245407402)
--(axis cs:23,124.111277646853)
--(axis cs:22,117.708725823805)
--(axis cs:21,110.569907748185)
--(axis cs:20,103.452259368173)
--(axis cs:19,96.0683863573373)
--(axis cs:18,88.2020886840377)
--(axis cs:17,80.0345531302143)
--(axis cs:16,71.1721220322785)
--(axis cs:15,61.1976921071663)
--(axis cs:14,51.284011948372)
--(axis cs:13,40.8508575843458)
--(axis cs:12,30.1273256213136)
--(axis cs:11,21.9658306526018)
--(axis cs:10,15.7412912912361)
--(axis cs:9,11.3432347168485)
--(axis cs:8,8.24240761499101)
--(axis cs:7,6.15754230831655)
--(axis cs:6,4.62038198869261)
--(axis cs:5,3.44147275418165)
--(axis cs:4,2.53810476186667)
--(axis cs:3,1.96637907007602)
--(axis cs:2,1.42198148744204)
--(axis cs:1,0.636057164792484)
--(axis cs:0,0)
--cycle;

\path [fill=color1, fill opacity=0.3, very thin]
(axis cs:0,0)
--(axis cs:0,0)
--(axis cs:1,0)
--(axis cs:2,0)
--(axis cs:3,0)
--(axis cs:4,0)
--(axis cs:5,0)
--(axis cs:6,0)
--(axis cs:7,0)
--(axis cs:8,0)
--(axis cs:9,0)
--(axis cs:10,0)
--(axis cs:11,0)
--(axis cs:12,0)
--(axis cs:13,25)
--(axis cs:14,31)
--(axis cs:15,35)
--(axis cs:16,38)
--(axis cs:17,51)
--(axis cs:18,61)
--(axis cs:19,70)
--(axis cs:20,78)
--(axis cs:21,86)
--(axis cs:22,87)
--(axis cs:23,92)
--(axis cs:24,101)
--(axis cs:25,113)
--(axis cs:26,117)
--(axis cs:27,124)
--(axis cs:28,130)
--(axis cs:29,130)
--(axis cs:30,134)
--(axis cs:31,142)
--(axis cs:32,146)
--(axis cs:33,162)
--(axis cs:34,176)
--(axis cs:35,178)
--(axis cs:36,180)
--(axis cs:37,180)
--(axis cs:38,188)
--(axis cs:39,196)
--(axis cs:40,204)
--(axis cs:41,208)
--(axis cs:42,211)
--(axis cs:43,213)
--(axis cs:44,213)
--(axis cs:45,222)
--(axis cs:46,228)
--(axis cs:47,236)
--(axis cs:48,235)
--(axis cs:49,237)
--(axis cs:50,238)
--(axis cs:51,240)
--(axis cs:52,244)
--(axis cs:53,247)
--(axis cs:54,254)
--(axis cs:55,259)
--(axis cs:56,273)
--(axis cs:57,273)
--(axis cs:58,274)
--(axis cs:59,280)
--(axis cs:60,285)
--(axis cs:61,283)
--(axis cs:62,290)
--(axis cs:63,292)
--(axis cs:64,292)
--(axis cs:65,294)
--(axis cs:66,304)
--(axis cs:67,310)
--(axis cs:68,313)
--(axis cs:69,314)
--(axis cs:70,317)
--(axis cs:71,318)
--(axis cs:72,319)
--(axis cs:73,319)
--(axis cs:74,324)
--(axis cs:75,331)
--(axis cs:76,336)
--(axis cs:77,340)
--(axis cs:78,340)
--(axis cs:79,340)
--(axis cs:79,340)
--(axis cs:79,340)
--(axis cs:78,340)
--(axis cs:77,340)
--(axis cs:76,336)
--(axis cs:75,331)
--(axis cs:74,324)
--(axis cs:73,319)
--(axis cs:72,319)
--(axis cs:71,318)
--(axis cs:70,317)
--(axis cs:69,314)
--(axis cs:68,313)
--(axis cs:67,310)
--(axis cs:66,304)
--(axis cs:65,294)
--(axis cs:64,292)
--(axis cs:63,292)
--(axis cs:62,290)
--(axis cs:61,283)
--(axis cs:60,285)
--(axis cs:59,280)
--(axis cs:58,274)
--(axis cs:57,273)
--(axis cs:56,273)
--(axis cs:55,259)
--(axis cs:54,254)
--(axis cs:53,247)
--(axis cs:52,244)
--(axis cs:51,240)
--(axis cs:50,238)
--(axis cs:49,237)
--(axis cs:48,235)
--(axis cs:47,236)
--(axis cs:46,228)
--(axis cs:45,222)
--(axis cs:44,213)
--(axis cs:43,213)
--(axis cs:42,211)
--(axis cs:41,208)
--(axis cs:40,204)
--(axis cs:39,196)
--(axis cs:38,188)
--(axis cs:37,180)
--(axis cs:36,180)
--(axis cs:35,178)
--(axis cs:34,176)
--(axis cs:33,162)
--(axis cs:32,146)
--(axis cs:31,142)
--(axis cs:30,134)
--(axis cs:29,130)
--(axis cs:28,130)
--(axis cs:27,124)
--(axis cs:26,117)
--(axis cs:25,113)
--(axis cs:24,101)
--(axis cs:23,92)
--(axis cs:22,87)
--(axis cs:21,86)
--(axis cs:20,78)
--(axis cs:19,70)
--(axis cs:18,61)
--(axis cs:17,51)
--(axis cs:16,38)
--(axis cs:15,35)
--(axis cs:14,31)
--(axis cs:13,25)
--(axis cs:12,0)
--(axis cs:11,0)
--(axis cs:10,0)
--(axis cs:9,0)
--(axis cs:8,0)
--(axis cs:7,0)
--(axis cs:6,0)
--(axis cs:5,0)
--(axis cs:4,0)
--(axis cs:3,0)
--(axis cs:2,0)
--(axis cs:1,0)
--(axis cs:0,0)
--cycle;

\addplot [semithick, color0]
table {%
0 0
1 0.214285714285714
2 0.647619047619048
3 0.980952380952381
4 1.34761904761905
5 1.97619047619048
6 2.85238095238095
7 4.02857142857143
8 5.6952380952381
9 8.1
10 11.5714285714286
11 16.3380952380952
12 23.147619047619
13 32.1809523809524
14 41.6142857142857
15 50.5285714285714
16 59.4
17 67.1047619047619
18 74.3047619047619
19 81.4047619047619
20 88.2523809523809
21 94.5571428571429
22 101.071428571429
23 107.166666666667
24 113.090476190476
25 119.280952380952
26 124.952380952381
27 130.909523809524
28 136.419047619048
29 141.890476190476
30 147.071428571429
31 152.266666666667
32 157.666666666667
33 162.990476190476
34 168.452380952381
35 173.680952380952
36 178.952380952381
37 184.080952380952
38 189.171428571429
39 194.42380952381
40 199.62380952381
41 204.814285714286
42 209.657142857143
43 214.390476190476
44 218.933333333333
45 223.642857142857
46 228.571428571429
47 233.22380952381
48 237.747619047619
49 242.233333333333
50 246.333333333333
51 250.733333333333
52 254.904761904762
53 258.990476190476
54 262.87619047619
55 266.752380952381
56 270.442857142857
57 274.004761904762
58 277.371428571429
59 280.82380952381
60 284.233333333333
61 287.509523809524
62 290.657142857143
63 293.757142857143
64 296.819047619048
65 299.552380952381
66 302.395238095238
67 304.890476190476
68 307.509523809524
69 310.138095238095
70 312.785714285714
71 315.119047619048
72 317.62380952381
73 319.871428571429
74 322.028571428571
75 324.047619047619
76 326.152380952381
77 328.266666666667
78 330.171428571429
79 332.047619047619
};
\addplot [semithick, color1]
table {%
0 0
1 0
2 0
3 0
4 0
5 0
6 0
7 0
8 0
9 0
10 0
11 0
12 0
13 25
14 31
15 35
16 38
17 51
18 61
19 70
20 78
21 86
22 87
23 92
24 101
25 113
26 117
27 124
28 130
29 130
30 134
31 142
32 146
33 162
34 176
35 178
36 180
37 180
38 188
39 196
40 204
41 208
42 211
43 213
44 213
45 222
46 228
47 236
48 235
49 237
50 238
51 240
52 244
53 247
54 254
55 259
56 273
57 273
58 274
59 280
60 285
61 283
62 290
63 292
64 292
65 294
66 304
67 310
68 313
69 314
70 317
71 318
72 319
73 319
74 324
75 331
76 336
77 340
78 340
79 340
};
\end{axis}

\end{tikzpicture}

%% file: pic/new198/198_daily_dea_cases.tex
\begin{tikzpicture}

\definecolor{color0}{rgb}{0.886274509803922,0.290196078431373,0.2}
\definecolor{color1}{rgb}{0.203921568627451,0.541176470588235,0.741176470588235}
\pgfmathsetlengthmacro\MajorTickLength{
      \pgfkeysvalueof{/pgfplots/major tick length} * 0.5
    }
\begin{axis}[
axis line style={white},
major tick length=\MajorTickLength,
legend cell align={left},
legend style={fill opacity=0.8, draw opacity=1, text opacity=1, at={(-0.603,0.97)}, anchor=north west, draw=white!80!black, fill=white!89.8039215686275!black},
tick align=inside,
tick pos=left,
x grid style={white},
font = \fontsize{7}{7.5}\selectfont,
height=1in,width=0.38\linewidth,
xlabel={days},
xmajorgrids,
xmin=0, xmax=80,
xtick style={color=white!33.3333333333333!black},
y grid style={white},
x label style={at={(0.5, 0.3)}},
ymajorgrids,
y label style={at={(0.55,0.5)}},
ymin=0, ymax=160,
xtick={0,20,40,60,80},
xticklabels={0,20,40,60,80},
ytick={0,75,150},
yticklabels={0,75,150},
yticklabel style={rotate=90},
ytick style={color=white!33.3333333333333!black}
]
\path [fill=color0, fill opacity=0.3, very thin]
(axis cs:0,0)
--(axis cs:0,0)
--(axis cs:1,0)
--(axis cs:2,-0.10438034089883)
--(axis cs:3,-0.165339807380932)
--(axis cs:4,-0.198305428712809)
--(axis cs:5,-0.257254886457143)
--(axis cs:6,-0.270517352541881)
--(axis cs:7,-0.232849199837609)
--(axis cs:8,-0.187931976820319)
--(axis cs:9,-0.138879660341377)
--(axis cs:10,-0.0512406076842271)
--(axis cs:11,0.165939373837011)
--(axis cs:12,0.542455218955153)
--(axis cs:13,1.12122344030286)
--(axis cs:14,1.91012435205056)
--(axis cs:15,2.89824912295159)
--(axis cs:16,4.24099643478013)
--(axis cs:17,5.6591879672851)
--(axis cs:18,7.21244991689716)
--(axis cs:19,9.11776939164556)
--(axis cs:20,10.867483613071)
--(axis cs:21,12.7136927232857)
--(axis cs:22,14.7589050454328)
--(axis cs:23,17.0207347165491)
--(axis cs:24,18.7750730238964)
--(axis cs:25,20.8193583320796)
--(axis cs:26,22.9769312201331)
--(axis cs:27,25.3053323308811)
--(axis cs:28,27.2564417134316)
--(axis cs:29,29.310273295149)
--(axis cs:30,31.5115586474599)
--(axis cs:31,33.434206699811)
--(axis cs:32,35.4898301319531)
--(axis cs:33,37.424615400255)
--(axis cs:34,39.4167935565555)
--(axis cs:35,41.593699112691)
--(axis cs:36,43.4607121947931)
--(axis cs:37,45.5554487296396)
--(axis cs:38,47.4433050834222)
--(axis cs:39,49.5013080017104)
--(axis cs:40,51.4846841189638)
--(axis cs:41,53.2345765308834)
--(axis cs:42,55.2448318087068)
--(axis cs:43,57.3687389967818)
--(axis cs:44,59.3275645327405)
--(axis cs:45,61.309014208776)
--(axis cs:46,63.1208826375152)
--(axis cs:47,65.119175366563)
--(axis cs:48,67.0041381711261)
--(axis cs:49,68.889714987934)
--(axis cs:50,70.4491886680847)
--(axis cs:51,72.2357361727013)
--(axis cs:52,74.2547351272755)
--(axis cs:53,75.8163087045031)
--(axis cs:54,77.2899591062474)
--(axis cs:55,79.0481316607592)
--(axis cs:56,80.7842503427723)
--(axis cs:57,82.3498838303492)
--(axis cs:58,84.0516685972982)
--(axis cs:59,85.6245381253728)
--(axis cs:60,87.1335563000207)
--(axis cs:61,88.6039972953224)
--(axis cs:62,90.1725070607465)
--(axis cs:63,91.4984253210932)
--(axis cs:64,92.911471016514)
--(axis cs:65,94.460634278399)
--(axis cs:66,96.0256367829023)
--(axis cs:67,97.4692114442433)
--(axis cs:68,98.8368551396301)
--(axis cs:69,99.943174191032)
--(axis cs:70,101.149183273657)
--(axis cs:71,102.436082711954)
--(axis cs:72,103.475643853524)
--(axis cs:73,104.748699509678)
--(axis cs:74,105.99724170953)
--(axis cs:75,106.954519089715)
--(axis cs:76,108.008374345369)
--(axis cs:77,109.122962805687)
--(axis cs:78,110.154797163004)
--(axis cs:79,111.185836349329)
--(axis cs:79,158.557020793528)
--(axis cs:79,158.557020793528)
--(axis cs:78,157.264250456044)
--(axis cs:77,155.829418146694)
--(axis cs:76,154.486863749869)
--(axis cs:75,153.064528529332)
--(axis cs:74,151.774186861899)
--(axis cs:73,150.184633823656)
--(axis cs:72,148.61007043219)
--(axis cs:71,147.011536335665)
--(axis cs:70,145.260340535866)
--(axis cs:69,143.713968666111)
--(axis cs:68,141.991716288941)
--(axis cs:67,140.273645698614)
--(axis cs:66,138.402934645669)
--(axis cs:65,136.672699054934)
--(axis cs:64,134.7742432692)
--(axis cs:63,133.015860393193)
--(axis cs:62,130.960826272587)
--(axis cs:61,128.967431276106)
--(axis cs:60,126.999777033313)
--(axis cs:59,124.308795207961)
--(axis cs:58,121.900712355083)
--(axis cs:57,119.583449502984)
--(axis cs:56,117.139559181037)
--(axis cs:55,114.789963577336)
--(axis cs:54,112.3290885128)
--(axis cs:53,109.564643676449)
--(axis cs:52,107.069074396534)
--(axis cs:51,104.611882874918)
--(axis cs:50,101.941287522392)
--(axis cs:49,99.4245707263518)
--(axis cs:48,96.6339570669691)
--(axis cs:47,94.0522532048656)
--(axis cs:46,91.3553078386753)
--(axis cs:45,88.5862238864621)
--(axis cs:44,86.0152926101167)
--(axis cs:43,83.5074514794087)
--(axis cs:42,80.7932634293884)
--(axis cs:41,78.4225663262595)
--(axis cs:40,75.6105539762743)
--(axis cs:39,72.9558348554325)
--(axis cs:38,70.0709806308635)
--(axis cs:37,67.2540750798842)
--(axis cs:36,64.4726211385402)
--(axis cs:35,61.8253485063566)
--(axis cs:34,59.2974921577302)
--(axis cs:33,56.6420512664116)
--(axis cs:32,53.7768365347136)
--(axis cs:31,50.6991266335223)
--(axis cs:30,47.6122508763497)
--(axis cs:29,44.5278219429462)
--(axis cs:28,42.1340344770446)
--(axis cs:27,39.4565724310236)
--(axis cs:26,36.965925922724)
--(axis cs:25,34.2473083345871)
--(axis cs:24,31.2154031665798)
--(axis cs:23,28.2554557596414)
--(axis cs:22,25.7077616212338)
--(axis cs:21,22.9624977529048)
--(axis cs:20,20.3039449583576)
--(axis cs:19,17.7584210845449)
--(axis cs:18,14.9589786545314)
--(axis cs:17,12.6646215565244)
--(axis cs:16,10.3970988033151)
--(axis cs:15,8.1969889722865)
--(axis cs:14,6.2517804098542)
--(axis cs:13,4.63115751207809)
--(axis cs:12,3.47659240009247)
--(axis cs:11,2.56739395949632)
--(axis cs:10,1.95600251244613)
--(axis cs:9,1.56745108891281)
--(axis cs:8,1.23555102443937)
--(axis cs:7,0.937611104599513)
--(axis cs:6,0.699088781113309)
--(axis cs:5,0.552492981695238)
--(axis cs:4,0.388781619188999)
--(axis cs:3,0.260577902619028)
--(axis cs:2,0.132951769470258)
--(axis cs:1,0)
--(axis cs:0,0)
--cycle;

\path [fill=color1, fill opacity=0.3, very thin]
(axis cs:0,0)
--(axis cs:0,0)
--(axis cs:1,0)
--(axis cs:2,0)
--(axis cs:3,0)
--(axis cs:4,0)
--(axis cs:5,0)
--(axis cs:6,0)
--(axis cs:7,0)
--(axis cs:8,1)
--(axis cs:9,1)
--(axis cs:10,1)
--(axis cs:11,2)
--(axis cs:12,2)
--(axis cs:13,2)
--(axis cs:14,2)
--(axis cs:15,2)
--(axis cs:16,2)
--(axis cs:17,2)
--(axis cs:18,2)
--(axis cs:19,2)
--(axis cs:20,2)
--(axis cs:21,2)
--(axis cs:22,3)
--(axis cs:23,4)
--(axis cs:24,4)
--(axis cs:25,6)
--(axis cs:26,10)
--(axis cs:27,12)
--(axis cs:28,18)
--(axis cs:29,19)
--(axis cs:30,19)
--(axis cs:31,21)
--(axis cs:32,24)
--(axis cs:33,26)
--(axis cs:34,38)
--(axis cs:35,43)
--(axis cs:36,47)
--(axis cs:37,50)
--(axis cs:38,55)
--(axis cs:39,67)
--(axis cs:40,74)
--(axis cs:41,69)
--(axis cs:42,71)
--(axis cs:43,73)
--(axis cs:44,73)
--(axis cs:45,75)
--(axis cs:46,83)
--(axis cs:47,82)
--(axis cs:48,90)
--(axis cs:49,95)
--(axis cs:50,98)
--(axis cs:51,98)
--(axis cs:52,98)
--(axis cs:53,105)
--(axis cs:54,107)
--(axis cs:55,113)
--(axis cs:56,115)
--(axis cs:57,116)
--(axis cs:58,118)
--(axis cs:59,119)
--(axis cs:60,123)
--(axis cs:61,135)
--(axis cs:62,135)
--(axis cs:63,137)
--(axis cs:64,139)
--(axis cs:65,139)
--(axis cs:66,139)
--(axis cs:67,140)
--(axis cs:68,140)
--(axis cs:69,140)
--(axis cs:70,142)
--(axis cs:71,154)
--(axis cs:72,146)
--(axis cs:73,146)
--(axis cs:74,146)
--(axis cs:75,147)
--(axis cs:76,147)
--(axis cs:77,147)
--(axis cs:78,147)
--(axis cs:79,147)
--(axis cs:79,147)
--(axis cs:79,147)
--(axis cs:78,147)
--(axis cs:77,147)
--(axis cs:76,147)
--(axis cs:75,147)
--(axis cs:74,146)
--(axis cs:73,146)
--(axis cs:72,146)
--(axis cs:71,154)
--(axis cs:70,142)
--(axis cs:69,140)
--(axis cs:68,140)
--(axis cs:67,140)
--(axis cs:66,139)
--(axis cs:65,139)
--(axis cs:64,139)
--(axis cs:63,137)
--(axis cs:62,135)
--(axis cs:61,135)
--(axis cs:60,123)
--(axis cs:59,119)
--(axis cs:58,118)
--(axis cs:57,116)
--(axis cs:56,115)
--(axis cs:55,113)
--(axis cs:54,107)
--(axis cs:53,105)
--(axis cs:52,98)
--(axis cs:51,98)
--(axis cs:50,98)
--(axis cs:49,95)
--(axis cs:48,90)
--(axis cs:47,82)
--(axis cs:46,83)
--(axis cs:45,75)
--(axis cs:44,73)
--(axis cs:43,73)
--(axis cs:42,71)
--(axis cs:41,69)
--(axis cs:40,74)
--(axis cs:39,67)
--(axis cs:38,55)
--(axis cs:37,50)
--(axis cs:36,47)
--(axis cs:35,43)
--(axis cs:34,38)
--(axis cs:33,26)
--(axis cs:32,24)
--(axis cs:31,21)
--(axis cs:30,19)
--(axis cs:29,19)
--(axis cs:28,18)
--(axis cs:27,12)
--(axis cs:26,10)
--(axis cs:25,6)
--(axis cs:24,4)
--(axis cs:23,4)
--(axis cs:22,3)
--(axis cs:21,2)
--(axis cs:20,2)
--(axis cs:19,2)
--(axis cs:18,2)
--(axis cs:17,2)
--(axis cs:16,2)
--(axis cs:15,2)
--(axis cs:14,2)
--(axis cs:13,2)
--(axis cs:12,2)
--(axis cs:11,2)
--(axis cs:10,1)
--(axis cs:9,1)
--(axis cs:8,1)
--(axis cs:7,0)
--(axis cs:6,0)
--(axis cs:5,0)
--(axis cs:4,0)
--(axis cs:3,0)
--(axis cs:2,0)
--(axis cs:1,0)
--(axis cs:0,0)
--cycle;

\addplot [semithick, color0]
table {%
0 0
1 0
2 0.0142857142857143
3 0.0476190476190476
4 0.0952380952380952
5 0.147619047619048
6 0.214285714285714
7 0.352380952380952
8 0.523809523809524
9 0.714285714285714
10 0.952380952380952
11 1.36666666666667
12 2.00952380952381
13 2.87619047619048
14 4.08095238095238
15 5.54761904761905
16 7.31904761904762
17 9.16190476190476
18 11.0857142857143
19 13.4380952380952
20 15.5857142857143
21 17.8380952380952
22 20.2333333333333
23 22.6380952380952
24 24.9952380952381
25 27.5333333333333
26 29.9714285714286
27 32.3809523809524
28 34.6952380952381
29 36.9190476190476
30 39.5619047619048
31 42.0666666666667
32 44.6333333333333
33 47.0333333333333
34 49.3571428571429
35 51.7095238095238
36 53.9666666666667
37 56.4047619047619
38 58.7571428571429
39 61.2285714285714
40 63.5476190476191
41 65.8285714285714
42 68.0190476190476
43 70.4380952380952
44 72.6714285714286
45 74.947619047619
46 77.2380952380952
47 79.5857142857143
48 81.8190476190476
49 84.1571428571429
50 86.1952380952381
51 88.4238095238095
52 90.6619047619048
53 92.6904761904762
54 94.8095238095238
55 96.9190476190476
56 98.9619047619048
57 100.966666666667
58 102.97619047619
59 104.966666666667
60 107.066666666667
61 108.785714285714
62 110.566666666667
63 112.257142857143
64 113.842857142857
65 115.566666666667
66 117.214285714286
67 118.871428571429
68 120.414285714286
69 121.828571428571
70 123.204761904762
71 124.72380952381
72 126.042857142857
73 127.466666666667
74 128.885714285714
75 130.009523809524
76 131.247619047619
77 132.47619047619
78 133.709523809524
79 134.871428571429
};
\addplot [semithick, color1]
table {%
0 0
1 0
2 0
3 0
4 0
5 0
6 0
7 0
8 1
9 1
10 1
11 2
12 2
13 2
14 2
15 2
16 2
17 2
18 2
19 2
20 2
21 2
22 3
23 4
24 4
25 6
26 10
27 12
28 18
29 19
30 19
31 21
32 24
33 26
34 38
35 43
36 47
37 50
38 55
39 67
40 74
41 69
42 71
43 73
44 73
45 75
46 83
47 82
48 90
49 95
50 98
51 98
52 98
53 105
54 107
55 113
56 115
57 116
58 118
59 119
60 123
61 135
62 135
63 137
64 139
65 139
66 139
67 140
68 140
69 140
70 142
71 154
72 146
73 146
74 146
75 147
76 147
77 147
78 147
79 147
};
\end{axis}

\end{tikzpicture}

%% file: pic/new217/217dic1k.tex
\begin{tikzpicture}

\definecolor{color0}{rgb}{0.886274509803922,0.290196078431373,0.2}
\definecolor{color1}{rgb}{0.203921568627451,0.541176470588235,0.741176470588235}
\pgfmathsetlengthmacro\MajorTickLength{
      \pgfkeysvalueof{/pgfplots/major tick length} * 0.5
    }
\begin{axis}[
width = 0.4\linewidth,
height = 1in,
major tick length=\MajorTickLength,
font =\fontsize{6}{6.5}\selectfont,
axis line style={white},
legend cell align={left},
legend style={fill opacity=0.8, draw opacity=1, text opacity=1, draw=white!80!black, fill=white!89.8039215686275!black},
tick align=inside,
tick pos=left,
x grid style={white},
yticklabel style={rotate=90},
xlabel={Days},
xmajorgrids,
xmin=0, xmax=80,
xlabel near ticks,
x label style={at={(0.5, -0.2)}},
ymajorgrids,
y label style={at={(0.45,0.5)}},
xtick style={color=white!33.3333333333333!black},
y grid style={white},
xtick={0,20,40,60,80},
xticklabels={0,20,40,60,80},
ylabel={Daily Cases},
ymajorgrids,
ytick={0,600,1200},
yticklabels={0,600,1.2k},
ymin=-57.5045703558751, ymax=1301.81984346191,
ytick style={color=white!33.3333333333333!black}
]
\path [fill=color0, fill opacity=0.3, very thin]
(axis cs:0,10.8750920068099)
--(axis cs:0,5.59157465985672)
--(axis cs:1,7.72864082367162)
--(axis cs:2,9.91085685893234)
--(axis cs:3,10.513553147154)
--(axis cs:4,16.9560386577209)
--(axis cs:5,26.1528434543228)
--(axis cs:6,38.0714982188861)
--(axis cs:7,56.8803806612314)
--(axis cs:8,84.0121966754838)
--(axis cs:9,119.083898258837)
--(axis cs:10,109.529095093741)
--(axis cs:11,80.871967496344)
--(axis cs:12,104.924179798176)
--(axis cs:13,119.152738027458)
--(axis cs:14,148.694763300592)
--(axis cs:15,169.512032816027)
--(axis cs:16,184.984017515235)
--(axis cs:17,198.971893766053)
--(axis cs:18,217.791641638384)
--(axis cs:19,233.499955791361)
--(axis cs:20,246.372359032524)
--(axis cs:21,263.808355093662)
--(axis cs:22,289.225994745432)
--(axis cs:23,297.545476444669)
--(axis cs:24,319.619123429413)
--(axis cs:25,346.591836416735)
--(axis cs:26,350.193822647646)
--(axis cs:27,361.080718737796)
--(axis cs:28,392.098117716139)
--(axis cs:29,401.006654753103)
--(axis cs:30,417.857335919667)
--(axis cs:31,418.241263789573)
--(axis cs:32,427.375589120964)
--(axis cs:33,436.050135675679)
--(axis cs:34,449.35921991936)
--(axis cs:35,434.187285353784)
--(axis cs:36,461.777985201966)
--(axis cs:37,472.146284254538)
--(axis cs:38,469.434716117625)
--(axis cs:39,461.769049741198)
--(axis cs:40,468.443988718083)
--(axis cs:41,481.891593463014)
--(axis cs:42,495.557374701237)
--(axis cs:43,493.926050924498)
--(axis cs:44,504.408050952107)
--(axis cs:45,518.355022883514)
--(axis cs:46,506.963459585388)
--(axis cs:47,533.776761380479)
--(axis cs:48,524.858210986814)
--(axis cs:49,529.30584081196)
--(axis cs:50,534.194721538897)
--(axis cs:51,551.460287854982)
--(axis cs:52,543.894562016686)
--(axis cs:53,557.300813880013)
--(axis cs:54,544.95953099632)
--(axis cs:55,550.448081978313)
--(axis cs:56,566.497508055224)
--(axis cs:57,571.410608737382)
--(axis cs:58,568.873644582973)
--(axis cs:59,576.102321442319)
--(axis cs:60,566.344044908386)
--(axis cs:61,580.340673553665)
--(axis cs:62,584.354454544625)
--(axis cs:63,575.928377091051)
--(axis cs:64,563.128396861418)
--(axis cs:65,570.605568031122)
--(axis cs:66,565.994874029189)
--(axis cs:67,570.239246731808)
--(axis cs:68,572.378613539736)
--(axis cs:69,579.7203303788)
--(axis cs:70,556.677748909952)
--(axis cs:71,583.584224764221)
--(axis cs:72,571.689819552566)
--(axis cs:73,572.215297485699)
--(axis cs:74,548.453957558717)
--(axis cs:75,564.472791397281)
--(axis cs:76,579.065152526153)
--(axis cs:77,568.866160698789)
--(axis cs:78,578.870155506923)
--(axis cs:79,581.367629893446)
--(axis cs:79,1240.03237010655)
--(axis cs:79,1240.03237010655)
--(axis cs:78,1211.99651115974)
--(axis cs:77,1192.53383930121)
--(axis cs:76,1180.80151414051)
--(axis cs:75,1157.32720860272)
--(axis cs:74,1146.81270910795)
--(axis cs:73,1138.45136918097)
--(axis cs:72,1127.7768471141)
--(axis cs:71,1118.61577523578)
--(axis cs:70,1091.25558442338)
--(axis cs:69,1094.14633628787)
--(axis cs:68,1065.7547197936)
--(axis cs:67,1066.09408660152)
--(axis cs:66,1036.80512597081)
--(axis cs:65,1018.39443196888)
--(axis cs:64,1034.00493647192)
--(axis cs:63,1023.40495624228)
--(axis cs:62,1017.17887878871)
--(axis cs:61,1020.525993113)
--(axis cs:60,992.455955091614)
--(axis cs:59,968.097678557681)
--(axis cs:58,963.126355417027)
--(axis cs:57,953.056057929285)
--(axis cs:56,939.702491944776)
--(axis cs:55,914.68525135502)
--(axis cs:54,905.44046900368)
--(axis cs:53,899.965852786653)
--(axis cs:52,859.172104649981)
--(axis cs:51,878.939712145018)
--(axis cs:50,845.738611794436)
--(axis cs:49,838.427492521373)
--(axis cs:48,837.608455679853)
--(axis cs:47,811.689905286188)
--(axis cs:46,808.036540414612)
--(axis cs:45,781.644977116486)
--(axis cs:44,763.85861571456)
--(axis cs:43,747.207282408835)
--(axis cs:42,745.842625298763)
--(axis cs:41,738.975073203653)
--(axis cs:40,714.422677948584)
--(axis cs:39,688.830950258802)
--(axis cs:38,666.298617215708)
--(axis cs:37,652.987049078795)
--(axis cs:36,621.2886814647)
--(axis cs:35,619.812714646216)
--(axis cs:34,582.907446747307)
--(axis cs:33,552.749864324321)
--(axis cs:32,524.624410879036)
--(axis cs:31,521.49206954376)
--(axis cs:30,520.275997413666)
--(axis cs:29,509.193345246897)
--(axis cs:28,481.235215617194)
--(axis cs:27,455.319281262204)
--(axis cs:26,432.939510685687)
--(axis cs:25,422.274830249931)
--(axis cs:24,399.780876570587)
--(axis cs:23,378.054523555331)
--(axis cs:22,363.574005254568)
--(axis cs:21,343.591644906338)
--(axis cs:20,333.227640967476)
--(axis cs:19,308.900044208639)
--(axis cs:18,281.608358361616)
--(axis cs:17,274.428106233947)
--(axis cs:16,250.349315818098)
--(axis cs:15,237.55463385064)
--(axis cs:14,213.838570032741)
--(axis cs:13,184.180595305875)
--(axis cs:12,187.475820201824)
--(axis cs:11,205.794699170323)
--(axis cs:10,249.004238239592)
--(axis cs:9,213.98276840783)
--(axis cs:8,144.854469991183)
--(axis cs:7,108.719619338769)
--(axis cs:6,74.8618351144472)
--(axis cs:5,51.6471565456772)
--(axis cs:4,39.4439613422791)
--(axis cs:3,26.486446852846)
--(axis cs:2,19.622476474401)
--(axis cs:1,17.2046925096617)
--(axis cs:0,10.8750920068099)
--cycle;

\path [fill=color1, fill opacity=0.3, very thin]
(axis cs:0,11.383763667188)
--(axis cs:0,4.28290299947864)
--(axis cs:1,6.58987895892336)
--(axis cs:2,8.88780613813593)
--(axis cs:3,13.6112177916641)
--(axis cs:4,18.5573452587113)
--(axis cs:5,29.2862923969226)
--(axis cs:6,37.3740723868913)
--(axis cs:7,53.2141174631069)
--(axis cs:8,78.2265026284327)
--(axis cs:9,118.990673780333)
--(axis cs:10,103.59583906288)
--(axis cs:11,55.7112902049158)
--(axis cs:12,79.9846182283521)
--(axis cs:13,96.7657675598148)
--(axis cs:14,121.955246585779)
--(axis cs:15,127.717308607088)
--(axis cs:16,138.56963137046)
--(axis cs:17,150.567442599065)
--(axis cs:18,155.580144100478)
--(axis cs:19,165.786310839833)
--(axis cs:20,172.837776674988)
--(axis cs:21,175.586053139931)
--(axis cs:22,188.913689662137)
--(axis cs:23,188.029845813988)
--(axis cs:24,200.108846751635)
--(axis cs:25,211.147958706526)
--(axis cs:26,222.616217977547)
--(axis cs:27,231.09791153424)
--(axis cs:28,243.453959349052)
--(axis cs:29,249.967908996175)
--(axis cs:30,261.463985918915)
--(axis cs:31,266.279626272393)
--(axis cs:32,282.836858634591)
--(axis cs:33,297.636263996249)
--(axis cs:34,304.114779302641)
--(axis cs:35,308.004928947666)
--(axis cs:36,317.58448160555)
--(axis cs:37,332.270508837744)
--(axis cs:38,347.99011847962)
--(axis cs:39,346.103041782322)
--(axis cs:40,359.750229085413)
--(axis cs:41,380.506594770172)
--(axis cs:42,367.012427817217)
--(axis cs:43,382.447334414114)
--(axis cs:44,380.300361870532)
--(axis cs:45,403.48870239532)
--(axis cs:46,403.614110119877)
--(axis cs:47,395.196302108431)
--(axis cs:48,400.176706129548)
--(axis cs:49,420.205523137846)
--(axis cs:50,427.568153261884)
--(axis cs:51,429.035972064294)
--(axis cs:52,418.507886194554)
--(axis cs:53,431.54042190529)
--(axis cs:54,448.990257672359)
--(axis cs:55,439.397091600992)
--(axis cs:56,438.287397959579)
--(axis cs:57,450.145478910291)
--(axis cs:58,449.044531933545)
--(axis cs:59,462.06013090647)
--(axis cs:60,461.56664374771)
--(axis cs:61,476.205569502678)
--(axis cs:62,477.667929513879)
--(axis cs:63,477.050348699453)
--(axis cs:64,490.735257602259)
--(axis cs:65,491.965436907974)
--(axis cs:66,498.145182724763)
--(axis cs:67,504.968740028047)
--(axis cs:68,494.44911971407)
--(axis cs:69,499.539161163173)
--(axis cs:70,494.18653496206)
--(axis cs:71,507.804340937322)
--(axis cs:72,512.062079090684)
--(axis cs:73,508.766537410499)
--(axis cs:74,513.901480974912)
--(axis cs:75,514.297599575086)
--(axis cs:76,514.500109649157)
--(axis cs:77,535.176747268495)
--(axis cs:78,524.615970343374)
--(axis cs:79,518.06486218322)
--(axis cs:79,875.201804483447)
--(axis cs:79,875.201804483447)
--(axis cs:78,881.650696323293)
--(axis cs:77,878.289919398171)
--(axis cs:76,869.16655701751)
--(axis cs:75,856.369067091581)
--(axis cs:74,850.765185691754)
--(axis cs:73,809.633462589501)
--(axis cs:72,801.737920909316)
--(axis cs:71,799.595659062679)
--(axis cs:70,796.546798371273)
--(axis cs:69,796.19417217016)
--(axis cs:68,773.084213619263)
--(axis cs:67,769.164593305287)
--(axis cs:66,762.18815060857)
--(axis cs:65,740.434563092026)
--(axis cs:64,744.798075731074)
--(axis cs:63,720.016317967213)
--(axis cs:62,713.465403819455)
--(axis cs:61,690.861097163988)
--(axis cs:60,703.966689585623)
--(axis cs:59,689.873202426863)
--(axis cs:58,657.888801399788)
--(axis cs:57,638.587854423042)
--(axis cs:56,643.645935373754)
--(axis cs:55,615.136241732341)
--(axis cs:54,605.476408994307)
--(axis cs:53,600.192911428043)
--(axis cs:52,582.358780472112)
--(axis cs:51,560.830694602372)
--(axis cs:50,558.365180071449)
--(axis cs:49,544.927810195488)
--(axis cs:48,543.489960537118)
--(axis cs:47,525.003697891569)
--(axis cs:46,509.119223213456)
--(axis cs:45,504.11129760468)
--(axis cs:44,487.766304796135)
--(axis cs:43,481.552665585886)
--(axis cs:42,461.05423884945)
--(axis cs:41,457.160071896494)
--(axis cs:40,453.116437581254)
--(axis cs:39,426.096958217678)
--(axis cs:38,427.60988152038)
--(axis cs:37,414.396157828922)
--(axis cs:36,407.81551839445)
--(axis cs:35,393.661737719)
--(axis cs:34,398.751887364026)
--(axis cs:33,377.630402670418)
--(axis cs:32,372.029808032076)
--(axis cs:31,363.320373727607)
--(axis cs:30,349.336014081085)
--(axis cs:29,340.298757670491)
--(axis cs:28,320.412707317615)
--(axis cs:27,311.435421799094)
--(axis cs:26,303.717115355786)
--(axis cs:25,285.585374626807)
--(axis cs:24,273.824486581698)
--(axis cs:23,262.903487519345)
--(axis cs:22,251.15297700453)
--(axis cs:21,236.147280193403)
--(axis cs:20,222.095556658346)
--(axis cs:19,217.7470224935)
--(axis cs:18,204.619855899522)
--(axis cs:17,186.365890734269)
--(axis cs:16,186.163701962873)
--(axis cs:15,174.882691392912)
--(axis cs:14,163.511420080888)
--(axis cs:13,147.434232440185)
--(axis cs:12,163.215381771648)
--(axis cs:11,190.822043128418)
--(axis cs:10,252.070827603786)
--(axis cs:9,206.209326219667)
--(axis cs:8,155.173497371567)
--(axis cs:7,106.719215870226)
--(axis cs:6,78.0259276131087)
--(axis cs:5,50.7803742697441)
--(axis cs:4,37.0426547412887)
--(axis cs:3,28.6554488750026)
--(axis cs:2,19.3788605285307)
--(axis cs:1,14.94345437441)
--(axis cs:0,11.383763667188)
--cycle;

\addplot [semithick, color0]
table {%
0 8.23333333333333
1 12.4666666666667
2 14.7666666666667
3 18.5
4 28.2
5 38.9
6 56.4666666666667
7 82.8
8 114.433333333333
9 166.533333333333
10 179.266666666667
11 143.333333333333
12 146.2
13 151.666666666667
14 181.266666666667
15 203.533333333333
16 217.666666666667
17 236.7
18 249.7
19 271.2
20 289.8
21 303.7
22 326.4
23 337.8
24 359.7
25 384.433333333333
26 391.566666666667
27 408.2
28 436.666666666667
29 455.1
30 469.066666666667
31 469.866666666667
32 476
33 494.4
34 516.133333333333
35 527
36 541.533333333333
37 562.566666666667
38 567.866666666667
39 575.3
40 591.433333333333
41 610.433333333333
42 620.7
43 620.566666666667
44 634.133333333333
45 650
46 657.5
47 672.733333333333
48 681.233333333333
49 683.866666666667
50 689.966666666667
51 715.2
52 701.533333333333
53 728.633333333333
54 725.2
55 732.566666666667
56 753.1
57 762.233333333333
58 766
59 772.1
60 779.4
61 800.433333333333
62 800.766666666667
63 799.666666666667
64 798.566666666667
65 794.5
66 801.4
67 818.166666666667
68 819.066666666667
69 836.933333333333
70 823.966666666667
71 851.1
72 849.733333333333
73 855.333333333333
74 847.633333333333
75 860.9
76 879.933333333333
77 880.7
78 895.433333333333
79 910.7
};
\addplot [semithick, color1]
table {%
0 7.83333333333333
1 10.7666666666667
2 14.1333333333333
3 21.1333333333333
4 27.8
5 40.0333333333333
6 57.7
7 79.9666666666667
8 116.7
9 162.6
10 177.833333333333
11 123.266666666667
12 121.6
13 122.1
14 142.733333333333
15 151.3
16 162.366666666667
17 168.466666666667
18 180.1
19 191.766666666667
20 197.466666666667
21 205.866666666667
22 220.033333333333
23 225.466666666667
24 236.966666666667
25 248.366666666667
26 263.166666666667
27 271.266666666667
28 281.933333333333
29 295.133333333333
30 305.4
31 314.8
32 327.433333333333
33 337.633333333333
34 351.433333333333
35 350.833333333333
36 362.7
37 373.333333333333
38 387.8
39 386.1
40 406.433333333333
41 418.833333333333
42 414.033333333333
43 432
44 434.033333333333
45 453.8
46 456.366666666667
47 460.1
48 471.833333333333
49 482.566666666667
50 492.966666666667
51 494.933333333333
52 500.433333333333
53 515.866666666667
54 527.233333333333
55 527.266666666667
56 540.966666666667
57 544.366666666667
58 553.466666666667
59 575.966666666667
60 582.766666666667
61 583.533333333333
62 595.566666666667
63 598.533333333333
64 617.766666666667
65 616.2
66 630.166666666667
67 637.066666666667
68 633.766666666667
69 647.866666666667
70 645.366666666667
71 653.7
72 656.9
73 659.2
74 682.333333333333
75 685.333333333333
76 691.833333333333
77 706.733333333333
78 703.133333333333
79 696.633333333333
};
\end{axis}
\end{tikzpicture}

%% file: pic/new217/217dic3k.tex
\begin{tikzpicture}

\definecolor{color0}{rgb}{0.886274509803922,0.290196078431373,0.2}
\definecolor{color1}{rgb}{0.203921568627451,0.541176470588235,0.741176470588235}
\pgfmathsetlengthmacro\MajorTickLength{
      \pgfkeysvalueof{/pgfplots/major tick length} * 0.5
    }
\begin{axis}[
width = 0.4\linewidth,
height = 1in,
major tick length=\MajorTickLength,
font =\fontsize{6}{6.5}\selectfont,
yticklabel style={rotate=90},
axis line style={white},
legend cell align={left},
legend style={fill opacity=0.8, draw opacity=1, text opacity=1, draw=white!80!black, fill=white!89.8039215686275!black},
tick align=inside,
tick pos=left,
x grid style={white},
xlabel={Days},
xmajorgrids,
xmin=0, xmax=80,
xlabel near ticks,
x label style={at={(0.5, -0.2)}},
y label style={at={(0.55,0.5)}},
xtick style={color=white!33.3333333333333!black},
y grid style={white},
xtick={0,20,40,60,80},
xticklabels={0,20,40,60,80},
ymajorgrids,
ytick={0,200,400},
yticklabels={0,200,400},
ymin=-16.4345523280843, ymax=428.282506585751,
ytick style={color=white!33.3333333333333!black}
]
\path [fill=color0, fill opacity=0.3, very thin]
(axis cs:0,10.4201405592736)
--(axis cs:0,3.77985944072644)
--(axis cs:1,7.01683136026503)
--(axis cs:2,9.84222276396341)
--(axis cs:3,12.1469153123196)
--(axis cs:4,16.2419375224712)
--(axis cs:5,25.763249555664)
--(axis cs:6,35.7130490449863)
--(axis cs:7,47.9001095052736)
--(axis cs:8,75.4668088134006)
--(axis cs:9,105.857286875938)
--(axis cs:10,110.743930505299)
--(axis cs:11,85.2084660460354)
--(axis cs:12,61.4258304062338)
--(axis cs:13,95.5794335283419)
--(axis cs:14,104.690261836075)
--(axis cs:15,118.613027202512)
--(axis cs:16,120.068903929105)
--(axis cs:17,123.25948415308)
--(axis cs:18,127.102165552814)
--(axis cs:19,129.181488280762)
--(axis cs:20,131.549730234756)
--(axis cs:21,136.495076151152)
--(axis cs:22,134.656795447821)
--(axis cs:23,137.727026739133)
--(axis cs:24,137.713636130198)
--(axis cs:25,145.990450718943)
--(axis cs:26,148.198226077713)
--(axis cs:27,146.894881566728)
--(axis cs:28,149.344354122755)
--(axis cs:29,152.38666880788)
--(axis cs:30,155.872799350898)
--(axis cs:31,157.680561566636)
--(axis cs:32,161.993010002473)
--(axis cs:33,167.872946157368)
--(axis cs:34,164.431432465685)
--(axis cs:35,170.316707347622)
--(axis cs:36,173.328656880008)
--(axis cs:37,179.490866995034)
--(axis cs:38,182.200422834871)
--(axis cs:39,185.343523064506)
--(axis cs:40,186.917514352078)
--(axis cs:41,192.623078880201)
--(axis cs:42,202.330204678816)
--(axis cs:43,207.765766802646)
--(axis cs:44,212.823815423682)
--(axis cs:45,220.673492452923)
--(axis cs:46,224.016743175484)
--(axis cs:47,220.506229994907)
--(axis cs:48,213.524736489209)
--(axis cs:49,210.60379758245)
--(axis cs:50,217.791979669211)
--(axis cs:51,221.838310705907)
--(axis cs:52,214.871200411769)
--(axis cs:53,224.723138052424)
--(axis cs:54,218.571900687764)
--(axis cs:55,215.826055590347)
--(axis cs:56,214.045278741918)
--(axis cs:57,213.854955489571)
--(axis cs:58,214.825897800428)
--(axis cs:59,222.147782276513)
--(axis cs:60,215.869632324404)
--(axis cs:61,222.402493657777)
--(axis cs:62,227.804472594955)
--(axis cs:63,217.861459800584)
--(axis cs:64,229.804924926504)
--(axis cs:65,222.965605420325)
--(axis cs:66,230.872054663466)
--(axis cs:67,236.580019514457)
--(axis cs:68,244.056171725746)
--(axis cs:69,241.136235927151)
--(axis cs:70,245.81998970476)
--(axis cs:71,254.519803610211)
--(axis cs:72,254.910643630811)
--(axis cs:73,256.901488829832)
--(axis cs:74,258.148770503604)
--(axis cs:75,256.646522018176)
--(axis cs:76,266.331430652111)
--(axis cs:77,251.683335310399)
--(axis cs:78,252.465238516393)
--(axis cs:79,257.213442810391)
--(axis cs:79,404.986557189609)
--(axis cs:79,404.986557189609)
--(axis cs:78,408.068094816941)
--(axis cs:77,392.183331356267)
--(axis cs:76,390.135236014556)
--(axis cs:75,395.220144648491)
--(axis cs:74,386.251229496396)
--(axis cs:73,394.965177836834)
--(axis cs:72,386.956023035856)
--(axis cs:71,380.346863056456)
--(axis cs:70,379.646676961907)
--(axis cs:69,386.597097406183)
--(axis cs:68,366.877161607588)
--(axis cs:67,370.953313818877)
--(axis cs:66,380.994612003201)
--(axis cs:65,378.967727913008)
--(axis cs:64,375.995075073496)
--(axis cs:63,378.338540199416)
--(axis cs:62,375.862194071712)
--(axis cs:61,370.330839675556)
--(axis cs:60,365.730367675596)
--(axis cs:59,367.318884390153)
--(axis cs:58,366.507435532905)
--(axis cs:57,358.078377843762)
--(axis cs:56,362.021387924748)
--(axis cs:55,354.307277742987)
--(axis cs:54,352.094765978903)
--(axis cs:53,355.076861947576)
--(axis cs:52,351.128799588231)
--(axis cs:51,343.495022627426)
--(axis cs:50,345.808020330789)
--(axis cs:49,336.99620241755)
--(axis cs:48,335.541930177458)
--(axis cs:47,322.693770005093)
--(axis cs:46,314.783256824516)
--(axis cs:45,313.126507547077)
--(axis cs:44,308.909517909651)
--(axis cs:43,295.434233197354)
--(axis cs:42,294.669795321184)
--(axis cs:41,290.043587786466)
--(axis cs:40,293.415818981255)
--(axis cs:39,276.056476935494)
--(axis cs:38,276.799577165129)
--(axis cs:37,271.109133004966)
--(axis cs:36,266.471343119992)
--(axis cs:35,259.083292652378)
--(axis cs:34,255.035234200982)
--(axis cs:33,249.260387175965)
--(axis cs:32,240.206989997527)
--(axis cs:31,251.386105100031)
--(axis cs:30,234.860533982436)
--(axis cs:29,235.41333119212)
--(axis cs:28,229.988979210578)
--(axis cs:27,228.305118433272)
--(axis cs:26,225.868440588954)
--(axis cs:25,217.209549281057)
--(axis cs:24,206.553030536469)
--(axis cs:23,208.2063065942)
--(axis cs:22,214.809871218845)
--(axis cs:21,198.104923848848)
--(axis cs:20,188.183603098578)
--(axis cs:19,179.618511719238)
--(axis cs:18,172.764501113853)
--(axis cs:17,174.007182513586)
--(axis cs:16,159.797762737561)
--(axis cs:15,161.586972797488)
--(axis cs:14,156.843071497258)
--(axis cs:13,145.087233138325)
--(axis cs:12,137.1075029271)
--(axis cs:11,248.991533953965)
--(axis cs:10,234.456069494701)
--(axis cs:9,203.476046457395)
--(axis cs:8,137.999857853266)
--(axis cs:7,98.6332238280597)
--(axis cs:6,74.9536176216803)
--(axis cs:5,49.5034171110027)
--(axis cs:4,34.9580624775288)
--(axis cs:3,23.5864180210138)
--(axis cs:2,18.9577772360366)
--(axis cs:1,15.4498353064016)
--(axis cs:0,10.4201405592736)
--cycle;

\path [fill=color1, fill opacity=0.3, very thin]
(axis cs:0,9.28516708042651)
--(axis cs:0,3.98149958624016)
--(axis cs:1,8.45651349372243)
--(axis cs:2,10.103789277355)
--(axis cs:3,12.4567454421654)
--(axis cs:4,15.9803995805644)
--(axis cs:5,27.4086542581241)
--(axis cs:6,38.2369822528649)
--(axis cs:7,57.8332304528709)
--(axis cs:8,81.5105111240782)
--(axis cs:9,108.941411906776)
--(axis cs:10,107.298001908738)
--(axis cs:11,23.3549527744921)
--(axis cs:12,36.3700108609169)
--(axis cs:13,77.5427404656168)
--(axis cs:14,84.9109048052783)
--(axis cs:15,89.4747451812099)
--(axis cs:16,91.6013166235831)
--(axis cs:17,97.2150861386383)
--(axis cs:18,99.1240397653839)
--(axis cs:19,98.5767961464052)
--(axis cs:20,97.2216949976205)
--(axis cs:21,97.0703931247031)
--(axis cs:22,102.125443961714)
--(axis cs:23,100.278915957839)
--(axis cs:24,99.6185199582079)
--(axis cs:25,100.153524512059)
--(axis cs:26,106.216397115648)
--(axis cs:27,109.641140762776)
--(axis cs:28,109.868140317632)
--(axis cs:29,110.509646001356)
--(axis cs:30,105.788517507617)
--(axis cs:31,110.953683140563)
--(axis cs:32,119.419727703967)
--(axis cs:33,121.394370722468)
--(axis cs:34,115.993312855236)
--(axis cs:35,119.414892571898)
--(axis cs:36,123.228013331987)
--(axis cs:37,120.284296689799)
--(axis cs:38,113.869498272547)
--(axis cs:39,117.312364324949)
--(axis cs:40,123.773841003884)
--(axis cs:41,122.986422802972)
--(axis cs:42,125.363441200092)
--(axis cs:43,122.198312644289)
--(axis cs:44,121.625916931299)
--(axis cs:45,122.47128782206)
--(axis cs:46,133.820573328157)
--(axis cs:47,130.228093367108)
--(axis cs:48,142.438264241813)
--(axis cs:49,131.450525030972)
--(axis cs:50,133.39945636814)
--(axis cs:51,130.61155962734)
--(axis cs:52,140.374807732241)
--(axis cs:53,147.94052172224)
--(axis cs:54,148.251421367108)
--(axis cs:55,158.94648052743)
--(axis cs:56,153.260507177611)
--(axis cs:57,153.272339371262)
--(axis cs:58,165.687882160235)
--(axis cs:59,160.012441748317)
--(axis cs:60,166.725286192898)
--(axis cs:61,171.848948099883)
--(axis cs:62,170.771059768887)
--(axis cs:63,172.629958830434)
--(axis cs:64,169.338269377562)
--(axis cs:65,170.330824580576)
--(axis cs:66,186.786376902669)
--(axis cs:67,185.237445248371)
--(axis cs:68,186.36667489712)
--(axis cs:69,188.940885976092)
--(axis cs:70,178.14582150991)
--(axis cs:71,186.357907338493)
--(axis cs:72,190.885283215379)
--(axis cs:73,186.915895311466)
--(axis cs:74,193.588190882255)
--(axis cs:75,188.485453393127)
--(axis cs:76,202.499079326116)
--(axis cs:77,201.2119025168)
--(axis cs:78,195.775915936996)
--(axis cs:79,205.079923448198)
--(axis cs:79,365.320076551802)
--(axis cs:79,365.320076551802)
--(axis cs:78,360.224084063004)
--(axis cs:77,350.1880974832)
--(axis cs:76,343.700920673884)
--(axis cs:75,337.98121327354)
--(axis cs:74,349.411809117745)
--(axis cs:73,333.484104688534)
--(axis cs:72,332.914716784621)
--(axis cs:71,327.042092661507)
--(axis cs:70,329.387511823424)
--(axis cs:69,321.925780690575)
--(axis cs:68,321.366658436213)
--(axis cs:67,316.829221418296)
--(axis cs:66,315.013623097331)
--(axis cs:65,315.602508752758)
--(axis cs:64,306.195063955771)
--(axis cs:63,307.3033745029)
--(axis cs:62,304.562273564446)
--(axis cs:61,297.28438523345)
--(axis cs:60,308.674713807102)
--(axis cs:59,300.387558251683)
--(axis cs:58,299.112117839765)
--(axis cs:57,302.127660628738)
--(axis cs:56,304.072826155722)
--(axis cs:55,301.25351947257)
--(axis cs:54,292.281911966225)
--(axis cs:53,291.992811611093)
--(axis cs:52,301.825192267759)
--(axis cs:51,294.38844037266)
--(axis cs:50,288.733876965193)
--(axis cs:49,292.816141635694)
--(axis cs:48,290.028402424853)
--(axis cs:47,285.238573299559)
--(axis cs:46,276.379426671843)
--(axis cs:45,280.195378844607)
--(axis cs:44,266.507416402035)
--(axis cs:43,267.068354022377)
--(axis cs:42,261.436558799908)
--(axis cs:41,264.413577197028)
--(axis cs:40,256.492825662782)
--(axis cs:39,251.287635675051)
--(axis cs:38,237.863835060787)
--(axis cs:37,232.449036643534)
--(axis cs:36,228.705320001347)
--(axis cs:35,223.851774094769)
--(axis cs:34,222.340020478097)
--(axis cs:33,211.405629277532)
--(axis cs:32,220.780272296033)
--(axis cs:31,204.846316859437)
--(axis cs:30,195.211482492383)
--(axis cs:29,193.090353998644)
--(axis cs:28,185.265193015702)
--(axis cs:27,186.625525903891)
--(axis cs:26,179.383602884352)
--(axis cs:25,174.313142154607)
--(axis cs:24,166.314813375125)
--(axis cs:23,155.054417375494)
--(axis cs:22,157.341222704952)
--(axis cs:21,151.929606875297)
--(axis cs:20,141.244971669046)
--(axis cs:19,138.823203853595)
--(axis cs:18,132.609293567949)
--(axis cs:17,130.784913861362)
--(axis cs:16,128.73201670975)
--(axis cs:15,119.72525481879)
--(axis cs:14,118.422428528055)
--(axis cs:13,114.857259534383)
--(axis cs:12,139.69665580575)
--(axis cs:11,194.578380558841)
--(axis cs:10,249.101998091262)
--(axis cs:9,200.258588093224)
--(axis cs:8,147.356155542588)
--(axis cs:7,111.833436213796)
--(axis cs:6,76.2963510804684)
--(axis cs:5,53.5913457418759)
--(axis cs:4,32.552933752769)
--(axis cs:3,25.0099212245012)
--(axis cs:2,20.3628773893116)
--(axis cs:1,19.4101531729442)
--(axis cs:0,9.28516708042651)
--cycle;

\addplot [semithick, color0]
table {%
0 7.1
1 11.2333333333333
2 14.4
3 17.8666666666667
4 25.6
5 37.6333333333333
6 55.3333333333333
7 73.2666666666667
8 106.733333333333
9 154.666666666667
10 172.6
11 167.1
12 99.2666666666667
13 120.333333333333
14 130.766666666667
15 140.1
16 139.933333333333
17 148.633333333333
18 149.933333333333
19 154.4
20 159.866666666667
21 167.3
22 174.733333333333
23 172.966666666667
24 172.133333333333
25 181.6
26 187.033333333333
27 187.6
28 189.666666666667
29 193.9
30 195.366666666667
31 204.533333333333
32 201.1
33 208.566666666667
34 209.733333333333
35 214.7
36 219.9
37 225.3
38 229.5
39 230.7
40 240.166666666667
41 241.333333333333
42 248.5
43 251.6
44 260.866666666667
45 266.9
46 269.4
47 271.6
48 274.533333333333
49 273.8
50 281.8
51 282.666666666667
52 283
53 289.9
54 285.333333333333
55 285.066666666667
56 288.033333333333
57 285.966666666667
58 290.666666666667
59 294.733333333333
60 290.8
61 296.366666666667
62 301.833333333333
63 298.1
64 302.9
65 300.966666666667
66 305.933333333333
67 303.766666666667
68 305.466666666667
69 313.866666666667
70 312.733333333333
71 317.433333333333
72 320.933333333333
73 325.933333333333
74 322.2
75 325.933333333333
76 328.233333333333
77 321.933333333333
78 330.266666666667
79 331.1
};
\addplot [semithick, color1]
table {%
0 6.63333333333333
1 13.9333333333333
2 15.2333333333333
3 18.7333333333333
4 24.2666666666667
5 40.5
6 57.2666666666667
7 84.8333333333333
8 114.433333333333
9 154.6
10 178.2
11 108.966666666667
12 88.0333333333333
13 96.2
14 101.666666666667
15 104.6
16 110.166666666667
17 114
18 115.866666666667
19 118.7
20 119.233333333333
21 124.5
22 129.733333333333
23 127.666666666667
24 132.966666666667
25 137.233333333333
26 142.8
27 148.133333333333
28 147.566666666667
29 151.8
30 150.5
31 157.9
32 170.1
33 166.4
34 169.166666666667
35 171.633333333333
36 175.966666666667
37 176.366666666667
38 175.866666666667
39 184.3
40 190.133333333333
41 193.7
42 193.4
43 194.633333333333
44 194.066666666667
45 201.333333333333
46 205.1
47 207.733333333333
48 216.233333333333
49 212.133333333333
50 211.066666666667
51 212.5
52 221.1
53 219.966666666667
54 220.266666666667
55 230.1
56 228.666666666667
57 227.7
58 232.4
59 230.2
60 237.7
61 234.566666666667
62 237.666666666667
63 239.966666666667
64 237.766666666667
65 242.966666666667
66 250.9
67 251.033333333333
68 253.866666666667
69 255.433333333333
70 253.766666666667
71 256.7
72 261.9
73 260.2
74 271.5
75 263.233333333333
76 273.1
77 275.7
78 278
79 285.2
};
\end{axis}
\end{tikzpicture}

%% file: pic/new217/217dic10k_2.tex
\begin{tikzpicture}

\definecolor{color0}{rgb}{0.886274509803922,0.290196078431373,0.2}
\definecolor{color1}{rgb}{0.203921568627451,0.541176470588235,0.741176470588235}
\pgfmathsetlengthmacro\MajorTickLength{
      \pgfkeysvalueof{/pgfplots/major tick length} * 0.5
    }
\begin{axis}[
width = 0.4\linewidth,
height = 1in,
major tick length=\MajorTickLength,
font =\fontsize{6}{6.5}\selectfont,
axis line style={white},
legend cell align={left},
yticklabel style={rotate=90},
legend style={fill opacity=0.8, draw opacity=1, text opacity=1, draw=white!80!black, fill=white!89.8039215686275!black},
tick align=inside,
tick pos=left,
x grid style={white},
xlabel={Days},
xmajorgrids,
xmin=0, xmax=80,
xlabel near ticks,
x label style={at={(0.5, -0.2)}},
ymajorgrids,
y label style={at={(0.45,0.5)}},
xtick style={color=white!33.3333333333333!black},
y grid style={white},
xtick={0,20,40,60,80},
xticklabels={0,20,40,60,80},
ymajorgrids,
ytick={0,120,240},
yticklabels={0,120,240},
ymin=-10.3556107367856, ymax=282.017102718771,
ytick style={color=white!33.3333333333333!black}
]
\path [fill=color0, fill opacity=0.3, very thin]
(axis cs:0,9.0976176963403)
--(axis cs:0,4.9023823036597)
--(axis cs:1,7.57031676438777)
--(axis cs:2,11.2089333101874)
--(axis cs:3,13.9528673243906)
--(axis cs:4,18.5464823980388)
--(axis cs:5,29.0559281919021)
--(axis cs:6,38.6557859371173)
--(axis cs:7,61.6970414733331)
--(axis cs:8,86.7417173440584)
--(axis cs:9,117.696278229453)
--(axis cs:10,114.006560173397)
--(axis cs:11,30.3431013263874)
--(axis cs:12,50.3128790064624)
--(axis cs:13,73.7663788611865)
--(axis cs:14,83.2956915096688)
--(axis cs:15,85.8255421492698)
--(axis cs:16,89.916614443523)
--(axis cs:17,88.5503421795194)
--(axis cs:18,90.2518311814022)
--(axis cs:19,86.7114568754518)
--(axis cs:20,93.3735836524981)
--(axis cs:21,83.9564875529815)
--(axis cs:22,86.4124570545459)
--(axis cs:23,88.5425750619299)
--(axis cs:24,81.6660687641515)
--(axis cs:25,82.5535253980684)
--(axis cs:26,86.2494897994599)
--(axis cs:27,89.7147178248388)
--(axis cs:28,88.638856647879)
--(axis cs:29,91.1848208314628)
--(axis cs:30,86.7953335845883)
--(axis cs:31,87.308253640772)
--(axis cs:32,79.5172173575871)
--(axis cs:33,85.4799355420131)
--(axis cs:34,93.2016640989639)
--(axis cs:35,81.7153497646262)
--(axis cs:36,80.3295263091671)
--(axis cs:37,94.2446004292319)
--(axis cs:38,87.966365172236)
--(axis cs:39,87.7828575955097)
--(axis cs:40,84.2494417771148)
--(axis cs:41,85.0571563968227)
--(axis cs:42,82.1667892851682)
--(axis cs:43,88.1760707161321)
--(axis cs:44,82.5419310915328)
--(axis cs:45,77.7579780203413)
--(axis cs:46,77.4359514214496)
--(axis cs:47,74.9130180982857)
--(axis cs:48,80.9824796598362)
--(axis cs:49,71.6115662262344)
--(axis cs:50,72.0902328417332)
--(axis cs:51,73.3367787311257)
--(axis cs:52,69.4445828192482)
--(axis cs:53,77.2170417232721)
--(axis cs:54,73.7881867580645)
--(axis cs:55,69.6240318747035)
--(axis cs:56,71.5921945748259)
--(axis cs:57,72.8259114124611)
--(axis cs:58,63.0724742989314)
--(axis cs:59,70.6542692171947)
--(axis cs:60,72.9521243747401)
--(axis cs:61,69.5427553373818)
--(axis cs:62,72.2594615188184)
--(axis cs:63,77.3440210041571)
--(axis cs:64,73.3603916077232)
--(axis cs:65,69.3103741836603)
--(axis cs:66,66.8839853987799)
--(axis cs:67,82.0627778889733)
--(axis cs:68,75.6336832691261)
--(axis cs:69,81.0860913230064)
--(axis cs:70,78.7282535725652)
--(axis cs:71,75.4817141850221)
--(axis cs:72,72.0752905843114)
--(axis cs:73,82.9436951321854)
--(axis cs:74,85.2725921076669)
--(axis cs:75,81.7204491765835)
--(axis cs:76,88.7101385330445)
--(axis cs:77,75.7078422420598)
--(axis cs:78,82.2002500003906)
--(axis cs:79,88.5145847327704)
--(axis cs:79,240.48541526723)
--(axis cs:79,240.48541526723)
--(axis cs:78,242.199749999609)
--(axis cs:77,226.025491091274)
--(axis cs:76,216.156528133622)
--(axis cs:75,210.746217490083)
--(axis cs:74,216.527407892333)
--(axis cs:73,220.189638201148)
--(axis cs:72,213.791376082355)
--(axis cs:71,222.851619148311)
--(axis cs:70,227.471746427435)
--(axis cs:69,211.713908676994)
--(axis cs:68,212.166316730874)
--(axis cs:67,218.07055544436)
--(axis cs:66,241.11601460122)
--(axis cs:65,233.48962581634)
--(axis cs:64,215.906275058943)
--(axis cs:63,201.255978995843)
--(axis cs:62,214.140538481182)
--(axis cs:61,205.857244662618)
--(axis cs:60,210.114542291927)
--(axis cs:59,207.612397449472)
--(axis cs:58,213.060859034402)
--(axis cs:57,196.640755254206)
--(axis cs:56,198.007805425174)
--(axis cs:55,199.242634791963)
--(axis cs:54,200.211813241936)
--(axis cs:53,202.382958276728)
--(axis cs:52,201.488750514085)
--(axis cs:51,202.329887935541)
--(axis cs:50,208.0431004916)
--(axis cs:49,211.855100440432)
--(axis cs:48,213.417520340164)
--(axis cs:47,203.353648568381)
--(axis cs:46,202.630715245217)
--(axis cs:45,205.775355312992)
--(axis cs:44,196.258068908467)
--(axis cs:43,186.823929283868)
--(axis cs:42,199.966544048165)
--(axis cs:41,185.542843603177)
--(axis cs:40,195.217224889552)
--(axis cs:39,175.950475737824)
--(axis cs:38,180.100301494431)
--(axis cs:37,182.888732904101)
--(axis cs:36,187.7371403575)
--(axis cs:35,185.284650235374)
--(axis cs:34,178.665002567703)
--(axis cs:33,179.25339779132)
--(axis cs:32,170.816115975746)
--(axis cs:31,170.091746359228)
--(axis cs:30,173.471333082078)
--(axis cs:29,170.548512501871)
--(axis cs:28,168.094476685454)
--(axis cs:27,162.551948841828)
--(axis cs:26,159.75051020054)
--(axis cs:25,163.313141268598)
--(axis cs:24,156.000597902515)
--(axis cs:23,139.990758271403)
--(axis cs:22,140.987542945454)
--(axis cs:21,141.710179113685)
--(axis cs:20,138.826416347502)
--(axis cs:19,139.555209791215)
--(axis cs:18,127.414835485264)
--(axis cs:17,141.316324487147)
--(axis cs:16,127.150052223144)
--(axis cs:15,131.041124517397)
--(axis cs:14,124.304308490331)
--(axis cs:13,116.233621138813)
--(axis cs:12,115.953787660204)
--(axis cs:11,165.923565340279)
--(axis cs:10,259.66010649327)
--(axis cs:9,205.770388437214)
--(axis cs:8,142.924949322608)
--(axis cs:7,102.63629186)
--(axis cs:6,71.3442140628827)
--(axis cs:5,47.2774051414312)
--(axis cs:4,32.0535176019612)
--(axis cs:3,26.4471326756094)
--(axis cs:2,19.6577333564793)
--(axis cs:1,18.8296832356122)
--(axis cs:0,9.0976176963403)
--cycle;

\path [fill=color1, fill opacity=0.3, very thin]
(axis cs:0,9.11778730404574)
--(axis cs:0,3.41554602928759)
--(axis cs:1,6.9860690779754)
--(axis cs:2,11.761318920268)
--(axis cs:3,14.8745892177223)
--(axis cs:4,20.5083371949344)
--(axis cs:5,28.5857209056018)
--(axis cs:6,39.914948645955)
--(axis cs:7,60.7497245725922)
--(axis cs:8,88.3982248986832)
--(axis cs:9,130.067710793281)
--(axis cs:10,78.5063168357084)
--(axis cs:11,11.6204217742384)
--(axis cs:12,43.0953425943109)
--(axis cs:13,64.4401023200455)
--(axis cs:14,73.7190337635882)
--(axis cs:15,75.6137639306194)
--(axis cs:16,71.0584477128581)
--(axis cs:17,74.0549255582644)
--(axis cs:18,73.4738388318195)
--(axis cs:19,69.3716607952113)
--(axis cs:20,75.3664993314994)
--(axis cs:21,79.7346870410014)
--(axis cs:22,71.1758358555541)
--(axis cs:23,74.2878011343434)
--(axis cs:24,72.7563601976605)
--(axis cs:25,74.7929091386608)
--(axis cs:26,80.1615708099323)
--(axis cs:27,76.8758307111254)
--(axis cs:28,81.8479543054148)
--(axis cs:29,78.0981284359025)
--(axis cs:30,75.1337923175445)
--(axis cs:31,79.5624327958366)
--(axis cs:32,77.6105936734483)
--(axis cs:33,80.020549384104)
--(axis cs:34,76.4058744752451)
--(axis cs:35,76.6832887019075)
--(axis cs:36,76.0907304574783)
--(axis cs:37,75.1541352951823)
--(axis cs:38,78.3984424242845)
--(axis cs:39,79.4428173331097)
--(axis cs:40,75.1177978889343)
--(axis cs:41,76.6026066695552)
--(axis cs:42,77.019980292801)
--(axis cs:43,83.4775933500043)
--(axis cs:44,81.3740147610343)
--(axis cs:45,80.8013605914424)
--(axis cs:46,79.398653244455)
--(axis cs:47,69.5799202385337)
--(axis cs:48,83.024624759629)
--(axis cs:49,83.9985746115285)
--(axis cs:50,82.6127189460031)
--(axis cs:51,84.4014245447862)
--(axis cs:52,81.9423730099369)
--(axis cs:53,86.1643953104311)
--(axis cs:54,80.4166238689569)
--(axis cs:55,85.7761405018774)
--(axis cs:56,82.8270811978193)
--(axis cs:57,84.5056058883927)
--(axis cs:58,88.6431882100365)
--(axis cs:59,81.7920251549725)
--(axis cs:60,81.162390035685)
--(axis cs:61,87.7683353456231)
--(axis cs:62,92.0806629708559)
--(axis cs:63,92.0830768404106)
--(axis cs:64,95.2816111420868)
--(axis cs:65,91.7066834434375)
--(axis cs:66,94.8172788918853)
--(axis cs:67,97.432522255156)
--(axis cs:68,90.6505965112271)
--(axis cs:69,91.6487484631374)
--(axis cs:70,81.6493340375065)
--(axis cs:71,89.2086977078838)
--(axis cs:72,89.0243997710835)
--(axis cs:73,80.0202534484595)
--(axis cs:74,82.9466071808791)
--(axis cs:75,83.6316576029026)
--(axis cs:76,77.6662037081689)
--(axis cs:77,77.8148943241307)
--(axis cs:78,79.7928584439241)
--(axis cs:79,79.0094817339761)
--(axis cs:79,122.123851599357)
--(axis cs:79,122.123851599357)
--(axis cs:78,119.007141556076)
--(axis cs:77,124.051772342536)
--(axis cs:76,125.667129625164)
--(axis cs:75,120.101675730431)
--(axis cs:74,124.786726152454)
--(axis cs:73,137.713079884874)
--(axis cs:72,128.70893356225)
--(axis cs:71,125.991302292116)
--(axis cs:70,129.61733262916)
--(axis cs:69,135.817918203529)
--(axis cs:68,137.682736822106)
--(axis cs:67,139.900811078177)
--(axis cs:66,133.382721108115)
--(axis cs:65,134.093316556563)
--(axis cs:64,131.251722191247)
--(axis cs:63,142.983589826256)
--(axis cs:62,135.252670362477)
--(axis cs:61,144.36499798771)
--(axis cs:60,139.104276630982)
--(axis cs:59,140.874641511694)
--(axis cs:58,135.42347845663)
--(axis cs:57,136.627727444941)
--(axis cs:56,129.572918802181)
--(axis cs:55,137.957192831456)
--(axis cs:54,125.85004279771)
--(axis cs:53,126.768938022902)
--(axis cs:52,131.590960323396)
--(axis cs:51,131.198575455214)
--(axis cs:50,121.387281053997)
--(axis cs:49,124.534758721805)
--(axis cs:48,125.775375240371)
--(axis cs:47,120.286746428133)
--(axis cs:46,119.001346755545)
--(axis cs:45,119.998639408558)
--(axis cs:44,124.159318572299)
--(axis cs:43,126.989073316662)
--(axis cs:42,124.380019707199)
--(axis cs:41,120.930726663778)
--(axis cs:40,116.615535444399)
--(axis cs:39,121.023849333557)
--(axis cs:38,121.201557575716)
--(axis cs:37,114.445864704818)
--(axis cs:36,115.775936209188)
--(axis cs:35,120.250044631426)
--(axis cs:34,121.794125524755)
--(axis cs:33,119.579450615896)
--(axis cs:32,117.456072993218)
--(axis cs:31,112.370900537497)
--(axis cs:30,113.866207682456)
--(axis cs:29,115.501871564098)
--(axis cs:28,117.018712361252)
--(axis cs:27,108.390835955541)
--(axis cs:26,110.038429190068)
--(axis cs:25,112.407090861339)
--(axis cs:24,106.510306469006)
--(axis cs:23,99.7121988656566)
--(axis cs:22,104.490830811113)
--(axis cs:21,105.998646292332)
--(axis cs:20,108.566834001834)
--(axis cs:19,102.961672538122)
--(axis cs:18,113.12616116818)
--(axis cs:17,102.078407775069)
--(axis cs:16,100.808218953809)
--(axis cs:15,111.119569402714)
--(axis cs:14,104.814299569745)
--(axis cs:13,114.826564346621)
--(axis cs:12,93.4379907390224)
--(axis cs:11,142.179578225762)
--(axis cs:10,261.960349830958)
--(axis cs:9,215.198955873386)
--(axis cs:8,156.001775101317)
--(axis cs:7,104.116942094074)
--(axis cs:6,70.1517180207117)
--(axis cs:5,59.4142790943982)
--(axis cs:4,38.4916628050656)
--(axis cs:3,25.5254107822777)
--(axis cs:2,19.0386810797321)
--(axis cs:1,15.6139309220246)
--(axis cs:0,9.11778730404574)
--cycle;

\addplot [semithick, color0]
table {%
0 7
1 13.2
2 15.4333333333333
3 20.2
4 25.3
5 38.1666666666667
6 55
7 82.1666666666667
8 114.833333333333
9 161.733333333333
10 186.833333333333
11 98.1333333333333
12 83.1333333333333
13 95
14 103.8
15 108.433333333333
16 108.533333333333
17 114.933333333333
18 108.833333333333
19 113.133333333333
20 116.1
21 112.833333333333
22 113.7
23 114.266666666667
24 118.833333333333
25 122.933333333333
26 123
27 126.133333333333
28 128.366666666667
29 130.866666666667
30 130.133333333333
31 128.7
32 125.166666666667
33 132.366666666667
34 135.933333333333
35 133.5
36 134.033333333333
37 138.566666666667
38 134.033333333333
39 131.866666666667
40 139.733333333333
41 135.3
42 141.066666666667
43 137.5
44 139.4
45 141.766666666667
46 140.033333333333
47 139.133333333333
48 147.2
49 141.733333333333
50 140.066666666667
51 137.833333333333
52 135.466666666667
53 139.8
54 137
55 134.433333333333
56 134.8
57 134.733333333333
58 138.066666666667
59 139.133333333333
60 141.533333333333
61 137.7
62 143.2
63 139.3
64 144.633333333333
65 151.4
66 154
67 150.066666666667
68 143.9
69 146.4
70 153.1
71 149.166666666667
72 142.933333333333
73 151.566666666667
74 150.9
75 146.233333333333
76 152.433333333333
77 150.866666666667
78 162.2
79 164.5
};
\addplot [semithick, color1]
table {%
0 6.26666666666667
1 11.3
2 15.4
3 20.2
4 29.5
5 44
6 55.0333333333333
7 82.4333333333333
8 122.2
9 172.633333333333
10 170.233333333333
11 76.9
12 68.2666666666667
13 89.6333333333333
14 89.2666666666667
15 93.3666666666667
16 85.9333333333333
17 88.0666666666667
18 93.3
19 86.1666666666667
20 91.9666666666667
21 92.8666666666667
22 87.8333333333333
23 87
24 89.6333333333333
25 93.6
26 95.1
27 92.6333333333333
28 99.4333333333333
29 96.8
30 94.5
31 95.9666666666667
32 97.5333333333333
33 99.8
34 99.1
35 98.4666666666667
36 95.9333333333333
37 94.8
38 99.8
39 100.233333333333
40 95.8666666666667
41 98.7666666666667
42 100.7
43 105.233333333333
44 102.766666666667
45 100.4
46 99.2
47 94.9333333333333
48 104.4
49 104.266666666667
50 102
51 107.8
52 106.766666666667
53 106.466666666667
54 103.133333333333
55 111.866666666667
56 106.2
57 110.566666666667
58 112.033333333333
59 111.333333333333
60 110.133333333333
61 116.066666666667
62 113.666666666667
63 117.533333333333
64 113.266666666667
65 112.9
66 114.1
67 118.666666666667
68 114.166666666667
69 113.733333333333
70 105.633333333333
71 107.6
72 108.866666666667
73 108.866666666667
74 103.866666666667
75 101.866666666667
76 101.666666666667
77 100.933333333333
78 99.4
79 100.566666666667
};
\end{axis}

\end{tikzpicture}

%% file: pic/new217/217dic25k.tex
\begin{tikzpicture}

\definecolor{color0}{rgb}{0.886274509803922,0.290196078431373,0.2}
\definecolor{color1}{rgb}{0.203921568627451,0.541176470588235,0.741176470588235}
\pgfmathsetlengthmacro\MajorTickLength{
      \pgfkeysvalueof{/pgfplots/major tick length} * 0.5
    }
\begin{axis}[
width = 0.4\linewidth,
height = 1in,
yticklabel style={rotate=90},
major tick length=\MajorTickLength,
font =\fontsize{6}{6.5}\selectfont,
axis line style={white},
legend style={inner xsep=1pt, inner ysep=-1pt, row sep=-3pt,at={(0.5,1.2)},anchor=north west},
tick align=inside,
tick pos=left,
x grid style={white},
xlabel={Days},
xmajorgrids,
xmin=0, xmax=80,
xlabel near ticks,
x label style={at={(0.5, -0.15)}},
y label style={at={(0.55,0.5)}},
xtick style={color=white!33.3333333333333!black},
y grid style={white},
xtick={0,20,40,60,80},
xticklabels={0,20,40,60,80},
ymajorgrids,
ytick={0,120,240},
yticklabels={0,120,240},
ymin=-9.99603369486922, ymax=288.496650993458,
ytick style={color=white!33.3333333333333!black}
]
\path [fill=color0, fill opacity=0.3, very thin]
(axis cs:0,9.52895703913776)
--(axis cs:0,5.47104296086224)
--(axis cs:1,9.13433595297262)
--(axis cs:2,9.4920014504221)
--(axis cs:3,12.8401850041959)
--(axis cs:4,17.2429569625934)
--(axis cs:5,31.0163165310258)
--(axis cs:6,43.2329773499693)
--(axis cs:7,56.5105323067535)
--(axis cs:8,82.9580808962198)
--(axis cs:9,115.986551148549)
--(axis cs:10,108.204531643891)
--(axis cs:11,14.0691346442752)
--(axis cs:12,52.3210684699719)
--(axis cs:13,57.5267179233782)
--(axis cs:14,65.1852807650334)
--(axis cs:15,68.4836656086851)
--(axis cs:16,66.1951840498092)
--(axis cs:17,63.8506247218883)
--(axis cs:18,62.7063486567296)
--(axis cs:19,60.6266597796162)
--(axis cs:20,61.0987673258154)
--(axis cs:21,59.1811366884008)
--(axis cs:22,60.0425270007077)
--(axis cs:23,56.9077688631281)
--(axis cs:24,59.3295052471182)
--(axis cs:25,59.3602721127516)
--(axis cs:26,57.8856657692354)
--(axis cs:27,54.6558226523789)
--(axis cs:28,56.9500630359847)
--(axis cs:29,58.4425270007077)
--(axis cs:30,58.2423969929716)
--(axis cs:31,59.386676528605)
--(axis cs:32,57.715529793265)
--(axis cs:33,59.7142443064965)
--(axis cs:34,58.8872421236545)
--(axis cs:35,59.7585450030306)
--(axis cs:36,58.153530485213)
--(axis cs:37,61.9740204872516)
--(axis cs:38,59.8824327163443)
--(axis cs:39,60.2494306539023)
--(axis cs:40,59.6562008691157)
--(axis cs:41,61.4630042157597)
--(axis cs:42,62.6488071157115)
--(axis cs:43,61.7683436868979)
--(axis cs:44,65.2487651253374)
--(axis cs:45,58.9461235234856)
--(axis cs:46,63.4200812003143)
--(axis cs:47,63.1103142577932)
--(axis cs:48,64.4934898159328)
--(axis cs:49,62.5994467591794)
--(axis cs:50,61.6590658478129)
--(axis cs:51,61.2774875906653)
--(axis cs:52,62.4175329860613)
--(axis cs:53,64.1904399677184)
--(axis cs:54,68.0202490780549)
--(axis cs:55,64.4253186885639)
--(axis cs:56,61.7172885674008)
--(axis cs:57,61.5249007509417)
--(axis cs:58,67.0140478217574)
--(axis cs:59,62.7883372569749)
--(axis cs:60,63.3927288748261)
--(axis cs:61,64.8191131019102)
--(axis cs:62,61.1023292463052)
--(axis cs:63,64.9180424712077)
--(axis cs:64,63.4858800621741)
--(axis cs:65,62.8787221410582)
--(axis cs:66,60.5992235314091)
--(axis cs:67,60.0692230542333)
--(axis cs:68,57.0650280814244)
--(axis cs:69,61.3933871110509)
--(axis cs:70,60.5069474499847)
--(axis cs:71,60.6673409272896)
--(axis cs:72,60.4856719577068)
--(axis cs:73,60.8252441675358)
--(axis cs:74,60.5574941617999)
--(axis cs:75,65.3827293063186)
--(axis cs:76,64.457440598124)
--(axis cs:77,62.4547411763805)
--(axis cs:78,61.479024015918)
--(axis cs:79,58.6212860272763)
--(axis cs:79,116.64538063939)
--(axis cs:79,116.64538063939)
--(axis cs:78,116.854309317415)
--(axis cs:77,116.878592156953)
--(axis cs:76,111.675892735209)
--(axis cs:75,112.750604027015)
--(axis cs:74,112.775839171533)
--(axis cs:73,113.441422499131)
--(axis cs:72,111.914328042293)
--(axis cs:71,113.399325739377)
--(axis cs:70,107.359719216682)
--(axis cs:69,111.806612888949)
--(axis cs:68,108.601638585242)
--(axis cs:67,103.530776945767)
--(axis cs:66,100.667443135258)
--(axis cs:65,97.5212778589418)
--(axis cs:64,96.3141199378259)
--(axis cs:63,96.6152908621257)
--(axis cs:62,97.8310040870281)
--(axis cs:61,99.4475535647565)
--(axis cs:60,96.4072711251739)
--(axis cs:59,97.4783294096918)
--(axis cs:58,100.852618844909)
--(axis cs:57,100.008432582392)
--(axis cs:56,99.6827114325992)
--(axis cs:55,99.5080146447694)
--(axis cs:54,105.313084255278)
--(axis cs:53,96.7428933656149)
--(axis cs:52,98.6491336806054)
--(axis cs:51,96.5891790760014)
--(axis cs:50,99.6742674855204)
--(axis cs:49,94.733886574154)
--(axis cs:48,97.4398435174006)
--(axis cs:47,95.9563524088735)
--(axis cs:46,95.7132521330191)
--(axis cs:45,92.9872098098478)
--(axis cs:44,93.8179015413293)
--(axis cs:43,94.2316563131021)
--(axis cs:42,95.0178595509552)
--(axis cs:41,85.7369957842403)
--(axis cs:40,88.5437991308843)
--(axis cs:39,87.0839026794311)
--(axis cs:38,89.1842339503223)
--(axis cs:37,86.2926461794151)
--(axis cs:36,84.9798028481204)
--(axis cs:35,83.7081216636361)
--(axis cs:34,81.9794245430121)
--(axis cs:33,81.6190890268368)
--(axis cs:32,79.9511368734017)
--(axis cs:31,77.413323471395)
--(axis cs:30,81.1576030070284)
--(axis cs:29,75.9574729992923)
--(axis cs:28,78.9832702973487)
--(axis cs:27,80.0775106809544)
--(axis cs:26,76.5810008974313)
--(axis cs:25,78.306394553915)
--(axis cs:24,79.0704947528818)
--(axis cs:23,83.3588978035386)
--(axis cs:22,77.5574729992923)
--(axis cs:21,81.8188633115992)
--(axis cs:20,83.6345660075179)
--(axis cs:19,86.4400068870505)
--(axis cs:18,85.6269846766038)
--(axis cs:17,87.4160419447784)
--(axis cs:16,89.2714826168575)
--(axis cs:15,89.3830010579816)
--(axis cs:14,88.1480525682999)
--(axis cs:13,84.4066154099552)
--(axis cs:12,76.6122648633614)
--(axis cs:11,150.930865355725)
--(axis cs:10,274.928801689443)
--(axis cs:9,214.480115518117)
--(axis cs:8,155.308585770447)
--(axis cs:7,104.289467693246)
--(axis cs:6,77.5670226500307)
--(axis cs:5,58.7836834689742)
--(axis cs:4,38.8237097040732)
--(axis cs:3,26.2931483291374)
--(axis cs:2,22.9746652162446)
--(axis cs:1,17.9989973803607)
--(axis cs:0,9.52895703913776)
--cycle;

\path [fill=color1, fill opacity=0.3, very thin]
(axis cs:0,8.9615177241877)
--(axis cs:0,3.57181560914563)
--(axis cs:1,8.31754580810136)
--(axis cs:2,10.1770794277827)
--(axis cs:3,11.419679698375)
--(axis cs:4,16.1144615741542)
--(axis cs:5,26.8056156948787)
--(axis cs:6,33.0303748268721)
--(axis cs:7,47.7856361776929)
--(axis cs:8,71.6922898473192)
--(axis cs:9,102.252955817361)
--(axis cs:10,103.473765500115)
--(axis cs:11,23.415811511877)
--(axis cs:12,6.63160768960211)
--(axis cs:13,27.3399900403639)
--(axis cs:14,58.5645399872579)
--(axis cs:15,63.6973121457529)
--(axis cs:16,64.9358589482865)
--(axis cs:17,63.3940491999414)
--(axis cs:18,64.5404440685062)
--(axis cs:19,62.7147415247239)
--(axis cs:20,62.7011575004496)
--(axis cs:21,57.8968233896754)
--(axis cs:22,61.2429512658796)
--(axis cs:23,57.4963186173979)
--(axis cs:24,55.281079576068)
--(axis cs:25,51.2388746571528)
--(axis cs:26,51.5569261635653)
--(axis cs:27,52.1905818510822)
--(axis cs:28,49.8732041856715)
--(axis cs:29,49.630683123147)
--(axis cs:30,52.8040534340068)
--(axis cs:31,50.8705774078449)
--(axis cs:32,46.4236067282882)
--(axis cs:33,46.899351316303)
--(axis cs:34,47.7935917179205)
--(axis cs:35,44.4572900118863)
--(axis cs:36,44.1376743472205)
--(axis cs:37,44.0987673258154)
--(axis cs:38,46.1500662429226)
--(axis cs:39,44.7962672240113)
--(axis cs:40,44.38442890597)
--(axis cs:41,44.6967823879791)
--(axis cs:42,43.3509291467686)
--(axis cs:43,40.8920638896343)
--(axis cs:44,42.9569445292259)
--(axis cs:45,40.8401061213822)
--(axis cs:46,41.2928706497503)
--(axis cs:47,42.0729310766944)
--(axis cs:48,42.680293044225)
--(axis cs:49,41.1836339625464)
--(axis cs:50,43.3683334677636)
--(axis cs:51,42.7386937760403)
--(axis cs:52,42.6215302985891)
--(axis cs:53,44.221764875533)
--(axis cs:54,41.9630388588494)
--(axis cs:55,41.6451185240297)
--(axis cs:56,39.3839125438268)
--(axis cs:57,40.8051689942572)
--(axis cs:58,38.5947296511209)
--(axis cs:59,42.3208512498699)
--(axis cs:60,39.1958707723255)
--(axis cs:61,39.5189661174844)
--(axis cs:62,39.8008795380927)
--(axis cs:63,40.6981184019343)
--(axis cs:64,38.4536329620909)
--(axis cs:65,40.6017255952906)
--(axis cs:66,39.6124622709376)
--(axis cs:67,38.5825392486611)
--(axis cs:68,35.9509855211014)
--(axis cs:69,37.5569312037457)
--(axis cs:70,39.0840549809687)
--(axis cs:71,37.1696873439608)
--(axis cs:72,38.3151751340551)
--(axis cs:73,35.7893777856003)
--(axis cs:74,37.1454722561778)
--(axis cs:75,35.7282727060359)
--(axis cs:76,35.4029898372651)
--(axis cs:77,35.5872139629221)
--(axis cs:78,38.3204461270471)
--(axis cs:79,37.324280786641)
--(axis cs:79,65.8090525466924)
--(axis cs:79,65.8090525466924)
--(axis cs:78,66.6128872062863)
--(axis cs:77,65.1461193704112)
--(axis cs:76,65.1303434960682)
--(axis cs:75,66.4717272939641)
--(axis cs:74,68.2545277438222)
--(axis cs:73,66.3439555477331)
--(axis cs:72,69.4181581992782)
--(axis cs:71,67.6969793227059)
--(axis cs:70,67.182611685698)
--(axis cs:69,72.3764021295877)
--(axis cs:68,71.1823478122319)
--(axis cs:67,67.6174607513389)
--(axis cs:66,64.9208710623957)
--(axis cs:65,68.9316077380427)
--(axis cs:64,65.7463670379092)
--(axis cs:63,69.635214931399)
--(axis cs:62,65.0657871285739)
--(axis cs:61,64.2810338825156)
--(axis cs:60,69.3374625610078)
--(axis cs:59,66.5458154167967)
--(axis cs:58,67.2719370155457)
--(axis cs:57,66.5948310057428)
--(axis cs:56,62.8827541228399)
--(axis cs:55,64.3548814759703)
--(axis cs:54,60.0369611411506)
--(axis cs:53,62.578235124467)
--(axis cs:52,61.1784697014109)
--(axis cs:51,63.4613062239597)
--(axis cs:50,64.0316665322364)
--(axis cs:49,64.2830327041202)
--(axis cs:48,64.519706955775)
--(axis cs:47,64.0604022566389)
--(axis cs:46,68.5737960169164)
--(axis cs:45,64.2932272119511)
--(axis cs:44,65.7097221374408)
--(axis cs:43,66.241269443699)
--(axis cs:42,62.7824041865648)
--(axis cs:41,67.8365509453543)
--(axis cs:40,66.9489044273633)
--(axis cs:39,69.8037327759887)
--(axis cs:38,67.1166004237441)
--(axis cs:37,66.6345660075179)
--(axis cs:36,69.7289923194461)
--(axis cs:35,75.276043321447)
--(axis cs:34,68.6064082820795)
--(axis cs:33,69.1673153503637)
--(axis cs:32,67.1097266050451)
--(axis cs:31,73.5960892588217)
--(axis cs:30,74.7292798993265)
--(axis cs:29,74.369316876853)
--(axis cs:28,77.0601291476618)
--(axis cs:27,71.0760848155844)
--(axis cs:26,76.776407169768)
--(axis cs:25,75.0944586761805)
--(axis cs:24,76.8522537572653)
--(axis cs:23,74.7036813826021)
--(axis cs:22,80.7570487341204)
--(axis cs:21,79.2365099436579)
--(axis cs:20,77.0988424995504)
--(axis cs:19,81.6185918086094)
--(axis cs:18,87.6595559314938)
--(axis cs:17,85.8059508000586)
--(axis cs:16,82.5308077183802)
--(axis cs:15,86.4360211875804)
--(axis cs:14,87.3021266794088)
--(axis cs:13,112.126676626303)
--(axis cs:12,106.435058977065)
--(axis cs:11,225.784188488123)
--(axis cs:10,239.326234499885)
--(axis cs:9,186.547044182639)
--(axis cs:8,130.641043486014)
--(axis cs:7,89.8810304889738)
--(axis cs:6,62.7029585064613)
--(axis cs:5,45.7943843051213)
--(axis cs:4,32.6188717591791)
--(axis cs:3,24.0469869682916)
--(axis cs:2,17.9562539055506)
--(axis cs:1,16.215787525232)
--(axis cs:0,8.9615177241877)
--cycle;

\addplot [semithick, color0]
table {%
0 7.5
1 13.5666666666667
2 16.2333333333333
3 19.5666666666667
4 28.0333333333333
5 44.9
6 60.4
7 80.4
8 119.133333333333
9 165.233333333333
10 191.566666666667
11 82.5
12 64.4666666666667
13 70.9666666666667
14 76.6666666666667
15 78.9333333333333
16 77.7333333333333
17 75.6333333333333
18 74.1666666666667
19 73.5333333333333
20 72.3666666666667
21 70.5
22 68.8
23 70.1333333333333
24 69.2
25 68.8333333333333
26 67.2333333333333
27 67.3666666666667
28 67.9666666666667
29 67.2
30 69.7
31 68.4
32 68.8333333333333
33 70.6666666666667
34 70.4333333333333
35 71.7333333333333
36 71.5666666666667
37 74.1333333333333
38 74.5333333333333
39 73.6666666666667
40 74.1
41 73.6
42 78.8333333333333
43 78
44 79.5333333333333
45 75.9666666666667
46 79.5666666666667
47 79.5333333333333
48 80.9666666666667
49 78.6666666666667
50 80.6666666666667
51 78.9333333333333
52 80.5333333333333
53 80.4666666666667
54 86.6666666666667
55 81.9666666666667
56 80.7
57 80.7666666666667
58 83.9333333333333
59 80.1333333333333
60 79.9
61 82.1333333333333
62 79.4666666666667
63 80.7666666666667
64 79.9
65 80.2
66 80.6333333333333
67 81.8
68 82.8333333333333
69 86.6
70 83.9333333333333
71 87.0333333333333
72 86.2
73 87.1333333333333
74 86.6666666666667
75 89.0666666666667
76 88.0666666666667
77 89.6666666666667
78 89.1666666666667
79 87.6333333333333
};
\addlegendentry{noid}
\addplot [semithick, color1]
table {%
0 6.26666666666667
1 12.2666666666667
2 14.0666666666667
3 17.7333333333333
4 24.3666666666667
5 36.3
6 47.8666666666667
7 68.8333333333333
8 101.166666666667
9 144.4
10 171.4
11 124.6
12 56.5333333333333
13 69.7333333333333
14 72.9333333333333
15 75.0666666666667
16 73.7333333333333
17 74.6
18 76.1
19 72.1666666666667
20 69.9
21 68.5666666666667
22 71
23 66.1
24 66.0666666666667
25 63.1666666666667
26 64.1666666666667
27 61.6333333333333
28 63.4666666666667
29 62
30 63.7666666666667
31 62.2333333333333
32 56.7666666666667
33 58.0333333333333
34 58.2
35 59.8666666666667
36 56.9333333333333
37 55.3666666666667
38 56.6333333333333
39 57.3
40 55.6666666666667
41 56.2666666666667
42 53.0666666666667
43 53.5666666666667
44 54.3333333333333
45 52.5666666666667
46 54.9333333333333
47 53.0666666666667
48 53.6
49 52.7333333333333
50 53.7
51 53.1
52 51.9
53 53.4
54 51
55 53
56 51.1333333333333
57 53.7
58 52.9333333333333
59 54.4333333333333
60 54.2666666666667
61 51.9
62 52.4333333333333
63 55.1666666666667
64 52.1
65 54.7666666666667
66 52.2666666666667
67 53.1
68 53.5666666666667
69 54.9666666666667
70 53.1333333333333
71 52.4333333333333
72 53.8666666666667
73 51.0666666666667
74 52.7
75 51.1
76 50.2666666666667
77 50.3666666666667
78 52.4666666666667
79 51.5666666666667
};
\addlegendentry{id}
\end{axis}

\end{tikzpicture}

%% file: pic/new217/217act1k.tex
\begin{tikzpicture}

\definecolor{color0}{rgb}{0.111111111111111,0,0.888888888888889}
\definecolor{color1}{rgb}{0.222222222222222,0,0.777777777777778}
\definecolor{color2}{rgb}{0.333333333333333,0,0.666666666666667}
\definecolor{color3}{rgb}{0.444444444444444,0,0.555555555555556}
\definecolor{color4}{rgb}{0.555555555555556,0,0.444444444444444}
\definecolor{color5}{rgb}{0.666666666666667,0,0.333333333333333}
\definecolor{color6}{rgb}{0.777777777777778,0,0.222222222222222}
\definecolor{color7}{rgb}{0.888888888888889,0,0.111111111111111}
\definecolor{color8}{rgb}{0.501960784313725,0,0.501960784313725}
\pgfmathsetlengthmacro\MajorTickLength{
      \pgfkeysvalueof{/pgfplots/major tick length} * 0.5
    }
\begin{axis}[
axis line style={white},
legend cell align={left},
width = 0.4\linewidth,
height = 1in,
font= \fontsize{6}{6.5}\selectfont,
major tick length=\MajorTickLength,
tick align=inside,
tick pos=left,
x grid style={white},
xlabel style={},
xmajorgrids,
xmin=0, xmax=9,
yticklabel style={rotate=90},
xtick style={color=white!33.3333333333333!black},
xtick={0,1,2,3,4,5,6,7,8,9},
xticklabels={0,.1,.2,.5,1,2,5,10,20},
y grid style={white},
ylabel={Full activity Lvl.},
xlabel near ticks,
x label style={at={(0.5, -0.15)}},
ymajorgrids,
y label style={at={(0.45,0.5)}},
ymin=0, ymax=1,
ytick style={font=\tiny,color=white!33.3333333333333!black},
ytick={0,.3,.6,.9},
yticklabels={0,.3,.6,.9}
]
\draw[draw=none,fill=blue,fill opacity=0.6,very thin] (axis cs:0,0) rectangle (axis cs:1,0.856208799929876);
\draw[draw=none,fill=color0,fill opacity=0.6,very thin] (axis cs:1,0) rectangle (axis cs:2,0.853883846900156);
\draw[draw=none,fill=color1,fill opacity=0.6,very thin] (axis cs:2,0) rectangle (axis cs:3,0.851230111563482);
\draw[draw=none,fill=color2,fill opacity=0.6,very thin] (axis cs:3,0) rectangle (axis cs:4,0.847820299620167);
\draw[draw=none,fill=color3,fill opacity=0.6,very thin] (axis cs:4,0) rectangle (axis cs:5,0.84171620030187);
\draw[draw=none,fill=color4,fill opacity=0.6,very thin] (axis cs:5,0) rectangle (axis cs:6,0.827436407714261);
\draw[draw=none,fill=color5,fill opacity=0.6,very thin] (axis cs:6,0) rectangle (axis cs:7,0.779943289490944);
\draw[draw=none,fill=color6,fill opacity=0.6,very thin] (axis cs:7,0) rectangle (axis cs:8,0.690885324488657);
\draw[draw=none,fill=color7,fill opacity=0.6,very thin] (axis cs:8,0) rectangle (axis cs:9,0.542865837025966);
\addplot [thin, black, dashed,dash pattern=on 2pt off 1pt]
table {%
-0.45 0.847841130952393
9.45 0.847841130952393
};
\addplot [thin, color8, dashed, dash pattern=on 2pt off 1pt]
table {%
-0.45 0.854473573931597
9.45 0.854473573931597
};
\end{axis}

\end{tikzpicture}

%% file: pic/new217/217act3k.tex
\begin{tikzpicture}

\definecolor{color0}{rgb}{0.111111111111111,0,0.888888888888889}
\definecolor{color1}{rgb}{0.222222222222222,0,0.777777777777778}
\definecolor{color2}{rgb}{0.333333333333333,0,0.666666666666667}
\definecolor{color3}{rgb}{0.444444444444444,0,0.555555555555556}
\definecolor{color4}{rgb}{0.555555555555556,0,0.444444444444444}
\definecolor{color5}{rgb}{0.666666666666667,0,0.333333333333333}
\definecolor{color6}{rgb}{0.777777777777778,0,0.222222222222222}
\definecolor{color7}{rgb}{0.888888888888889,0,0.111111111111111}
\definecolor{color8}{rgb}{0.501960784313725,0,0.501960784313725}
\pgfmathsetlengthmacro\MajorTickLength{
      \pgfkeysvalueof{/pgfplots/major tick length} * 0.5
    }
\begin{axis}[
axis line style={white},
legend cell align={left},
width = 0.4\linewidth,
height = 1in,
yticklabel style={rotate=90},
font =\fontsize{6}{6.5}\selectfont,
major tick length=\MajorTickLength,
tick align=inside,
tick pos=left,
x grid style={white},
xmajorgrids,
xmin=0, xmax=9,
xtick style={color=white!33.3333333333333!black},
xtick={0,1,2,3,4,5,6,7,8,9},
xticklabels={0,.1,.2,.5,1,2,5,10,20},
y grid style={white},
xlabel near ticks,
x label style={at={(0.5, -0.15)}},
ymajorgrids,
y label style={at={(0.55,0.5)}},
ymin=0, ymax=1,
ytick style={color=white!33.3333333333333!black},
ytick={-1,2},
]
\draw[draw=none,fill=blue,fill opacity=0.6,very thin] (axis cs:0,0) rectangle (axis cs:1,0.852105473156729);
\draw[draw=none,fill=color0,fill opacity=0.6,very thin] (axis cs:1,0) rectangle (axis cs:2,0.850185858158674);
\draw[draw=none,fill=color1,fill opacity=0.6,very thin] (axis cs:2,0) rectangle (axis cs:3,0.848016553895397);
\draw[draw=none,fill=color2,fill opacity=0.6,very thin] (axis cs:3,0) rectangle (axis cs:4,0.843365960162074);
\draw[draw=none,fill=color3,fill opacity=0.6,very thin] (axis cs:4,0) rectangle (axis cs:5,0.83538864405421);
\draw[draw=none,fill=color4,fill opacity=0.6,very thin] (axis cs:5,0) rectangle (axis cs:6,0.817933846144222);
\draw[draw=none,fill=color5,fill opacity=0.6,very thin] (axis cs:6,0) rectangle (axis cs:7,0.791429223491308);
\draw[draw=none,fill=color6,fill opacity=0.6,very thin] (axis cs:7,0) rectangle (axis cs:8,0.752786590109089);
\draw[draw=none,fill=color7,fill opacity=0.6,very thin] (axis cs:8,0) rectangle (axis cs:9,0.693246748785878);
\addplot [thin, black, dashed,dash pattern=on 2pt off 1pt]
table {%
-0.45 0.848620610495095
9.45 0.848620610495095
};
\addplot [thin, color8, dashed,dash pattern=on 2pt off 1pt]
table {%
-0.45 0.849684538062079
9.45 0.849684538062079
};
\end{axis}
\end{tikzpicture}

%% file: pic/new217/217act10k_2.tex
\begin{tikzpicture}

\definecolor{color0}{rgb}{0.111111111111111,0,0.888888888888889}
\definecolor{color1}{rgb}{0.222222222222222,0,0.777777777777778}
\definecolor{color2}{rgb}{0.333333333333333,0,0.666666666666667}
\definecolor{color3}{rgb}{0.444444444444444,0,0.555555555555556}
\definecolor{color4}{rgb}{0.555555555555556,0,0.444444444444444}
\definecolor{color5}{rgb}{0.666666666666667,0,0.333333333333333}
\definecolor{color6}{rgb}{0.777777777777778,0,0.222222222222222}
\definecolor{color7}{rgb}{0.888888888888889,0,0.111111111111111}
\definecolor{color8}{rgb}{0.501960784313725,0,0.501960784313725}
\pgfmathsetlengthmacro\MajorTickLength{
      \pgfkeysvalueof{/pgfplots/major tick length} * 0.5
    }
\begin{axis}[
axis line style={white},
legend cell align={left},
width = 0.4\linewidth,
height = 1in,
font = \fontsize{6}{6.5}\selectfont,
yticklabel style={rotate=90},
major tick length=\MajorTickLength,
tick align=inside,
tick pos=left,
x grid style={white},
xmajorgrids,
xmin=0, xmax=9,
xtick style={color=white!33.3333333333333!black},
xtick={0,1,2,3,4,5,6,7,8,9},
xticklabels={0,.1,.2,.5,1,2,5,10,20},
y grid style={white},
xlabel near ticks,
x label style={at={(0.5, -0.15)}},
ymajorgrids,
y label style={at={(0.55,0.5)}},
ymin=0, ymax=1,
ytick style={color=white!33.3333333333333!black},
ytick={-1,2},
]
\draw[draw=none,fill=blue,fill opacity=0.6,very thin] (axis cs:0,0) rectangle (axis cs:1,0.830338850054117);
\draw[draw=none,fill=color0,fill opacity=0.6,very thin] (axis cs:1,0) rectangle (axis cs:2,0.825204965043571);
\draw[draw=none,fill=color1,fill opacity=0.6,very thin] (axis cs:2,0) rectangle (axis cs:3,0.819722434985597);
\draw[draw=none,fill=color2,fill opacity=0.6,very thin] (axis cs:3,0) rectangle (axis cs:4,0.809971120850772);
\draw[draw=none,fill=color3,fill opacity=0.6,very thin] (axis cs:4,0) rectangle (axis cs:5,0.791759743765616);
\draw[draw=none,fill=color4,fill opacity=0.6,very thin] (axis cs:5,0) rectangle (axis cs:6,0.7376244110691);
\draw[draw=none,fill=color5,fill opacity=0.6,very thin] (axis cs:6,0) rectangle (axis cs:7,0.603768430039754);
\draw[draw=none,fill=color6,fill opacity=0.6,very thin] (axis cs:7,0) rectangle (axis cs:8,0.438245925564452);
\draw[draw=none,fill=color7,fill opacity=0.6,very thin] (axis cs:8,0) rectangle (axis cs:9,0.352977832004563);
\addplot [thin, black, dashed,dash pattern=on 2pt off 1pt]
table {%
-0.45 0.766117643183739
9.45 0.766117643183739
};
\addplot [thin, color8, dashed,dash pattern=on 2pt off 1pt]
table {%
-0.45 0.817239366239485
9.45 0.817239366239485
};
\end{axis}

\end{tikzpicture}

%% file: pic/new217/217act25k.tex
\begin{tikzpicture}

\definecolor{color0}{rgb}{0.111111111111111,0,0.888888888888889}
\definecolor{color1}{rgb}{0.222222222222222,0,0.777777777777778}
\definecolor{color2}{rgb}{0.333333333333333,0,0.666666666666667}
\definecolor{color3}{rgb}{0.444444444444444,0,0.555555555555556}
\definecolor{color4}{rgb}{0.555555555555556,0,0.444444444444444}
\definecolor{color5}{rgb}{0.666666666666667,0,0.333333333333333}
\definecolor{color6}{rgb}{0.777777777777778,0,0.222222222222222}
\definecolor{color7}{rgb}{0.888888888888889,0,0.111111111111111}
\definecolor{color8}{rgb}{0.501960784313725,0,0.501960784313725}
\pgfmathsetlengthmacro\MajorTickLength{
      \pgfkeysvalueof{/pgfplots/major tick length} * 0.5
    }
\begin{axis}[
axis line style={white},
legend cell align={left},
width = 0.4\linewidth,
height = 1in,
font= \fontsize{6}{6.5}\selectfont,
major tick length=\MajorTickLength,
legend style={inner xsep=1pt, inner ysep=-1pt, row sep=-3pt,at={(0,1.15)},anchor=north west}, 
tick align=inside,
tick pos=left,
yticklabel style={rotate=90},
x grid style={white},
xmajorgrids,
xmin=-0.45, xmax=9.45,
xtick style={color=white!33.3333333333333!black},
xtick={0,1,2,3,4,5,6,7,8,9},
xticklabels={0,.1,.2,.5,1,2,5,10,20},
y grid style={white},
xlabel near ticks,
x label style={at={(0.5, -0.15)}},
ymajorgrids,
y label style={at={(0.55,0.5)}},
ymin=0, ymax=1,
ytick style={color=white!33.3333333333333!black},
ytick={-1,2},
]
\draw[draw=none,fill=blue,fill opacity=0.6,very thin] (axis cs:0,0) rectangle (axis cs:1,0.824364070394453);
\draw[draw=none,fill=color0,fill opacity=0.6,very thin] (axis cs:1,0) rectangle (axis cs:2,0.81111290251211);
\draw[draw=none,fill=color1,fill opacity=0.6,very thin] (axis cs:2,0) rectangle (axis cs:3,0.795071239630559);
\draw[draw=none,fill=color2,fill opacity=0.6,very thin] (axis cs:3,0) rectangle (axis cs:4,0.75281727836145);
\draw[draw=none,fill=color3,fill opacity=0.6,very thin] (axis cs:4,0) rectangle (axis cs:5,0.677664480375168);
\draw[draw=none,fill=color4,fill opacity=0.6,very thin] (axis cs:5,0) rectangle (axis cs:6,0.564437821069465);
\draw[draw=none,fill=color5,fill opacity=0.6,very thin] (axis cs:6,0) rectangle (axis cs:7,0.422587923457984);
\draw[draw=none,fill=color6,fill opacity=0.6,very thin] (axis cs:7,0) rectangle (axis cs:8,0.267611826617851);
\draw[draw=none,fill=color7,fill opacity=0.6,very thin] (axis cs:8,0) rectangle (axis cs:9,0.182843444826914);
\addplot [thin, black, dashed,dash pattern=on 2pt off 1pt]
table {%
-0.45 0.690191361400584
9.45 0.690191361400584
};
\addplot [thin, color8, dashed,dash pattern=on 2pt off 1pt]
table {%
-0.45 0.791494870948766
9.45 0.791494870948766
};
\end{axis}

\end{tikzpicture}

%% file: pic/new217/217mar1k.tex
\begin{tikzpicture}

\definecolor{color0}{rgb}{0.111111111111111,0,0.888888888888889}
\definecolor{color1}{rgb}{0.222222222222222,0,0.777777777777778}
\definecolor{color2}{rgb}{0.333333333333333,0,0.666666666666667}
\definecolor{color3}{rgb}{0.444444444444444,0,0.555555555555556}
\definecolor{color4}{rgb}{0.555555555555556,0,0.444444444444444}
\definecolor{color5}{rgb}{0.666666666666667,0,0.333333333333333}
\definecolor{color6}{rgb}{0.777777777777778,0,0.222222222222222}
\definecolor{color7}{rgb}{0.888888888888889,0,0.111111111111111}
\definecolor{color8}{rgb}{0.501960784313725,0,0.501960784313725}
\pgfmathsetlengthmacro\MajorTickLength{
      \pgfkeysvalueof{/pgfplots/major tick length} * 0.5
    }
\begin{axis}[
axis line style={white},
legend cell align={left},
width = 0.4\linewidth,
major tick length=\MajorTickLength,
height = 1in,
font= \fontsize{6}{6.5}\selectfont,
legend style={fill opacity=0.8, draw opacity=1, text opacity=1, draw=white!80!black, fill=white!89.8039215686275!black},
yticklabel style={rotate=90},
tick align=inside,
tick pos=left,
x grid style={white},
xmajorgrids,
xmin=0, xmax=9,
xtick style={color=white!33.3333333333333!black},
xtick={0,1,2,3,4,5,6,7,8,9},
xticklabels={0,.1,.2,0.5,1,2,5,10,20},
y grid style={white},
ylabel={Offline shopping},
x label style={at={(0.5, 0.4)}},
ymajorgrids,
y label style={at={(0.45,0.5)}},
ymin=0, ymax=0.15,
ytick={0, .05, .1, .15, .2},
yticklabels={0, .05, .1, .15, .2},
ytick style={color=white!33.3333333333333!black}
]
\draw[draw=none,fill=blue,fill opacity=0.6,very thin] (axis cs:0,0) rectangle (axis cs:1,0.115557367594423);
\draw[draw=none,fill=color0,fill opacity=0.6,very thin] (axis cs:1,0) rectangle (axis cs:2,0.132275489372331);
\draw[draw=none,fill=color1,fill opacity=0.6,very thin] (axis cs:2,0) rectangle (axis cs:3,0.125047969438312);
\draw[draw=none,fill=color2,fill opacity=0.6,very thin] (axis cs:3,0) rectangle (axis cs:4,0.117563623319068);
\draw[draw=none,fill=color3,fill opacity=0.6,very thin] (axis cs:4,0) rectangle (axis cs:5,0.10768103765066);
\draw[draw=none,fill=color4,fill opacity=0.6,very thin] (axis cs:5,0) rectangle (axis cs:6,0.090942580736791);
\draw[draw=none,fill=color5,fill opacity=0.6,very thin] (axis cs:6,0) rectangle (axis cs:7,0.0724903532118473);
\draw[draw=none,fill=color6,fill opacity=0.6,very thin] (axis cs:7,0) rectangle (axis cs:8,0.0651570719692373);
\draw[draw=none,fill=color7,fill opacity=0.6,very thin] (axis cs:8,0) rectangle (axis cs:9,0.0694855729201335);
\addplot [thin, black, dashed,dash pattern=on 2pt off 1pt]
table {%
-0.45 0.0919987941501785
9.45 0.0919987941501785
};
\addplot [thin, color8, dashed,dash pattern=on 2pt off 1pt]
table {%
-0.45 0.108901399378296
9.45 0.108901399378296
};
\end{axis}

\end{tikzpicture}

%% file: pic/new217/217mar3k.tex
\begin{tikzpicture}

\definecolor{color0}{rgb}{0.111111111111111,0,0.888888888888889}
\definecolor{color1}{rgb}{0.222222222222222,0,0.777777777777778}
\definecolor{color2}{rgb}{0.333333333333333,0,0.666666666666667}
\definecolor{color3}{rgb}{0.444444444444444,0,0.555555555555556}
\definecolor{color4}{rgb}{0.555555555555556,0,0.444444444444444}
\definecolor{color5}{rgb}{0.666666666666667,0,0.333333333333333}
\definecolor{color6}{rgb}{0.777777777777778,0,0.222222222222222}
\definecolor{color7}{rgb}{0.888888888888889,0,0.111111111111111}
\definecolor{color8}{rgb}{0.501960784313725,0,0.501960784313725}
\pgfmathsetlengthmacro\MajorTickLength{
      \pgfkeysvalueof{/pgfplots/major tick length} * 0.5
    }
\begin{axis}[
axis line style={white},
legend cell align={left},
width = 0.4\linewidth,
major tick length=\MajorTickLength,
yticklabel style={rotate=90},
height = 1in,
font= \fontsize{6}{6.5}\selectfont,
tick align=inside,
tick pos=left,
x grid style={white},
xmajorgrids,
xmin=0, xmax=9,
xtick style={color=white!33.3333333333333!black},
xtick={0,1,2,3,4,5,6,7,8,9},
xticklabels={0,.1,.2,.5,1,2,5,10,20},
y grid style={white},
x label style={at={(0.5, 0.4)}},
ymajorgrids,
y label style={at={(0.55,0.5)}},
ymin=0, ymax=0.15,
ytick={-1,1},
ytick style={color=white!33.3333333333333!black}
]
\draw[draw=none,fill=blue,fill opacity=0.6,very thin] (axis cs:0,0) rectangle (axis cs:1,0.115793131436663);
\draw[draw=none,fill=color0,fill opacity=0.6,very thin] (axis cs:1,0) rectangle (axis cs:2,0.123743869383669);
\draw[draw=none,fill=color1,fill opacity=0.6,very thin] (axis cs:2,0) rectangle (axis cs:3,0.118019298970613);
\draw[draw=none,fill=color2,fill opacity=0.6,very thin] (axis cs:3,0) rectangle (axis cs:4,0.112093355280492);
\draw[draw=none,fill=color3,fill opacity=0.6,very thin] (axis cs:4,0) rectangle (axis cs:5,0.10428059015386);
\draw[draw=none,fill=color4,fill opacity=0.6,very thin] (axis cs:5,0) rectangle (axis cs:6,0.0918429524325166);
\draw[draw=none,fill=color5,fill opacity=0.6,very thin] (axis cs:6,0) rectangle (axis cs:7,0.0758055690199236);
\draw[draw=none,fill=color6,fill opacity=0.6,very thin] (axis cs:7,0) rectangle (axis cs:8,0.0610843356160983);
\draw[draw=none,fill=color7,fill opacity=0.6,very thin] (axis cs:8,0) rectangle (axis cs:9,0.0500051682617137);
\addplot [thin, black, dashed,dash pattern=on 2pt off 1pt]
table {%
-0.45 0.0927918406209105
9.45 0.0927918406209105
};
\addplot [thin, color8, dashed,dash pattern=on 2pt off 1pt]
table {%
-0.45 0.104530583725343
9.45 0.104530583725343
};
\end{axis}

\end{tikzpicture}

%% file: pic/new217/217mar10k_2.tex
\begin{tikzpicture}

\definecolor{color0}{rgb}{0.111111111111111,0,0.888888888888889}
\definecolor{color1}{rgb}{0.222222222222222,0,0.777777777777778}
\definecolor{color2}{rgb}{0.333333333333333,0,0.666666666666667}
\definecolor{color3}{rgb}{0.444444444444444,0,0.555555555555556}
\definecolor{color4}{rgb}{0.555555555555556,0,0.444444444444444}
\definecolor{color5}{rgb}{0.666666666666667,0,0.333333333333333}
\definecolor{color6}{rgb}{0.777777777777778,0,0.222222222222222}
\definecolor{color7}{rgb}{0.888888888888889,0,0.111111111111111}
\definecolor{color8}{rgb}{0.501960784313725,0,0.501960784313725}
\pgfmathsetlengthmacro\MajorTickLength{
      \pgfkeysvalueof{/pgfplots/major tick length} * 0.5
    }
\begin{axis}[
axis line style={white},
legend cell align={left},
yticklabel style={rotate=90},
width = 0.4\linewidth,
height = 1in,
font= \fontsize{6}{6.5}\selectfont,
tick align=inside,
major tick length=\MajorTickLength,
tick pos=left,
x grid style={white},
xmajorgrids,
xmin=0, xmax=9,
xtick style={color=white!33.3333333333333!black},
xtick={0,1,2,3,4,5,6,7,8,9},
xticklabels={0,.1,.2,0.5,1,2,5,10,20},
y grid style={white},
x label style={at={(0.5, 0.4)}},
ymajorgrids,
y label style={at={(0.35,0.5)}},
ymin=0, ymax=0.15,
ytick={-1,1},
ytick style={color=white!33.3333333333333!black}
]
\draw[draw=none,fill=blue,fill opacity=0.6,very thin] (axis cs:0,0) rectangle (axis cs:1,0.0908576624728252);
\draw[draw=none,fill=color0,fill opacity=0.6,very thin] (axis cs:1,0) rectangle (axis cs:2,0.0927804475685723);
\draw[draw=none,fill=color1,fill opacity=0.6,very thin] (axis cs:2,0) rectangle (axis cs:3,0.0902494224287278);
\draw[draw=none,fill=color2,fill opacity=0.6,very thin] (axis cs:3,0) rectangle (axis cs:4,0.0870557172022903);
\draw[draw=none,fill=color3,fill opacity=0.6,very thin] (axis cs:4,0) rectangle (axis cs:5,0.0829662014238647);
\draw[draw=none,fill=color4,fill opacity=0.6,very thin] (axis cs:5,0) rectangle (axis cs:6,0.0773992510431492);
\draw[draw=none,fill=color5,fill opacity=0.6,very thin] (axis cs:6,0) rectangle (axis cs:7,0.0691001044829628);
\draw[draw=none,fill=color6,fill opacity=0.6,very thin] (axis cs:7,0) rectangle (axis cs:8,0.0618100889703471);
\draw[draw=none,fill=color7,fill opacity=0.6,very thin] (axis cs:8,0) rectangle (axis cs:9,0.0570059110904316);
\addplot [thin, black, dashed,dash pattern=on 2pt off 1pt]
table {%
-0.45 0.0703655666759402
9.45 0.0703655666759402
};
\addplot [thin, color8, dashed,dash pattern=on 2pt off 1pt]
table {%
-0.45 0.0781899650576067
9.45 0.0781899650576067
};
\end{axis}

\end{tikzpicture}

%% file: pic/new217/217mar25k.tex
\begin{tikzpicture}

\definecolor{color0}{rgb}{0.111111111111111,0,0.888888888888889}
\definecolor{color1}{rgb}{0.222222222222222,0,0.777777777777778}
\definecolor{color2}{rgb}{0.333333333333333,0,0.666666666666667}
\definecolor{color3}{rgb}{0.444444444444444,0,0.555555555555556}
\definecolor{color4}{rgb}{0.555555555555556,0,0.444444444444444}
\definecolor{color5}{rgb}{0.666666666666667,0,0.333333333333333}
\definecolor{color6}{rgb}{0.777777777777778,0,0.222222222222222}
\definecolor{color7}{rgb}{0.888888888888889,0,0.111111111111111}
\definecolor{color8}{rgb}{0.501960784313725,0,0.501960784313725}
\pgfmathsetlengthmacro\MajorTickLength{
      \pgfkeysvalueof{/pgfplots/major tick length} * 0.5
    }
\begin{axis}[
axis line style={white},
major tick length=\MajorTickLength,
legend style={inner xsep=1pt, inner ysep=-1pt, row sep=-3pt,at={(0,1.1)},anchor=north west}, 
yticklabel style={rotate=90},
legend cell align={left},
width = 0.4\linewidth,
height = 1in,
font= \fontsize{6}{6.5}\selectfont,
tick align=inside,
tick pos=left,
x grid style={white},
xmajorgrids,
xmin=0, xmax=9,
xtick style={color=white!33.3333333333333!black},
xtick={0,1,2,3,4,5,6,7,8,9},
xticklabels={0,.1,.2,.5,1,2,5,10,20},
y grid style={white},
x label style={at={(0.5, 0.4)}},
ymajorgrids,
y label style={at={(0.55,0.5)}},
ymin=0, ymax=0.15,
ytick={-1,1},
ytick style={color=white!33.3333333333333!black}
]
\draw[draw=none,fill=blue,fill opacity=0.6,very thin] (axis cs:0,0) rectangle (axis cs:1,0.100917505930061);
\draw[draw=none,fill=color0,fill opacity=0.6,very thin] (axis cs:1,0) rectangle (axis cs:2,0.105691983977157);
\draw[draw=none,fill=color1,fill opacity=0.6,very thin] (axis cs:2,0) rectangle (axis cs:3,0.100652131039585);
\draw[draw=none,fill=color2,fill opacity=0.6,very thin] (axis cs:3,0) rectangle (axis cs:4,0.0951513822520619);
\draw[draw=none,fill=color3,fill opacity=0.6,very thin] (axis cs:4,0) rectangle (axis cs:5,0.0888027881938692);
\draw[draw=none,fill=color4,fill opacity=0.6,very thin] (axis cs:5,0) rectangle (axis cs:6,0.07315194877797);
\draw[draw=none,fill=color5,fill opacity=0.6,very thin] (axis cs:6,0) rectangle (axis cs:7,0.0560932308480465);
\draw[draw=none,fill=color6,fill opacity=0.6,very thin] (axis cs:7,0) rectangle (axis cs:8,0.043333583629916);
\draw[draw=none,fill=color7,fill opacity=0.6,very thin] (axis cs:8,0) rectangle (axis cs:9,0.0404500117197933);
\addplot [thin, black, dashed,dash pattern=on 2pt off 1pt]
table {%
-0.45 0.0682686605315264
9.45 0.0682686605315264
};
\addlegendentry{noid average level}
\addplot [thin, color8, dashed,dash pattern=on 2pt off 1pt]
table {%
-0.45 0.076552328209949
9.45 0.076552328209949
};
\addlegendentry{id average level}
\end{axis}

\end{tikzpicture}

%% file: pic/new217/217mas1k.tex
\begin{tikzpicture}

\definecolor{color0}{rgb}{0.111111111111111,0,0.888888888888889}
\definecolor{color1}{rgb}{0.222222222222222,0,0.777777777777778}
\definecolor{color2}{rgb}{0.333333333333333,0,0.666666666666667}
\definecolor{color3}{rgb}{0.444444444444444,0,0.555555555555556}
\definecolor{color4}{rgb}{0.555555555555556,0,0.444444444444444}
\definecolor{color5}{rgb}{0.666666666666667,0,0.333333333333333}
\definecolor{color6}{rgb}{0.777777777777778,0,0.222222222222222}
\definecolor{color7}{rgb}{0.888888888888889,0,0.111111111111111}
\definecolor{color8}{rgb}{0.501960784313725,0,0.501960784313725}
\pgfmathsetlengthmacro\MajorTickLength{
      \pgfkeysvalueof{/pgfplots/major tick length} * 0.5
    }
\begin{axis}[
axis line style={white},
legend cell align={left},
width = 0.4\linewidth,
major tick length=\MajorTickLength,
height = 1in,
yticklabel style={rotate=90},
font= \fontsize{6}{6.5}\selectfont,
tick align=inside,
tick pos=left,
x grid style={white},
xlabel={Penalty Expectation},
xmajorgrids,
xmin=0, xmax=9,
xtick style={color=white!33.3333333333333!black},
xtick={0,1,2,3,4,5,6,7,8,9},
xticklabels={0,.1,.2,.5,1,2,5,10,20},
y grid style={white},
ylabel={Not wearing masks},
x label style={at={(0.5, 0.3)}},
ymajorgrids,
y label style={at={(0.45,0.5)}},
ymin=0, ymax=0.6,
ytick={0, 0.2, 0.4, 0.6},
yticklabels={0, .2,.4, .6},
ytick style={color=white!33.3333333333333!black}
]
\draw[draw=none,fill=blue,fill opacity=0.6,very thin] (axis cs:0,0) rectangle (axis cs:1,0.536008966791416);
\draw[draw=none,fill=color0,fill opacity=0.6,very thin] (axis cs:1,0) rectangle (axis cs:2,0.509050958622336);
\draw[draw=none,fill=color1,fill opacity=0.6,very thin] (axis cs:2,0) rectangle (axis cs:3,0.478878388673521);
\draw[draw=none,fill=color2,fill opacity=0.6,very thin] (axis cs:3,0) rectangle (axis cs:4,0.432067527841094);
\draw[draw=none,fill=color3,fill opacity=0.6,very thin] (axis cs:4,0) rectangle (axis cs:5,0.401057743870317);
\draw[draw=none,fill=color4,fill opacity=0.6,very thin] (axis cs:5,0) rectangle (axis cs:6,0.372182061104952);
\draw[draw=none,fill=color5,fill opacity=0.6,very thin] (axis cs:6,0) rectangle (axis cs:7,0.339622580684181);
\draw[draw=none,fill=color6,fill opacity=0.6,very thin] (axis cs:7,0) rectangle (axis cs:8,0.299638869513762);
\draw[draw=none,fill=color7,fill opacity=0.6,very thin] (axis cs:8,0) rectangle (axis cs:9,0.225166848281258);
\addplot [thin, black, dashed,dash pattern=on 2pt off 1pt]
table {%
-0.45 0.284143432594184
9.45 0.284143432594184
};
\addplot [thin, color8, dashed,dash pattern=on 2pt off 1pt]
table {%
-0.45 0.317775851177883
9.45 0.317775851177883
};
\end{axis}

\end{tikzpicture}

%% file: pic/new217/217mas3k.tex
\begin{tikzpicture}

\definecolor{color0}{rgb}{0.111111111111111,0,0.888888888888889}
\definecolor{color1}{rgb}{0.222222222222222,0,0.777777777777778}
\definecolor{color2}{rgb}{0.333333333333333,0,0.666666666666667}
\definecolor{color3}{rgb}{0.444444444444444,0,0.555555555555556}
\definecolor{color4}{rgb}{0.555555555555556,0,0.444444444444444}
\definecolor{color5}{rgb}{0.666666666666667,0,0.333333333333333}
\definecolor{color6}{rgb}{0.777777777777778,0,0.222222222222222}
\definecolor{color7}{rgb}{0.888888888888889,0,0.111111111111111}
\definecolor{color8}{rgb}{0.501960784313725,0,0.501960784313725}
\pgfmathsetlengthmacro\MajorTickLength{
      \pgfkeysvalueof{/pgfplots/major tick length} * 0.5
    }
\begin{axis}[
axis line style={white},
legend cell align={left},
major tick length=\MajorTickLength,
width = 0.4\linewidth,
height = 1in,
font= \fontsize{6}{6.5}\selectfont,
tick align=inside,
tick pos=left,
x grid style={white},
xlabel={Penalty Expectation},
xmajorgrids,
xmin=0, xmax=9,
xtick style={color=white!33.3333333333333!black},
xtick={0,1,2,3,4,5,6,7,8,9},
xticklabels={0,.1,.2,.5,1,2,5,10,20},
y grid style={white},
x label style={at={(0.5, 0.3)}},
ymajorgrids,
y label style={at={(0.55,0.5)}},
ymin=0, ymax=0.6,
ytick={-1,1},
]
\draw[draw=none,fill=blue,fill opacity=0.6,very thin] (axis cs:0,0) rectangle (axis cs:1,0.52520049914643);
\draw[draw=none,fill=color0,fill opacity=0.6,very thin] (axis cs:1,0) rectangle (axis cs:2,0.448699622556029);
\draw[draw=none,fill=color1,fill opacity=0.6,very thin] (axis cs:2,0) rectangle (axis cs:3,0.446850956383751);
\draw[draw=none,fill=color2,fill opacity=0.6,very thin] (axis cs:3,0) rectangle (axis cs:4,0.410706230896126);
\draw[draw=none,fill=color3,fill opacity=0.6,very thin] (axis cs:4,0) rectangle (axis cs:5,0.370083501162143);
\draw[draw=none,fill=color4,fill opacity=0.6,very thin] (axis cs:5,0) rectangle (axis cs:6,0.324992118638926);
\draw[draw=none,fill=color5,fill opacity=0.6,very thin] (axis cs:6,0) rectangle (axis cs:7,0.28221898122693);
\draw[draw=none,fill=color6,fill opacity=0.6,very thin] (axis cs:7,0) rectangle (axis cs:8,0.244470122642689);
\draw[draw=none,fill=color7,fill opacity=0.6,very thin] (axis cs:8,0) rectangle (axis cs:9,0.199437660352847);
\addplot [thin, black, dashed,dash pattern=on 2pt off 1pt]
table {%
-0.45 0.284143432594184
9.45 0.284143432594184
};
\addplot [thin, color8, dashed,dash pattern=on 2pt off 1pt]
table {%
-0.45 0.317775851177883
9.45 0.317775851177883
};
\end{axis}

\end{tikzpicture}

%% file: pic/new217/217mas10k_2.tex
\begin{tikzpicture}

\definecolor{color0}{rgb}{0.111111111111111,0,0.888888888888889}
\definecolor{color1}{rgb}{0.222222222222222,0,0.777777777777778}
\definecolor{color2}{rgb}{0.333333333333333,0,0.666666666666667}
\definecolor{color3}{rgb}{0.444444444444444,0,0.555555555555556}
\definecolor{color4}{rgb}{0.555555555555556,0,0.444444444444444}
\definecolor{color5}{rgb}{0.666666666666667,0,0.333333333333333}
\definecolor{color6}{rgb}{0.777777777777778,0,0.222222222222222}
\definecolor{color7}{rgb}{0.888888888888889,0,0.111111111111111}
\definecolor{color8}{rgb}{0.501960784313725,0,0.501960784313725}
\pgfmathsetlengthmacro\MajorTickLength{
      \pgfkeysvalueof{/pgfplots/major tick length} * 0.5
    }
\begin{axis}[
axis line style={white},
major tick length=\MajorTickLength,
legend cell align={left},
width = 0.4\linewidth,
height = 1in,
font= \fontsize{6}{6.5}\selectfont,
tick align=inside,
tick pos=left,
x grid style={white},
xlabel={Penalty Expectation},
xmajorgrids,
xmin=0, xmax=9,
xtick style={color=white!33.3333333333333!black},
xtick={0,1,2,3,4,5,6,7,8,9},
xticklabels={0,.1,.2,.5,1,2,5,10,20},
y grid style={white},
x label style={at={(0.5, 0.3)}},
ymajorgrids,
y label style={at={(0.55,0.5)}},
ymin=0, ymax=0.6,
ytick={-1,1},
]
\draw[draw=none,fill=blue,fill opacity=0.6,very thin] (axis cs:0,0) rectangle (axis cs:1,0.49699850431834);
\draw[draw=none,fill=color0,fill opacity=0.6,very thin] (axis cs:1,0) rectangle (axis cs:2,0.30167574082576);
\draw[draw=none,fill=color1,fill opacity=0.6,very thin] (axis cs:2,0) rectangle (axis cs:3,0.286185735823078);
\draw[draw=none,fill=color2,fill opacity=0.6,very thin] (axis cs:3,0) rectangle (axis cs:4,0.289167011444677);
\draw[draw=none,fill=color3,fill opacity=0.6,very thin] (axis cs:4,0) rectangle (axis cs:5,0.276930411701842);
\draw[draw=none,fill=color4,fill opacity=0.6,very thin] (axis cs:5,0) rectangle (axis cs:6,0.23915765534658);
\draw[draw=none,fill=color5,fill opacity=0.6,very thin] (axis cs:6,0) rectangle (axis cs:7,0.207176333987742);
\draw[draw=none,fill=color6,fill opacity=0.6,very thin] (axis cs:7,0) rectangle (axis cs:8,0.195207229356437);
\draw[draw=none,fill=color7,fill opacity=0.6,very thin] (axis cs:8,0) rectangle (axis cs:9,0.200568264155379);
\addplot [thin, black, dashed,dash pattern=on 2pt off 1pt]
table {%
-0.45 0.260435395595498
9.45 0.260435395595498
};
\addplot [thin, color8, dashed,dash pattern=on 2pt off 1pt]
table {%
-0.45 0.305283177953065
9.45 0.305283177953065
};
\end{axis}

\end{tikzpicture}

%% file: pic/new217/217mas25k.tex
\begin{tikzpicture}

\definecolor{color0}{rgb}{0.111111111111111,0,0.888888888888889}
\definecolor{color1}{rgb}{0.222222222222222,0,0.777777777777778}
\definecolor{color2}{rgb}{0.333333333333333,0,0.666666666666667}
\definecolor{color3}{rgb}{0.444444444444444,0,0.555555555555556}
\definecolor{color4}{rgb}{0.555555555555556,0,0.444444444444444}
\definecolor{color5}{rgb}{0.666666666666667,0,0.333333333333333}
\definecolor{color6}{rgb}{0.777777777777778,0,0.222222222222222}
\definecolor{color7}{rgb}{0.888888888888889,0,0.111111111111111}
\definecolor{color8}{rgb}{0.501960784313725,0,0.501960784313725}
\pgfmathsetlengthmacro\MajorTickLength{
      \pgfkeysvalueof{/pgfplots/major tick length} * 0.5
    }
\begin{axis}[
axis line style={white},
legend cell align={left},
width = 0.4\linewidth,
major tick length=\MajorTickLength,
height = 1in,
font= \fontsize{6}{6.5}\selectfont,
legend style={inner xsep=1pt, inner ysep=-1pt, row sep=-3pt,at={(1,0.7)},anchor= north west}, 
tick align=inside,
tick pos=left,
x grid style={white},
xlabel={Penalty Expectation},
xmajorgrids,
xmin=0, xmax=9,
xtick style={color=white!33.3333333333333!black},
xtick={0,1,2,3,4,5,6,7,8,9},
xticklabels={0,.1,.2,.5,1,2,5,10,20},
y grid style={white},
x label style={at={(0.5, 0.3)}},
ymajorgrids,
y label style={at={(0.55,0.5)}},
ymin=0, ymax=0.6,
ytick={-1,1},
]
\draw[draw=none,fill=blue,fill opacity=0.6,very thin] (axis cs:0,0) rectangle (axis cs:1,0.474133228709691);
\draw[draw=none,fill=color0,fill opacity=0.6,very thin] (axis cs:1,0) rectangle (axis cs:2,0.280601761089304);
\draw[draw=none,fill=color1,fill opacity=0.6,very thin] (axis cs:2,0) rectangle (axis cs:3,0.252574250151293);
\draw[draw=none,fill=color2,fill opacity=0.6,very thin] (axis cs:3,0) rectangle (axis cs:4,0.222732797552049);
\draw[draw=none,fill=color3,fill opacity=0.6,very thin] (axis cs:4,0) rectangle (axis cs:5,0.215842610951774);
\draw[draw=none,fill=color4,fill opacity=0.6,very thin] (axis cs:5,0) rectangle (axis cs:6,0.195512883414815);
\draw[draw=none,fill=color5,fill opacity=0.6,very thin] (axis cs:6,0) rectangle (axis cs:7,0.172327748615815);
\draw[draw=none,fill=color6,fill opacity=0.6,very thin] (axis cs:7,0) rectangle (axis cs:8,0.165218996408204);
\draw[draw=none,fill=color7,fill opacity=0.6,very thin] (axis cs:8,0) rectangle (axis cs:9,0.149174316619252);
\addplot [thin, black, dashed,dash pattern=on 2pt off 1pt]
table {%
-0.45 0.284143432594184
9.45 0.284143432594184
};
\addplot [thin, color8, dashed,dash pattern=on 2pt off 1pt]
table {%
-0.45 0.317775851177883
9.45 0.317775851177883
};
\end{axis}

\end{tikzpicture}

%% file: pic/new22/22dic1k.tex
\begin{tikzpicture}

\definecolor{color0}{rgb}{0.886274509803922,0.290196078431373,0.2}
\definecolor{color1}{rgb}{0.203921568627451,0.541176470588235,0.741176470588235}
\definecolor{color2}{rgb}{0.596078431372549,0.556862745098039,0.835294117647059}
\pgfmathsetlengthmacro\MajorTickLength{
      \pgfkeysvalueof{/pgfplots/major tick length} * 0.5
    }
\begin{axis}[
axis line style={white},
legend cell align={left},
width = 0.37\linewidth,
height = 1in,
major tick length=\MajorTickLength,
font= \fontsize{7}{7.5}\selectfont,
tick align=inside,
tick pos=left,
x grid style={white},
xmajorgrids,
xmin=0, xmax=80,
xtick style={color=white!33.3333333333333!black},
y grid style={white},
ylabel={Daily cases},
yticklabel style={rotate=90},
x label style={at={(0.5, 0.4)}},
ymajorgrids,
y label style={at={(0.4,0.5)}},
xtick={0,20,40,60,80},
xticklabels={0,20,40,60,80},
ytick={0,450,900},
yticklabels={0,450,900},
ymin=-39.5854866667121, ymax=925.519085989483,
ytick style={color=white!33.3333333333333!black}
]
\path [fill=color0, fill opacity=0.3, very thin]
(axis cs:0,11.383763667188)
--(axis cs:0,4.28290299947864)
--(axis cs:1,6.58987895892336)
--(axis cs:2,8.88780613813593)
--(axis cs:3,13.6112177916641)
--(axis cs:4,18.5573452587113)
--(axis cs:5,29.2862923969226)
--(axis cs:6,37.3740723868913)
--(axis cs:7,53.2141174631069)
--(axis cs:8,78.2265026284327)
--(axis cs:9,118.990673780333)
--(axis cs:10,103.59583906288)
--(axis cs:11,55.7112902049158)
--(axis cs:12,79.9846182283521)
--(axis cs:13,96.7657675598148)
--(axis cs:14,121.955246585779)
--(axis cs:15,127.717308607088)
--(axis cs:16,138.56963137046)
--(axis cs:17,150.567442599065)
--(axis cs:18,155.580144100478)
--(axis cs:19,165.786310839833)
--(axis cs:20,172.837776674988)
--(axis cs:21,175.586053139931)
--(axis cs:22,188.913689662137)
--(axis cs:23,188.029845813988)
--(axis cs:24,200.108846751635)
--(axis cs:25,211.147958706526)
--(axis cs:26,222.616217977547)
--(axis cs:27,231.09791153424)
--(axis cs:28,243.453959349052)
--(axis cs:29,249.967908996175)
--(axis cs:30,261.463985918915)
--(axis cs:31,266.279626272393)
--(axis cs:32,282.836858634591)
--(axis cs:33,297.636263996249)
--(axis cs:34,304.114779302641)
--(axis cs:35,308.004928947666)
--(axis cs:36,317.58448160555)
--(axis cs:37,332.270508837744)
--(axis cs:38,347.99011847962)
--(axis cs:39,346.103041782322)
--(axis cs:40,359.750229085413)
--(axis cs:41,380.506594770172)
--(axis cs:42,367.012427817217)
--(axis cs:43,382.447334414114)
--(axis cs:44,380.300361870532)
--(axis cs:45,403.48870239532)
--(axis cs:46,403.614110119877)
--(axis cs:47,395.196302108431)
--(axis cs:48,400.176706129548)
--(axis cs:49,420.205523137846)
--(axis cs:50,427.568153261884)
--(axis cs:51,429.035972064294)
--(axis cs:52,418.507886194554)
--(axis cs:53,431.54042190529)
--(axis cs:54,448.990257672359)
--(axis cs:55,439.397091600992)
--(axis cs:56,438.287397959579)
--(axis cs:57,450.145478910291)
--(axis cs:58,449.044531933545)
--(axis cs:59,462.06013090647)
--(axis cs:60,461.56664374771)
--(axis cs:61,476.205569502678)
--(axis cs:62,477.667929513879)
--(axis cs:63,477.050348699453)
--(axis cs:64,490.735257602259)
--(axis cs:65,491.965436907974)
--(axis cs:66,498.145182724763)
--(axis cs:67,504.968740028047)
--(axis cs:68,494.44911971407)
--(axis cs:69,499.539161163173)
--(axis cs:70,494.18653496206)
--(axis cs:71,507.804340937322)
--(axis cs:72,512.062079090684)
--(axis cs:73,508.766537410499)
--(axis cs:74,513.901480974912)
--(axis cs:75,514.297599575086)
--(axis cs:76,514.500109649157)
--(axis cs:77,535.176747268495)
--(axis cs:78,524.615970343374)
--(axis cs:79,518.06486218322)
--(axis cs:79,875.201804483447)
--(axis cs:79,875.201804483447)
--(axis cs:78,881.650696323293)
--(axis cs:77,878.289919398171)
--(axis cs:76,869.16655701751)
--(axis cs:75,856.369067091581)
--(axis cs:74,850.765185691754)
--(axis cs:73,809.633462589501)
--(axis cs:72,801.737920909316)
--(axis cs:71,799.595659062679)
--(axis cs:70,796.546798371273)
--(axis cs:69,796.19417217016)
--(axis cs:68,773.084213619263)
--(axis cs:67,769.164593305287)
--(axis cs:66,762.18815060857)
--(axis cs:65,740.434563092026)
--(axis cs:64,744.798075731074)
--(axis cs:63,720.016317967213)
--(axis cs:62,713.465403819455)
--(axis cs:61,690.861097163988)
--(axis cs:60,703.966689585623)
--(axis cs:59,689.873202426863)
--(axis cs:58,657.888801399788)
--(axis cs:57,638.587854423042)
--(axis cs:56,643.645935373754)
--(axis cs:55,615.136241732341)
--(axis cs:54,605.476408994307)
--(axis cs:53,600.192911428043)
--(axis cs:52,582.358780472112)
--(axis cs:51,560.830694602372)
--(axis cs:50,558.365180071449)
--(axis cs:49,544.927810195488)
--(axis cs:48,543.489960537118)
--(axis cs:47,525.003697891569)
--(axis cs:46,509.119223213456)
--(axis cs:45,504.11129760468)
--(axis cs:44,487.766304796135)
--(axis cs:43,481.552665585886)
--(axis cs:42,461.05423884945)
--(axis cs:41,457.160071896494)
--(axis cs:40,453.116437581254)
--(axis cs:39,426.096958217678)
--(axis cs:38,427.60988152038)
--(axis cs:37,414.396157828922)
--(axis cs:36,407.81551839445)
--(axis cs:35,393.661737719)
--(axis cs:34,398.751887364026)
--(axis cs:33,377.630402670418)
--(axis cs:32,372.029808032076)
--(axis cs:31,363.320373727607)
--(axis cs:30,349.336014081085)
--(axis cs:29,340.298757670491)
--(axis cs:28,320.412707317615)
--(axis cs:27,311.435421799094)
--(axis cs:26,303.717115355786)
--(axis cs:25,285.585374626807)
--(axis cs:24,273.824486581698)
--(axis cs:23,262.903487519345)
--(axis cs:22,251.15297700453)
--(axis cs:21,236.147280193403)
--(axis cs:20,222.095556658346)
--(axis cs:19,217.7470224935)
--(axis cs:18,204.619855899522)
--(axis cs:17,186.365890734269)
--(axis cs:16,186.163701962873)
--(axis cs:15,174.882691392912)
--(axis cs:14,163.511420080888)
--(axis cs:13,147.434232440185)
--(axis cs:12,163.215381771648)
--(axis cs:11,190.822043128418)
--(axis cs:10,252.070827603786)
--(axis cs:9,206.209326219667)
--(axis cs:8,155.173497371567)
--(axis cs:7,106.719215870226)
--(axis cs:6,78.0259276131087)
--(axis cs:5,50.7803742697441)
--(axis cs:4,37.0426547412887)
--(axis cs:3,28.6554488750026)
--(axis cs:2,19.3788605285307)
--(axis cs:1,14.94345437441)
--(axis cs:0,11.383763667188)
--cycle;

\path [fill=color1, fill opacity=0.3, very thin]
(axis cs:0,9.34604279963717)
--(axis cs:0,4.78729053369617)
--(axis cs:1,8.70645224675662)
--(axis cs:2,8.06712645865147)
--(axis cs:3,12.0238984450134)
--(axis cs:4,18.7639554639919)
--(axis cs:5,20.049219424708)
--(axis cs:6,36.9228717959843)
--(axis cs:7,51.149460706467)
--(axis cs:8,72.7490225428745)
--(axis cs:9,106.492724762386)
--(axis cs:10,111.876886789945)
--(axis cs:11,51.6789310852862)
--(axis cs:12,60.9006775129949)
--(axis cs:13,84.8623430979544)
--(axis cs:14,98.9172403666009)
--(axis cs:15,114.53262109888)
--(axis cs:16,114.04563134915)
--(axis cs:17,114.158117848524)
--(axis cs:18,117.222024018649)
--(axis cs:19,109.772666500834)
--(axis cs:20,114.974909456272)
--(axis cs:21,123.115165946888)
--(axis cs:22,117.670232409682)
--(axis cs:23,115.971681667309)
--(axis cs:24,122.935229488735)
--(axis cs:25,122.205398553137)
--(axis cs:26,122.403856979832)
--(axis cs:27,126.611320250895)
--(axis cs:28,123.791972010019)
--(axis cs:29,126.086573203411)
--(axis cs:30,130.271995997303)
--(axis cs:31,128.473355451266)
--(axis cs:32,134.67567521653)
--(axis cs:33,139.097052707083)
--(axis cs:34,137.450612291925)
--(axis cs:35,134.811211543455)
--(axis cs:36,145.732585476808)
--(axis cs:37,138.059926032967)
--(axis cs:38,149.616018557293)
--(axis cs:39,148.545552493973)
--(axis cs:40,146.101724551365)
--(axis cs:41,157.063416880819)
--(axis cs:42,155.590429283582)
--(axis cs:43,155.901814460702)
--(axis cs:44,168.338966828151)
--(axis cs:45,158.705391519425)
--(axis cs:46,162.713910350097)
--(axis cs:47,163.980504194606)
--(axis cs:48,161.478264939826)
--(axis cs:49,167.006406414567)
--(axis cs:50,176.078566217718)
--(axis cs:51,173.009180787138)
--(axis cs:52,175.002267367223)
--(axis cs:53,182.962098800593)
--(axis cs:54,184.406624055588)
--(axis cs:55,182.737757964235)
--(axis cs:56,189.546622794298)
--(axis cs:57,196.043027829796)
--(axis cs:58,199.413870342322)
--(axis cs:59,202.328740881345)
--(axis cs:60,211.626327633272)
--(axis cs:61,212.636455184159)
--(axis cs:62,219.607186145158)
--(axis cs:63,230.107115658219)
--(axis cs:64,226.097226096019)
--(axis cs:65,239.262360652896)
--(axis cs:66,245.076349739326)
--(axis cs:67,243.082785744881)
--(axis cs:68,250.819515676593)
--(axis cs:69,253.146840808183)
--(axis cs:70,250.988259374021)
--(axis cs:71,252.330591400449)
--(axis cs:72,250.853692675872)
--(axis cs:73,248.911426926757)
--(axis cs:74,236.964830590061)
--(axis cs:75,245.017165487613)
--(axis cs:76,223.670118744993)
--(axis cs:77,204.246924349903)
--(axis cs:78,214.225906933103)
--(axis cs:79,187.238674016561)
--(axis cs:79,318.094659316772)
--(axis cs:79,318.094659316772)
--(axis cs:78,347.840759733564)
--(axis cs:77,362.753075650097)
--(axis cs:76,367.863214588341)
--(axis cs:75,381.449501179054)
--(axis cs:74,393.701836076605)
--(axis cs:73,381.821906406576)
--(axis cs:72,392.346307324128)
--(axis cs:71,386.869408599551)
--(axis cs:70,397.878407292646)
--(axis cs:69,391.986492525151)
--(axis cs:68,391.31381765674)
--(axis cs:67,382.583880921785)
--(axis cs:66,401.123650260674)
--(axis cs:65,380.670972680437)
--(axis cs:64,371.502773903981)
--(axis cs:63,368.292884341781)
--(axis cs:62,371.126147188176)
--(axis cs:61,363.896878149175)
--(axis cs:60,358.973672366728)
--(axis cs:59,352.337925785322)
--(axis cs:58,338.786129657678)
--(axis cs:57,329.556972170204)
--(axis cs:56,327.320043872369)
--(axis cs:55,316.328908702432)
--(axis cs:54,308.526709277745)
--(axis cs:53,308.237901199407)
--(axis cs:52,296.53106596611)
--(axis cs:51,293.057485879528)
--(axis cs:50,301.188100448949)
--(axis cs:49,289.793593585433)
--(axis cs:48,289.455068393507)
--(axis cs:47,286.419495805394)
--(axis cs:46,266.619422983237)
--(axis cs:45,266.894608480575)
--(axis cs:44,268.527699838516)
--(axis cs:43,258.164852205964)
--(axis cs:42,261.942904049752)
--(axis cs:41,258.269916452514)
--(axis cs:40,245.964942115302)
--(axis cs:39,244.921114172694)
--(axis cs:38,240.050648109374)
--(axis cs:37,239.6067406337)
--(axis cs:36,228.934081189859)
--(axis cs:35,231.922121789878)
--(axis cs:34,217.749387708075)
--(axis cs:33,223.169613959584)
--(axis cs:32,219.52432478347)
--(axis cs:31,210.859977882068)
--(axis cs:30,211.994670669364)
--(axis cs:29,202.646760129922)
--(axis cs:28,204.074694656648)
--(axis cs:27,205.122013082438)
--(axis cs:26,186.929476353501)
--(axis cs:25,188.06126811353)
--(axis cs:24,181.531437177932)
--(axis cs:23,176.894984999358)
--(axis cs:22,171.263100923651)
--(axis cs:21,174.018167386445)
--(axis cs:20,157.158423877061)
--(axis cs:19,160.694000165833)
--(axis cs:18,159.577975981351)
--(axis cs:17,152.508548818143)
--(axis cs:16,153.15436865085)
--(axis cs:15,145.734045567787)
--(axis cs:14,145.882759633399)
--(axis cs:13,141.404323568712)
--(axis cs:12,134.699322487005)
--(axis cs:11,221.987735581381)
--(axis cs:10,259.723113210055)
--(axis cs:9,190.040608570947)
--(axis cs:8,156.584310790459)
--(axis cs:7,110.383872626866)
--(axis cs:6,74.7437948706823)
--(axis cs:5,59.550780575292)
--(axis cs:4,35.5693778693414)
--(axis cs:3,29.1094348883199)
--(axis cs:2,17.7328735413485)
--(axis cs:1,17.8268810865767)
--(axis cs:0,9.34604279963717)
--cycle;

\path [fill=color2, fill opacity=0.3, very thin]
(axis cs:0,8.83472174030374)
--(axis cs:0,4.83194492636292)
--(axis cs:1,7.05182030423458)
--(axis cs:2,10.5266708281331)
--(axis cs:3,11.533030309291)
--(axis cs:4,17.3762027147381)
--(axis cs:5,26.0582518404196)
--(axis cs:6,31.8722135736095)
--(axis cs:7,53.3412518315869)
--(axis cs:8,75.4126459758722)
--(axis cs:9,109.059614159525)
--(axis cs:10,95.0037858794983)
--(axis cs:11,39.890091488125)
--(axis cs:12,52.4939691948059)
--(axis cs:13,72.7503083467174)
--(axis cs:14,85.9253304071511)
--(axis cs:15,90.5296152869018)
--(axis cs:16,87.1675438839494)
--(axis cs:17,90.127895361864)
--(axis cs:18,93.7942868799907)
--(axis cs:19,92.2854928031487)
--(axis cs:20,90.6828441199137)
--(axis cs:21,91.2315485889085)
--(axis cs:22,89.6293270856909)
--(axis cs:23,102.034714871559)
--(axis cs:24,91.8725113519473)
--(axis cs:25,96.324612724116)
--(axis cs:26,100.191702787733)
--(axis cs:27,100.400527432763)
--(axis cs:28,100.441774183152)
--(axis cs:29,97.8320070466249)
--(axis cs:30,99.9961678524247)
--(axis cs:31,101.273760012259)
--(axis cs:32,104.976635192948)
--(axis cs:33,104.474075189647)
--(axis cs:34,106.653234111899)
--(axis cs:35,107.554377389154)
--(axis cs:36,104.036226197927)
--(axis cs:37,107.471812271766)
--(axis cs:38,106.928566564904)
--(axis cs:39,113.442929134584)
--(axis cs:40,111.527040726784)
--(axis cs:41,111.30418115274)
--(axis cs:42,112.460786856666)
--(axis cs:43,116.595984200693)
--(axis cs:44,113.25671305663)
--(axis cs:45,118.254809237522)
--(axis cs:46,117.998028924615)
--(axis cs:47,114.124906479376)
--(axis cs:48,120.857383611225)
--(axis cs:49,120.375705134325)
--(axis cs:50,125.094413054641)
--(axis cs:51,130.714997022598)
--(axis cs:52,124.880833037507)
--(axis cs:53,129.453271188303)
--(axis cs:54,140.339106476527)
--(axis cs:55,134.245689646231)
--(axis cs:56,139.249089510797)
--(axis cs:57,142.94531826009)
--(axis cs:58,137.552777877091)
--(axis cs:59,147.804435648371)
--(axis cs:60,148.991828508622)
--(axis cs:61,147.382126405981)
--(axis cs:62,161.935933041565)
--(axis cs:63,155.040492795748)
--(axis cs:64,161.367989442496)
--(axis cs:65,150.899791145075)
--(axis cs:66,165.540150678742)
--(axis cs:67,165.723810802173)
--(axis cs:68,166.927322399834)
--(axis cs:69,169.525614890222)
--(axis cs:70,178.566990458322)
--(axis cs:71,170.187270133129)
--(axis cs:72,167.291768703347)
--(axis cs:73,167.259318762383)
--(axis cs:74,178.309319811377)
--(axis cs:75,175.641194011486)
--(axis cs:76,173.918064572071)
--(axis cs:77,173.045475222397)
--(axis cs:78,168.742338023643)
--(axis cs:79,167.806526964641)
--(axis cs:79,240.660139702026)
--(axis cs:79,240.660139702026)
--(axis cs:78,233.390995309691)
--(axis cs:77,233.887858110936)
--(axis cs:76,240.881935427929)
--(axis cs:75,242.492139321847)
--(axis cs:74,240.890680188623)
--(axis cs:73,242.87401457095)
--(axis cs:72,233.308231296653)
--(axis cs:71,232.346063200205)
--(axis cs:70,243.766342875011)
--(axis cs:69,239.207718443111)
--(axis cs:68,232.739344266832)
--(axis cs:67,236.676189197827)
--(axis cs:66,230.726515987925)
--(axis cs:65,230.700208854925)
--(axis cs:64,232.765343890838)
--(axis cs:63,225.026173870919)
--(axis cs:62,214.930733625102)
--(axis cs:61,224.484540260686)
--(axis cs:60,218.074838158045)
--(axis cs:59,212.928897684963)
--(axis cs:58,215.247222122909)
--(axis cs:57,210.05468173991)
--(axis cs:56,207.284243822536)
--(axis cs:55,198.887643687103)
--(axis cs:54,203.460893523473)
--(axis cs:53,212.413395478363)
--(axis cs:52,197.652500295826)
--(axis cs:51,204.418336310735)
--(axis cs:50,194.905586945359)
--(axis cs:49,201.624294865675)
--(axis cs:48,195.075949722108)
--(axis cs:47,190.608426853957)
--(axis cs:46,181.135304408718)
--(axis cs:45,181.611857429145)
--(axis cs:44,182.476620276703)
--(axis cs:43,177.470682465973)
--(axis cs:42,177.472546476668)
--(axis cs:41,170.029152180593)
--(axis cs:40,175.806292606549)
--(axis cs:39,169.823737532083)
--(axis cs:38,167.538100101762)
--(axis cs:37,167.928187728234)
--(axis cs:36,162.030440468739)
--(axis cs:35,164.978955944179)
--(axis cs:34,158.613432554767)
--(axis cs:33,153.659258143686)
--(axis cs:32,151.223364807052)
--(axis cs:31,145.126239987741)
--(axis cs:30,149.870498814242)
--(axis cs:29,142.234659620042)
--(axis cs:28,142.424892483515)
--(axis cs:27,142.532805900571)
--(axis cs:26,132.874963878933)
--(axis cs:25,132.008720609217)
--(axis cs:24,129.127488648053)
--(axis cs:23,126.965285128441)
--(axis cs:22,122.637339580976)
--(axis cs:21,135.435118077758)
--(axis cs:20,124.183822546753)
--(axis cs:19,127.914507196851)
--(axis cs:18,121.805713120009)
--(axis cs:17,118.738771304803)
--(axis cs:16,118.832456116051)
--(axis cs:15,124.003718046432)
--(axis cs:14,124.808002926182)
--(axis cs:13,142.116358319949)
--(axis cs:12,124.906030805194)
--(axis cs:11,174.576575178542)
--(axis cs:10,239.396214120502)
--(axis cs:9,208.007052507142)
--(axis cs:8,144.987354024128)
--(axis cs:7,102.72541483508)
--(axis cs:6,71.1277864263905)
--(axis cs:5,52.4750814929137)
--(axis cs:4,37.7571306185952)
--(axis cs:3,26.2003030240423)
--(axis cs:2,21.2066625052002)
--(axis cs:1,15.7481796957654)
--(axis cs:0,8.83472174030374)
--cycle;

\addplot [semithick, color0]
table {%
0 7.83333333333333
1 10.7666666666667
2 14.1333333333333
3 21.1333333333333
4 27.8
5 40.0333333333333
6 57.7
7 79.9666666666667
8 116.7
9 162.6
10 177.833333333333
11 123.266666666667
12 121.6
13 122.1
14 142.733333333333
15 151.3
16 162.366666666667
17 168.466666666667
18 180.1
19 191.766666666667
20 197.466666666667
21 205.866666666667
22 220.033333333333
23 225.466666666667
24 236.966666666667
25 248.366666666667
26 263.166666666667
27 271.266666666667
28 281.933333333333
29 295.133333333333
30 305.4
31 314.8
32 327.433333333333
33 337.633333333333
34 351.433333333333
35 350.833333333333
36 362.7
37 373.333333333333
38 387.8
39 386.1
40 406.433333333333
41 418.833333333333
42 414.033333333333
43 432
44 434.033333333333
45 453.8
46 456.366666666667
47 460.1
48 471.833333333333
49 482.566666666667
50 492.966666666667
51 494.933333333333
52 500.433333333333
53 515.866666666667
54 527.233333333333
55 527.266666666667
56 540.966666666667
57 544.366666666667
58 553.466666666667
59 575.966666666667
60 582.766666666667
61 583.533333333333
62 595.566666666667
63 598.533333333333
64 617.766666666667
65 616.2
66 630.166666666667
67 637.066666666667
68 633.766666666667
69 647.866666666667
70 645.366666666667
71 653.7
72 656.9
73 659.2
74 682.333333333333
75 685.333333333333
76 691.833333333333
77 706.733333333333
78 703.133333333333
79 696.633333333333
};
\addplot [semithick, color1]
table {%
0 7.06666666666667
1 13.2666666666667
2 12.9
3 20.5666666666667
4 27.1666666666667
5 39.8
6 55.8333333333333
7 80.7666666666667
8 114.666666666667
9 148.266666666667
10 185.8
11 136.833333333333
12 97.8
13 113.133333333333
14 122.4
15 130.133333333333
16 133.6
17 133.333333333333
18 138.4
19 135.233333333333
20 136.066666666667
21 148.566666666667
22 144.466666666667
23 146.433333333333
24 152.233333333333
25 155.133333333333
26 154.666666666667
27 165.866666666667
28 163.933333333333
29 164.366666666667
30 171.133333333333
31 169.666666666667
32 177.1
33 181.133333333333
34 177.6
35 183.366666666667
36 187.333333333333
37 188.833333333333
38 194.833333333333
39 196.733333333333
40 196.033333333333
41 207.666666666667
42 208.766666666667
43 207.033333333333
44 218.433333333333
45 212.8
46 214.666666666667
47 225.2
48 225.466666666667
49 228.4
50 238.633333333333
51 233.033333333333
52 235.766666666667
53 245.6
54 246.466666666667
55 249.533333333333
56 258.433333333333
57 262.8
58 269.1
59 277.333333333333
60 285.3
61 288.266666666667
62 295.366666666667
63 299.2
64 298.8
65 309.966666666667
66 323.1
67 312.833333333333
68 321.066666666667
69 322.566666666667
70 324.433333333333
71 319.6
72 321.6
73 315.366666666667
74 315.333333333333
75 313.233333333333
76 295.766666666667
77 283.5
78 281.033333333333
79 252.666666666667
};
\addplot [semithick, color2]
table {%
0 6.83333333333333
1 11.4
2 15.8666666666667
3 18.8666666666667
4 27.5666666666667
5 39.2666666666667
6 51.5
7 78.0333333333333
8 110.2
9 158.533333333333
10 167.2
11 107.233333333333
12 88.7
13 107.433333333333
14 105.366666666667
15 107.266666666667
16 103
17 104.433333333333
18 107.8
19 110.1
20 107.433333333333
21 113.333333333333
22 106.133333333333
23 114.5
24 110.5
25 114.166666666667
26 116.533333333333
27 121.466666666667
28 121.433333333333
29 120.033333333333
30 124.933333333333
31 123.2
32 128.1
33 129.066666666667
34 132.633333333333
35 136.266666666667
36 133.033333333333
37 137.7
38 137.233333333333
39 141.633333333333
40 143.666666666667
41 140.666666666667
42 144.966666666667
43 147.033333333333
44 147.866666666667
45 149.933333333333
46 149.566666666667
47 152.366666666667
48 157.966666666667
49 161
50 160
51 167.566666666667
52 161.266666666667
53 170.933333333333
54 171.9
55 166.566666666667
56 173.266666666667
57 176.5
58 176.4
59 180.366666666667
60 183.533333333333
61 185.933333333333
62 188.433333333333
63 190.033333333333
64 197.066666666667
65 190.8
66 198.133333333333
67 201.2
68 199.833333333333
69 204.366666666667
70 211.166666666667
71 201.266666666667
72 200.3
73 205.066666666667
74 209.6
75 209.066666666667
76 207.4
77 203.466666666667
78 201.066666666667
79 204.233333333333
};
\end{axis}

\end{tikzpicture}

%% file: pic/new22/22dic3k.tex
\begin{tikzpicture}

\definecolor{color0}{rgb}{0.886274509803922,0.290196078431373,0.2}
\definecolor{color1}{rgb}{0.203921568627451,0.541176470588235,0.741176470588235}
\definecolor{color2}{rgb}{0.596078431372549,0.556862745098039,0.835294117647059}
\pgfmathsetlengthmacro\MajorTickLength{
      \pgfkeysvalueof{/pgfplots/major tick length} * 0.5
    }
\begin{axis}[
axis line style={white},
legend cell align={left},
width = 0.37\linewidth,
height = 1in,
major tick length=\MajorTickLength,
font= \fontsize{7}{7.5}\selectfont,
tick align=inside,
tick pos=left,
x grid style={white},
xmajorgrids,
xmin=0, xmax=80,
yticklabel style={rotate=90},
xtick style={color=white!33.3333333333333!black},
y grid style={white},
x label style={at={(0.5, 0.4)}},
ymajorgrids,
y label style={at={(0.55,0.5)}},
xtick={0,20,40,60,80},
xticklabels={0,20,40,60,80},
ytick={0,150,300},
yticklabels={0,150,300},
ymin=-14.2504432196881, ymax=383.394863207588,
ytick style={color=white!33.3333333333333!black}
]
\path [fill=color0, fill opacity=0.3, very thin]
(axis cs:0,9.28516708042651)
--(axis cs:0,3.98149958624016)
--(axis cs:1,8.45651349372243)
--(axis cs:2,10.103789277355)
--(axis cs:3,12.4567454421654)
--(axis cs:4,15.9803995805644)
--(axis cs:5,27.4086542581241)
--(axis cs:6,38.2369822528649)
--(axis cs:7,57.8332304528709)
--(axis cs:8,81.5105111240782)
--(axis cs:9,108.941411906776)
--(axis cs:10,107.298001908738)
--(axis cs:11,23.3549527744921)
--(axis cs:12,36.3700108609169)
--(axis cs:13,77.5427404656168)
--(axis cs:14,84.9109048052783)
--(axis cs:15,89.4747451812099)
--(axis cs:16,91.6013166235831)
--(axis cs:17,97.2150861386383)
--(axis cs:18,99.1240397653839)
--(axis cs:19,98.5767961464052)
--(axis cs:20,97.2216949976205)
--(axis cs:21,97.0703931247031)
--(axis cs:22,102.125443961714)
--(axis cs:23,100.278915957839)
--(axis cs:24,99.6185199582079)
--(axis cs:25,100.153524512059)
--(axis cs:26,106.216397115648)
--(axis cs:27,109.641140762776)
--(axis cs:28,109.868140317632)
--(axis cs:29,110.509646001356)
--(axis cs:30,105.788517507617)
--(axis cs:31,110.953683140563)
--(axis cs:32,119.419727703967)
--(axis cs:33,121.394370722468)
--(axis cs:34,115.993312855236)
--(axis cs:35,119.414892571898)
--(axis cs:36,123.228013331987)
--(axis cs:37,120.284296689799)
--(axis cs:38,113.869498272547)
--(axis cs:39,117.312364324949)
--(axis cs:40,123.773841003884)
--(axis cs:41,122.986422802972)
--(axis cs:42,125.363441200092)
--(axis cs:43,122.198312644289)
--(axis cs:44,121.625916931299)
--(axis cs:45,122.47128782206)
--(axis cs:46,133.820573328157)
--(axis cs:47,130.228093367108)
--(axis cs:48,142.438264241813)
--(axis cs:49,131.450525030972)
--(axis cs:50,133.39945636814)
--(axis cs:51,130.61155962734)
--(axis cs:52,140.374807732241)
--(axis cs:53,147.94052172224)
--(axis cs:54,148.251421367108)
--(axis cs:55,158.94648052743)
--(axis cs:56,153.260507177611)
--(axis cs:57,153.272339371262)
--(axis cs:58,165.687882160235)
--(axis cs:59,160.012441748317)
--(axis cs:60,166.725286192898)
--(axis cs:61,171.848948099883)
--(axis cs:62,170.771059768887)
--(axis cs:63,172.629958830434)
--(axis cs:64,169.338269377562)
--(axis cs:65,170.330824580576)
--(axis cs:66,186.786376902669)
--(axis cs:67,185.237445248371)
--(axis cs:68,186.36667489712)
--(axis cs:69,188.940885976092)
--(axis cs:70,178.14582150991)
--(axis cs:71,186.357907338493)
--(axis cs:72,190.885283215379)
--(axis cs:73,186.915895311466)
--(axis cs:74,193.588190882255)
--(axis cs:75,188.485453393127)
--(axis cs:76,202.499079326116)
--(axis cs:77,201.2119025168)
--(axis cs:78,195.775915936996)
--(axis cs:79,205.079923448198)
--(axis cs:79,365.320076551802)
--(axis cs:79,365.320076551802)
--(axis cs:78,360.224084063004)
--(axis cs:77,350.1880974832)
--(axis cs:76,343.700920673884)
--(axis cs:75,337.98121327354)
--(axis cs:74,349.411809117745)
--(axis cs:73,333.484104688534)
--(axis cs:72,332.914716784621)
--(axis cs:71,327.042092661507)
--(axis cs:70,329.387511823424)
--(axis cs:69,321.925780690575)
--(axis cs:68,321.366658436213)
--(axis cs:67,316.829221418296)
--(axis cs:66,315.013623097331)
--(axis cs:65,315.602508752758)
--(axis cs:64,306.195063955771)
--(axis cs:63,307.3033745029)
--(axis cs:62,304.562273564446)
--(axis cs:61,297.28438523345)
--(axis cs:60,308.674713807102)
--(axis cs:59,300.387558251683)
--(axis cs:58,299.112117839765)
--(axis cs:57,302.127660628738)
--(axis cs:56,304.072826155722)
--(axis cs:55,301.25351947257)
--(axis cs:54,292.281911966225)
--(axis cs:53,291.992811611093)
--(axis cs:52,301.825192267759)
--(axis cs:51,294.38844037266)
--(axis cs:50,288.733876965193)
--(axis cs:49,292.816141635694)
--(axis cs:48,290.028402424853)
--(axis cs:47,285.238573299559)
--(axis cs:46,276.379426671843)
--(axis cs:45,280.195378844607)
--(axis cs:44,266.507416402035)
--(axis cs:43,267.068354022377)
--(axis cs:42,261.436558799908)
--(axis cs:41,264.413577197028)
--(axis cs:40,256.492825662782)
--(axis cs:39,251.287635675051)
--(axis cs:38,237.863835060787)
--(axis cs:37,232.449036643534)
--(axis cs:36,228.705320001347)
--(axis cs:35,223.851774094769)
--(axis cs:34,222.340020478097)
--(axis cs:33,211.405629277532)
--(axis cs:32,220.780272296033)
--(axis cs:31,204.846316859437)
--(axis cs:30,195.211482492383)
--(axis cs:29,193.090353998644)
--(axis cs:28,185.265193015702)
--(axis cs:27,186.625525903891)
--(axis cs:26,179.383602884352)
--(axis cs:25,174.313142154607)
--(axis cs:24,166.314813375125)
--(axis cs:23,155.054417375494)
--(axis cs:22,157.341222704952)
--(axis cs:21,151.929606875297)
--(axis cs:20,141.244971669046)
--(axis cs:19,138.823203853595)
--(axis cs:18,132.609293567949)
--(axis cs:17,130.784913861362)
--(axis cs:16,128.73201670975)
--(axis cs:15,119.72525481879)
--(axis cs:14,118.422428528055)
--(axis cs:13,114.857259534383)
--(axis cs:12,139.69665580575)
--(axis cs:11,194.578380558841)
--(axis cs:10,249.101998091262)
--(axis cs:9,200.258588093224)
--(axis cs:8,147.356155542588)
--(axis cs:7,111.833436213796)
--(axis cs:6,76.2963510804684)
--(axis cs:5,53.5913457418759)
--(axis cs:4,32.552933752769)
--(axis cs:3,25.0099212245012)
--(axis cs:2,20.3628773893116)
--(axis cs:1,19.4101531729442)
--(axis cs:0,9.28516708042651)
--cycle;

\path [fill=color1, fill opacity=0.3, very thin]
(axis cs:0,11.2180627470025)
--(axis cs:0,5.04860391966419)
--(axis cs:1,7.45584431364505)
--(axis cs:2,10.3368665822807)
--(axis cs:3,13.1631324486492)
--(axis cs:4,21.6069362200022)
--(axis cs:5,28.1995024967976)
--(axis cs:6,42.6248391041071)
--(axis cs:7,65.4940536422725)
--(axis cs:8,92.3729216915575)
--(axis cs:9,127.215721069646)
--(axis cs:10,84.4604916194855)
--(axis cs:11,27.8052402359201)
--(axis cs:12,51.3686671650868)
--(axis cs:13,78.8209622926301)
--(axis cs:14,86.4011828295478)
--(axis cs:15,83.954275662774)
--(axis cs:16,87.174828470609)
--(axis cs:17,85.1526505499782)
--(axis cs:18,80.4582979574536)
--(axis cs:19,84.8936972863401)
--(axis cs:20,89.7679487692103)
--(axis cs:21,81.8143212197546)
--(axis cs:22,84.3214499722072)
--(axis cs:23,87.2043135165059)
--(axis cs:24,81.1859355527005)
--(axis cs:25,83.6532195076945)
--(axis cs:26,85.2071096419863)
--(axis cs:27,85.1894374118183)
--(axis cs:28,84.0504978419354)
--(axis cs:29,84.8344153097055)
--(axis cs:30,85.9836789056759)
--(axis cs:31,81.3537863264207)
--(axis cs:32,84.0682171066734)
--(axis cs:33,83.5790736642637)
--(axis cs:34,83.4235580645219)
--(axis cs:35,80.4414221721208)
--(axis cs:36,81.4257224175197)
--(axis cs:37,82.2986309858488)
--(axis cs:38,84.0472722570938)
--(axis cs:39,86.8790862254353)
--(axis cs:40,78.365619470837)
--(axis cs:41,82.3633562213988)
--(axis cs:42,85.6229564069436)
--(axis cs:43,81.4979300045363)
--(axis cs:44,82.3760259818184)
--(axis cs:45,81.7780176485609)
--(axis cs:46,83.1818292615386)
--(axis cs:47,79.4919768543909)
--(axis cs:48,81.2611057344521)
--(axis cs:49,80.230311114962)
--(axis cs:50,87.3642885291093)
--(axis cs:51,88.4805021796663)
--(axis cs:52,87.385859629621)
--(axis cs:53,87.0502310545912)
--(axis cs:54,86.1298964110115)
--(axis cs:55,78.2073363914374)
--(axis cs:56,78.7966728669515)
--(axis cs:57,85.7863682677021)
--(axis cs:58,84.110077912271)
--(axis cs:59,78.5038290269565)
--(axis cs:60,81.2699001970921)
--(axis cs:61,77.1230011231047)
--(axis cs:62,80.7588916801275)
--(axis cs:63,74.2542327540141)
--(axis cs:64,72.6854570127348)
--(axis cs:65,72.3291243569163)
--(axis cs:66,73.9942880455012)
--(axis cs:67,69.1087141546446)
--(axis cs:68,66.9336136819559)
--(axis cs:69,68.1947525879786)
--(axis cs:70,69.022927485078)
--(axis cs:71,66.0065629985534)
--(axis cs:72,72.274824988921)
--(axis cs:73,66.4290051888409)
--(axis cs:74,66.3851469908804)
--(axis cs:75,69.2711289028937)
--(axis cs:76,67.8231037643146)
--(axis cs:77,67.9491427892323)
--(axis cs:78,68.9951241156627)
--(axis cs:79,56.9073693235527)
--(axis cs:79,130.025964009781)
--(axis cs:79,130.025964009781)
--(axis cs:78,130.071542551004)
--(axis cs:77,127.250857210768)
--(axis cs:76,132.243562902352)
--(axis cs:75,136.995537763773)
--(axis cs:74,138.21485300912)
--(axis cs:73,138.570994811159)
--(axis cs:72,132.191841677746)
--(axis cs:71,138.460103668113)
--(axis cs:70,143.243739181589)
--(axis cs:69,136.805247412021)
--(axis cs:68,146.199719651377)
--(axis cs:67,143.691285845355)
--(axis cs:66,142.472378621165)
--(axis cs:65,144.13754230975)
--(axis cs:64,150.247876320599)
--(axis cs:63,156.479100579319)
--(axis cs:62,155.507774986539)
--(axis cs:61,152.410332210229)
--(axis cs:60,156.863433136241)
--(axis cs:59,152.496170973043)
--(axis cs:58,145.289922087729)
--(axis cs:57,147.746965065631)
--(axis cs:56,152.269993799715)
--(axis cs:55,153.925996941896)
--(axis cs:54,154.670103588988)
--(axis cs:53,141.149768945409)
--(axis cs:52,149.947473703712)
--(axis cs:51,150.786164487)
--(axis cs:50,145.769044804224)
--(axis cs:49,143.836355551705)
--(axis cs:48,144.405560932215)
--(axis cs:47,147.774689812276)
--(axis cs:46,149.484837405128)
--(axis cs:45,139.421982351439)
--(axis cs:44,138.757307351515)
--(axis cs:43,144.302069995464)
--(axis cs:42,140.777043593056)
--(axis cs:41,141.103310445268)
--(axis cs:40,140.967713862496)
--(axis cs:39,140.187580441231)
--(axis cs:38,143.419394409573)
--(axis cs:37,138.301369014151)
--(axis cs:36,131.507610915814)
--(axis cs:35,134.691911161212)
--(axis cs:34,139.243108602145)
--(axis cs:33,129.687593002403)
--(axis cs:32,129.931782893327)
--(axis cs:31,137.179547006913)
--(axis cs:30,135.016321094324)
--(axis cs:29,133.098918023628)
--(axis cs:28,122.549502158065)
--(axis cs:27,127.277229254848)
--(axis cs:26,131.45955702468)
--(axis cs:25,128.946780492305)
--(axis cs:24,122.280731113966)
--(axis cs:23,123.262353150161)
--(axis cs:22,120.611883361126)
--(axis cs:21,114.385678780245)
--(axis cs:20,121.83205123079)
--(axis cs:19,121.039636046993)
--(axis cs:18,125.608368709213)
--(axis cs:17,117.114016116688)
--(axis cs:16,119.158504862724)
--(axis cs:15,128.712391003893)
--(axis cs:14,127.732150503786)
--(axis cs:13,123.112371040703)
--(axis cs:12,148.564666168247)
--(axis cs:11,159.261426430747)
--(axis cs:10,253.606175047181)
--(axis cs:9,224.45094559702)
--(axis cs:8,155.427078308443)
--(axis cs:7,110.905946357728)
--(axis cs:6,80.1751608958929)
--(axis cs:5,55.0004975032024)
--(axis cs:4,38.2597304466645)
--(axis cs:3,24.1702008846841)
--(axis cs:2,20.3964667510526)
--(axis cs:1,18.144155686355)
--(axis cs:0,11.2180627470025)
--cycle;

\path [fill=color2, fill opacity=0.3, very thin]
(axis cs:0,9.64232323056952)
--(axis cs:0,3.82434343609714)
--(axis cs:1,9.20330194927097)
--(axis cs:2,9.49563718660405)
--(axis cs:3,12.0795007548019)
--(axis cs:4,19.1121245776296)
--(axis cs:5,25.6513581773221)
--(axis cs:6,42.8818953089951)
--(axis cs:7,56.8742192917457)
--(axis cs:8,83.7092077369946)
--(axis cs:9,127.160578704085)
--(axis cs:10,117.002317326448)
--(axis cs:11,13.7615110021869)
--(axis cs:12,54.9891022671043)
--(axis cs:13,70.3869586780323)
--(axis cs:14,76.1736156753005)
--(axis cs:15,70.8525364470534)
--(axis cs:16,67.688241366432)
--(axis cs:17,61.666409764271)
--(axis cs:18,62.6432891095051)
--(axis cs:19,63.1428211726689)
--(axis cs:20,60.0238438427108)
--(axis cs:21,55.7867174691815)
--(axis cs:22,58.5267847542544)
--(axis cs:23,53.2949531086984)
--(axis cs:24,53.1689324835671)
--(axis cs:25,52.9460148835941)
--(axis cs:26,45.9461774132858)
--(axis cs:27,53.8030134391944)
--(axis cs:28,50.7490271015864)
--(axis cs:29,51.1020715061193)
--(axis cs:30,47.3042529392167)
--(axis cs:31,49.467467140557)
--(axis cs:32,47.5482982317612)
--(axis cs:33,46.5683411263121)
--(axis cs:34,45.7258936604791)
--(axis cs:35,49.8189293959329)
--(axis cs:36,45.6611559971047)
--(axis cs:37,47.8276858292952)
--(axis cs:38,48.8347071461541)
--(axis cs:39,44.0860881718074)
--(axis cs:40,50.3814126006513)
--(axis cs:41,47.1631968296806)
--(axis cs:42,49.1846387675344)
--(axis cs:43,46.7278874371462)
--(axis cs:44,48.0501141555662)
--(axis cs:45,48.5327965716399)
--(axis cs:46,43.4627056550882)
--(axis cs:47,46.6050159829993)
--(axis cs:48,48.7236565116996)
--(axis cs:49,47.1575908972689)
--(axis cs:50,43.8365142266228)
--(axis cs:51,40.4342388824772)
--(axis cs:52,44.0201964109275)
--(axis cs:53,41.3113796267226)
--(axis cs:54,45.9202310243408)
--(axis cs:55,43.4653394993679)
--(axis cs:56,42.1890658012609)
--(axis cs:57,42.9433156108103)
--(axis cs:58,43.0384258156989)
--(axis cs:59,40.5429553469175)
--(axis cs:60,41.3138389610716)
--(axis cs:61,43.5328523378601)
--(axis cs:62,42.8961580988342)
--(axis cs:63,39.9562330713272)
--(axis cs:64,40.1342234248562)
--(axis cs:65,41.4040163881802)
--(axis cs:66,38.9203518881509)
--(axis cs:67,41.708679690997)
--(axis cs:68,39.2969011381808)
--(axis cs:69,37.6475375743942)
--(axis cs:70,39.4369522394832)
--(axis cs:71,41.2047317244703)
--(axis cs:72,37.8557210792763)
--(axis cs:73,39.3989696279109)
--(axis cs:74,39.2601121851418)
--(axis cs:75,38.980961011502)
--(axis cs:76,34.7639786786291)
--(axis cs:77,41.6873196944318)
--(axis cs:78,41.5479834197412)
--(axis cs:79,39.0288911083077)
--(axis cs:79,81.3044422250256)
--(axis cs:79,81.3044422250256)
--(axis cs:78,82.1853499135921)
--(axis cs:77,88.5793469722348)
--(axis cs:76,74.0360213213709)
--(axis cs:75,81.019038988498)
--(axis cs:74,79.2732211481915)
--(axis cs:73,82.5343637054225)
--(axis cs:72,78.8776122540571)
--(axis cs:71,82.528601608863)
--(axis cs:70,81.4963810938501)
--(axis cs:69,79.9524624256058)
--(axis cs:68,84.8364321951525)
--(axis cs:67,88.4246536423364)
--(axis cs:66,89.9463147785158)
--(axis cs:65,91.1959836118198)
--(axis cs:64,90.0657765751438)
--(axis cs:63,90.9771002620061)
--(axis cs:62,86.5705085678325)
--(axis cs:61,89.7338143288066)
--(axis cs:60,93.0861610389284)
--(axis cs:59,86.7237113197491)
--(axis cs:58,87.7615741843011)
--(axis cs:57,87.4566843891897)
--(axis cs:56,88.9442675320725)
--(axis cs:55,87.0013271672987)
--(axis cs:54,89.0131023089925)
--(axis cs:53,87.2886203732774)
--(axis cs:52,91.3131369224058)
--(axis cs:51,85.8324277841895)
--(axis cs:50,87.4968191067106)
--(axis cs:49,85.9757424360644)
--(axis cs:48,85.4763434883004)
--(axis cs:47,82.928317350334)
--(axis cs:46,89.4706276782451)
--(axis cs:45,89.9338700950268)
--(axis cs:44,89.4165525111005)
--(axis cs:43,87.5387792295205)
--(axis cs:42,88.548694565799)
--(axis cs:41,91.0368031703193)
--(axis cs:40,88.4185873993488)
--(axis cs:39,84.1139118281926)
--(axis cs:38,91.6986261871792)
--(axis cs:37,87.7723141707047)
--(axis cs:36,87.605510669562)
--(axis cs:35,89.7144039374005)
--(axis cs:34,89.9407730061876)
--(axis cs:33,89.6983255403546)
--(axis cs:32,91.7850351015721)
--(axis cs:31,88.332532859443)
--(axis cs:30,89.36241372745)
--(axis cs:29,85.9645951605474)
--(axis cs:28,83.5843062317469)
--(axis cs:27,89.1969865608056)
--(axis cs:26,89.1871559200475)
--(axis cs:25,93.5873184497392)
--(axis cs:24,88.9644008497662)
--(axis cs:23,88.3050468913016)
--(axis cs:22,96.5398819124123)
--(axis cs:21,94.2799491974852)
--(axis cs:20,95.0428228239559)
--(axis cs:19,95.9238454939978)
--(axis cs:18,95.5567108904949)
--(axis cs:17,96.2669235690623)
--(axis cs:16,101.245091966901)
--(axis cs:15,100.747463552947)
--(axis cs:14,107.026384324699)
--(axis cs:13,104.479707988634)
--(axis cs:12,86.8108977328957)
--(axis cs:11,177.771822331146)
--(axis cs:10,266.597682673552)
--(axis cs:9,195.972754629248)
--(axis cs:8,137.157458929672)
--(axis cs:7,98.0591140415876)
--(axis cs:6,72.0514380243382)
--(axis cs:5,51.2153084893446)
--(axis cs:4,32.754542089037)
--(axis cs:3,24.1871659118648)
--(axis cs:2,17.9043628133959)
--(axis cs:1,15.9300313840624)
--(axis cs:0,9.64232323056952)
--cycle;

\addplot [semithick, color0]
table {%
0 6.63333333333333
1 13.9333333333333
2 15.2333333333333
3 18.7333333333333
4 24.2666666666667
5 40.5
6 57.2666666666667
7 84.8333333333333
8 114.433333333333
9 154.6
10 178.2
11 108.966666666667
12 88.0333333333333
13 96.2
14 101.666666666667
15 104.6
16 110.166666666667
17 114
18 115.866666666667
19 118.7
20 119.233333333333
21 124.5
22 129.733333333333
23 127.666666666667
24 132.966666666667
25 137.233333333333
26 142.8
27 148.133333333333
28 147.566666666667
29 151.8
30 150.5
31 157.9
32 170.1
33 166.4
34 169.166666666667
35 171.633333333333
36 175.966666666667
37 176.366666666667
38 175.866666666667
39 184.3
40 190.133333333333
41 193.7
42 193.4
43 194.633333333333
44 194.066666666667
45 201.333333333333
46 205.1
47 207.733333333333
48 216.233333333333
49 212.133333333333
50 211.066666666667
51 212.5
52 221.1
53 219.966666666667
54 220.266666666667
55 230.1
56 228.666666666667
57 227.7
58 232.4
59 230.2
60 237.7
61 234.566666666667
62 237.666666666667
63 239.966666666667
64 237.766666666667
65 242.966666666667
66 250.9
67 251.033333333333
68 253.866666666667
69 255.433333333333
70 253.766666666667
71 256.7
72 261.9
73 260.2
74 271.5
75 263.233333333333
76 273.1
77 275.7
78 278
79 285.2
};
\addplot [semithick, color1]
table {%
0 8.13333333333333
1 12.8
2 15.3666666666667
3 18.6666666666667
4 29.9333333333333
5 41.6
6 61.4
7 88.2
8 123.9
9 175.833333333333
10 169.033333333333
11 93.5333333333333
12 99.9666666666667
13 100.966666666667
14 107.066666666667
15 106.333333333333
16 103.166666666667
17 101.133333333333
18 103.033333333333
19 102.966666666667
20 105.8
21 98.1
22 102.466666666667
23 105.233333333333
24 101.733333333333
25 106.3
26 108.333333333333
27 106.233333333333
28 103.3
29 108.966666666667
30 110.5
31 109.266666666667
32 107
33 106.633333333333
34 111.333333333333
35 107.566666666667
36 106.466666666667
37 110.3
38 113.733333333333
39 113.533333333333
40 109.666666666667
41 111.733333333333
42 113.2
43 112.9
44 110.566666666667
45 110.6
46 116.333333333333
47 113.633333333333
48 112.833333333333
49 112.033333333333
50 116.566666666667
51 119.633333333333
52 118.666666666667
53 114.1
54 120.4
55 116.066666666667
56 115.533333333333
57 116.766666666667
58 114.7
59 115.5
60 119.066666666667
61 114.766666666667
62 118.133333333333
63 115.366666666667
64 111.466666666667
65 108.233333333333
66 108.233333333333
67 106.4
68 106.566666666667
69 102.5
70 106.133333333333
71 102.233333333333
72 102.233333333333
73 102.5
74 102.3
75 103.133333333333
76 100.033333333333
77 97.6
78 99.5333333333333
79 93.4666666666667
};
\addplot [semithick, color2]
table {%
0 6.73333333333333
1 12.5666666666667
2 13.7
3 18.1333333333333
4 25.9333333333333
5 38.4333333333333
6 57.4666666666667
7 77.4666666666667
8 110.433333333333
9 161.566666666667
10 191.8
11 95.7666666666667
12 70.9
13 87.4333333333333
14 91.6
15 85.8
16 84.4666666666667
17 78.9666666666667
18 79.1
19 79.5333333333333
20 77.5333333333333
21 75.0333333333333
22 77.5333333333333
23 70.8
24 71.0666666666667
25 73.2666666666667
26 67.5666666666667
27 71.5
28 67.1666666666667
29 68.5333333333333
30 68.3333333333333
31 68.9
32 69.6666666666667
33 68.1333333333333
34 67.8333333333333
35 69.7666666666667
36 66.6333333333333
37 67.8
38 70.2666666666667
39 64.1
40 69.4
41 69.1
42 68.8666666666667
43 67.1333333333333
44 68.7333333333333
45 69.2333333333333
46 66.4666666666667
47 64.7666666666667
48 67.1
49 66.5666666666667
50 65.6666666666667
51 63.1333333333333
52 67.6666666666667
53 64.3
54 67.4666666666667
55 65.2333333333333
56 65.5666666666667
57 65.2
58 65.4
59 63.6333333333333
60 67.2
61 66.6333333333333
62 64.7333333333333
63 65.4666666666667
64 65.1
65 66.3
66 64.4333333333333
67 65.0666666666667
68 62.0666666666667
69 58.8
70 60.4666666666667
71 61.8666666666667
72 58.3666666666667
73 60.9666666666667
74 59.2666666666667
75 60
76 54.4
77 65.1333333333333
78 61.8666666666667
79 60.1666666666667
};
\end{axis}

\end{tikzpicture}

%% file: pic/new22/22dic10k.tex
\begin{tikzpicture}

\definecolor{color0}{rgb}{0.886274509803922,0.290196078431373,0.2}
\definecolor{color1}{rgb}{0.203921568627451,0.541176470588235,0.741176470588235}
\definecolor{color2}{rgb}{0.596078431372549,0.556862745098039,0.835294117647059}
\pgfmathsetlengthmacro\MajorTickLength{
      \pgfkeysvalueof{/pgfplots/major tick length} * 0.5
    }
\begin{axis}[
axis line style={white},
legend cell align={left},
width = 0.37\linewidth,
height = 1in,
font= \fontsize{7}{7.5}\selectfont,
tick align=inside,
tick pos=left,
major tick length=\MajorTickLength,
x grid style={white},
xmajorgrids,
xmin=0, xmax=80,
xtick style={color=white!33.3333333333333!black},
y grid style={white},
x label style={at={(0.5, 0.4)}},
yticklabel style={rotate=90},
ymajorgrids,
y label style={at={(0.55,0.5)}},
xtick={0,20,40,60,80},
xticklabels={0,20,40,60,80},
ytick={0,120,240},
yticklabels={0,120,240},
ymin=-11.2132545625069, ymax=278.573363822753,
ytick style={color=white!33.3333333333333!black}
]
\path [fill=color0, fill opacity=0.3, very thin]
(axis cs:0,9.06594194335118)
--(axis cs:0,2.93405805664882)
--(axis cs:1,8.62293226985726)
--(axis cs:2,12.0121435554453)
--(axis cs:3,16.1822572007694)
--(axis cs:4,21.835626067594)
--(axis cs:5,30.1204838788353)
--(axis cs:6,43.5581351880034)
--(axis cs:7,62.1264452226143)
--(axis cs:8,98.7038309410506)
--(axis cs:9,146.847370733106)
--(axis cs:10,72.598755194759)
--(axis cs:11,7.46513903223571)
--(axis cs:12,42.0287453802363)
--(axis cs:13,66.0400534760314)
--(axis cs:14,71.3986755547733)
--(axis cs:15,70.9665214871942)
--(axis cs:16,66.9787071060383)
--(axis cs:17,71.528160402154)
--(axis cs:18,68.9666996699993)
--(axis cs:19,65.6451421766236)
--(axis cs:20,70.8359065268507)
--(axis cs:21,75.2250244618629)
--(axis cs:22,67.7394368240785)
--(axis cs:23,70.8)
--(axis cs:24,70.4523758210193)
--(axis cs:25,70.2340247609267)
--(axis cs:26,76.3406414642877)
--(axis cs:27,74.5448426771417)
--(axis cs:28,79.3629062837067)
--(axis cs:29,73.37763061155)
--(axis cs:30,73.977805145088)
--(axis cs:31,75.231786125308)
--(axis cs:32,71.7821757230843)
--(axis cs:33,78.3873768076308)
--(axis cs:34,72.1974028164079)
--(axis cs:35,73.8080671402723)
--(axis cs:36,75.5816285924492)
--(axis cs:37,76.9601061015423)
--(axis cs:38,76.7020662921307)
--(axis cs:39,81.380033222683)
--(axis cs:40,70.8541531270666)
--(axis cs:41,73.5863077902948)
--(axis cs:42,74.2265133429588)
--(axis cs:43,82.8303495502627)
--(axis cs:44,79.7218443323355)
--(axis cs:45,78.366710642117)
--(axis cs:46,82.0508865805692)
--(axis cs:47,68.1610373207469)
--(axis cs:48,82.032479952582)
--(axis cs:49,84.9591855578799)
--(axis cs:50,86.5643988785579)
--(axis cs:51,84.6088530254894)
--(axis cs:52,80.786020818794)
--(axis cs:53,85.6137347374025)
--(axis cs:54,75.3201930047422)
--(axis cs:55,86.0872877894867)
--(axis cs:56,83.3042064343033)
--(axis cs:57,80.9485821495612)
--(axis cs:58,85.0884150401091)
--(axis cs:59,84.7687880541236)
--(axis cs:60,80.9218280692822)
--(axis cs:61,81.8854957672493)
--(axis cs:62,90.2283973904889)
--(axis cs:63,85.888969992044)
--(axis cs:64,97.23467318412)
--(axis cs:65,88.3791817153117)
--(axis cs:66,90.9867451848101)
--(axis cs:67,93.0031416196276)
--(axis cs:68,88.2)
--(axis cs:69,86.2142267199903)
--(axis cs:70,76.4939470098361)
--(axis cs:71,84.0904396557427)
--(axis cs:72,83.0934792879151)
--(axis cs:73,72.5155987672273)
--(axis cs:74,77.4514916998847)
--(axis cs:75,79.362308824676)
--(axis cs:76,70.8042020516226)
--(axis cs:77,73.9369666639642)
--(axis cs:78,76.9773328299124)
--(axis cs:79,75.9921197561144)
--(axis cs:79,116.607880243886)
--(axis cs:79,116.607880243886)
--(axis cs:78,115.822667170088)
--(axis cs:77,122.863033336036)
--(axis cs:76,118.395797948377)
--(axis cs:75,111.237691175324)
--(axis cs:74,119.348508300115)
--(axis cs:73,134.684401232773)
--(axis cs:72,124.506520712085)
--(axis cs:71,120.709560344257)
--(axis cs:70,121.106052990164)
--(axis cs:69,131.18577328001)
--(axis cs:68,129.4)
--(axis cs:67,131.196858380372)
--(axis cs:66,127.21325481519)
--(axis cs:65,117.220818284688)
--(axis cs:64,117.96532681588)
--(axis cs:63,128.511030007956)
--(axis cs:62,118.371602609511)
--(axis cs:61,126.714504232751)
--(axis cs:60,110.678171930718)
--(axis cs:59,110.431211945876)
--(axis cs:58,115.311584959891)
--(axis cs:57,114.451417850439)
--(axis cs:56,104.695793565697)
--(axis cs:55,108.912712210513)
--(axis cs:54,108.479806995258)
--(axis cs:53,105.986265262597)
--(axis cs:52,109.413979181206)
--(axis cs:51,107.191146974511)
--(axis cs:50,97.8356011214422)
--(axis cs:49,102.64081444212)
--(axis cs:48,104.167520047418)
--(axis cs:47,92.8389626792531)
--(axis cs:46,96.1491134194308)
--(axis cs:45,103.633289357883)
--(axis cs:44,104.878155667664)
--(axis cs:43,102.569650449737)
--(axis cs:42,104.373486657041)
--(axis cs:41,102.813692209705)
--(axis cs:40,100.345846872933)
--(axis cs:39,93.419966777317)
--(axis cs:38,100.897933707869)
--(axis cs:37,92.0398938984577)
--(axis cs:36,93.0183714075508)
--(axis cs:35,98.5919328597277)
--(axis cs:34,103.002597183592)
--(axis cs:33,97.4126231923692)
--(axis cs:32,103.217824276916)
--(axis cs:31,100.368213874692)
--(axis cs:30,92.022194854912)
--(axis cs:29,100.22236938845)
--(axis cs:28,107.437093716293)
--(axis cs:27,93.4551573228583)
--(axis cs:26,100.659358535712)
--(axis cs:25,100.765975239073)
--(axis cs:24,92.5476241789807)
--(axis cs:23,93.4)
--(axis cs:22,96.8605631759215)
--(axis cs:21,100.774975538137)
--(axis cs:20,99.7640934731493)
--(axis cs:19,93.9548578233764)
--(axis cs:18,102.633300330001)
--(axis cs:17,93.871839597846)
--(axis cs:16,91.4212928939617)
--(axis cs:15,109.233478512806)
--(axis cs:14,101.601324445227)
--(axis cs:13,95.9599465239686)
--(axis cs:12,79.5712546197637)
--(axis cs:11,152.934860967764)
--(axis cs:10,265.401244805241)
--(axis cs:9,206.752629266894)
--(axis cs:8,148.296169058949)
--(axis cs:7,104.073554777386)
--(axis cs:6,68.4418648119966)
--(axis cs:5,59.8795161211647)
--(axis cs:4,39.764373932406)
--(axis cs:3,25.2177427992306)
--(axis cs:2,18.5878564445547)
--(axis cs:1,15.5770677301427)
--(axis cs:0,9.06594194335118)
--cycle;

\path [fill=color1, fill opacity=0.3, very thin]
(axis cs:0,9.21444737824533)
--(axis cs:0,4.58555262175467)
--(axis cs:1,6.63677896501555)
--(axis cs:2,9.98725133282073)
--(axis cs:3,12.1677134449412)
--(axis cs:4,16.33380448762)
--(axis cs:5,26.5880089719913)
--(axis cs:6,37.928944353115)
--(axis cs:7,55.5799688097614)
--(axis cs:8,79.8548963363676)
--(axis cs:9,108.966452847638)
--(axis cs:10,82.0711279653762)
--(axis cs:11,21.0935296056209)
--(axis cs:12,25.4742481117188)
--(axis cs:13,42.6669577885921)
--(axis cs:14,62.897853401526)
--(axis cs:15,65.0160239813394)
--(axis cs:16,59.1637390862007)
--(axis cs:17,57.4721969796969)
--(axis cs:18,54.4520116493001)
--(axis cs:19,48.5216730399062)
--(axis cs:20,49.1488298455005)
--(axis cs:21,47.9464429034077)
--(axis cs:22,43.9624004207213)
--(axis cs:23,41.382805916776)
--(axis cs:24,40.5906594124851)
--(axis cs:25,36.9981353197098)
--(axis cs:26,38.9365466512485)
--(axis cs:27,39.0931307245736)
--(axis cs:28,35.2076072849395)
--(axis cs:29,33.6584387213652)
--(axis cs:30,32.8234589970524)
--(axis cs:31,33.7926301635833)
--(axis cs:32,31.6331489686796)
--(axis cs:33,31.9976387383288)
--(axis cs:34,29.243526960356)
--(axis cs:35,27.7828462704041)
--(axis cs:36,29.0751177354687)
--(axis cs:37,30.2811601714043)
--(axis cs:38,26.0556042557099)
--(axis cs:39,26.7140895042806)
--(axis cs:40,28.3688371924678)
--(axis cs:41,25.1088268030442)
--(axis cs:42,24.8673992895175)
--(axis cs:43,21.5442409298324)
--(axis cs:44,21.9228997463705)
--(axis cs:45,22.0343183953236)
--(axis cs:46,23.2636639682649)
--(axis cs:47,20.6381020093799)
--(axis cs:48,21.0317191250044)
--(axis cs:49,17.3558086220591)
--(axis cs:50,20.0930538543924)
--(axis cs:51,21.5869088094498)
--(axis cs:52,19.9487374274104)
--(axis cs:53,20.281491120272)
--(axis cs:54,19.2556729600434)
--(axis cs:55,19.2495828543504)
--(axis cs:56,18.3027852177247)
--(axis cs:57,21.3940491999414)
--(axis cs:58,18.8033007345447)
--(axis cs:59,19.4874680559769)
--(axis cs:60,20.6005249452323)
--(axis cs:61,20.5967936393533)
--(axis cs:62,20.3216246074434)
--(axis cs:63,23.2493678326286)
--(axis cs:64,22.5012446641921)
--(axis cs:65,19.396078996603)
--(axis cs:66,21.2881892845646)
--(axis cs:67,23.2412703246138)
--(axis cs:68,21.046701124459)
--(axis cs:69,21.7876928899939)
--(axis cs:70,21.1631303902483)
--(axis cs:71,22.6894474492675)
--(axis cs:72,19.364457093408)
--(axis cs:73,19.7452599223205)
--(axis cs:74,18.8864670033485)
--(axis cs:75,20.1998596497449)
--(axis cs:76,16.5594543999684)
--(axis cs:77,17.4491674087428)
--(axis cs:78,20.0049293731792)
--(axis cs:79,17.7734913449074)
--(axis cs:79,56.0265086550926)
--(axis cs:79,56.0265086550926)
--(axis cs:78,54.2617372934874)
--(axis cs:77,55.8174992579239)
--(axis cs:76,47.8405456000316)
--(axis cs:75,51.8668070169218)
--(axis cs:74,48.0468663299848)
--(axis cs:73,48.0547400776795)
--(axis cs:72,49.035542906592)
--(axis cs:71,47.4438858840658)
--(axis cs:70,44.1035362764183)
--(axis cs:69,47.0789737766728)
--(axis cs:68,45.553298875541)
--(axis cs:67,44.0920630087195)
--(axis cs:66,48.1118107154354)
--(axis cs:65,46.603921003397)
--(axis cs:64,43.0320886691412)
--(axis cs:63,47.4172988340381)
--(axis cs:62,46.1450420592233)
--(axis cs:61,46.93653969398)
--(axis cs:60,45.9994750547677)
--(axis cs:59,49.3791986106898)
--(axis cs:58,45.7300325987886)
--(axis cs:57,43.8059508000586)
--(axis cs:56,44.2305481156087)
--(axis cs:55,47.8837504789829)
--(axis cs:54,45.5443270399566)
--(axis cs:53,43.0518422130613)
--(axis cs:52,49.6512625725896)
--(axis cs:51,44.3464245238835)
--(axis cs:50,46.1736128122743)
--(axis cs:49,48.5775247112742)
--(axis cs:48,46.5016142083289)
--(axis cs:47,47.4285646572867)
--(axis cs:46,52.8696693650684)
--(axis cs:45,50.2323482713431)
--(axis cs:44,48.3437669202962)
--(axis cs:43,50.0557590701676)
--(axis cs:42,49.1326007104825)
--(axis cs:41,48.8911731969558)
--(axis cs:40,55.4978294741989)
--(axis cs:39,50.7525771623861)
--(axis cs:38,59.0110624109567)
--(axis cs:37,52.652173161929)
--(axis cs:36,53.2582155978647)
--(axis cs:35,59.1504870629293)
--(axis cs:34,56.156473039644)
--(axis cs:33,56.4690279283378)
--(axis cs:32,61.7668510313204)
--(axis cs:31,61.54070316975)
--(axis cs:30,58.3765410029476)
--(axis cs:29,60.6748946119681)
--(axis cs:28,62.6590593817271)
--(axis cs:27,61.9068692754264)
--(axis cs:26,61.0634533487515)
--(axis cs:25,65.6018646802902)
--(axis cs:24,65.3426739208483)
--(axis cs:23,70.3505274165573)
--(axis cs:22,66.1042662459453)
--(axis cs:21,70.9868904299256)
--(axis cs:20,72.6511701544995)
--(axis cs:19,69.8783269600938)
--(axis cs:18,80.5479883506999)
--(axis cs:17,85.5944696869697)
--(axis cs:16,83.8362609137993)
--(axis cs:15,90.5173093519939)
--(axis cs:14,91.3688132651407)
--(axis cs:13,118.999708878075)
--(axis cs:12,142.125751888281)
--(axis cs:11,169.639803727712)
--(axis cs:10,241.462205367957)
--(axis cs:9,207.700213819028)
--(axis cs:8,148.211770330299)
--(axis cs:7,99.7533645235719)
--(axis cs:6,70.3377223135517)
--(axis cs:5,52.545324361342)
--(axis cs:4,35.1995288457133)
--(axis cs:3,25.9656198883921)
--(axis cs:2,19.4127486671793)
--(axis cs:1,16.6298877016511)
--(axis cs:0,9.21444737824533)
--cycle;

\path [fill=color2, fill opacity=0.3, very thin]
(axis cs:0,10.1741766747562)
--(axis cs:0,4.69248999191044)
--(axis cs:1,8.20314605470034)
--(axis cs:2,10.5902963294812)
--(axis cs:3,12.9493775380492)
--(axis cs:4,20.080454462664)
--(axis cs:5,26.9566546873842)
--(axis cs:6,40.1723250402285)
--(axis cs:7,54.8046223899303)
--(axis cs:8,81.4373017995485)
--(axis cs:9,113.530637864489)
--(axis cs:10,89.8099159511855)
--(axis cs:11,3.88551232605015)
--(axis cs:12,37.742150238944)
--(axis cs:13,48.1215954832616)
--(axis cs:14,49.9430556399865)
--(axis cs:15,44.3990303543146)
--(axis cs:16,44.1752130889518)
--(axis cs:17,39.8741720631232)
--(axis cs:18,37.3167339797565)
--(axis cs:19,32.8800737761946)
--(axis cs:20,33.6220794070638)
--(axis cs:21,30.2766952514721)
--(axis cs:22,28.1724762816396)
--(axis cs:23,28.4270722195337)
--(axis cs:24,25.7059543802744)
--(axis cs:25,24.1530648060861)
--(axis cs:26,23.8676442258094)
--(axis cs:27,23.3009921220817)
--(axis cs:28,23.1909029852167)
--(axis cs:29,21.4326477016393)
--(axis cs:30,21.0610376417415)
--(axis cs:31,21.0268326793799)
--(axis cs:32,21.1997333375999)
--(axis cs:33,19.3646249311157)
--(axis cs:34,16.2598659681511)
--(axis cs:35,16.850690634087)
--(axis cs:36,15.8642546730231)
--(axis cs:37,15.3548257475413)
--(axis cs:38,14.5389655715032)
--(axis cs:39,15.2995564117296)
--(axis cs:40,13.4240403527716)
--(axis cs:41,13.8438682368115)
--(axis cs:42,14.7585955186749)
--(axis cs:43,13.1125085034053)
--(axis cs:44,13.3569277796591)
--(axis cs:45,12.098833401381)
--(axis cs:46,13.2942122893471)
--(axis cs:47,12.8148515131678)
--(axis cs:48,11.4336550896509)
--(axis cs:49,11.2346731841201)
--(axis cs:50,12.2153391929602)
--(axis cs:51,9.20943816410102)
--(axis cs:52,8.193569065362)
--(axis cs:53,8.93701219428828)
--(axis cs:54,10.1321513745463)
--(axis cs:55,7.34314149902465)
--(axis cs:56,8.53214716881896)
--(axis cs:57,8.61119198738946)
--(axis cs:58,7.76961201801912)
--(axis cs:59,7.24892645107407)
--(axis cs:60,8.30534371080632)
--(axis cs:61,7.88838787233453)
--(axis cs:62,6.36157214380976)
--(axis cs:63,6.34318942888646)
--(axis cs:64,4.74712852610333)
--(axis cs:65,6.30248518536556)
--(axis cs:66,4.21799572168533)
--(axis cs:67,4.14833925374179)
--(axis cs:68,3.62469716416255)
--(axis cs:69,4.94735132027746)
--(axis cs:70,3.03286101219828)
--(axis cs:71,3.44235403300912)
--(axis cs:72,3.36033020909011)
--(axis cs:73,3.61862031792673)
--(axis cs:74,2.05312770820167)
--(axis cs:75,3.10331990957766)
--(axis cs:76,2.29771105451871)
--(axis cs:77,2.09975124839881)
--(axis cs:78,2.41334636690721)
--(axis cs:79,1.95886445500492)
--(axis cs:79,14.6411355449951)
--(axis cs:79,14.6411355449951)
--(axis cs:78,16.0533202997595)
--(axis cs:77,15.5002487516012)
--(axis cs:76,14.4356222788146)
--(axis cs:75,18.8300134237557)
--(axis cs:74,18.4802056251317)
--(axis cs:73,19.1147130154066)
--(axis cs:72,18.9730031242432)
--(axis cs:71,18.0909793003242)
--(axis cs:70,19.5004723211351)
--(axis cs:69,20.7859820130559)
--(axis cs:68,23.5753028358374)
--(axis cs:67,23.3849940795915)
--(axis cs:66,21.9820042783147)
--(axis cs:65,24.6308481479678)
--(axis cs:64,22.4528714738967)
--(axis cs:63,25.7234772377802)
--(axis cs:62,24.9050945228569)
--(axis cs:61,24.6449454609988)
--(axis cs:60,23.8946562891937)
--(axis cs:59,23.5510735489259)
--(axis cs:58,25.4970546486475)
--(axis cs:57,28.4554746792772)
--(axis cs:56,27.267852831181)
--(axis cs:55,29.9901918343087)
--(axis cs:54,28.9345152921203)
--(axis cs:53,28.2629878057117)
--(axis cs:52,31.006430934638)
--(axis cs:51,30.390561835899)
--(axis cs:50,29.5846608070398)
--(axis cs:49,31.96532681588)
--(axis cs:48,28.6996782436825)
--(axis cs:47,32.2518151534989)
--(axis cs:46,32.3057877106529)
--(axis cs:45,32.1011665986191)
--(axis cs:44,29.9097388870075)
--(axis cs:43,31.4874914965947)
--(axis cs:42,31.9747378146585)
--(axis cs:41,33.5561317631885)
--(axis cs:40,36.5092929805617)
--(axis cs:39,32.8337769216037)
--(axis cs:38,34.2610344284968)
--(axis cs:37,34.978507585792)
--(axis cs:36,36.1357453269769)
--(axis cs:35,34.949309365913)
--(axis cs:34,39.1401340318489)
--(axis cs:33,39.502041735551)
--(axis cs:32,37.8669333290668)
--(axis cs:31,41.3731673206201)
--(axis cs:30,40.8056290249251)
--(axis cs:29,42.500685631694)
--(axis cs:28,42.7424303481166)
--(axis cs:27,40.0990078779183)
--(axis cs:26,46.5990224408573)
--(axis cs:25,44.1802685272472)
--(axis cs:24,43.4273789530589)
--(axis cs:23,49.9062611137996)
--(axis cs:22,50.1608570516938)
--(axis cs:21,49.3233047485279)
--(axis cs:20,50.9112539262696)
--(axis cs:19,59.3865928904721)
--(axis cs:18,57.9499326869102)
--(axis cs:17,60.5924946035435)
--(axis cs:16,65.3581202443815)
--(axis cs:15,62.7343029790187)
--(axis cs:14,72.7902776933468)
--(axis cs:13,72.7450711834051)
--(axis cs:12,65.457849761056)
--(axis cs:11,149.847821007283)
--(axis cs:10,255.190084048814)
--(axis cs:9,205.869362135511)
--(axis cs:8,155.629364867118)
--(axis cs:7,111.528710943403)
--(axis cs:6,76.4943416264382)
--(axis cs:5,55.7100119792824)
--(axis cs:4,41.9862122040027)
--(axis cs:3,26.7839557952841)
--(axis cs:2,20.6763703371855)
--(axis cs:1,15.2635206119663)
--(axis cs:0,10.1741766747562)
--cycle;

\addplot [semithick, color0]
table {%
0 6
1 12.1
2 15.3
3 20.7
4 30.8
5 45
6 56
7 83.1
8 123.5
9 176.8
10 169
11 80.2
12 60.8
13 81
14 86.5
15 90.1
16 79.2
17 82.7
18 85.8
19 79.8
20 85.3
21 88
22 82.3
23 82.1
24 81.5
25 85.5
26 88.5
27 84
28 93.4
29 86.8
30 83
31 87.8
32 87.5
33 87.9
34 87.6
35 86.2
36 84.3
37 84.5
38 88.8
39 87.4
40 85.6
41 88.2
42 89.3
43 92.7
44 92.3
45 91
46 89.1
47 80.5
48 93.1
49 93.8
50 92.2
51 95.9
52 95.1
53 95.8
54 91.9
55 97.5
56 94
57 97.7
58 100.2
59 97.6
60 95.8
61 104.3
62 104.3
63 107.2
64 107.6
65 102.8
66 109.1
67 112.1
68 108.8
69 108.7
70 98.8
71 102.4
72 103.8
73 103.6
74 98.4
75 95.3
76 94.6
77 98.4
78 96.4
79 96.3
};
\addplot [semithick, color1]
table {%
0 6.9
1 11.6333333333333
2 14.7
3 19.0666666666667
4 25.7666666666667
5 39.5666666666667
6 54.1333333333333
7 77.6666666666667
8 114.033333333333
9 158.333333333333
10 161.766666666667
11 95.3666666666667
12 83.8
13 80.8333333333333
14 77.1333333333333
15 77.7666666666667
16 71.5
17 71.5333333333333
18 67.5
19 59.2
20 60.9
21 59.4666666666667
22 55.0333333333333
23 55.8666666666667
24 52.9666666666667
25 51.3
26 50
27 50.5
28 48.9333333333333
29 47.1666666666667
30 45.6
31 47.6666666666667
32 46.7
33 44.2333333333333
34 42.7
35 43.4666666666667
36 41.1666666666667
37 41.4666666666667
38 42.5333333333333
39 38.7333333333333
40 41.9333333333333
41 37
42 37
43 35.8
44 35.1333333333333
45 36.1333333333333
46 38.0666666666667
47 34.0333333333333
48 33.7666666666667
49 32.9666666666667
50 33.1333333333333
51 32.9666666666667
52 34.8
53 31.6666666666667
54 32.4
55 33.5666666666667
56 31.2666666666667
57 32.6
58 32.2666666666667
59 34.4333333333333
60 33.3
61 33.7666666666667
62 33.2333333333333
63 35.3333333333333
64 32.7666666666667
65 33
66 34.7
67 33.6666666666667
68 33.3
69 34.4333333333333
70 32.6333333333333
71 35.0666666666667
72 34.2
73 33.9
74 33.4666666666667
75 36.0333333333333
76 32.2
77 36.6333333333333
78 37.1333333333333
79 36.9
};
\addplot [semithick, color2]
table {%
0 7.43333333333333
1 11.7333333333333
2 15.6333333333333
3 19.8666666666667
4 31.0333333333333
5 41.3333333333333
6 58.3333333333333
7 83.1666666666667
8 118.533333333333
9 159.7
10 172.5
11 76.8666666666667
12 51.6
13 60.4333333333333
14 61.3666666666667
15 53.5666666666667
16 54.7666666666667
17 50.2333333333333
18 47.6333333333333
19 46.1333333333333
20 42.2666666666667
21 39.8
22 39.1666666666667
23 39.1666666666667
24 34.5666666666667
25 34.1666666666667
26 35.2333333333333
27 31.7
28 32.9666666666667
29 31.9666666666667
30 30.9333333333333
31 31.2
32 29.5333333333333
33 29.4333333333333
34 27.7
35 25.9
36 26
37 25.1666666666667
38 24.4
39 24.0666666666667
40 24.9666666666667
41 23.7
42 23.3666666666667
43 22.3
44 21.6333333333333
45 22.1
46 22.8
47 22.5333333333333
48 20.0666666666667
49 21.6
50 20.9
51 19.8
52 19.6
53 18.6
54 19.5333333333333
55 18.6666666666667
56 17.9
57 18.5333333333333
58 16.6333333333333
59 15.4
60 16.1
61 16.2666666666667
62 15.6333333333333
63 16.0333333333333
64 13.6
65 15.4666666666667
66 13.1
67 13.7666666666667
68 13.6
69 12.8666666666667
70 11.2666666666667
71 10.7666666666667
72 11.1666666666667
73 11.3666666666667
74 10.2666666666667
75 10.9666666666667
76 8.36666666666667
77 8.8
78 9.23333333333333
79 8.3
};
\end{axis}

\end{tikzpicture}

%% file: pic/new22/22dic25k.tex
\begin{tikzpicture}

\definecolor{color0}{rgb}{0.886274509803922,0.290196078431373,0.2}
\definecolor{color1}{rgb}{0.203921568627451,0.541176470588235,0.741176470588235}
\definecolor{color2}{rgb}{0.596078431372549,0.556862745098039,0.835294117647059}
\pgfmathsetlengthmacro\MajorTickLength{
      \pgfkeysvalueof{/pgfplots/major tick length} * 0.5
    }
\begin{axis}[
axis line style={white},
legend cell align={left},
width = 0.37\linewidth,
height = 1in,
major tick length=\MajorTickLength,
font= \fontsize{7}{7.5}\selectfont,
legend style={inner xsep=1pt, inner ysep=-1pt, row sep=-3pt,at={(0.25,1.3)},anchor=north west},
tick align=inside,
tick pos=left,
x grid style={white},
xmajorgrids,
xmin=0, xmax=80,
yticklabel style={rotate=90},
xtick style={color=white!33.3333333333333!black},
y grid style={white},
x label style={at={(0.5, 0.4)}},
ymajorgrids,
y label style={at={(0.55,0.5)}},
xtick={0,20,40,60,80},
xticklabels={0,20,40,60,80},
ytick={0,120,240},
yticklabels={0,120,240},
ymin=-16.2998146735476, ymax=258.212972491561,
ytick style={color=white!33.3333333333333!black}
]
\path [fill=color0, fill opacity=0.3, very thin]
(axis cs:0,8.9615177241877)
--(axis cs:0,3.57181560914563)
--(axis cs:1,8.31754580810136)
--(axis cs:2,10.1770794277827)
--(axis cs:3,11.419679698375)
--(axis cs:4,16.1144615741542)
--(axis cs:5,26.8056156948787)
--(axis cs:6,33.0303748268721)
--(axis cs:7,47.7856361776929)
--(axis cs:8,71.6922898473192)
--(axis cs:9,102.252955817361)
--(axis cs:10,103.473765500115)
--(axis cs:11,23.415811511877)
--(axis cs:12,6.63160768960211)
--(axis cs:13,27.3399900403639)
--(axis cs:14,58.5645399872579)
--(axis cs:15,63.6973121457529)
--(axis cs:16,64.9358589482865)
--(axis cs:17,63.3940491999414)
--(axis cs:18,64.5404440685062)
--(axis cs:19,62.7147415247239)
--(axis cs:20,62.7011575004496)
--(axis cs:21,57.8968233896754)
--(axis cs:22,61.2429512658796)
--(axis cs:23,57.4963186173979)
--(axis cs:24,55.281079576068)
--(axis cs:25,51.2388746571528)
--(axis cs:26,51.5569261635653)
--(axis cs:27,52.1905818510822)
--(axis cs:28,49.8732041856715)
--(axis cs:29,49.630683123147)
--(axis cs:30,52.8040534340068)
--(axis cs:31,50.8705774078449)
--(axis cs:32,46.4236067282882)
--(axis cs:33,46.899351316303)
--(axis cs:34,47.7935917179205)
--(axis cs:35,44.4572900118863)
--(axis cs:36,44.1376743472205)
--(axis cs:37,44.0987673258154)
--(axis cs:38,46.1500662429226)
--(axis cs:39,44.7962672240113)
--(axis cs:40,44.38442890597)
--(axis cs:41,44.6967823879791)
--(axis cs:42,43.3509291467686)
--(axis cs:43,40.8920638896343)
--(axis cs:44,42.9569445292259)
--(axis cs:45,40.8401061213822)
--(axis cs:46,41.2928706497503)
--(axis cs:47,42.0729310766944)
--(axis cs:48,42.680293044225)
--(axis cs:49,41.1836339625464)
--(axis cs:50,43.3683334677636)
--(axis cs:51,42.7386937760403)
--(axis cs:52,42.6215302985891)
--(axis cs:53,44.221764875533)
--(axis cs:54,41.9630388588494)
--(axis cs:55,41.6451185240297)
--(axis cs:56,39.3839125438268)
--(axis cs:57,40.8051689942572)
--(axis cs:58,38.5947296511209)
--(axis cs:59,42.3208512498699)
--(axis cs:60,39.1958707723255)
--(axis cs:61,39.5189661174844)
--(axis cs:62,39.8008795380927)
--(axis cs:63,40.6981184019343)
--(axis cs:64,38.4536329620909)
--(axis cs:65,40.6017255952906)
--(axis cs:66,39.6124622709376)
--(axis cs:67,38.5825392486611)
--(axis cs:68,35.9509855211014)
--(axis cs:69,37.5569312037457)
--(axis cs:70,39.0840549809687)
--(axis cs:71,37.1696873439608)
--(axis cs:72,38.3151751340551)
--(axis cs:73,35.7893777856003)
--(axis cs:74,37.1454722561778)
--(axis cs:75,35.7282727060359)
--(axis cs:76,35.4029898372651)
--(axis cs:77,35.5872139629221)
--(axis cs:78,38.3204461270471)
--(axis cs:79,37.324280786641)
--(axis cs:79,65.8090525466924)
--(axis cs:79,65.8090525466924)
--(axis cs:78,66.6128872062863)
--(axis cs:77,65.1461193704112)
--(axis cs:76,65.1303434960682)
--(axis cs:75,66.4717272939641)
--(axis cs:74,68.2545277438222)
--(axis cs:73,66.3439555477331)
--(axis cs:72,69.4181581992782)
--(axis cs:71,67.6969793227059)
--(axis cs:70,67.182611685698)
--(axis cs:69,72.3764021295877)
--(axis cs:68,71.1823478122319)
--(axis cs:67,67.6174607513389)
--(axis cs:66,64.9208710623957)
--(axis cs:65,68.9316077380427)
--(axis cs:64,65.7463670379092)
--(axis cs:63,69.635214931399)
--(axis cs:62,65.0657871285739)
--(axis cs:61,64.2810338825156)
--(axis cs:60,69.3374625610078)
--(axis cs:59,66.5458154167967)
--(axis cs:58,67.2719370155457)
--(axis cs:57,66.5948310057428)
--(axis cs:56,62.8827541228399)
--(axis cs:55,64.3548814759703)
--(axis cs:54,60.0369611411506)
--(axis cs:53,62.578235124467)
--(axis cs:52,61.1784697014109)
--(axis cs:51,63.4613062239597)
--(axis cs:50,64.0316665322364)
--(axis cs:49,64.2830327041202)
--(axis cs:48,64.519706955775)
--(axis cs:47,64.0604022566389)
--(axis cs:46,68.5737960169164)
--(axis cs:45,64.2932272119511)
--(axis cs:44,65.7097221374408)
--(axis cs:43,66.241269443699)
--(axis cs:42,62.7824041865648)
--(axis cs:41,67.8365509453543)
--(axis cs:40,66.9489044273633)
--(axis cs:39,69.8037327759887)
--(axis cs:38,67.1166004237441)
--(axis cs:37,66.6345660075179)
--(axis cs:36,69.7289923194461)
--(axis cs:35,75.276043321447)
--(axis cs:34,68.6064082820795)
--(axis cs:33,69.1673153503637)
--(axis cs:32,67.1097266050451)
--(axis cs:31,73.5960892588217)
--(axis cs:30,74.7292798993265)
--(axis cs:29,74.369316876853)
--(axis cs:28,77.0601291476618)
--(axis cs:27,71.0760848155844)
--(axis cs:26,76.776407169768)
--(axis cs:25,75.0944586761805)
--(axis cs:24,76.8522537572653)
--(axis cs:23,74.7036813826021)
--(axis cs:22,80.7570487341204)
--(axis cs:21,79.2365099436579)
--(axis cs:20,77.0988424995504)
--(axis cs:19,81.6185918086094)
--(axis cs:18,87.6595559314938)
--(axis cs:17,85.8059508000586)
--(axis cs:16,82.5308077183802)
--(axis cs:15,86.4360211875804)
--(axis cs:14,87.3021266794088)
--(axis cs:13,112.126676626303)
--(axis cs:12,106.435058977065)
--(axis cs:11,225.784188488123)
--(axis cs:10,239.326234499885)
--(axis cs:9,186.547044182639)
--(axis cs:8,130.641043486014)
--(axis cs:7,89.8810304889738)
--(axis cs:6,62.7029585064613)
--(axis cs:5,45.7943843051213)
--(axis cs:4,32.6188717591791)
--(axis cs:3,24.0469869682916)
--(axis cs:2,17.9562539055506)
--(axis cs:1,16.215787525232)
--(axis cs:0,8.9615177241877)
--cycle;

\path [fill=color1, fill opacity=0.3, very thin]
(axis cs:0,10.4706798203088)
--(axis cs:0,4.92932017969115)
--(axis cs:1,7.02648125062592)
--(axis cs:2,10.2093331464873)
--(axis cs:3,14.1687060055863)
--(axis cs:4,17.9153134842161)
--(axis cs:5,26.1244585797791)
--(axis cs:6,39.8914562303688)
--(axis cs:7,57.532578341129)
--(axis cs:8,85.6338771976528)
--(axis cs:9,118.582902975403)
--(axis cs:10,78.289954109052)
--(axis cs:11,-3.82196071149725)
--(axis cs:12,19.5900484265376)
--(axis cs:13,52.3241112507453)
--(axis cs:14,50.3745045579778)
--(axis cs:15,47.1182770255726)
--(axis cs:16,46.3669510849149)
--(axis cs:17,44.3897965468443)
--(axis cs:18,39.6063649004676)
--(axis cs:19,35.3360335702052)
--(axis cs:20,33.6306602065441)
--(axis cs:21,32.0902143183062)
--(axis cs:22,32.1507193309258)
--(axis cs:23,28.5547689590201)
--(axis cs:24,24.5626561850602)
--(axis cs:25,23.9359811257355)
--(axis cs:26,23.530458002665)
--(axis cs:27,21.7733734638481)
--(axis cs:28,18.6234899302637)
--(axis cs:29,20.3174344966747)
--(axis cs:30,18.3973013869272)
--(axis cs:31,18.7471765193928)
--(axis cs:32,17.2185423949359)
--(axis cs:33,17.737203383896)
--(axis cs:34,16.714008894806)
--(axis cs:35,15.5097656157622)
--(axis cs:36,15.4918135890388)
--(axis cs:37,14.0868952580316)
--(axis cs:38,13.1625401343802)
--(axis cs:39,12.4957377015867)
--(axis cs:40,11.653884431564)
--(axis cs:41,12.4377493289113)
--(axis cs:42,11.3157747926499)
--(axis cs:43,9.81026400760151)
--(axis cs:44,9.53480840400075)
--(axis cs:45,11.2514388559474)
--(axis cs:46,9.59687576256715)
--(axis cs:47,10.8402111651708)
--(axis cs:48,10.9833647799862)
--(axis cs:49,8.82202451874853)
--(axis cs:50,9.62383346397074)
--(axis cs:51,8.77647411800146)
--(axis cs:52,9.01504739220787)
--(axis cs:53,8.72365158552486)
--(axis cs:54,9.22288485673528)
--(axis cs:55,9.07157128060221)
--(axis cs:56,6.48733974589129)
--(axis cs:57,6.58426493263742)
--(axis cs:58,7.25855929804108)
--(axis cs:59,7.18729583412817)
--(axis cs:60,6.40289946451752)
--(axis cs:61,6.26416110902721)
--(axis cs:62,4.50220624025706)
--(axis cs:63,5.86900656836637)
--(axis cs:64,6.07580767148084)
--(axis cs:65,4.70063617865788)
--(axis cs:66,6.38656916644663)
--(axis cs:67,3.93373738857925)
--(axis cs:68,4.69117115812125)
--(axis cs:69,4.43393289250888)
--(axis cs:70,4.34241362984282)
--(axis cs:71,4.1308464634415)
--(axis cs:72,4.34527484277547)
--(axis cs:73,4.5770568724513)
--(axis cs:74,4.18198940620979)
--(axis cs:75,4.38689691885328)
--(axis cs:76,3.59549755009568)
--(axis cs:77,5.58668881512202)
--(axis cs:78,3.69565926734319)
--(axis cs:79,3.50234227791174)
--(axis cs:79,11.5643243887549)
--(axis cs:79,11.5643243887549)
--(axis cs:78,13.1710073993235)
--(axis cs:77,13.613311184878)
--(axis cs:76,13.471169116571)
--(axis cs:75,12.8797697478134)
--(axis cs:74,15.5513439271235)
--(axis cs:73,16.2229431275487)
--(axis cs:72,12.5213918238912)
--(axis cs:71,13.0691535365585)
--(axis cs:70,15.5909197034905)
--(axis cs:69,13.6994004408245)
--(axis cs:68,14.6421621752121)
--(axis cs:67,14.9329292780874)
--(axis cs:66,13.68009750022)
--(axis cs:65,13.4326971546755)
--(axis cs:64,15.7908589951858)
--(axis cs:63,13.464326764967)
--(axis cs:62,18.0977937597429)
--(axis cs:61,15.1358388909728)
--(axis cs:60,15.5971005354825)
--(axis cs:59,15.9460374992052)
--(axis cs:58,14.9414407019589)
--(axis cs:57,17.6157350673626)
--(axis cs:56,17.645993587442)
--(axis cs:55,18.1284287193978)
--(axis cs:54,18.9104484765981)
--(axis cs:53,20.2096817478085)
--(axis cs:52,18.4516192744588)
--(axis cs:51,17.4901925486652)
--(axis cs:50,20.0428332026959)
--(axis cs:49,21.3113088145848)
--(axis cs:48,19.8166352200138)
--(axis cs:47,22.6264555014959)
--(axis cs:46,22.4031242374328)
--(axis cs:45,21.4818944773859)
--(axis cs:44,24.5985249293326)
--(axis cs:43,23.6564026590652)
--(axis cs:42,22.2175585406835)
--(axis cs:41,24.7622506710887)
--(axis cs:40,24.8794489017693)
--(axis cs:39,26.8375956317466)
--(axis cs:38,27.4374598656198)
--(axis cs:37,29.4464380753017)
--(axis cs:36,27.3081864109612)
--(axis cs:35,28.3569010509045)
--(axis cs:34,33.7526577718607)
--(axis cs:33,32.662796616104)
--(axis cs:32,31.5814576050641)
--(axis cs:31,34.7861568139405)
--(axis cs:30,34.8693652797395)
--(axis cs:29,35.149232169992)
--(axis cs:28,37.7098434030696)
--(axis cs:27,39.4932932028186)
--(axis cs:26,40.5362086640017)
--(axis cs:25,42.3973522075978)
--(axis cs:24,44.2373438149397)
--(axis cs:23,48.6452310409799)
--(axis cs:22,47.9826140024075)
--(axis cs:21,48.1097856816938)
--(axis cs:20,53.1693397934558)
--(axis cs:19,56.7306330964615)
--(axis cs:18,55.660301766199)
--(axis cs:17,65.743536786489)
--(axis cs:16,65.8997155817517)
--(axis cs:15,70.748389641094)
--(axis cs:14,76.0254954420222)
--(axis cs:13,77.8758887492547)
--(axis cs:12,95.0766182401291)
--(axis cs:11,156.688627378164)
--(axis cs:10,235.910045890948)
--(axis cs:9,210.883763691264)
--(axis cs:8,148.966122802347)
--(axis cs:7,116.400754992204)
--(axis cs:6,71.2418771029646)
--(axis cs:5,52.6755414202209)
--(axis cs:4,37.4180198491172)
--(axis cs:3,26.6979606610804)
--(axis cs:2,19.724000186846)
--(axis cs:1,17.5735187493741)
--(axis cs:0,10.4706798203088)
--cycle;

\path [fill=color2, fill opacity=0.3, very thin]
(axis cs:0,10.3519689920734)
--(axis cs:0,4.84803100792663)
--(axis cs:1,8.78188768463829)
--(axis cs:2,10.9860690779754)
--(axis cs:3,11.5614092501027)
--(axis cs:4,18.8151645813178)
--(axis cs:5,29.7687770890824)
--(axis cs:6,38.1341477776902)
--(axis cs:7,57.4510179648637)
--(axis cs:8,80.1441174471257)
--(axis cs:9,118.251291847014)
--(axis cs:10,86.3982148038231)
--(axis cs:11,8.60299790678503)
--(axis cs:12,17.5018731632194)
--(axis cs:13,42.4860421115864)
--(axis cs:14,47.5819798857559)
--(axis cs:15,44.2186063877082)
--(axis cs:16,42.2124922305135)
--(axis cs:17,37.6178797305842)
--(axis cs:18,31.0731278876955)
--(axis cs:19,33.3391554568393)
--(axis cs:20,27.8955010120446)
--(axis cs:21,26.2489406078482)
--(axis cs:22,25.0897343432203)
--(axis cs:23,24.7627334708888)
--(axis cs:24,23.2604544302749)
--(axis cs:25,20.2368391317282)
--(axis cs:26,21.8232665918414)
--(axis cs:27,17.4402303844354)
--(axis cs:28,16.534523904443)
--(axis cs:29,16.6601995528274)
--(axis cs:30,15.6735150454064)
--(axis cs:31,15.8260183560849)
--(axis cs:32,14.0023813321182)
--(axis cs:33,12.3857446752824)
--(axis cs:34,12.6341445864542)
--(axis cs:35,13.0302799642894)
--(axis cs:36,11.318210032228)
--(axis cs:37,12.4702909258813)
--(axis cs:38,11.8815302145872)
--(axis cs:39,9.77486185505708)
--(axis cs:40,10.6533466400665)
--(axis cs:41,10.5988406673657)
--(axis cs:42,8.54302596527083)
--(axis cs:43,9.45063101465288)
--(axis cs:44,8.46301923494049)
--(axis cs:45,8.48802427642881)
--(axis cs:46,8.71187744259768)
--(axis cs:47,8.10173642519158)
--(axis cs:48,8.88521123830163)
--(axis cs:49,6.75646190550941)
--(axis cs:50,6.92099972716552)
--(axis cs:51,5.86788442424963)
--(axis cs:52,6.7770568724513)
--(axis cs:53,4.74111019668485)
--(axis cs:54,4.50557984013134)
--(axis cs:55,6.19415504358323)
--(axis cs:56,5.22049323228652)
--(axis cs:57,5.810183625931)
--(axis cs:58,3.98975224494828)
--(axis cs:59,5.60597281152818)
--(axis cs:60,4.04778212846193)
--(axis cs:61,3.61930011995832)
--(axis cs:62,3.24652692719618)
--(axis cs:63,4.2642692054003)
--(axis cs:64,3.86819067527067)
--(axis cs:65,4.40193831214618)
--(axis cs:66,3.61964568164778)
--(axis cs:67,2.68347554245746)
--(axis cs:68,3.21559841600043)
--(axis cs:69,1.99903388406135)
--(axis cs:70,2.85192803625046)
--(axis cs:71,3.82701562323159)
--(axis cs:72,1.66016569401471)
--(axis cs:73,2.62581607156274)
--(axis cs:74,3.58510959995746)
--(axis cs:75,2.35180025927468)
--(axis cs:76,3.24339137681025)
--(axis cs:77,1.52542285525853)
--(axis cs:78,2.41440234076118)
--(axis cs:79,1.82474806791967)
--(axis cs:79,12.5752519320803)
--(axis cs:79,12.5752519320803)
--(axis cs:78,15.0522643259055)
--(axis cs:77,16.1412438114081)
--(axis cs:76,12.6232752898564)
--(axis cs:75,14.8481997407253)
--(axis cs:74,13.7482237333759)
--(axis cs:73,12.2408505951039)
--(axis cs:72,15.3398343059853)
--(axis cs:71,13.5063177101017)
--(axis cs:70,12.2147386304162)
--(axis cs:69,15.8009661159386)
--(axis cs:68,12.3177349173329)
--(axis cs:67,12.4498577908759)
--(axis cs:66,13.7803543183522)
--(axis cs:65,15.8647283545205)
--(axis cs:64,15.5318093247293)
--(axis cs:63,13.5357307945997)
--(axis cs:62,16.2868064061372)
--(axis cs:61,15.114033213375)
--(axis cs:60,17.9522178715381)
--(axis cs:59,18.9940271884718)
--(axis cs:58,16.1435810883851)
--(axis cs:57,18.4564830407357)
--(axis cs:56,18.6461734343801)
--(axis cs:55,17.6058449564168)
--(axis cs:54,21.027753493202)
--(axis cs:53,19.5922231366485)
--(axis cs:52,18.4229431275487)
--(axis cs:51,20.4654489090837)
--(axis cs:50,19.5456669395011)
--(axis cs:49,20.3768714278239)
--(axis cs:48,22.0481220950317)
--(axis cs:47,17.6982635748084)
--(axis cs:46,20.954789224069)
--(axis cs:45,19.9119757235712)
--(axis cs:44,21.8703140983928)
--(axis cs:43,22.2160356520138)
--(axis cs:42,24.1236407013958)
--(axis cs:41,25.9344926659676)
--(axis cs:40,24.0133200266002)
--(axis cs:39,25.3584714782763)
--(axis cs:38,26.5184697854128)
--(axis cs:37,24.7297090741187)
--(axis cs:36,26.6151233011053)
--(axis cs:35,25.7697200357106)
--(axis cs:34,27.6991887468792)
--(axis cs:33,29.5475886580509)
--(axis cs:32,28.9309520012152)
--(axis cs:31,28.4406483105818)
--(axis cs:30,30.5931516212603)
--(axis cs:29,32.473133780506)
--(axis cs:28,31.465476095557)
--(axis cs:27,35.9597696155646)
--(axis cs:26,33.6434000748253)
--(axis cs:25,35.7631608682718)
--(axis cs:24,39.0062122363917)
--(axis cs:23,40.3039331957778)
--(axis cs:22,41.443598990113)
--(axis cs:21,43.8177260588184)
--(axis cs:20,47.9044989879554)
--(axis cs:19,50.1275112098274)
--(axis cs:18,50.3268721123045)
--(axis cs:17,55.2487869360824)
--(axis cs:16,57.8541744361531)
--(axis cs:15,64.9813936122918)
--(axis cs:14,64.6180201142441)
--(axis cs:13,74.5806245550802)
--(axis cs:12,98.7647935034473)
--(axis cs:11,195.397002093215)
--(axis cs:10,245.73511852951)
--(axis cs:9,212.215374819653)
--(axis cs:8,152.255882552874)
--(axis cs:7,105.748982035136)
--(axis cs:6,76.3325188889765)
--(axis cs:5,55.5645562442509)
--(axis cs:4,36.3848354186822)
--(axis cs:3,26.5719240832307)
--(axis cs:2,19.6139309220246)
--(axis cs:1,19.351445648695)
--(axis cs:0,10.3519689920734)
--cycle;

\addplot [semithick, color0]
table {%
0 6.26666666666667
1 12.2666666666667
2 14.0666666666667
3 17.7333333333333
4 24.3666666666667
5 36.3
6 47.8666666666667
7 68.8333333333333
8 101.166666666667
9 144.4
10 171.4
11 124.6
12 56.5333333333333
13 69.7333333333333
14 72.9333333333333
15 75.0666666666667
16 73.7333333333333
17 74.6
18 76.1
19 72.1666666666667
20 69.9
21 68.5666666666667
22 71
23 66.1
24 66.0666666666667
25 63.1666666666667
26 64.1666666666667
27 61.6333333333333
28 63.4666666666667
29 62
30 63.7666666666667
31 62.2333333333333
32 56.7666666666667
33 58.0333333333333
34 58.2
35 59.8666666666667
36 56.9333333333333
37 55.3666666666667
38 56.6333333333333
39 57.3
40 55.6666666666667
41 56.2666666666667
42 53.0666666666667
43 53.5666666666667
44 54.3333333333333
45 52.5666666666667
46 54.9333333333333
47 53.0666666666667
48 53.6
49 52.7333333333333
50 53.7
51 53.1
52 51.9
53 53.4
54 51
55 53
56 51.1333333333333
57 53.7
58 52.9333333333333
59 54.4333333333333
60 54.2666666666667
61 51.9
62 52.4333333333333
63 55.1666666666667
64 52.1
65 54.7666666666667
66 52.2666666666667
67 53.1
68 53.5666666666667
69 54.9666666666667
70 53.1333333333333
71 52.4333333333333
72 53.8666666666667
73 51.0666666666667
74 52.7
75 51.1
76 50.2666666666667
77 50.3666666666667
78 52.4666666666667
79 51.5666666666667
};
\addlegendentry{None}
\addplot [semithick, color1]
table {%
0 7.7
1 12.3
2 14.9666666666667
3 20.4333333333333
4 27.6666666666667
5 39.4
6 55.5666666666667
7 86.9666666666667
8 117.3
9 164.733333333333
10 157.1
11 76.4333333333333
12 57.3333333333333
13 65.1
14 63.2
15 58.9333333333333
16 56.1333333333333
17 55.0666666666667
18 47.6333333333333
19 46.0333333333333
20 43.4
21 40.1
22 40.0666666666667
23 38.6
24 34.4
25 33.1666666666667
26 32.0333333333333
27 30.6333333333333
28 28.1666666666667
29 27.7333333333333
30 26.6333333333333
31 26.7666666666667
32 24.4
33 25.2
34 25.2333333333333
35 21.9333333333333
36 21.4
37 21.7666666666667
38 20.3
39 19.6666666666667
40 18.2666666666667
41 18.6
42 16.7666666666667
43 16.7333333333333
44 17.0666666666667
45 16.3666666666667
46 16
47 16.7333333333333
48 15.4
49 15.0666666666667
50 14.8333333333333
51 13.1333333333333
52 13.7333333333333
53 14.4666666666667
54 14.0666666666667
55 13.6
56 12.0666666666667
57 12.1
58 11.1
59 11.5666666666667
60 11
61 10.7
62 11.3
63 9.66666666666667
64 10.9333333333333
65 9.06666666666667
66 10.0333333333333
67 9.43333333333333
68 9.66666666666667
69 9.06666666666667
70 9.96666666666667
71 8.6
72 8.43333333333333
73 10.4
74 9.86666666666667
75 8.63333333333333
76 8.53333333333333
77 9.6
78 8.43333333333333
79 7.53333333333333
};
\addlegendentry{weak QT}
\addplot [semithick, color2]
table {%
0 7.6
1 14.0666666666667
2 15.3
3 19.0666666666667
4 27.6
5 42.6666666666667
6 57.2333333333333
7 81.6
8 116.2
9 165.233333333333
10 166.066666666667
11 102
12 58.1333333333333
13 58.5333333333333
14 56.1
15 54.6
16 50.0333333333333
17 46.4333333333333
18 40.7
19 41.7333333333333
20 37.9
21 35.0333333333333
22 33.2666666666667
23 32.5333333333333
24 31.1333333333333
25 28
26 27.7333333333333
27 26.7
28 24
29 24.5666666666667
30 23.1333333333333
31 22.1333333333333
32 21.4666666666667
33 20.9666666666667
34 20.1666666666667
35 19.4
36 18.9666666666667
37 18.6
38 19.2
39 17.5666666666667
40 17.3333333333333
41 18.2666666666667
42 16.3333333333333
43 15.8333333333333
44 15.1666666666667
45 14.2
46 14.8333333333333
47 12.9
48 15.4666666666667
49 13.5666666666667
50 13.2333333333333
51 13.1666666666667
52 12.6
53 12.1666666666667
54 12.7666666666667
55 11.9
56 11.9333333333333
57 12.1333333333333
58 10.0666666666667
59 12.3
60 11
61 9.36666666666667
62 9.76666666666667
63 8.9
64 9.7
65 10.1333333333333
66 8.7
67 7.56666666666667
68 7.76666666666667
69 8.9
70 7.53333333333333
71 8.66666666666667
72 8.5
73 7.43333333333333
74 8.66666666666667
75 8.6
76 7.93333333333333
77 8.83333333333333
78 8.73333333333333
79 7.2
};
\addlegendentry{strong QT}
\end{axis}

\end{tikzpicture}

%% file: pic/new22/22quct013_1k.tex
\begin{tikzpicture}

\definecolor{color0}{rgb}{0.886274509803922,0.290196078431373,0.2}
\definecolor{color1}{rgb}{0.203921568627451,0.541176470588235,0.741176470588235}
\pgfmathsetlengthmacro\MajorTickLength{
      \pgfkeysvalueof{/pgfplots/major tick length} * 0.5
    }
\begin{axis}[
axis line style={white},
legend cell align={left},
width = 0.37\linewidth,
height = 1in,
major tick length=\MajorTickLength,
font= \fontsize{7}{7.5}\selectfont,
tick align=inside,
tick pos=left,
x grid style={white},
xmajorgrids,
xmin=0, xmax=80,
xtick style={color=white!33.3333333333333!black},
y grid style={white},
ytick={0,2400,4800},
yticklabels={0,2.4k,4.8k},
yticklabel style={rotate=90},
xtick={0,20,40,60,80},
xticklabels={0,20,40,60,80},
ylabel={Weak QT},
x label style={at={(0.5, 0.4)}},
ymajorgrids,
y label style={at={(0.4,0.5)}},
ymin=-245.486973211548, ymax=5049.38636414202,
ytick style={color=white!33.3333333333333!black}
]
\path [fill=color0, fill opacity=0.3, very thin]
(axis cs:0,20.2457924158074)
--(axis cs:0,14.487540917526)
--(axis cs:1,24.4825321419199)
--(axis cs:2,34.4984868186394)
--(axis cs:3,47.9219803277726)
--(axis cs:4,65.4022092568693)
--(axis cs:5,88.2344968479796)
--(axis cs:6,123.221123895381)
--(axis cs:7,172.750804515143)
--(axis cs:8,246.711388369428)
--(axis cs:9,356.833209471315)
--(axis cs:10,517.172763274692)
--(axis cs:11,644.308433472347)
--(axis cs:12,702.758278333523)
--(axis cs:13,755.52982889576)
--(axis cs:14,810.532704323378)
--(axis cs:15,866.356001376001)
--(axis cs:16,913.540113644872)
--(axis cs:17,956.408896568254)
--(axis cs:18,991.605684384874)
--(axis cs:19,1021.88949864194)
--(axis cs:20,1052.53619490983)
--(axis cs:21,1078.87594458081)
--(axis cs:22,1105.1033229873)
--(axis cs:23,1126.53531166964)
--(axis cs:24,1145.20812994387)
--(axis cs:25,1162.81819004774)
--(axis cs:26,1182.18951296282)
--(axis cs:27,1199.8617789715)
--(axis cs:28,1214.95841501429)
--(axis cs:29,1230.69158657192)
--(axis cs:30,1246.90086582403)
--(axis cs:31,1257.57528013384)
--(axis cs:32,1274.33374372442)
--(axis cs:33,1290.50499499262)
--(axis cs:34,1302.60171290994)
--(axis cs:35,1314.11702353751)
--(axis cs:36,1328.71448513481)
--(axis cs:37,1340.58432929182)
--(axis cs:38,1355.05980322083)
--(axis cs:39,1366.66472935521)
--(axis cs:40,1378.13822557269)
--(axis cs:41,1393.21063662044)
--(axis cs:42,1405.17717333328)
--(axis cs:43,1414.98474594795)
--(axis cs:44,1429.94111065)
--(axis cs:45,1439.47072962565)
--(axis cs:46,1449.69145417068)
--(axis cs:47,1459.08497922802)
--(axis cs:48,1470.64448140372)
--(axis cs:49,1481.68705948563)
--(axis cs:50,1492.0708192853)
--(axis cs:51,1502.75274706755)
--(axis cs:52,1517.79380690397)
--(axis cs:53,1532.15751133932)
--(axis cs:54,1545.40921340442)
--(axis cs:55,1558.16321450114)
--(axis cs:56,1571.38329606482)
--(axis cs:57,1587.70957933917)
--(axis cs:58,1605.59055008613)
--(axis cs:59,1621.79568752821)
--(axis cs:60,1640.20600011158)
--(axis cs:61,1661.4801063483)
--(axis cs:62,1684.93807606749)
--(axis cs:63,1706.76742869866)
--(axis cs:64,1729.60677396021)
--(axis cs:65,1749.24016968081)
--(axis cs:66,1769.93452700295)
--(axis cs:67,1791.98801199064)
--(axis cs:68,1808.74781784779)
--(axis cs:69,1827.93786057755)
--(axis cs:70,1846.50568120052)
--(axis cs:71,1863.80554961816)
--(axis cs:72,1878.60295027101)
--(axis cs:73,1890.6992204563)
--(axis cs:74,1896.52132870475)
--(axis cs:75,1898.60625765379)
--(axis cs:76,1895.29251257595)
--(axis cs:77,1884.36759574583)
--(axis cs:78,1871.24629645642)
--(axis cs:79,1846.76582357841)
--(axis cs:79,4565.01512880254)
--(axis cs:79,4565.01512880254)
--(axis cs:78,4635.69656068643)
--(axis cs:77,4684.03240425417)
--(axis cs:76,4736.13605885262)
--(axis cs:75,4771.58421853668)
--(axis cs:74,4794.6596236762)
--(axis cs:73,4808.71030335323)
--(axis cs:72,4799.39704972899)
--(axis cs:71,4775.15635514374)
--(axis cs:70,4742.04669975187)
--(axis cs:69,4694.52880608912)
--(axis cs:68,4637.39503929506)
--(axis cs:67,4580.91674991412)
--(axis cs:66,4518.37023490181)
--(axis cs:65,4458.683639843)
--(axis cs:64,4406.38370223026)
--(axis cs:63,4356.1754284442)
--(axis cs:62,4288.36668583727)
--(axis cs:61,4212.93894127075)
--(axis cs:60,4125.12733322175)
--(axis cs:59,4048.38526485274)
--(axis cs:58,3974.56183086625)
--(axis cs:57,3901.03327780369)
--(axis cs:56,3840.11194203042)
--(axis cs:55,3781.86535692743)
--(axis cs:54,3724.11459611939)
--(axis cs:53,3664.39486961306)
--(axis cs:52,3606.42524071508)
--(axis cs:51,3555.31391959911)
--(axis cs:50,3501.67203785755)
--(axis cs:49,3447.45579765723)
--(axis cs:48,3395.77456621532)
--(axis cs:47,3333.09597315294)
--(axis cs:46,3279.25140297217)
--(axis cs:45,3230.21498466007)
--(axis cs:44,3174.00174649286)
--(axis cs:43,3119.35811119491)
--(axis cs:42,3059.82282666672)
--(axis cs:41,3005.43698242718)
--(axis cs:40,2951.0998696654)
--(axis cs:39,2893.0590801686)
--(axis cs:38,2837.30210154108)
--(axis cs:37,2771.29186118437)
--(axis cs:36,2706.85694343662)
--(axis cs:35,2637.04488122439)
--(axis cs:34,2562.34114423292)
--(axis cs:33,2490.99024310261)
--(axis cs:32,2413.22816103748)
--(axis cs:31,2335.86281510425)
--(axis cs:30,2258.37532465217)
--(axis cs:29,2179.94650866618)
--(axis cs:28,2100.14634689047)
--(axis cs:27,2022.7763162666)
--(axis cs:26,1944.74382037051)
--(axis cs:25,1870.29609566655)
--(axis cs:24,1791.0585367228)
--(axis cs:23,1713.89325975894)
--(axis cs:22,1636.62048653651)
--(axis cs:21,1557.66691256205)
--(axis cs:20,1479.90190032827)
--(axis cs:19,1404.05335850092)
--(axis cs:18,1330.26098228179)
--(axis cs:17,1256.46729390794)
--(axis cs:16,1180.34560064084)
--(axis cs:15,1106.58685576686)
--(axis cs:14,1032.60062900996)
--(axis cs:13,950.013028247097)
--(axis cs:12,876.717912142667)
--(axis cs:11,821.853471289558)
--(axis cs:10,760.798665296736)
--(axis cs:9,619.176314338209)
--(axis cs:8,440.088611630572)
--(axis cs:7,309.344433580095)
--(axis cs:6,217.797923723667)
--(axis cs:5,154.203598390116)
--(axis cs:4,104.197790743131)
--(axis cs:3,75.0970672912751)
--(axis cs:2,51.3586560385034)
--(axis cs:1,35.174610715223)
--(axis cs:0,20.2457924158074)
--cycle;

\path [fill=color1, fill opacity=0.3, very thin]
(axis cs:0,0)
--(axis cs:0,0)
--(axis cs:1,0)
--(axis cs:2,0)
--(axis cs:3,0)
--(axis cs:4,0)
--(axis cs:5,0)
--(axis cs:6,0)
--(axis cs:7,0)
--(axis cs:8,0)
--(axis cs:9,0)
--(axis cs:10,-1.91655607317521)
--(axis cs:11,-4.81091242274972)
--(axis cs:12,3.75015077924962)
--(axis cs:13,26.1164955140085)
--(axis cs:14,52.5949880669531)
--(axis cs:15,88.6954591406341)
--(axis cs:16,135.494783556178)
--(axis cs:17,189.500101339446)
--(axis cs:18,249.161646603363)
--(axis cs:19,311.329692183203)
--(axis cs:20,372.544418896268)
--(axis cs:21,429.695878448935)
--(axis cs:22,487.364355852535)
--(axis cs:23,540.54810931294)
--(axis cs:24,588.048641071934)
--(axis cs:25,630.324081495755)
--(axis cs:26,670.139229107766)
--(axis cs:27,705.966050757831)
--(axis cs:28,740.340315449069)
--(axis cs:29,770.491207056694)
--(axis cs:30,796.824534097014)
--(axis cs:31,820.945011362509)
--(axis cs:32,843.093856994195)
--(axis cs:33,863.213171160408)
--(axis cs:34,882.768856681506)
--(axis cs:35,902.308917681791)
--(axis cs:36,920.72499889166)
--(axis cs:37,936.849862612333)
--(axis cs:38,953.802261905633)
--(axis cs:39,967.953404937625)
--(axis cs:40,981.316366685168)
--(axis cs:41,997.532103511629)
--(axis cs:42,1011.14677985794)
--(axis cs:43,1024.84659922222)
--(axis cs:44,1038.63246992535)
--(axis cs:45,1051.88401055483)
--(axis cs:46,1066.97019940993)
--(axis cs:47,1080.73862650032)
--(axis cs:48,1093.21624548288)
--(axis cs:49,1108.0814909983)
--(axis cs:50,1122.86755656468)
--(axis cs:51,1135.83931362949)
--(axis cs:52,1148.5913312638)
--(axis cs:53,1159.83080504449)
--(axis cs:54,1173.22599471057)
--(axis cs:55,1185.8549146618)
--(axis cs:56,1197.95115215597)
--(axis cs:57,1212.04475923562)
--(axis cs:58,1224.3378744635)
--(axis cs:59,1236.1394753379)
--(axis cs:60,1248.00912538119)
--(axis cs:61,1262.96576921817)
--(axis cs:62,1276.76303621946)
--(axis cs:63,1291.46601413384)
--(axis cs:64,1305.13704129737)
--(axis cs:65,1318.96491148109)
--(axis cs:66,1335.91158333839)
--(axis cs:67,1353.05591193361)
--(axis cs:68,1369.34994965213)
--(axis cs:69,1384.65906223824)
--(axis cs:70,1402.3501683805)
--(axis cs:71,1423.52092669154)
--(axis cs:72,1444.95370141404)
--(axis cs:73,1464.35183838034)
--(axis cs:74,1483.1295829276)
--(axis cs:75,1504.75800527901)
--(axis cs:76,1524.33945487372)
--(axis cs:77,1541.44854670217)
--(axis cs:78,1561.41551180986)
--(axis cs:79,1577.41928344503)
--(axis cs:79,3901.66643084069)
--(axis cs:79,3901.66643084069)
--(axis cs:78,3860.89877390443)
--(axis cs:77,3809.86573901212)
--(axis cs:76,3757.74625941199)
--(axis cs:75,3702.62294710194)
--(axis cs:74,3642.08946469145)
--(axis cs:73,3582.59101876251)
--(axis cs:72,3526.13201287167)
--(axis cs:71,3466.80288283226)
--(axis cs:70,3406.40221257188)
--(axis cs:69,3343.00760442842)
--(axis cs:68,3282.90719320502)
--(axis cs:67,3222.42027854259)
--(axis cs:66,3162.30746428066)
--(axis cs:65,3104.68270756653)
--(axis cs:64,3043.50105394073)
--(axis cs:63,2989.2292239614)
--(axis cs:62,2936.73220187577)
--(axis cs:61,2887.65327840088)
--(axis cs:60,2837.01944604738)
--(axis cs:59,2788.21290561448)
--(axis cs:58,2740.23355410793)
--(axis cs:57,2690.602859812)
--(axis cs:56,2639.40122879641)
--(axis cs:55,2590.16413295725)
--(axis cs:54,2538.49781481324)
--(axis cs:53,2487.06443305075)
--(axis cs:52,2433.70390683144)
--(axis cs:51,2381.97021018004)
--(axis cs:50,2330.47530057818)
--(axis cs:49,2281.19469947789)
--(axis cs:48,2225.36470689808)
--(axis cs:47,2170.64232588063)
--(axis cs:46,2116.74408630436)
--(axis cs:45,2057.60170373088)
--(axis cs:44,1996.05324436036)
--(axis cs:43,1938.25816268254)
--(axis cs:42,1876.22464871349)
--(axis cs:41,1815.01075363123)
--(axis cs:40,1749.7502999815)
--(axis cs:39,1683.66564268142)
--(axis cs:38,1621.18821428484)
--(axis cs:37,1554.58823262576)
--(axis cs:36,1492.30357253691)
--(axis cs:35,1431.18632041345)
--(axis cs:34,1369.77400046135)
--(axis cs:33,1308.10111455388)
--(axis cs:32,1245.15376205342)
--(axis cs:31,1184.55975054225)
--(axis cs:30,1125.08022780775)
--(axis cs:29,1064.83260246712)
--(axis cs:28,1001.70730359855)
--(axis cs:27,940.157758765978)
--(axis cs:26,881.222675654139)
--(axis cs:25,819.085442313769)
--(axis cs:24,759.065644642351)
--(axis cs:23,698.12808116325)
--(axis cs:22,632.607072718893)
--(axis cs:21,566.66602631297)
--(axis cs:20,500.255581103732)
--(axis cs:19,432.946498292987)
--(axis cs:18,365.457401015685)
--(axis cs:17,298.023708184364)
--(axis cs:16,232.27664501525)
--(axis cs:15,170.352159906985)
--(axis cs:14,118.528821456856)
--(axis cs:13,75.3692187717058)
--(axis cs:12,41.5355635064647)
--(axis cs:11,15.896626708464)
--(axis cs:10,2.44036559698473)
--(axis cs:9,0)
--(axis cs:8,0)
--(axis cs:7,0)
--(axis cs:6,0)
--(axis cs:5,0)
--(axis cs:4,0)
--(axis cs:3,0)
--(axis cs:2,0)
--(axis cs:1,0)
--(axis cs:0,0)
--cycle;

\addplot [semithick, color0]
table {%
0 17.3666666666667
1 29.8285714285714
2 42.9285714285714
3 61.5095238095238
4 84.8
5 121.219047619048
6 170.509523809524
7 241.047619047619
8 343.4
9 488.004761904762
10 638.985714285714
11 733.080952380952
12 789.738095238095
13 852.771428571429
14 921.566666666667
15 986.471428571429
16 1046.94285714286
17 1106.4380952381
18 1160.93333333333
19 1212.97142857143
20 1266.21904761905
21 1318.27142857143
22 1370.8619047619
23 1420.21428571429
24 1468.13333333333
25 1516.55714285714
26 1563.46666666667
27 1611.31904761905
28 1657.55238095238
29 1705.31904761905
30 1752.6380952381
31 1796.71904761905
32 1843.78095238095
33 1890.74761904762
34 1932.47142857143
35 1975.58095238095
36 2017.78571428571
37 2055.9380952381
38 2096.18095238095
39 2129.8619047619
40 2164.61904761905
41 2199.32380952381
42 2232.5
43 2267.17142857143
44 2301.97142857143
45 2334.84285714286
46 2364.47142857143
47 2396.09047619048
48 2433.20952380952
49 2464.57142857143
50 2496.87142857143
51 2529.03333333333
52 2562.10952380952
53 2598.27619047619
54 2634.7619047619
55 2670.01428571429
56 2705.74761904762
57 2744.37142857143
58 2790.07619047619
59 2835.09047619048
60 2882.66666666667
61 2937.20952380952
62 2986.65238095238
63 3031.47142857143
64 3067.99523809524
65 3103.9619047619
66 3144.15238095238
67 3186.45238095238
68 3223.07142857143
69 3261.23333333333
70 3294.27619047619
71 3319.48095238095
72 3339
73 3349.70476190476
74 3345.59047619048
75 3335.09523809524
76 3315.71428571429
77 3284.2
78 3253.47142857143
79 3205.89047619048
};
\addplot [semithick, color1]
table {%
0 0
1 0
2 0
3 0
4 0
5 0
6 0
7 0
8 0
9 0
10 0.261904761904762
11 5.54285714285714
12 22.6428571428571
13 50.7428571428571
14 85.5619047619048
15 129.52380952381
16 183.885714285714
17 243.761904761905
18 307.309523809524
19 372.138095238095
20 436.4
21 498.180952380952
22 559.985714285714
23 619.338095238095
24 673.557142857143
25 724.704761904762
26 775.680952380952
27 823.061904761905
28 871.02380952381
29 917.661904761905
30 960.952380952381
31 1002.75238095238
32 1044.12380952381
33 1085.65714285714
34 1126.27142857143
35 1166.74761904762
36 1206.51428571429
37 1245.71904761905
38 1287.49523809524
39 1325.80952380952
40 1365.53333333333
41 1406.27142857143
42 1443.68571428571
43 1481.55238095238
44 1517.34285714286
45 1554.74285714286
46 1591.85714285714
47 1625.69047619048
48 1659.29047619048
49 1694.6380952381
50 1726.67142857143
51 1758.90476190476
52 1791.14761904762
53 1823.44761904762
54 1855.8619047619
55 1888.00952380952
56 1918.67619047619
57 1951.32380952381
58 1982.28571428571
59 2012.17619047619
60 2042.51428571429
61 2075.30952380952
62 2106.74761904762
63 2140.34761904762
64 2174.31904761905
65 2211.82380952381
66 2249.10952380952
67 2287.7380952381
68 2326.12857142857
69 2363.83333333333
70 2404.37619047619
71 2445.1619047619
72 2485.54285714286
73 2523.47142857143
74 2562.60952380952
75 2603.69047619048
76 2641.04285714286
77 2675.65714285714
78 2711.15714285714
79 2739.54285714286
};
\end{axis}

\end{tikzpicture}

%% file: pic/new22/22quct013_3k.tex
\begin{tikzpicture}

\definecolor{color0}{rgb}{0.886274509803922,0.290196078431373,0.2}
\definecolor{color1}{rgb}{0.203921568627451,0.541176470588235,0.741176470588235}
\pgfmathsetlengthmacro\MajorTickLength{
      \pgfkeysvalueof{/pgfplots/major tick length} * 0.5
    }
\begin{axis}[
axis line style={white},
legend cell align={left},
width = 0.37\linewidth,
height = 1in,
major tick length=\MajorTickLength,
font= \fontsize{7}{7.5}\selectfont,
tick align=inside,
tick pos=left,
x grid style={white},
xmajorgrids,
xmin=0, xmax=80,
xtick style={color=white!33.3333333333333!black},
y grid style={white},
xtick={0,20,40,60,80},
xticklabels={0,20,40,60,80},
yticklabel style={rotate=90},
ytick={0,1000,2000},
yticklabels={0,1k,2k},
x label style={at={(0.5, 0.4)}},
ymajorgrids,
y label style={at={(0.55,0.5)}},
ymin=-94.3465258010096, ymax=2000,
ytick style={color=white!33.3333333333333!black}
]
\path [fill=color0, fill opacity=0.3, very thin]
(axis cs:0,19.6599069357456)
--(axis cs:0,14.4067597309211)
--(axis cs:1,24.0302067381666)
--(axis cs:2,34.5769612166575)
--(axis cs:3,47.4872431881705)
--(axis cs:4,66.5187587973119)
--(axis cs:5,92.094469957316)
--(axis cs:6,130.886727468823)
--(axis cs:7,182.654367159064)
--(axis cs:8,258.236398975265)
--(axis cs:9,371.731138228261)
--(axis cs:10,529.958676914728)
--(axis cs:11,646.22656194941)
--(axis cs:12,701.190847535994)
--(axis cs:13,751.027065915064)
--(axis cs:14,802.081696861598)
--(axis cs:15,848.625151727953)
--(axis cs:16,885.348768667478)
--(axis cs:17,915.975032449422)
--(axis cs:18,942.289284417338)
--(axis cs:19,963.278235524823)
--(axis cs:20,982.648131724472)
--(axis cs:21,999.529604291184)
--(axis cs:22,1011.37952515943)
--(axis cs:23,1023.12584957053)
--(axis cs:24,1029.31194553523)
--(axis cs:25,1038.19250282952)
--(axis cs:26,1044.82671376533)
--(axis cs:27,1050.03991577522)
--(axis cs:28,1056.31806114033)
--(axis cs:29,1058.46786250788)
--(axis cs:30,1062.37833501387)
--(axis cs:31,1065.10291510841)
--(axis cs:32,1068.84018632512)
--(axis cs:33,1070.50462498789)
--(axis cs:34,1075.4052675145)
--(axis cs:35,1077.15351055971)
--(axis cs:36,1078.48555723148)
--(axis cs:37,1085.16498252177)
--(axis cs:38,1087.17301907671)
--(axis cs:39,1090.55428312284)
--(axis cs:40,1093.33312781827)
--(axis cs:41,1095.95991549314)
--(axis cs:42,1096.81080206059)
--(axis cs:43,1099.62925163893)
--(axis cs:44,1102.06725369414)
--(axis cs:45,1104.23620697128)
--(axis cs:46,1106.54696629272)
--(axis cs:47,1108.40792366909)
--(axis cs:48,1112.08064109477)
--(axis cs:49,1112.51086122554)
--(axis cs:50,1113.47775641555)
--(axis cs:51,1115.12848639457)
--(axis cs:52,1115.23255755121)
--(axis cs:53,1114.53025782239)
--(axis cs:54,1114.9048684425)
--(axis cs:55,1112.18274614895)
--(axis cs:56,1110.24042617719)
--(axis cs:57,1108.39326233184)
--(axis cs:58,1107.59438029449)
--(axis cs:59,1102.02777483259)
--(axis cs:60,1101.87422699409)
--(axis cs:61,1100.04023401883)
--(axis cs:62,1097.81010727396)
--(axis cs:63,1093.02562436672)
--(axis cs:64,1088.7574169746)
--(axis cs:65,1081.77425632248)
--(axis cs:66,1079.16542190817)
--(axis cs:67,1072.92488778746)
--(axis cs:68,1066.68253370696)
--(axis cs:69,1058.40586770105)
--(axis cs:70,1055.60332708684)
--(axis cs:71,1049.4402386392)
--(axis cs:72,1046.75296214571)
--(axis cs:73,1042.93186752562)
--(axis cs:74,1038.41019966318)
--(axis cs:75,1035.16954275701)
--(axis cs:76,1033.64788621121)
--(axis cs:77,1027.17160996235)
--(axis cs:78,1025.38368238633)
--(axis cs:79,1019.54987511567)
--(axis cs:79,1646.29774393195)
--(axis cs:79,1646.29774393195)
--(axis cs:78,1651.06393666129)
--(axis cs:77,1659.72362813288)
--(axis cs:76,1671.9140185507)
--(axis cs:75,1684.48760010013)
--(axis cs:74,1695.20884795586)
--(axis cs:73,1706.89670390295)
--(axis cs:72,1715.95179975905)
--(axis cs:71,1724.86452326556)
--(axis cs:70,1735.56810148458)
--(axis cs:69,1743.3750846799)
--(axis cs:68,1753.10794248351)
--(axis cs:67,1758.66558840302)
--(axis cs:66,1765.27267332993)
--(axis cs:65,1769.67336272514)
--(axis cs:64,1771.63305921588)
--(axis cs:63,1773.51723277614)
--(axis cs:62,1776.68513082128)
--(axis cs:61,1777.55976598117)
--(axis cs:60,1776.83053491067)
--(axis cs:59,1776.78174897693)
--(axis cs:58,1777.90085780075)
--(axis cs:57,1775.37816623958)
--(axis cs:56,1776.61671667995)
--(axis cs:55,1774.2363014701)
--(axis cs:54,1772.16179822417)
--(axis cs:53,1768.48878979666)
--(axis cs:52,1765.85315673451)
--(axis cs:51,1760.72865646257)
--(axis cs:50,1758.03652929874)
--(axis cs:49,1754.33675782208)
--(axis cs:48,1751.88126366714)
--(axis cs:47,1746.39207633091)
--(axis cs:46,1742.85303370728)
--(axis cs:45,1737.9161739811)
--(axis cs:44,1731.70417487729)
--(axis cs:43,1721.68503407535)
--(axis cs:42,1714.64634079656)
--(axis cs:41,1705.99246545924)
--(axis cs:40,1694.88591980078)
--(axis cs:39,1684.16952640097)
--(axis cs:38,1673.07459997091)
--(axis cs:37,1660.13025557347)
--(axis cs:36,1646.77158562566)
--(axis cs:35,1631.27506086886)
--(axis cs:34,1616.49949439026)
--(axis cs:33,1597.82870834545)
--(axis cs:32,1583.01695653203)
--(axis cs:31,1567.14470393921)
--(axis cs:30,1548.16452212898)
--(axis cs:29,1525.77023273022)
--(axis cs:28,1502.78670076443)
--(axis cs:27,1479.25532232002)
--(axis cs:26,1460.87804813943)
--(axis cs:25,1435.93130669429)
--(axis cs:24,1408.18329256001)
--(axis cs:23,1379.70272185804)
--(axis cs:22,1352.21095103105)
--(axis cs:21,1324.73706237548)
--(axis cs:20,1296.50424922791)
--(axis cs:19,1261.23605018946)
--(axis cs:18,1220.93928701123)
--(axis cs:17,1177.72020564582)
--(axis cs:16,1129.241707523)
--(axis cs:15,1074.07961017681)
--(axis cs:14,1012.82306504316)
--(axis cs:13,945.687219799222)
--(axis cs:12,877.542485797339)
--(axis cs:11,828.087723764875)
--(axis cs:10,777.917513561462)
--(axis cs:9,624.449814152692)
--(axis cs:8,441.487410548545)
--(axis cs:7,309.936109031412)
--(axis cs:6,217.275177293082)
--(axis cs:5,151.372196709351)
--(axis cs:4,106.481241202688)
--(axis cs:3,73.58894728802)
--(axis cs:2,51.3754197357235)
--(axis cs:1,34.788840880881)
--(axis cs:0,19.6599069357456)
--cycle;

\path [fill=color1, fill opacity=0.3, very thin]
(axis cs:0,0)
--(axis cs:0,0)
--(axis cs:1,0)
--(axis cs:2,0)
--(axis cs:3,0)
--(axis cs:4,0)
--(axis cs:5,0)
--(axis cs:6,0)
--(axis cs:7,0)
--(axis cs:8,0)
--(axis cs:9,0)
--(axis cs:10,-2.27790383664575)
--(axis cs:11,-5.1918884866402)
--(axis cs:12,5.24547508185887)
--(axis cs:13,27.158198558991)
--(axis cs:14,56.1852487276441)
--(axis cs:15,94.0567175962709)
--(axis cs:16,141.911783917008)
--(axis cs:17,196.741347727926)
--(axis cs:18,256.644579973044)
--(axis cs:19,318.060824013035)
--(axis cs:20,377.519155073487)
--(axis cs:21,433.136278407634)
--(axis cs:22,485.765950161257)
--(axis cs:23,532.261534588376)
--(axis cs:24,573.982959601337)
--(axis cs:25,612.531962640535)
--(axis cs:26,646.762019069848)
--(axis cs:27,674.834212661674)
--(axis cs:28,682.131760150781)
--(axis cs:29,722.158682826083)
--(axis cs:30,740.613715987794)
--(axis cs:31,757.286664675752)
--(axis cs:32,771.565660905407)
--(axis cs:33,784.756315577472)
--(axis cs:34,796.544743525791)
--(axis cs:35,806.392000599026)
--(axis cs:36,815.491271653295)
--(axis cs:37,823.008183445415)
--(axis cs:38,829.356422542201)
--(axis cs:39,837.127446810167)
--(axis cs:40,843.566881551082)
--(axis cs:41,849.195014828807)
--(axis cs:42,852.180834755156)
--(axis cs:43,859.197066335713)
--(axis cs:44,864.858794383809)
--(axis cs:45,870.872530506552)
--(axis cs:46,877.489452046226)
--(axis cs:47,883.390736503454)
--(axis cs:48,889.32566898462)
--(axis cs:49,894.138296901674)
--(axis cs:50,899.711463363821)
--(axis cs:51,903.699023929731)
--(axis cs:52,908.159237371356)
--(axis cs:53,912.922728672243)
--(axis cs:54,918.58935944014)
--(axis cs:55,922.092570219381)
--(axis cs:56,925.7691653553)
--(axis cs:57,932.408019251755)
--(axis cs:58,935.166038081896)
--(axis cs:59,938.440068672839)
--(axis cs:60,942.138278509037)
--(axis cs:61,946.696972899414)
--(axis cs:62,949.692613124569)
--(axis cs:63,952.405545945765)
--(axis cs:64,954.101068951995)
--(axis cs:65,956.120932017297)
--(axis cs:66,959.133165894984)
--(axis cs:67,963.114435188921)
--(axis cs:68,965.242939290341)
--(axis cs:69,966.714753107563)
--(axis cs:70,952.775531714638)
--(axis cs:71,970.137842240655)
--(axis cs:72,954.173422009436)
--(axis cs:73,956.475843034634)
--(axis cs:74,970.459220294939)
--(axis cs:75,970.145205035034)
--(axis cs:76,937.466209882503)
--(axis cs:77,954.752477006568)
--(axis cs:78,948.353994111641)
--(axis cs:79,952.580284699132)
--(axis cs:79,1545.41971530087)
--(axis cs:79,1545.41971530087)
--(axis cs:78,1544.18886303122)
--(axis cs:77,1544.93323727915)
--(axis cs:76,1553.02902821273)
--(axis cs:75,1539.26431877449)
--(axis cs:74,1537.04554160982)
--(axis cs:73,1541.77177601298)
--(axis cs:72,1539.08372084771)
--(axis cs:71,1527.35739585458)
--(axis cs:70,1531.23399209489)
--(axis cs:69,1517.26619927339)
--(axis cs:68,1513.17610832871)
--(axis cs:67,1508.63794576346)
--(axis cs:66,1503.00969124787)
--(axis cs:65,1497.34573464937)
--(axis cs:64,1490.64178819086)
--(axis cs:63,1483.2611207209)
--(axis cs:62,1476.46929163734)
--(axis cs:61,1470.32207471963)
--(axis cs:60,1462.41410244334)
--(axis cs:59,1456.24564561288)
--(axis cs:58,1449.40539048953)
--(axis cs:57,1443.81102836729)
--(axis cs:56,1435.4308346447)
--(axis cs:55,1426.28838216157)
--(axis cs:54,1416.71540246462)
--(axis cs:53,1407.20108085157)
--(axis cs:52,1394.52647691436)
--(axis cs:51,1383.22478559408)
--(axis cs:50,1372.06948901713)
--(axis cs:49,1356.84265547928)
--(axis cs:48,1343.75052149157)
--(axis cs:47,1326.9521206394)
--(axis cs:46,1313.33911938235)
--(axis cs:45,1300.72746949345)
--(axis cs:44,1284.34120561619)
--(axis cs:43,1267.31721937857)
--(axis cs:42,1249.30487953056)
--(axis cs:41,1228.73831850453)
--(axis cs:40,1212.16645178225)
--(axis cs:39,1192.82493414221)
--(axis cs:38,1172.27214888637)
--(axis cs:37,1151.21086417363)
--(axis cs:36,1128.68968072766)
--(axis cs:35,1107.04609463907)
--(axis cs:34,1084.56001837897)
--(axis cs:33,1058.50082727967)
--(axis cs:32,1031.94862480888)
--(axis cs:31,1002.84666865758)
--(axis cs:30,972.614855440778)
--(axis cs:29,941.041317173917)
--(axis cs:28,916.792049373028)
--(axis cs:27,868.57531114785)
--(axis cs:26,826.885599977771)
--(axis cs:25,784.715656407084)
--(axis cs:24,737.969421351044)
--(axis cs:23,683.995608268767)
--(axis cs:22,625.491192695885)
--(axis cs:21,564.025626354271)
--(axis cs:20,500.547511593179)
--(axis cs:19,434.110604558393)
--(axis cs:18,365.583991455527)
--(axis cs:17,298.477699891122)
--(axis cs:16,233.059644654421)
--(axis cs:15,172.333758594205)
--(axis cs:14,118.481417939023)
--(axis cs:13,76.1084681076757)
--(axis cs:12,42.011667775284)
--(axis cs:11,15.553793248545)
--(axis cs:10,2.85885621759813)
--(axis cs:9,0)
--(axis cs:8,0)
--(axis cs:7,0)
--(axis cs:6,0)
--(axis cs:5,0)
--(axis cs:4,0)
--(axis cs:3,0)
--(axis cs:2,0)
--(axis cs:1,0)
--(axis cs:0,0)
--cycle;

\addplot [semithick, color0]
table {%
0 17.0333333333333
1 29.4095238095238
2 42.9761904761905
3 60.5380952380952
4 86.5
5 121.733333333333
6 174.080952380952
7 246.295238095238
8 349.861904761905
9 498.090476190476
10 653.938095238095
11 737.157142857143
12 789.366666666667
13 848.357142857143
14 907.452380952381
15 961.352380952381
16 1007.29523809524
17 1046.84761904762
18 1081.61428571429
19 1112.25714285714
20 1139.57619047619
21 1162.13333333333
22 1181.79523809524
23 1201.41428571429
24 1218.74761904762
25 1237.0619047619
26 1252.85238095238
27 1264.64761904762
28 1279.55238095238
29 1292.11904761905
30 1305.27142857143
31 1316.12380952381
32 1325.92857142857
33 1334.16666666667
34 1345.95238095238
35 1354.21428571429
36 1362.62857142857
37 1372.64761904762
38 1380.12380952381
39 1387.3619047619
40 1394.10952380952
41 1400.97619047619
42 1405.72857142857
43 1410.65714285714
44 1416.88571428571
45 1421.07619047619
46 1424.7
47 1427.4
48 1431.98095238095
49 1433.42380952381
50 1435.75714285714
51 1437.92857142857
52 1440.54285714286
53 1441.50952380952
54 1443.53333333333
55 1443.20952380952
56 1443.42857142857
57 1441.88571428571
58 1442.74761904762
59 1439.40476190476
60 1439.35238095238
61 1438.8
62 1437.24761904762
63 1433.27142857143
64 1430.19523809524
65 1425.72380952381
66 1422.21904761905
67 1415.79523809524
68 1409.89523809524
69 1400.89047619048
70 1395.58571428571
71 1387.15238095238
72 1381.35238095238
73 1374.91428571429
74 1366.80952380952
75 1359.82857142857
76 1352.78095238095
77 1343.44761904762
78 1338.22380952381
79 1332.92380952381
};
\addplot [semithick, color1]
table {%
0 0
1 0
2 0
3 0
4 0
5 0
6 0
7 0
8 0
9 0
10 0.29047619047619
11 5.18095238095238
12 23.6285714285714
13 51.6333333333333
14 87.3333333333333
15 133.195238095238
16 187.485714285714
17 247.609523809524
18 311.114285714286
19 376.085714285714
20 439.033333333333
21 498.580952380952
22 555.628571428571
23 608.128571428571
24 655.97619047619
25 698.62380952381
26 736.823809523809
27 771.704761904762
28 799.461904761905
29 831.6
30 856.614285714286
31 880.066666666667
32 901.757142857143
33 921.628571428571
34 940.552380952381
35 956.719047619048
36 972.090476190476
37 987.109523809524
38 1000.81428571429
39 1014.97619047619
40 1027.86666666667
41 1038.96666666667
42 1050.74285714286
43 1063.25714285714
44 1074.6
45 1085.8
46 1095.41428571429
47 1105.17142857143
48 1116.5380952381
49 1125.49047619048
50 1135.89047619048
51 1143.4619047619
52 1151.34285714286
53 1160.0619047619
54 1167.65238095238
55 1174.19047619048
56 1180.6
57 1188.10952380952
58 1192.28571428571
59 1197.34285714286
60 1202.27619047619
61 1208.50952380952
62 1213.08095238095
63 1217.83333333333
64 1222.37142857143
65 1226.73333333333
66 1231.07142857143
67 1235.87619047619
68 1239.20952380952
69 1241.99047619048
70 1242.00476190476
71 1248.74761904762
72 1246.62857142857
73 1249.12380952381
74 1253.75238095238
75 1254.70476190476
76 1245.24761904762
77 1249.84285714286
78 1246.27142857143
79 1249
};
\end{axis}

\end{tikzpicture}

%% file: pic/new22/22quct013_10k.tex
\begin{tikzpicture}

\definecolor{color0}{rgb}{0.886274509803922,0.290196078431373,0.2}
\definecolor{color1}{rgb}{0.203921568627451,0.541176470588235,0.741176470588235}
\pgfmathsetlengthmacro\MajorTickLength{
      \pgfkeysvalueof{/pgfplots/major tick length} * 0.5
    }
\begin{axis}[
axis line style={white},
legend cell align={left},
width = 0.37\linewidth,
height = 1in,
major tick length=\MajorTickLength,
font= \fontsize{7}{7.5}\selectfont,
tick align=inside,
tick pos=left,
x grid style={white},
xmajorgrids,
xmin=0, xmax=80,
xtick style={color=white!33.3333333333333!black},
y grid style={white},
xtick={0,20,40,60,80},
xticklabels={0,20,40,60,80},
yticklabel style={rotate=90},
ytick={0,500,1000},
yticklabels={0,500,1k},
x label style={at={(0.5, 0.4)}},
ymajorgrids,
y label style={at={(0.55,0.5)}},
ymin=-50.988944351419, ymax=1000,
ytick style={color=white!33.3333333333333!black}
]
\path [fill=color0, fill opacity=0.3, very thin]
(axis cs:0,19.8822503969567)
--(axis cs:0,14.4129876982814)
--(axis cs:1,23.9370752269112)
--(axis cs:2,34.3398060004391)
--(axis cs:3,48.1083657090702)
--(axis cs:4,66.5715287549786)
--(axis cs:5,93.1096751936595)
--(axis cs:6,130.099511683256)
--(axis cs:7,185.690337132892)
--(axis cs:8,260.803160345433)
--(axis cs:9,375.013535050635)
--(axis cs:10,531.258896341727)
--(axis cs:11,629.694807041517)
--(axis cs:12,676.922795604164)
--(axis cs:13,706.712271090531)
--(axis cs:14,730.400257749016)
--(axis cs:15,747.905232328494)
--(axis cs:16,754.849421876849)
--(axis cs:17,756.82331457096)
--(axis cs:18,753.031227764605)
--(axis cs:19,743.933035482109)
--(axis cs:20,735.603094860418)
--(axis cs:21,720.534499194752)
--(axis cs:22,704.206639728379)
--(axis cs:23,688.31180431776)
--(axis cs:24,672.709486855727)
--(axis cs:25,655.159670723918)
--(axis cs:26,637.515515913385)
--(axis cs:27,621.023614044902)
--(axis cs:28,602.805514118327)
--(axis cs:29,586.552429579041)
--(axis cs:30,570.243664667549)
--(axis cs:31,554.969125510761)
--(axis cs:32,538.292822449413)
--(axis cs:33,523.112903751436)
--(axis cs:34,507.826154126389)
--(axis cs:35,495.732588506289)
--(axis cs:36,482.938050226212)
--(axis cs:37,471.647318658146)
--(axis cs:38,458.70789978571)
--(axis cs:39,446.776544965196)
--(axis cs:40,435.602257929871)
--(axis cs:41,424.646574899374)
--(axis cs:42,416.011790660296)
--(axis cs:43,404.30978660365)
--(axis cs:44,394.13657918687)
--(axis cs:45,383.417114513633)
--(axis cs:46,374.34915957075)
--(axis cs:47,364.744417152675)
--(axis cs:48,357.466395943022)
--(axis cs:49,348.425806711731)
--(axis cs:50,341.76123159624)
--(axis cs:51,334.188105346016)
--(axis cs:52,327.888742246636)
--(axis cs:53,323.071589925614)
--(axis cs:54,316.260569399263)
--(axis cs:55,310.563470884451)
--(axis cs:56,304.435759725674)
--(axis cs:57,300.176545047302)
--(axis cs:58,295.181853645573)
--(axis cs:59,291.91741425144)
--(axis cs:60,288.661981481227)
--(axis cs:61,286.62624726677)
--(axis cs:62,284.410932205904)
--(axis cs:63,283.315123331184)
--(axis cs:64,281.448697681141)
--(axis cs:65,278.479155084737)
--(axis cs:66,276.369950809893)
--(axis cs:67,274.698910925152)
--(axis cs:68,273.033936617049)
--(axis cs:69,271.034785495152)
--(axis cs:70,269.966832822305)
--(axis cs:71,267.559513591096)
--(axis cs:72,265.716488621271)
--(axis cs:73,263.13667046937)
--(axis cs:74,260.790859648981)
--(axis cs:75,259.61840743398)
--(axis cs:76,258.119161568068)
--(axis cs:77,257.889915111291)
--(axis cs:78,256.490433200782)
--(axis cs:79,254.309265063598)
--(axis cs:79,598.109782555449)
--(axis cs:79,598.109782555449)
--(axis cs:78,591.433376323028)
--(axis cs:77,586.66246584109)
--(axis cs:76,581.890362241456)
--(axis cs:75,576.562544946973)
--(axis cs:74,570.656759398638)
--(axis cs:73,566.596662863964)
--(axis cs:72,560.521606616824)
--(axis cs:71,557.707153075571)
--(axis cs:70,554.785548130076)
--(axis cs:69,552.184262123896)
--(axis cs:68,550.68987290676)
--(axis cs:67,549.434422408181)
--(axis cs:66,549.506239666298)
--(axis cs:65,550.835130629549)
--(axis cs:64,551.798921366478)
--(axis cs:63,551.865829049769)
--(axis cs:62,553.608115413143)
--(axis cs:61,558.640419399896)
--(axis cs:60,562.061828042583)
--(axis cs:59,564.730204796179)
--(axis cs:58,567.008622544903)
--(axis cs:57,570.375835905079)
--(axis cs:56,573.535668845754)
--(axis cs:55,576.684148163168)
--(axis cs:54,580.434668695975)
--(axis cs:53,583.480791026767)
--(axis cs:52,588.016019658126)
--(axis cs:51,592.735704177794)
--(axis cs:50,597.095911260903)
--(axis cs:49,602.783717097793)
--(axis cs:48,610.085985009359)
--(axis cs:47,614.465106656849)
--(axis cs:46,621.031792810202)
--(axis cs:45,625.859075962557)
--(axis cs:44,634.453897003606)
--(axis cs:43,640.994975301112)
--(axis cs:42,647.883447434942)
--(axis cs:41,654.934377481579)
--(axis cs:40,663.931075403463)
--(axis cs:39,673.156788368137)
--(axis cs:38,682.377814500004)
--(axis cs:37,690.562205151378)
--(axis cs:36,701.071473583312)
--(axis cs:35,711.400744827044)
--(axis cs:34,720.392893492658)
--(axis cs:33,729.925191486659)
--(axis cs:32,742.707177550587)
--(axis cs:31,755.040398298763)
--(axis cs:30,767.508716284832)
--(axis cs:29,780.542808516197)
--(axis cs:28,793.956390643578)
--(axis cs:27,809.75733833605)
--(axis cs:26,824.789245991377)
--(axis cs:25,839.487948323701)
--(axis cs:24,853.728608382368)
--(axis cs:23,871.116767110811)
--(axis cs:22,883.031455509716)
--(axis cs:21,895.97026271001)
--(axis cs:20,908.36833371101)
--(axis cs:19,914.466964517891)
--(axis cs:18,917.40686747349)
--(axis cs:17,917.310018762373)
--(axis cs:16,910.712482885056)
--(axis cs:15,899.647148623886)
--(axis cs:14,878.199742250984)
--(axis cs:13,850.335347957088)
--(axis cs:12,817.372442491074)
--(axis cs:11,792.77185962515)
--(axis cs:10,760.874436991606)
--(axis cs:9,639.481703044603)
--(axis cs:8,455.187315845044)
--(axis cs:7,316.633472390918)
--(axis cs:6,223.205250221506)
--(axis cs:5,155.871277187293)
--(axis cs:4,109.12370934026)
--(axis cs:3,76.1678247671203)
--(axis cs:2,52.2982892376561)
--(axis cs:1,34.7200676302317)
--(axis cs:0,19.8822503969567)
--cycle;

\path [fill=color1, fill opacity=0.3, very thin]
(axis cs:0,0)
--(axis cs:0,0)
--(axis cs:1,0)
--(axis cs:2,0)
--(axis cs:3,0)
--(axis cs:4,0)
--(axis cs:5,0)
--(axis cs:6,0)
--(axis cs:7,0)
--(axis cs:8,0)
--(axis cs:9,0)
--(axis cs:10,-2.40249200587919)
--(axis cs:11,-4.87485807404237)
--(axis cs:12,6.58372680612229)
--(axis cs:13,27.6067648789965)
--(axis cs:14,56.6165865064122)
--(axis cs:15,95.0064436320304)
--(axis cs:16,144.225885996897)
--(axis cs:17,193.07928576392)
--(axis cs:18,253.988092592968)
--(axis cs:19,310.846104187066)
--(axis cs:20,364.370639443716)
--(axis cs:21,410.723109760484)
--(axis cs:22,449.264860963693)
--(axis cs:23,483.096732576637)
--(axis cs:24,493.873578698925)
--(axis cs:25,528.590249119017)
--(axis cs:26,542.565916442861)
--(axis cs:27,550.742936897818)
--(axis cs:28,554.317728284859)
--(axis cs:29,553.98067657763)
--(axis cs:30,552.186809345928)
--(axis cs:31,549.032947154974)
--(axis cs:32,543.168064269625)
--(axis cs:33,536.304633486415)
--(axis cs:34,527.921225454762)
--(axis cs:35,518.667911085855)
--(axis cs:36,508.884188683452)
--(axis cs:37,499.102012080535)
--(axis cs:38,488.989107548449)
--(axis cs:39,478.898469285539)
--(axis cs:40,469.767683026788)
--(axis cs:41,460.035632897048)
--(axis cs:42,451.434200732857)
--(axis cs:43,440.731123369231)
--(axis cs:44,432.69158308967)
--(axis cs:45,423.271055218889)
--(axis cs:46,406.252852549716)
--(axis cs:47,406.781660311678)
--(axis cs:48,389.816099660013)
--(axis cs:49,382.543862800195)
--(axis cs:50,382.750664854922)
--(axis cs:51,375.905754521478)
--(axis cs:52,360.888529059733)
--(axis cs:53,361.570430830713)
--(axis cs:54,354.501764518015)
--(axis cs:55,348.374487389971)
--(axis cs:56,342.801928498882)
--(axis cs:57,337.171754554095)
--(axis cs:58,326.69991848578)
--(axis cs:59,321.61765578288)
--(axis cs:60,322.628605863394)
--(axis cs:61,318.512978290606)
--(axis cs:62,309.38641919196)
--(axis cs:63,309.380583119726)
--(axis cs:64,306.497759626879)
--(axis cs:65,297.055126797089)
--(axis cs:66,290.14987080682)
--(axis cs:67,298.071271010737)
--(axis cs:68,295.780593965079)
--(axis cs:69,286.92942019439)
--(axis cs:70,291.370425880731)
--(axis cs:71,290.130538146231)
--(axis cs:72,280.406853608703)
--(axis cs:73,288.725161816917)
--(axis cs:74,288.204942033332)
--(axis cs:75,287.264223275448)
--(axis cs:76,288.000575824769)
--(axis cs:77,287.798035495698)
--(axis cs:78,282.671257025024)
--(axis cs:79,288.236289131381)
--(axis cs:79,524.373234678143)
--(axis cs:79,524.373234678143)
--(axis cs:78,522.585885832119)
--(axis cs:77,519.116250218588)
--(axis cs:76,517.504186079993)
--(axis cs:75,517.421491010266)
--(axis cs:74,517.947438919049)
--(axis cs:73,517.779600087845)
--(axis cs:72,520.669336867488)
--(axis cs:71,518.059938044245)
--(axis cs:70,518.877193166888)
--(axis cs:69,521.042008377038)
--(axis cs:68,521.009882225397)
--(axis cs:67,523.014443274977)
--(axis cs:66,528.459653002704)
--(axis cs:65,528.306777964816)
--(axis cs:64,527.683192754073)
--(axis cs:63,528.486083546941)
--(axis cs:62,531.508818903278)
--(axis cs:61,531.496545518918)
--(axis cs:60,533.352346517558)
--(axis cs:59,538.953772788549)
--(axis cs:58,542.290557704696)
--(axis cs:57,542.237769255429)
--(axis cs:56,544.712357215404)
--(axis cs:55,547.787417371933)
--(axis cs:54,550.926806910556)
--(axis cs:53,554.115283455001)
--(axis cs:52,561.930518559315)
--(axis cs:51,562.132340716617)
--(axis cs:50,565.363620859364)
--(axis cs:49,573.751375295043)
--(axis cs:48,580.088662244749)
--(axis cs:47,581.542149212132)
--(axis cs:46,590.842385545522)
--(axis cs:45,591.690849543016)
--(axis cs:44,597.498893100806)
--(axis cs:43,602.687924249817)
--(axis cs:42,610.50865641)
--(axis cs:41,617.65960519819)
--(axis cs:40,623.956126497021)
--(axis cs:39,630.282483095413)
--(axis cs:38,637.572797213456)
--(axis cs:37,644.040845062322)
--(axis cs:36,651.277716078453)
--(axis cs:35,658.465422247478)
--(axis cs:34,665.097822164286)
--(axis cs:33,671.209652227871)
--(axis cs:32,678.27003096847)
--(axis cs:31,680.738481416455)
--(axis cs:30,683.117952558834)
--(axis cs:29,680.924085327132)
--(axis cs:28,677.225128857998)
--(axis cs:27,669.018967864087)
--(axis cs:26,656.06265498571)
--(axis cs:25,638.438322309554)
--(axis cs:24,627.059754634408)
--(axis cs:23,587.303267423363)
--(axis cs:22,551.487519988688)
--(axis cs:21,509.753080715707)
--(axis cs:20,463.105551032474)
--(axis cs:19,410.839610098649)
--(axis cs:18,354.716669311794)
--(axis cs:17,294.730238045604)
--(axis cs:16,232.012209241198)
--(axis cs:15,172.279270653684)
--(axis cs:14,119.345318255493)
--(axis cs:13,77.7456160733844)
--(axis cs:12,44.8638922414968)
--(axis cs:11,18.541524740709)
--(axis cs:10,3.05963486302205)
--(axis cs:9,0)
--(axis cs:8,0)
--(axis cs:7,0)
--(axis cs:6,0)
--(axis cs:5,0)
--(axis cs:4,0)
--(axis cs:3,0)
--(axis cs:2,0)
--(axis cs:1,0)
--(axis cs:0,0)
--cycle;

\addplot [semithick, color0]
table {%
0 17.147619047619
1 29.3285714285714
2 43.3190476190476
3 62.1380952380952
4 87.847619047619
5 124.490476190476
6 176.652380952381
7 251.161904761905
8 357.995238095238
9 507.247619047619
10 646.066666666667
11 711.233333333333
12 747.147619047619
13 778.52380952381
14 804.3
15 823.77619047619
16 832.780952380952
17 837.066666666667
18 835.219047619048
19 829.2
20 821.985714285714
21 808.252380952381
22 793.619047619048
23 779.714285714286
24 763.219047619048
25 747.323809523809
26 731.152380952381
27 715.390476190476
28 698.380952380952
29 683.547619047619
30 668.87619047619
31 655.004761904762
32 640.5
33 626.519047619048
34 614.109523809524
35 603.566666666667
36 592.004761904762
37 581.104761904762
38 570.542857142857
39 559.966666666667
40 549.766666666667
41 539.790476190476
42 531.947619047619
43 522.652380952381
44 514.295238095238
45 504.638095238095
46 497.690476190476
47 489.604761904762
48 483.77619047619
49 475.604761904762
50 469.428571428571
51 463.461904761905
52 457.952380952381
53 453.27619047619
54 448.347619047619
55 443.62380952381
56 438.985714285714
57 435.27619047619
58 431.095238095238
59 428.32380952381
60 425.361904761905
61 422.633333333333
62 419.009523809524
63 417.590476190476
64 416.62380952381
65 414.657142857143
66 412.938095238095
67 412.066666666667
68 411.861904761905
69 411.609523809524
70 412.37619047619
71 412.633333333333
72 413.119047619048
73 414.866666666667
74 415.72380952381
75 418.090476190476
76 420.004761904762
77 422.27619047619
78 423.961904761905
79 426.209523809524
};
\addplot [semithick, color1]
table {%
0 0
1 0
2 0
3 0
4 0
5 0
6 0
7 0
8 0
9 0
10 0.328571428571429
11 6.83333333333333
12 25.7238095238095
13 52.6761904761905
14 87.9809523809524
15 133.642857142857
16 188.119047619048
17 243.904761904762
18 304.352380952381
19 360.842857142857
20 413.738095238095
21 460.238095238095
22 500.37619047619
23 535.2
24 560.466666666667
25 583.514285714286
26 599.314285714286
27 609.880952380952
28 615.771428571429
29 617.452380952381
30 617.652380952381
31 614.885714285714
32 610.719047619048
33 603.757142857143
34 596.509523809524
35 588.566666666667
36 580.080952380952
37 571.571428571429
38 563.280952380952
39 554.590476190476
40 546.861904761905
41 538.847619047619
42 530.971428571429
43 521.709523809524
44 515.095238095238
45 507.480952380952
46 498.547619047619
47 494.161904761905
48 484.952380952381
49 478.147619047619
50 474.057142857143
51 469.019047619048
52 461.409523809524
53 457.842857142857
54 452.714285714286
55 448.080952380952
56 443.757142857143
57 439.704761904762
58 434.495238095238
59 430.285714285714
60 427.990476190476
61 425.004761904762
62 420.447619047619
63 418.933333333333
64 417.090476190476
65 412.680952380952
66 409.304761904762
67 410.542857142857
68 408.395238095238
69 403.985714285714
70 405.12380952381
71 404.095238095238
72 400.538095238095
73 403.252380952381
74 403.076190476191
75 402.342857142857
76 402.752380952381
77 403.457142857143
78 402.628571428571
79 406.304761904762
};
\end{axis}

\end{tikzpicture}

%% file: pic/new22/22quct013_25k.tex
\begin{tikzpicture}

\definecolor{color0}{rgb}{0.886274509803922,0.290196078431373,0.2}
\definecolor{color1}{rgb}{0.203921568627451,0.541176470588235,0.741176470588235}
\pgfmathsetlengthmacro\MajorTickLength{
      \pgfkeysvalueof{/pgfplots/major tick length} * 0.5
    }
\begin{axis}[
axis line style={white},
legend cell align={left},
width = 0.37\linewidth,
height = 1in,
major tick length=\MajorTickLength,
legend style={inner xsep=1pt, inner ysep=-1pt, row sep=-3pt,at={(0,1.4)},anchor=north west},
font= \fontsize{7}{7.5}\selectfont,
tick align=inside,
tick pos=left,
x grid style={white},
xmajorgrids,
xmin=0, xmax=80,
xtick style={color=white!33.3333333333333!black},
y grid style={white},
xtick={0,20,40,60,80},
yticklabel style={rotate=90},
xticklabels={0,20,40,60,80},
ytick={0,500,1000},
yticklabels={0,500,1k},
x label style={at={(0.5, 0.4)}},
ymajorgrids,
y label style={at={(0.55,0.5)}},
ymin=-47.2548724983418, ymax=1000,
ytick style={color=white!33.3333333333333!black}
]
\path [fill=color0, fill opacity=0.3, very thin]
(axis cs:0,20.3522562557546)
--(axis cs:0,15.0202927638532)
--(axis cs:1,24.6838866659887)
--(axis cs:2,36.4150993543542)
--(axis cs:3,50.6303939929436)
--(axis cs:4,70.3500359770439)
--(axis cs:5,96.4552382025223)
--(axis cs:6,137.247956160446)
--(axis cs:7,192.350626890458)
--(axis cs:8,271.510132699284)
--(axis cs:9,387.748635594705)
--(axis cs:10,535.485599649625)
--(axis cs:11,620.096923550325)
--(axis cs:12,653.452281144357)
--(axis cs:13,673.833903300549)
--(axis cs:14,688.512409908709)
--(axis cs:15,695.318026841368)
--(axis cs:16,697.434637271728)
--(axis cs:17,693.140094480306)
--(axis cs:18,682.57770298855)
--(axis cs:19,669.164984997698)
--(axis cs:20,652.582527300447)
--(axis cs:21,635.070550313094)
--(axis cs:22,617.613561032544)
--(axis cs:23,597.615939898512)
--(axis cs:24,577.583409903889)
--(axis cs:25,556.237306270279)
--(axis cs:26,536.413493342186)
--(axis cs:27,515.534085951822)
--(axis cs:28,494.236911415745)
--(axis cs:29,476.620189529338)
--(axis cs:30,458.306361369468)
--(axis cs:31,441.07724051277)
--(axis cs:32,423.609053187207)
--(axis cs:33,406.758263015035)
--(axis cs:34,389.196092069695)
--(axis cs:35,372.583737107704)
--(axis cs:36,356.955145177639)
--(axis cs:37,343.461673492792)
--(axis cs:38,329.904139555771)
--(axis cs:39,317.520219247161)
--(axis cs:40,305.208513103887)
--(axis cs:41,292.761550067157)
--(axis cs:42,282.005605799639)
--(axis cs:43,269.585308832407)
--(axis cs:44,259.56184973616)
--(axis cs:45,249.908000297325)
--(axis cs:46,241.194523584718)
--(axis cs:47,233.274780691993)
--(axis cs:48,225.372864487408)
--(axis cs:49,217.424535384344)
--(axis cs:50,209.714342681873)
--(axis cs:51,201.802115601882)
--(axis cs:52,195.634964584875)
--(axis cs:53,189.744328417441)
--(axis cs:54,184.042382669026)
--(axis cs:55,178.801142845328)
--(axis cs:56,173.170998514247)
--(axis cs:57,166.47408381008)
--(axis cs:58,161.596943261044)
--(axis cs:59,157.179684302992)
--(axis cs:60,151.699020502581)
--(axis cs:61,147.306607917001)
--(axis cs:62,142.832288173836)
--(axis cs:63,139.347647541018)
--(axis cs:64,135.701980995078)
--(axis cs:65,131.323425952091)
--(axis cs:66,128.205703351657)
--(axis cs:67,124.608042556876)
--(axis cs:68,120.670472063395)
--(axis cs:69,118.002808295409)
--(axis cs:70,115.860789402727)
--(axis cs:71,112.688692096854)
--(axis cs:72,109.824933803499)
--(axis cs:73,107.543958199942)
--(axis cs:74,106.133270009068)
--(axis cs:75,104.493334847159)
--(axis cs:76,103.245384668338)
--(axis cs:77,102.746589062426)
--(axis cs:78,101.526622419769)
--(axis cs:79,100.879156293399)
--(axis cs:79,240.277706451699)
--(axis cs:79,240.277706451699)
--(axis cs:78,242.12043640376)
--(axis cs:77,243.018116819927)
--(axis cs:76,244.421281998329)
--(axis cs:75,246.869410250881)
--(axis cs:74,249.327514304657)
--(axis cs:73,251.465845721627)
--(axis cs:72,253.684870118069)
--(axis cs:71,257.144641236479)
--(axis cs:70,260.15881844041)
--(axis cs:69,265.497191704591)
--(axis cs:68,269.133449505233)
--(axis cs:67,273.058624109791)
--(axis cs:66,278.823708413049)
--(axis cs:65,282.755005420458)
--(axis cs:64,288.17056802453)
--(axis cs:63,293.152352458982)
--(axis cs:62,298.3441824144)
--(axis cs:61,304.860058749666)
--(axis cs:60,311.761763811145)
--(axis cs:59,317.888943147989)
--(axis cs:58,324.108939091897)
--(axis cs:57,331.712190699724)
--(axis cs:56,336.49566815242)
--(axis cs:55,343.453759115456)
--(axis cs:54,350.173303605484)
--(axis cs:53,357.657632366873)
--(axis cs:52,367.070917768066)
--(axis cs:51,375.227296162824)
--(axis cs:50,383.05036320048)
--(axis cs:49,390.604876380362)
--(axis cs:48,398.43105708122)
--(axis cs:47,407.13698401389)
--(axis cs:46,414.903515630968)
--(axis cs:45,425.591999702675)
--(axis cs:44,433.516581636389)
--(axis cs:43,443.924495089162)
--(axis cs:42,452.876747141538)
--(axis cs:41,464.679626403431)
--(axis cs:40,476.467957484348)
--(axis cs:39,487.460172909702)
--(axis cs:38,499.811546718739)
--(axis cs:37,512.685385330738)
--(axis cs:36,526.55465874393)
--(axis cs:35,540.46528250014)
--(axis cs:34,554.813711851874)
--(axis cs:33,571.231933063396)
--(axis cs:32,586.920358577499)
--(axis cs:31,603.834524193112)
--(axis cs:30,623.272070003081)
--(axis cs:29,642.938634000073)
--(axis cs:28,663.01799054504)
--(axis cs:27,682.534541499159)
--(axis cs:26,701.96885959899)
--(axis cs:25,722.468576082662)
--(axis cs:24,743.593060684346)
--(axis cs:23,763.962491474037)
--(axis cs:22,783.386438967456)
--(axis cs:21,801.998077137886)
--(axis cs:20,819.260609954455)
--(axis cs:19,834.403642453282)
--(axis cs:18,845.745826423215)
--(axis cs:17,856.154023166753)
--(axis cs:16,856.751637238076)
--(axis cs:15,852.27020845275)
--(axis cs:14,841.556217542271)
--(axis cs:13,820.715116307294)
--(axis cs:12,796.028111012506)
--(axis cs:11,777.971703900655)
--(axis cs:10,760.494792507238)
--(axis cs:9,637.604305581765)
--(axis cs:8,452.764377104638)
--(axis cs:7,318.610157423267)
--(axis cs:6,222.78145560426)
--(axis cs:5,155.358487287674)
--(axis cs:4,109.061728728838)
--(axis cs:3,76.2813707129388)
--(axis cs:2,53.6339202534889)
--(axis cs:1,35.8161133340113)
--(axis cs:0,20.3522562557546)
--cycle;

\path [fill=color1, fill opacity=0.3, very thin]
(axis cs:0,0)
--(axis cs:0,0)
--(axis cs:1,0)
--(axis cs:2,0)
--(axis cs:3,0)
--(axis cs:4,0)
--(axis cs:5,0)
--(axis cs:6,0)
--(axis cs:7,0)
--(axis cs:8,0)
--(axis cs:9,0)
--(axis cs:10,-1.97219232457229)
--(axis cs:11,-4.20694346327425)
--(axis cs:12,8.19875005024325)
--(axis cs:13,31.1727155912805)
--(axis cs:14,60.3074072244743)
--(axis cs:15,98.5353077961341)
--(axis cs:16,146.455027732284)
--(axis cs:17,190.67080244682)
--(axis cs:18,247.777872007535)
--(axis cs:19,307.018616665902)
--(axis cs:20,354.945727813227)
--(axis cs:21,397.077426449668)
--(axis cs:22,433.080866301637)
--(axis cs:23,462.357847195565)
--(axis cs:24,482.356520066829)
--(axis cs:25,496.798286997712)
--(axis cs:26,507.024198707667)
--(axis cs:27,510.727400703446)
--(axis cs:28,510.45413121671)
--(axis cs:29,506.949034537501)
--(axis cs:30,501.051913619278)
--(axis cs:31,491.186792100374)
--(axis cs:32,481.224278581757)
--(axis cs:33,470.944662330331)
--(axis cs:34,458.016386549695)
--(axis cs:35,447.158319657984)
--(axis cs:36,433.973425182502)
--(axis cs:37,420.868159163482)
--(axis cs:38,408.713396690536)
--(axis cs:39,395.182931356148)
--(axis cs:40,383.083919091888)
--(axis cs:41,372.11065239978)
--(axis cs:42,360.239442926724)
--(axis cs:43,347.797820061582)
--(axis cs:44,336.894737570755)
--(axis cs:45,326.169119503931)
--(axis cs:46,316.849101852836)
--(axis cs:47,307.188105949907)
--(axis cs:48,297.466301727975)
--(axis cs:49,288.237857172614)
--(axis cs:50,279.360346883334)
--(axis cs:51,272.161933107375)
--(axis cs:52,263.530317399316)
--(axis cs:53,255.763668713101)
--(axis cs:54,244.109884949175)
--(axis cs:55,242.375731864851)
--(axis cs:56,236.621458364644)
--(axis cs:57,230.529170640725)
--(axis cs:58,225.702203756361)
--(axis cs:59,219.959245334931)
--(axis cs:60,215.477774653799)
--(axis cs:61,211.145671956111)
--(axis cs:62,198.838286279159)
--(axis cs:63,201.954869073983)
--(axis cs:64,197.676697419006)
--(axis cs:65,194.448245380951)
--(axis cs:66,189.362123559514)
--(axis cs:67,185.503604631094)
--(axis cs:68,178.893458033458)
--(axis cs:69,175.880554781472)
--(axis cs:70,172.611244327127)
--(axis cs:71,172.161856931915)
--(axis cs:72,169.974223338996)
--(axis cs:73,164.24121904768)
--(axis cs:74,158.523223574451)
--(axis cs:75,162.735989525027)
--(axis cs:76,160.032998635997)
--(axis cs:77,157.653462314456)
--(axis cs:78,155.993934695325)
--(axis cs:79,154.649367053442)
--(axis cs:79,283.380044711264)
--(axis cs:79,283.380044711264)
--(axis cs:78,287.094300598793)
--(axis cs:77,289.277910234563)
--(axis cs:76,292.231707246356)
--(axis cs:75,295.077735965169)
--(axis cs:74,298.545403876529)
--(axis cs:73,301.631329971928)
--(axis cs:72,304.702247249239)
--(axis cs:71,307.906770519065)
--(axis cs:70,311.702481163069)
--(axis cs:69,316.570425610685)
--(axis cs:68,320.204581182229)
--(axis cs:67,323.290513015965)
--(axis cs:66,328.000621538525)
--(axis cs:65,331.649793834735)
--(axis cs:64,336.450753561386)
--(axis cs:63,340.898072102487)
--(axis cs:62,348.24014509339)
--(axis cs:61,351.177857455654)
--(axis cs:60,355.845754757966)
--(axis cs:59,360.60938211605)
--(axis cs:58,366.278188400502)
--(axis cs:57,372.637496025942)
--(axis cs:56,378.221678890258)
--(axis cs:55,383.76152303711)
--(axis cs:54,392.909722893963)
--(axis cs:53,396.255939130036)
--(axis cs:52,403.146153188919)
--(axis cs:51,409.857674735762)
--(axis cs:50,417.914162920588)
--(axis cs:49,425.938613415621)
--(axis cs:48,432.906247291633)
--(axis cs:47,441.066796010877)
--(axis cs:46,449.562662853046)
--(axis cs:45,458.811272652932)
--(axis cs:44,467.526831056696)
--(axis cs:43,477.986493663908)
--(axis cs:42,488.848792367394)
--(axis cs:41,499.556014266887)
--(axis cs:40,510.621963261053)
--(axis cs:39,522.464127467382)
--(axis cs:38,534.110132721228)
--(axis cs:37,547.229880052204)
--(axis cs:36,558.752065013576)
--(axis cs:35,569.998543087114)
--(axis cs:34,581.44439776403)
--(axis cs:33,592.084749434375)
--(axis cs:32,601.687486124125)
--(axis cs:31,610.313207899626)
--(axis cs:30,616.212792263075)
--(axis cs:29,619.227436050734)
--(axis cs:28,621.536064861721)
--(axis cs:27,620.576520865182)
--(axis cs:26,614.465997370764)
--(axis cs:25,602.642889472877)
--(axis cs:24,585.643479933171)
--(axis cs:23,563.014701824043)
--(axis cs:22,533.00736899248)
--(axis cs:21,498.128455903273)
--(axis cs:20,455.211134931871)
--(axis cs:19,404.96177549096)
--(axis cs:18,353.633892698347)
--(axis cs:17,296.770374023768)
--(axis cs:16,233.927325208892)
--(axis cs:15,174.847045145042)
--(axis cs:14,122.320043755918)
--(axis cs:13,79.0429706832293)
--(axis cs:12,45.0365440674038)
--(axis cs:11,18.1677277769997)
--(axis cs:10,2.4035648735919)
--(axis cs:9,0)
--(axis cs:8,0)
--(axis cs:7,0)
--(axis cs:6,0)
--(axis cs:5,0)
--(axis cs:4,0)
--(axis cs:3,0)
--(axis cs:2,0)
--(axis cs:1,0)
--(axis cs:0,0)
--cycle;

\addplot [semithick, color0]
table {%
0 17.6862745098039
1 30.25
2 45.0245098039216
3 63.4558823529412
4 89.7058823529412
5 125.906862745098
6 180.014705882353
7 255.480392156863
8 362.137254901961
9 512.676470588235
10 647.990196078431
11 699.03431372549
12 724.740196078431
13 747.274509803922
14 765.03431372549
15 773.794117647059
16 777.093137254902
17 774.647058823529
18 764.161764705882
19 751.78431372549
20 735.921568627451
21 718.53431372549
22 700.5
23 680.789215686274
24 660.588235294118
25 639.352941176471
26 619.191176470588
27 599.03431372549
28 578.627450980392
29 559.779411764706
30 540.789215686274
31 522.455882352941
32 505.264705882353
33 488.995098039216
34 472.004901960784
35 456.524509803922
36 441.754901960784
37 428.073529411765
38 414.857843137255
39 402.490196078431
40 390.838235294118
41 378.720588235294
42 367.441176470588
43 356.754901960784
44 346.539215686275
45 337.75
46 328.049019607843
47 320.205882352941
48 311.901960784314
49 304.014705882353
50 296.382352941176
51 288.514705882353
52 281.352941176471
53 273.700980392157
54 267.107843137255
55 261.127450980392
56 254.833333333333
57 249.093137254902
58 242.852941176471
59 237.53431372549
60 231.730392156863
61 226.083333333333
62 220.588235294118
63 216.25
64 211.936274509804
65 207.039215686275
66 203.514705882353
67 198.833333333333
68 194.901960784314
69 191.75
70 188.009803921569
71 184.916666666667
72 181.754901960784
73 179.504901960784
74 177.730392156863
75 175.68137254902
76 173.833333333333
77 172.882352941176
78 171.823529411765
79 170.578431372549
};
\addlegendentry{current cases}
\addplot [semithick, color1]
table {%
0 0
1 0
2 0
3 0
4 0
5 0
6 0
7 0
8 0
9 0
10 0.215686274509804
11 6.98039215686275
12 26.6176470588235
13 55.1078431372549
14 91.3137254901961
15 136.691176470588
16 190.191176470588
17 243.720588235294
18 300.705882352941
19 355.990196078431
20 405.078431372549
21 447.602941176471
22 483.044117647059
23 512.686274509804
24 534
25 549.720588235294
26 560.745098039216
27 565.651960784314
28 565.995098039216
29 563.088235294118
30 558.632352941176
31 550.75
32 541.455882352941
33 531.514705882353
34 519.730392156863
35 508.578431372549
36 496.362745098039
37 484.049019607843
38 471.411764705882
39 458.823529411765
40 446.852941176471
41 435.833333333333
42 424.544117647059
43 412.892156862745
44 402.210784313725
45 392.490196078431
46 383.205882352941
47 374.127450980392
48 365.186274509804
49 357.088235294118
50 348.637254901961
51 341.009803921569
52 333.338235294118
53 326.009803921569
54 318.509803921569
55 313.06862745098
56 307.421568627451
57 301.583333333333
58 295.990196078431
59 290.28431372549
60 285.661764705882
61 281.161764705882
62 273.539215686275
63 271.426470588235
64 267.063725490196
65 263.049019607843
66 258.68137254902
67 254.397058823529
68 249.549019607843
69 246.225490196078
70 242.156862745098
71 240.03431372549
72 237.338235294118
73 232.936274509804
74 228.53431372549
75 228.906862745098
76 226.132352941176
77 223.46568627451
78 221.544117647059
79 219.014705882353
};
\addlegendentry{isolated}
\end{axis}

\end{tikzpicture}

%% file: pic/new22/22quct041_1k.tex
\begin{tikzpicture}

\definecolor{color0}{rgb}{0.886274509803922,0.290196078431373,0.2}
\definecolor{color1}{rgb}{0.203921568627451,0.541176470588235,0.741176470588235}
\pgfmathsetlengthmacro\MajorTickLength{
      \pgfkeysvalueof{/pgfplots/major tick length} * 0.5
    }
\begin{axis}[
axis line style={white},
legend cell align={left},
width = 0.37\linewidth,
height = 1in,
major tick length=\MajorTickLength,
font= \fontsize{7}{7.5}\selectfont,
tick align=inside,
tick pos=left,
x grid style={white},
xlabel={Days},
xmajorgrids,
xmin=0, xmax=80,
xtick style={color=white!33.3333333333333!black},
y grid style={white},
xtick={0,20,40,60,80},
xticklabels={0,20,40,60,80},
ytick={0,1000,2000,3000},
yticklabel style={rotate=90},
yticklabels={0,1k,2k,3k},
ylabel={Strong QT},
x label style={at={(0.5, 0.3)}},
ymajorgrids,
y label style={at={(0.4,0.5)}},
ymin=-149.207704175957, ymax=3000,
ytick style={color=white!33.3333333333333!black}
]
\path [fill=color0, fill opacity=0.3, very thin]
(axis cs:0,19.745926642839)
--(axis cs:0,14.484073357161)
--(axis cs:1,24.441958298414)
--(axis cs:2,35.2864470101197)
--(axis cs:3,48.6032791292085)
--(axis cs:4,66.8095119977174)
--(axis cs:5,93.0322245614286)
--(axis cs:6,130.129196859657)
--(axis cs:7,183.939104371963)
--(axis cs:8,262.541570322509)
--(axis cs:9,369.405466362433)
--(axis cs:10,525.228749903463)
--(axis cs:11,647.56194322293)
--(axis cs:12,702.670553967244)
--(axis cs:13,758.536023111635)
--(axis cs:14,808.40677191872)
--(axis cs:15,850.967737100789)
--(axis cs:16,884.751824810814)
--(axis cs:17,914.708502999874)
--(axis cs:18,942.576090638238)
--(axis cs:19,962.883035234904)
--(axis cs:20,979.678180949332)
--(axis cs:21,995.80548870277)
--(axis cs:22,1011.3445994196)
--(axis cs:23,1023.39094733729)
--(axis cs:24,1032.15654151656)
--(axis cs:25,1043.4817383901)
--(axis cs:26,1059.28923995095)
--(axis cs:27,1070.1180564455)
--(axis cs:28,1082.62214918643)
--(axis cs:29,1092.46375386982)
--(axis cs:30,1102.67152213687)
--(axis cs:31,1110.66520883457)
--(axis cs:32,1120.34554569671)
--(axis cs:33,1129.45913154147)
--(axis cs:34,1138.73784858869)
--(axis cs:35,1147.83763685397)
--(axis cs:36,1155.78852295691)
--(axis cs:37,1167.3047700511)
--(axis cs:38,1176.79280817441)
--(axis cs:39,1185.60223345542)
--(axis cs:40,1192.90975862348)
--(axis cs:41,1201.37858564363)
--(axis cs:42,1209.80322620177)
--(axis cs:43,1217.91232162235)
--(axis cs:44,1221.79984450881)
--(axis cs:45,1226.84959969764)
--(axis cs:46,1228.97945184041)
--(axis cs:47,1235.50378659029)
--(axis cs:48,1243.78398341046)
--(axis cs:49,1249.30455178126)
--(axis cs:50,1256.43974327512)
--(axis cs:51,1263.66361583885)
--(axis cs:52,1268.51222626244)
--(axis cs:53,1275.57409642366)
--(axis cs:54,1285.93686136682)
--(axis cs:55,1291.64845475828)
--(axis cs:56,1297.70072695725)
--(axis cs:57,1305.57892253764)
--(axis cs:58,1313.42710226458)
--(axis cs:59,1321.85190103571)
--(axis cs:60,1329.77019953289)
--(axis cs:61,1337.68100364423)
--(axis cs:62,1348.98852185692)
--(axis cs:63,1357.93563215333)
--(axis cs:64,1370.39807855131)
--(axis cs:65,1385.96752473375)
--(axis cs:66,1396.79581931211)
--(axis cs:67,1409.05026449813)
--(axis cs:68,1421.62593682702)
--(axis cs:69,1431.10205470176)
--(axis cs:70,1442.41402292222)
--(axis cs:71,1444.10442139775)
--(axis cs:72,1441.17571238911)
--(axis cs:73,1435.78650962728)
--(axis cs:74,1426.34160367492)
--(axis cs:75,1420.02689648074)
--(axis cs:76,1409.44372649861)
--(axis cs:77,1397.58914296852)
--(axis cs:78,1383.64608089363)
--(axis cs:79,1365.72848117705)
--(axis cs:79,2696.43151882295)
--(axis cs:79,2696.43151882295)
--(axis cs:78,2700.26391910637)
--(axis cs:77,2699.70085703148)
--(axis cs:76,2699.72627350139)
--(axis cs:75,2693.13310351926)
--(axis cs:74,2681.48839632508)
--(axis cs:73,2670.01349037272)
--(axis cs:72,2663.01428761089)
--(axis cs:71,2652.14557860225)
--(axis cs:70,2637.58597707778)
--(axis cs:69,2626.01794529824)
--(axis cs:68,2609.06406317298)
--(axis cs:67,2595.06973550187)
--(axis cs:66,2577.69418068789)
--(axis cs:65,2565.78247526625)
--(axis cs:64,2553.75192144869)
--(axis cs:63,2538.38436784667)
--(axis cs:62,2522.86147814308)
--(axis cs:61,2501.43899635577)
--(axis cs:60,2485.71980046711)
--(axis cs:59,2464.64809896429)
--(axis cs:58,2444.50289773542)
--(axis cs:57,2421.72107746236)
--(axis cs:56,2402.57927304275)
--(axis cs:55,2379.31154524172)
--(axis cs:54,2354.30313863318)
--(axis cs:53,2332.01590357634)
--(axis cs:52,2305.15777373756)
--(axis cs:51,2274.65638416115)
--(axis cs:50,2246.41025672488)
--(axis cs:49,2221.76544821874)
--(axis cs:48,2192.81601658954)
--(axis cs:47,2165.0962134097)
--(axis cs:46,2134.92054815959)
--(axis cs:45,2106.06040030236)
--(axis cs:44,2078.44015549119)
--(axis cs:43,2047.43767837765)
--(axis cs:42,2015.40677379823)
--(axis cs:41,1983.89141435637)
--(axis cs:40,1954.79024137652)
--(axis cs:39,1919.97776654458)
--(axis cs:38,1888.01719182559)
--(axis cs:37,1855.1752299489)
--(axis cs:36,1823.44147704309)
--(axis cs:35,1791.45236314603)
--(axis cs:34,1761.12215141131)
--(axis cs:33,1727.30086845853)
--(axis cs:32,1690.30445430329)
--(axis cs:31,1655.23479116543)
--(axis cs:30,1622.26847786313)
--(axis cs:29,1590.30624613018)
--(axis cs:28,1555.07785081357)
--(axis cs:27,1520.7119435545)
--(axis cs:26,1487.88076004905)
--(axis cs:25,1453.4282616099)
--(axis cs:24,1418.80345848344)
--(axis cs:23,1383.39905266271)
--(axis cs:22,1347.0454005804)
--(axis cs:21,1309.13451129723)
--(axis cs:20,1268.84181905067)
--(axis cs:19,1230.4869647651)
--(axis cs:18,1188.23390936176)
--(axis cs:17,1143.80149700013)
--(axis cs:16,1095.75817518919)
--(axis cs:15,1045.74226289921)
--(axis cs:14,991.72322808128)
--(axis cs:13,932.733976888365)
--(axis cs:12,870.369446032756)
--(axis cs:11,819.568056777071)
--(axis cs:10,768.011250096537)
--(axis cs:9,604.254533637567)
--(axis cs:8,423.048429677491)
--(axis cs:7,296.430895628037)
--(axis cs:6,208.260803140343)
--(axis cs:5,145.017775438571)
--(axis cs:4,102.670488002283)
--(axis cs:3,72.9967208707915)
--(axis cs:2,51.4735529898803)
--(axis cs:1,33.938041701586)
--(axis cs:0,19.745926642839)
--cycle;

\path [fill=color1, fill opacity=0.3, very thin]
(axis cs:0,0)
--(axis cs:0,0)
--(axis cs:1,0)
--(axis cs:2,0)
--(axis cs:3,0)
--(axis cs:4,0)
--(axis cs:5,0)
--(axis cs:6,0)
--(axis cs:7,0)
--(axis cs:8,0)
--(axis cs:9,0)
--(axis cs:10,-6.63563947312671)
--(axis cs:11,-2.5933255519884)
--(axis cs:12,60.0338849687877)
--(axis cs:13,133.598717967874)
--(axis cs:14,185.974880158226)
--(axis cs:15,246.260990546432)
--(axis cs:16,318.708318235014)
--(axis cs:17,394.943002642812)
--(axis cs:18,471.275251550501)
--(axis cs:19,544.590654165651)
--(axis cs:20,611.542004097318)
--(axis cs:21,671.407168384741)
--(axis cs:22,726.012918063862)
--(axis cs:23,772.003567695946)
--(axis cs:24,813.811220178183)
--(axis cs:25,848.274900108647)
--(axis cs:26,877.050197826338)
--(axis cs:27,903.062710564272)
--(axis cs:28,927.594526941929)
--(axis cs:29,948.351432918552)
--(axis cs:30,968.888582243571)
--(axis cs:31,986.727296261687)
--(axis cs:32,1003.0430983243)
--(axis cs:33,1019.1775458288)
--(axis cs:34,1033.79407424872)
--(axis cs:35,1049.07446696797)
--(axis cs:36,1061.47000272884)
--(axis cs:37,1075.99952044318)
--(axis cs:38,1087.34784046845)
--(axis cs:39,1099.94807238458)
--(axis cs:40,1110.05631151737)
--(axis cs:41,1125.32981996139)
--(axis cs:42,1137.83569313768)
--(axis cs:43,1148.9178687451)
--(axis cs:44,1162.60521450541)
--(axis cs:45,1174.64516427527)
--(axis cs:46,1188.6379971289)
--(axis cs:47,1201.63443505333)
--(axis cs:48,1213.70385286464)
--(axis cs:49,1223.71540359277)
--(axis cs:50,1235.91626476419)
--(axis cs:51,1246.58809854922)
--(axis cs:52,1257.93255974904)
--(axis cs:53,1267.21940775479)
--(axis cs:54,1274.89826067795)
--(axis cs:55,1283.59807785827)
--(axis cs:56,1293.49216064935)
--(axis cs:57,1304.83004408805)
--(axis cs:58,1315.80385505099)
--(axis cs:59,1325.87096729797)
--(axis cs:60,1336.1126378434)
--(axis cs:61,1346.37860424416)
--(axis cs:62,1358.50801609073)
--(axis cs:63,1370.15289246082)
--(axis cs:64,1382.13149417)
--(axis cs:65,1393.73391486745)
--(axis cs:66,1404.72914054887)
--(axis cs:67,1420.59590769175)
--(axis cs:68,1431.99029267003)
--(axis cs:69,1445.08188964367)
--(axis cs:70,1459.78088028383)
--(axis cs:71,1475.01858520906)
--(axis cs:72,1490.92873022214)
--(axis cs:73,1505.54817849594)
--(axis cs:74,1521.60837912936)
--(axis cs:75,1535.40829742813)
--(axis cs:76,1549.33766114564)
--(axis cs:77,1562.94737715545)
--(axis cs:78,1573.64484706743)
--(axis cs:79,1581.44434541653)
--(axis cs:79,2844.80565458347)
--(axis cs:79,2844.80565458347)
--(axis cs:78,2826.97515293257)
--(axis cs:77,2807.76262284455)
--(axis cs:76,2782.42233885436)
--(axis cs:75,2759.85170257187)
--(axis cs:74,2737.93162087064)
--(axis cs:73,2715.40182150406)
--(axis cs:72,2692.95126977786)
--(axis cs:71,2669.40141479094)
--(axis cs:70,2644.90911971617)
--(axis cs:69,2621.59811035633)
--(axis cs:68,2598.14970732997)
--(axis cs:67,2571.67409230825)
--(axis cs:66,2544.41085945113)
--(axis cs:65,2515.32608513255)
--(axis cs:64,2485.51850583)
--(axis cs:63,2455.62710753918)
--(axis cs:62,2426.45198390927)
--(axis cs:61,2395.81139575584)
--(axis cs:60,2364.3173621566)
--(axis cs:59,2333.72903270203)
--(axis cs:58,2304.38614494901)
--(axis cs:57,2268.80995591195)
--(axis cs:56,2236.37783935065)
--(axis cs:55,2200.57192214173)
--(axis cs:54,2166.74173932205)
--(axis cs:53,2129.70059224521)
--(axis cs:52,2095.70744025096)
--(axis cs:51,2059.52190145078)
--(axis cs:50,2025.09373523581)
--(axis cs:49,1991.13459640723)
--(axis cs:48,1953.52614713536)
--(axis cs:47,1918.29556494667)
--(axis cs:46,1883.8420028711)
--(axis cs:45,1849.27483572473)
--(axis cs:44,1813.05478549459)
--(axis cs:43,1778.5021312549)
--(axis cs:42,1742.22430686232)
--(axis cs:41,1704.65018003861)
--(axis cs:40,1668.66368848263)
--(axis cs:39,1631.33192761542)
--(axis cs:38,1595.12215953155)
--(axis cs:37,1557.59047955682)
--(axis cs:36,1519.53999727116)
--(axis cs:35,1482.58553303203)
--(axis cs:34,1446.94592575128)
--(axis cs:33,1409.3524541712)
--(axis cs:32,1370.2869016757)
--(axis cs:31,1331.99270373831)
--(axis cs:30,1292.04141775643)
--(axis cs:29,1252.79856708145)
--(axis cs:28,1210.80547305807)
--(axis cs:27,1165.95728943573)
--(axis cs:26,1119.93980217366)
--(axis cs:25,1069.26509989135)
--(axis cs:24,1017.46877982182)
--(axis cs:23,963.776432304054)
--(axis cs:22,905.067081936138)
--(axis cs:21,839.052831615259)
--(axis cs:20,768.717995902682)
--(axis cs:19,697.609345834349)
--(axis cs:18,619.654748449499)
--(axis cs:17,534.566997357188)
--(axis cs:16,448.781681764986)
--(axis cs:15,364.629009453568)
--(axis cs:14,283.395119841774)
--(axis cs:13,219.401282032126)
--(axis cs:12,172.386115031212)
--(axis cs:11,88.7933255519884)
--(axis cs:10,18.8156394731267)
--(axis cs:9,0)
--(axis cs:8,0)
--(axis cs:7,0)
--(axis cs:6,0)
--(axis cs:5,0)
--(axis cs:4,0)
--(axis cs:3,0)
--(axis cs:2,0)
--(axis cs:1,0)
--(axis cs:0,0)
--cycle;

\addplot [semithick, color0]
table {%
0 17.115
1 29.19
2 43.38
3 60.8
4 84.74
5 119.025
6 169.195
7 240.185
8 342.795
9 486.83
10 646.62
11 733.565
12 786.52
13 845.635
14 900.065
15 948.355
16 990.255
17 1029.255
18 1065.405
19 1096.685
20 1124.26
21 1152.47
22 1179.195
23 1203.395
24 1225.48
25 1248.455
26 1273.585
27 1295.415
28 1318.85
29 1341.385
30 1362.47
31 1382.95
32 1405.325
33 1428.38
34 1449.93
35 1469.645
36 1489.615
37 1511.24
38 1532.405
39 1552.79
40 1573.85
41 1592.635
42 1612.605
43 1632.675
44 1650.12
45 1666.455
46 1681.95
47 1700.3
48 1718.3
49 1735.535
50 1751.425
51 1769.16
52 1786.835
53 1803.795
54 1820.12
55 1835.48
56 1850.14
57 1863.65
58 1878.965
59 1893.25
60 1907.745
61 1919.56
62 1935.925
63 1948.16
64 1962.075
65 1975.875
66 1987.245
67 2002.06
68 2015.345
69 2028.56
70 2040
71 2048.125
72 2052.095
73 2052.9
74 2053.915
75 2056.58
76 2054.585
77 2048.645
78 2041.955
79 2031.08
};
\addplot [semithick, color1]
table {%
0 0
1 0
2 0
3 0
4 0
5 0
6 0
7 0
8 0
9 0
10 6.09
11 43.1
12 116.21
13 176.5
14 234.685
15 305.445
16 383.745
17 464.755
18 545.465
19 621.1
20 690.13
21 755.23
22 815.54
23 867.89
24 915.64
25 958.77
26 998.495
27 1034.51
28 1069.2
29 1100.575
30 1130.465
31 1159.36
32 1186.665
33 1214.265
34 1240.37
35 1265.83
36 1290.505
37 1316.795
38 1341.235
39 1365.64
40 1389.36
41 1414.99
42 1440.03
43 1463.71
44 1487.83
45 1511.96
46 1536.24
47 1559.965
48 1583.615
49 1607.425
50 1630.505
51 1653.055
52 1676.82
53 1698.46
54 1720.82
55 1742.085
56 1764.935
57 1786.82
58 1810.095
59 1829.8
60 1850.215
61 1871.095
62 1892.48
63 1912.89
64 1933.825
65 1954.53
66 1974.57
67 1996.135
68 2015.07
69 2033.34
70 2052.345
71 2072.21
72 2091.94
73 2110.475
74 2129.77
75 2147.63
76 2165.88
77 2185.355
78 2200.31
79 2213.125
};
\end{axis}

\end{tikzpicture}

%% file: pic/new22/22quct041_3k.tex
\begin{tikzpicture}

\definecolor{color0}{rgb}{0.886274509803922,0.290196078431373,0.2}
\definecolor{color1}{rgb}{0.203921568627451,0.541176470588235,0.741176470588235}
\pgfmathsetlengthmacro\MajorTickLength{
      \pgfkeysvalueof{/pgfplots/major tick length} * 0.5
    }
\begin{axis}[
axis line style={white},
legend cell align={left},
width = 0.37\linewidth,
height = 1in,
major tick length=\MajorTickLength,
font= \fontsize{7}{7.5}\selectfont,
tick align=inside,
tick pos=left,
x grid style={white},
xlabel={Days},
xmajorgrids,
xmin=0, xmax=80,
xtick style={color=white!33.3333333333333!black},
y grid style={white},
xtick={0,20,40,60,80},
xticklabels={0,20,40,60,80},
ytick={0,700,1400},
yticklabel style={rotate=90},
yticklabels={0,800,1.6k},
x label style={at={(0.5, 0.3)}},
ymajorgrids,
y label style={at={(0.55,0.5)}},
ymin=-71.6316150697779, ymax=1400,
ytick style={color=white!33.3333333333333!black}
]
\path [fill=color0, fill opacity=0.3, very thin]
(axis cs:0,20.0124723592281)
--(axis cs:0,13.9494324026767)
--(axis cs:1,23.5780344158377)
--(axis cs:2,34.9610962268748)
--(axis cs:3,48.3303821399321)
--(axis cs:4,66.5061221798846)
--(axis cs:5,92.6122525463683)
--(axis cs:6,130.546362622282)
--(axis cs:7,184.375228377066)
--(axis cs:8,260.754229815869)
--(axis cs:9,376.049780297627)
--(axis cs:10,530.644910440583)
--(axis cs:11,636.130137275294)
--(axis cs:12,689.892819129189)
--(axis cs:13,730.160692597175)
--(axis cs:14,769.174460868197)
--(axis cs:15,798.246117651795)
--(axis cs:16,817.870195313326)
--(axis cs:17,832.001675611075)
--(axis cs:18,843.484219933036)
--(axis cs:19,849.356358637547)
--(axis cs:20,852.484227077901)
--(axis cs:21,852.316886961669)
--(axis cs:22,851.808366441967)
--(axis cs:23,848.272346535856)
--(axis cs:24,844.125707767664)
--(axis cs:25,839.532607799867)
--(axis cs:26,830.742511941618)
--(axis cs:27,825.268443638334)
--(axis cs:28,819.286648306481)
--(axis cs:29,813.665635644619)
--(axis cs:30,807.080642574893)
--(axis cs:31,800.538559666617)
--(axis cs:32,793.677402815793)
--(axis cs:33,786.594701198391)
--(axis cs:34,778.806297803452)
--(axis cs:35,775.612651946958)
--(axis cs:36,771.783207132541)
--(axis cs:37,766.308507079328)
--(axis cs:38,763.307882518561)
--(axis cs:39,758.248737163993)
--(axis cs:40,754.159368422671)
--(axis cs:41,748.990985305999)
--(axis cs:42,745.006861927531)
--(axis cs:43,743.289307462872)
--(axis cs:44,741.10610094786)
--(axis cs:45,739.158578561481)
--(axis cs:46,736.941888614412)
--(axis cs:47,733.031774556034)
--(axis cs:48,731.066482294088)
--(axis cs:49,726.978603266253)
--(axis cs:50,722.599201397841)
--(axis cs:51,718.188256446053)
--(axis cs:52,716.080278569377)
--(axis cs:53,713.063107538039)
--(axis cs:54,712.171927449324)
--(axis cs:55,708.946535531205)
--(axis cs:56,704.444712308032)
--(axis cs:57,700.210014545105)
--(axis cs:58,698.569567010452)
--(axis cs:59,697.436909922125)
--(axis cs:60,697.896376723144)
--(axis cs:61,696.124712970922)
--(axis cs:62,696.001075864458)
--(axis cs:63,695.060512532685)
--(axis cs:64,695.396461047883)
--(axis cs:65,695.247863883597)
--(axis cs:66,694.411587493741)
--(axis cs:67,695.451364872982)
--(axis cs:68,693.848199828567)
--(axis cs:69,692.755403633758)
--(axis cs:70,690.217635108061)
--(axis cs:71,689.206781547368)
--(axis cs:72,685.282170012069)
--(axis cs:73,683.200388245514)
--(axis cs:74,679.758884271836)
--(axis cs:75,676.82342328816)
--(axis cs:76,671.077942407663)
--(axis cs:77,668.296722916113)
--(axis cs:78,665.694398617329)
--(axis cs:79,666.063510973565)
--(axis cs:79,1053.51744140739)
--(axis cs:79,1053.51744140739)
--(axis cs:78,1051.96274423981)
--(axis cs:77,1052.83661041722)
--(axis cs:76,1054.92205759234)
--(axis cs:75,1060.84324337851)
--(axis cs:74,1065.12683001388)
--(axis cs:73,1071.24723080211)
--(axis cs:72,1079.06068713079)
--(axis cs:71,1083.57417083358)
--(axis cs:70,1089.62998393956)
--(axis cs:69,1095.09221541386)
--(axis cs:68,1099.27560969524)
--(axis cs:67,1105.03434941273)
--(axis cs:66,1109.93126964912)
--(axis cs:65,1112.3902313545)
--(axis cs:64,1117.16544371402)
--(axis cs:63,1121.02520175303)
--(axis cs:62,1122.84654318316)
--(axis cs:61,1126.3229060767)
--(axis cs:60,1129.09409946733)
--(axis cs:59,1130.55356626835)
--(axis cs:58,1131.58281394193)
--(axis cs:57,1134.56141402632)
--(axis cs:56,1135.87909721578)
--(axis cs:55,1139.74870256403)
--(axis cs:54,1140.16140588401)
--(axis cs:53,1139.04165436672)
--(axis cs:52,1135.83400714491)
--(axis cs:51,1132.61174355395)
--(axis cs:50,1131.59127479264)
--(axis cs:49,1130.27853959089)
--(axis cs:48,1132.0097081821)
--(axis cs:47,1129.80632068206)
--(axis cs:46,1129.52477805225)
--(axis cs:45,1129.1938023909)
--(axis cs:44,1130.93199429024)
--(axis cs:43,1130.69164491808)
--(axis cs:42,1127.90742378675)
--(axis cs:41,1126.68520517019)
--(axis cs:40,1130.47872681542)
--(axis cs:39,1130.67507235982)
--(axis cs:38,1131.04449843382)
--(axis cs:37,1129.4629214921)
--(axis cs:36,1131.81679286746)
--(axis cs:35,1131.96830043399)
--(axis cs:34,1129.75560695845)
--(axis cs:33,1128.14815594447)
--(axis cs:32,1125.11307337468)
--(axis cs:31,1125.17572604767)
--(axis cs:30,1124.08126218701)
--(axis cs:29,1119.99150721252)
--(axis cs:28,1116.63716121733)
--(axis cs:27,1115.39822302833)
--(axis cs:26,1110.83844043933)
--(axis cs:25,1104.19120172394)
--(axis cs:24,1099.61714937519)
--(axis cs:23,1089.78479632129)
--(axis cs:22,1083.00115736756)
--(axis cs:21,1072.82597018119)
--(axis cs:20,1062.59196339829)
--(axis cs:19,1045.77697469579)
--(axis cs:18,1027.94435149554)
--(axis cs:17,1007.49356248416)
--(axis cs:16,986.386947543817)
--(axis cs:15,958.763406157729)
--(axis cs:14,928.873158179422)
--(axis cs:13,885.877402640921)
--(axis cs:12,839.469085632715)
--(axis cs:11,804.765100819944)
--(axis cs:10,762.745565749893)
--(axis cs:9,633.226410178563)
--(axis cs:8,451.836246374608)
--(axis cs:7,317.281914480077)
--(axis cs:6,222.501256425337)
--(axis cs:5,156.263937929822)
--(axis cs:4,107.77959210583)
--(axis cs:3,74.764855955306)
--(axis cs:2,51.6769990112204)
--(axis cs:1,34.9267274889242)
--(axis cs:0,20.0124723592281)
--cycle;

\path [fill=color1, fill opacity=0.3, very thin]
(axis cs:0,0)
--(axis cs:0,0)
--(axis cs:1,0)
--(axis cs:2,0)
--(axis cs:3,0)
--(axis cs:4,0)
--(axis cs:5,0)
--(axis cs:6,0)
--(axis cs:7,0)
--(axis cs:8,0)
--(axis cs:9,-3.50344529405405)
--(axis cs:10,-9.6226388647541)
--(axis cs:11,0.170579504015869)
--(axis cs:12,62.4604697353579)
--(axis cs:13,132.625445768595)
--(axis cs:14,186.014819842052)
--(axis cs:15,244.836912713493)
--(axis cs:16,314.229455207657)
--(axis cs:17,388.274511286703)
--(axis cs:18,460.146315056417)
--(axis cs:19,527.8155771538)
--(axis cs:20,586.946518871725)
--(axis cs:21,639.740279547237)
--(axis cs:22,684.573474020655)
--(axis cs:23,698.155160019909)
--(axis cs:24,747.249079804293)
--(axis cs:25,769.049099044806)
--(axis cs:26,786.94751839148)
--(axis cs:27,799.400077891047)
--(axis cs:28,809.122400689429)
--(axis cs:29,814.814995014255)
--(axis cs:30,820.244231232096)
--(axis cs:31,822.421796660375)
--(axis cs:32,823.093823760827)
--(axis cs:33,822.31180702995)
--(axis cs:34,821.495399635664)
--(axis cs:35,818.640750313146)
--(axis cs:36,815.146825749671)
--(axis cs:37,812.188468473919)
--(axis cs:38,809.098503272142)
--(axis cs:39,806.435085898577)
--(axis cs:40,802.39675989153)
--(axis cs:41,799.658030866996)
--(axis cs:42,797.894199052616)
--(axis cs:43,795.6081935639)
--(axis cs:44,793.850693305101)
--(axis cs:45,791.482453445169)
--(axis cs:46,790.42545334485)
--(axis cs:47,789.448940612825)
--(axis cs:48,788.909809631116)
--(axis cs:49,787.979156148632)
--(axis cs:50,786.395027142516)
--(axis cs:51,769.325364317698)
--(axis cs:52,786.011431782395)
--(axis cs:53,786.734861027079)
--(axis cs:54,785.862397530824)
--(axis cs:55,785.751298933795)
--(axis cs:56,784.137355363671)
--(axis cs:57,782.372747198989)
--(axis cs:58,782.739767337605)
--(axis cs:59,765.594837854889)
--(axis cs:60,781.082535739471)
--(axis cs:61,779.996917472706)
--(axis cs:62,779.63508947895)
--(axis cs:63,780.215753887959)
--(axis cs:64,765.258304416226)
--(axis cs:65,781.295311759158)
--(axis cs:66,780.810646659294)
--(axis cs:67,780.585354748293)
--(axis cs:68,782.851229845061)
--(axis cs:69,786.736528807171)
--(axis cs:70,787.221745996541)
--(axis cs:71,789.303579882344)
--(axis cs:72,791.410795403435)
--(axis cs:73,778.576448097613)
--(axis cs:74,796.439133596133)
--(axis cs:75,782.153452408829)
--(axis cs:76,785.003788724072)
--(axis cs:77,802.414431196304)
--(axis cs:78,804.37681483869)
--(axis cs:79,789.016454887471)
--(axis cs:79,1226.48830701729)
--(axis cs:79,1226.48830701729)
--(axis cs:78,1221.70889944702)
--(axis cs:77,1222.7855688037)
--(axis cs:76,1229.97716365688)
--(axis cs:75,1230.0560714007)
--(axis cs:74,1224.02753307053)
--(axis cs:73,1230.55688523572)
--(axis cs:72,1223.59872840609)
--(axis cs:71,1221.32499154623)
--(axis cs:70,1224.50206352727)
--(axis cs:69,1223.26347119283)
--(axis cs:68,1223.22496063113)
--(axis cs:67,1220.87178810885)
--(axis cs:66,1221.21792476928)
--(axis cs:65,1218.4094501456)
--(axis cs:64,1223.16074320282)
--(axis cs:63,1214.63186515966)
--(axis cs:62,1210.70776766391)
--(axis cs:61,1205.33641586063)
--(axis cs:60,1200.25079759386)
--(axis cs:59,1202.69087643082)
--(axis cs:58,1193.95547075763)
--(axis cs:57,1191.3034432772)
--(axis cs:56,1186.70073987442)
--(axis cs:55,1180.91536773287)
--(axis cs:54,1178.15665008822)
--(axis cs:53,1174.96990087768)
--(axis cs:52,1171.31237774142)
--(axis cs:51,1171.03654044421)
--(axis cs:50,1161.00497285748)
--(axis cs:49,1157.41132004184)
--(axis cs:48,1153.1378094165)
--(axis cs:47,1149.08439272051)
--(axis cs:46,1146.33645141705)
--(axis cs:45,1140.90802274531)
--(axis cs:44,1136.33025907585)
--(axis cs:43,1132.23942548372)
--(axis cs:42,1128.30580094738)
--(axis cs:41,1123.44673103777)
--(axis cs:40,1118.86990677514)
--(axis cs:39,1113.47919981571)
--(axis cs:38,1107.63483006119)
--(axis cs:37,1102.00200771656)
--(axis cs:36,1095.41507901223)
--(axis cs:35,1086.26401159162)
--(axis cs:34,1078.09507655481)
--(axis cs:33,1069.35485963672)
--(axis cs:32,1057.68712862012)
--(axis cs:31,1047.04487000629)
--(axis cs:30,1033.47005448219)
--(axis cs:29,1019.83262403336)
--(axis cs:28,1000.40140883438)
--(axis cs:27,979.895160204192)
--(axis cs:26,957.033433989472)
--(axis cs:25,931.884234288528)
--(axis cs:24,902.169967814755)
--(axis cs:23,882.882935218186)
--(axis cs:22,828.11224026506)
--(axis cs:21,779.12638711943)
--(axis cs:20,724.196338271133)
--(axis cs:19,666.965375227153)
--(axis cs:18,598.167970657869)
--(axis cs:17,523.315964903773)
--(axis cs:16,445.903878125677)
--(axis cs:15,367.639277762697)
--(axis cs:14,289.813751586519)
--(axis cs:13,227.345982802833)
--(axis cs:12,179.472863597975)
--(axis cs:11,107.638944305508)
--(axis cs:10,31.4988293409446)
--(axis cs:9,4.41773100833977)
--(axis cs:8,0)
--(axis cs:7,0)
--(axis cs:6,0)
--(axis cs:5,0)
--(axis cs:4,0)
--(axis cs:3,0)
--(axis cs:2,0)
--(axis cs:1,0)
--(axis cs:0,0)
--cycle;

\addplot [semithick, color0]
table {%
0 16.9809523809524
1 29.252380952381
2 43.3190476190476
3 61.5476190476191
4 87.1428571428571
5 124.438095238095
6 176.52380952381
7 250.828571428571
8 356.295238095238
9 504.638095238095
10 646.695238095238
11 720.447619047619
12 764.680952380952
13 808.019047619048
14 849.02380952381
15 878.504761904762
16 902.128571428571
17 919.747619047619
18 935.714285714286
19 947.566666666667
20 957.538095238095
21 962.571428571429
22 967.404761904762
23 969.028571428571
24 971.871428571429
25 971.861904761905
26 970.790476190476
27 970.333333333333
28 967.961904761905
29 966.828571428571
30 965.580952380952
31 962.857142857143
32 959.395238095238
33 957.371428571429
34 954.280952380952
35 953.790476190476
36 951.8
37 947.885714285714
38 947.176190476191
39 944.461904761905
40 942.319047619048
41 937.838095238095
42 936.457142857143
43 936.990476190476
44 936.019047619048
45 934.176190476191
46 933.233333333333
47 931.419047619048
48 931.538095238095
49 928.628571428571
50 927.095238095238
51 925.4
52 925.957142857143
53 926.052380952381
54 926.166666666667
55 924.347619047619
56 920.161904761905
57 917.385714285714
58 915.076190476191
59 913.995238095238
60 913.495238095238
61 911.22380952381
62 909.423809523809
63 908.042857142857
64 906.280952380952
65 903.819047619048
66 902.171428571429
67 900.242857142857
68 896.561904761905
69 893.923809523809
70 889.923809523809
71 886.390476190476
72 882.171428571429
73 877.22380952381
74 872.442857142857
75 868.833333333333
76 863
77 860.566666666667
78 858.828571428571
79 859.790476190476
};
\addplot [semithick, color1]
table {%
0 0
1 0
2 0
3 0
4 0
5 0
6 0
7 0
8 0
9 0.457142857142857
10 10.9380952380952
11 53.9047619047619
12 120.966666666667
13 179.985714285714
14 237.914285714286
15 306.238095238095
16 380.066666666667
17 455.795238095238
18 529.157142857143
19 597.390476190476
20 655.571428571429
21 709.433333333333
22 756.342857142857
23 790.519047619048
24 824.709523809524
25 850.466666666667
26 871.990476190476
27 889.647619047619
28 904.761904761905
29 917.323809523809
30 926.857142857143
31 934.733333333333
32 940.390476190476
33 945.833333333333
34 949.795238095238
35 952.452380952381
36 955.280952380952
37 957.095238095238
38 958.366666666667
39 959.957142857143
40 960.633333333333
41 961.552380952381
42 963.1
43 963.923809523809
44 965.090476190476
45 966.195238095238
46 968.380952380952
47 969.266666666667
48 971.02380952381
49 972.695238095238
50 973.7
51 970.180952380952
52 978.661904761905
53 980.852380952381
54 982.009523809524
55 983.333333333333
56 985.419047619048
57 986.838095238095
58 988.347619047619
59 984.142857142857
60 990.666666666667
61 992.666666666667
62 995.171428571429
63 997.423809523809
64 994.209523809524
65 999.852380952381
66 1001.01428571429
67 1000.72857142857
68 1003.0380952381
69 1005
70 1005.8619047619
71 1005.31428571429
72 1007.50476190476
73 1004.56666666667
74 1010.23333333333
75 1006.10476190476
76 1007.49047619048
77 1012.6
78 1013.04285714286
79 1007.75238095238
};
\end{axis}

\end{tikzpicture}

%% file: pic/new22/22quct041_10k.tex
\begin{tikzpicture}

\definecolor{color0}{rgb}{0.886274509803922,0.290196078431373,0.2}
\definecolor{color1}{rgb}{0.203921568627451,0.541176470588235,0.741176470588235}
\pgfmathsetlengthmacro\MajorTickLength{
      \pgfkeysvalueof{/pgfplots/major tick length} * 0.5
    }
\begin{axis}[
axis line style={white},
legend cell align={left},
width = 0.37\linewidth,
height = 1in,
major tick length=\MajorTickLength,
font= \fontsize{7}{7.5}\selectfont,
tick align=inside,
tick pos=left,
x grid style={white},
xlabel={Days},
xmajorgrids,
xmin=0, xmax=80,
xtick style={color=white!33.3333333333333!black},
y grid style={white},
xtick={0,20,40,60,80},
xticklabels={0,20,40,60,80},
ytick={0,400,800},
yticklabels={0,400,800},
x label style={at={(0.5, 0.3)}},
yticklabel style={rotate=90},
ymajorgrids,
y label style={at={(0.55,0.5)}},
ymin=-53.2602873434306, ymax=887.140253197472,
ytick style={color=white!33.3333333333333!black}
]
\path [fill=color0, fill opacity=0.3, very thin]
(axis cs:0,19.9496585712667)
--(axis cs:0,14.2217700001619)
--(axis cs:1,23.9129520217915)
--(axis cs:2,34.7466609320914)
--(axis cs:3,48.241869244887)
--(axis cs:4,66.7546103484016)
--(axis cs:5,93.3961729555537)
--(axis cs:6,130.004668273369)
--(axis cs:7,182.830342403498)
--(axis cs:8,259.417832440445)
--(axis cs:9,373.270294348471)
--(axis cs:10,525.519121002277)
--(axis cs:11,620.777749392409)
--(axis cs:12,652.654387935533)
--(axis cs:13,674.070101774342)
--(axis cs:14,688.549883963615)
--(axis cs:15,692.462368775166)
--(axis cs:16,689.648120364921)
--(axis cs:17,682.386365598112)
--(axis cs:18,670.576306047999)
--(axis cs:19,656.637336127034)
--(axis cs:20,641.071519814979)
--(axis cs:21,625.161279271185)
--(axis cs:22,607.402153495048)
--(axis cs:23,587.673414920181)
--(axis cs:24,568.820789601292)
--(axis cs:25,551.676689515361)
--(axis cs:26,535.802466287656)
--(axis cs:27,518.288974995529)
--(axis cs:28,501.054446727333)
--(axis cs:29,482.871550168109)
--(axis cs:30,466.722070241845)
--(axis cs:31,450.301856724747)
--(axis cs:32,435.484060514483)
--(axis cs:33,420.333016237475)
--(axis cs:34,404.85757670765)
--(axis cs:35,391.851986087693)
--(axis cs:36,378.627426295847)
--(axis cs:37,365.809001824738)
--(axis cs:38,353.341874370011)
--(axis cs:39,341.119168781412)
--(axis cs:40,328.758437550621)
--(axis cs:41,318.076422396899)
--(axis cs:42,308.538122312333)
--(axis cs:43,298.537012557503)
--(axis cs:44,288.904647435958)
--(axis cs:45,280.300118535239)
--(axis cs:46,271.74568804622)
--(axis cs:47,263.491361739432)
--(axis cs:48,254.31386376379)
--(axis cs:49,245.872871790101)
--(axis cs:50,237.559942712226)
--(axis cs:51,229.537411425614)
--(axis cs:52,220.703579271212)
--(axis cs:53,212.681160856255)
--(axis cs:54,205.805304540323)
--(axis cs:55,198.084160033025)
--(axis cs:56,190.448529192721)
--(axis cs:57,182.904030220557)
--(axis cs:58,176.349738893406)
--(axis cs:59,168.705046307959)
--(axis cs:60,161.692528465075)
--(axis cs:61,155.789529104747)
--(axis cs:62,150.612648083738)
--(axis cs:63,146.045628641663)
--(axis cs:64,139.276952017121)
--(axis cs:65,133.006131753637)
--(axis cs:66,126.472084115613)
--(axis cs:67,121.159840252176)
--(axis cs:68,115.309772472968)
--(axis cs:69,110.157148198037)
--(axis cs:70,106.040531088177)
--(axis cs:71,100.9009975468)
--(axis cs:72,95.9303890136991)
--(axis cs:73,91.5970819302735)
--(axis cs:74,87.2428549148362)
--(axis cs:75,83.2345666875332)
--(axis cs:76,79.2723264202647)
--(axis cs:77,75.5938686325172)
--(axis cs:78,71.7609916323401)
--(axis cs:79,68.1488683932748)
--(axis cs:79,317.003512559106)
--(axis cs:79,317.003512559106)
--(axis cs:78,322.277103605755)
--(axis cs:77,327.82517898653)
--(axis cs:76,333.822911674973)
--(axis cs:75,338.6987666458)
--(axis cs:74,341.680954608973)
--(axis cs:73,345.73625140306)
--(axis cs:72,349.498182414872)
--(axis cs:71,352.651383405581)
--(axis cs:70,356.454707007061)
--(axis cs:69,360.652375611487)
--(axis cs:68,362.709275146079)
--(axis cs:67,366.535397843062)
--(axis cs:66,370.50886826534)
--(axis cs:65,373.774820627315)
--(axis cs:64,377.008762268593)
--(axis cs:63,381.96389516786)
--(axis cs:62,387.101637630547)
--(axis cs:61,392.343804228586)
--(axis cs:60,397.555090582544)
--(axis cs:59,402.475906072993)
--(axis cs:58,408.11692777326)
--(axis cs:57,411.867398350872)
--(axis cs:56,418.065756521565)
--(axis cs:55,425.039649490785)
--(axis cs:54,431.251838316819)
--(axis cs:53,438.595029619935)
--(axis cs:52,444.963087395454)
--(axis cs:51,451.7483028601)
--(axis cs:50,460.10672395444)
--(axis cs:49,467.393794876566)
--(axis cs:48,475.724231474305)
--(axis cs:47,483.680066831996)
--(axis cs:46,491.282883382351)
--(axis cs:45,499.337976702857)
--(axis cs:44,506.809638278328)
--(axis cs:43,516.586796966306)
--(axis cs:42,523.785687211477)
--(axis cs:41,534.742625222148)
--(axis cs:40,543.851086258903)
--(axis cs:39,551.290355028111)
--(axis cs:38,561.924792296656)
--(axis cs:37,570.505283889548)
--(axis cs:36,578.944002275582)
--(axis cs:35,588.538490102783)
--(axis cs:34,600.332899482826)
--(axis cs:33,611.409840905382)
--(axis cs:32,622.620701390279)
--(axis cs:31,634.04100041811)
--(axis cs:30,644.554120234345)
--(axis cs:29,657.233211736653)
--(axis cs:28,671.116981844096)
--(axis cs:27,686.244358337805)
--(axis cs:26,701.149914664725)
--(axis cs:25,715.437596198925)
--(axis cs:24,731.950638970137)
--(axis cs:23,747.640870794104)
--(axis cs:22,764.29308460019)
--(axis cs:21,779.743482633577)
--(axis cs:20,798.880861137402)
--(axis cs:19,812.686473396775)
--(axis cs:18,826.166551094858)
--(axis cs:17,837.07077725903)
--(axis cs:16,842.351879635079)
--(axis cs:15,844.394774081977)
--(axis cs:14,838.973925560195)
--(axis cs:13,821.853707749468)
--(axis cs:12,800.650373969229)
--(axis cs:11,786.584155369496)
--(axis cs:10,760.671355188199)
--(axis cs:9,620.396372318196)
--(axis cs:8,443.410738988127)
--(axis cs:7,311.074419501264)
--(axis cs:6,216.947712679012)
--(axis cs:5,151.584779425399)
--(axis cs:4,106.216818223027)
--(axis cs:3,73.8819402789225)
--(axis cs:2,51.1961962107658)
--(axis cs:1,34.1156194067799)
--(axis cs:0,19.9496585712667)
--cycle;

\path [fill=color1, fill opacity=0.3, very thin]
(axis cs:0,0)
--(axis cs:0,0)
--(axis cs:1,0)
--(axis cs:2,0)
--(axis cs:3,0)
--(axis cs:4,0)
--(axis cs:5,0)
--(axis cs:6,0)
--(axis cs:7,0)
--(axis cs:8,0)
--(axis cs:9,-3.507281779228)
--(axis cs:10,-10.514808227935)
--(axis cs:11,-1.48357058454503)
--(axis cs:12,62.1200093764658)
--(axis cs:13,132.952077997044)
--(axis cs:14,181.73326072954)
--(axis cs:15,231.690478484435)
--(axis cs:16,308.781269669917)
--(axis cs:17,376.724991252972)
--(axis cs:18,440.168650291584)
--(axis cs:19,498.153375035057)
--(axis cs:20,542.483762609641)
--(axis cs:21,576.982542049895)
--(axis cs:22,604.489940430601)
--(axis cs:23,622.307068773121)
--(axis cs:24,630.33835171489)
--(axis cs:25,633.389330216102)
--(axis cs:26,632.175451706799)
--(axis cs:27,626.76473106772)
--(axis cs:28,617.877539053658)
--(axis cs:29,608.710393009468)
--(axis cs:30,597.689062320371)
--(axis cs:31,585.804440072104)
--(axis cs:32,571.89675148618)
--(axis cs:33,557.972995930241)
--(axis cs:34,543.819694882055)
--(axis cs:35,530.936184442921)
--(axis cs:36,517.617707916649)
--(axis cs:37,503.732197207702)
--(axis cs:38,489.031474480305)
--(axis cs:39,475.746435932499)
--(axis cs:40,462.613322088757)
--(axis cs:41,449.892334644593)
--(axis cs:42,437.745820055232)
--(axis cs:43,425.951421004025)
--(axis cs:44,415.142589517255)
--(axis cs:45,405.32659055149)
--(axis cs:46,394.16243130268)
--(axis cs:47,384.74819806705)
--(axis cs:48,374.606749785666)
--(axis cs:49,365.870249438698)
--(axis cs:50,356.430614565053)
--(axis cs:51,348.395940009259)
--(axis cs:52,339.949149064143)
--(axis cs:53,331.94481080587)
--(axis cs:54,325.001705726565)
--(axis cs:55,317.601398709289)
--(axis cs:56,310.140601069292)
--(axis cs:57,303.514564877896)
--(axis cs:58,296.139723445991)
--(axis cs:59,288.877097063103)
--(axis cs:60,281.806379934542)
--(axis cs:61,275.033057817761)
--(axis cs:62,268.351794745414)
--(axis cs:63,262.282612063216)
--(axis cs:64,252.964752654658)
--(axis cs:65,249.665417744991)
--(axis cs:66,241.075060705615)
--(axis cs:67,238.356053056337)
--(axis cs:68,231.81512996688)
--(axis cs:69,225.907101866119)
--(axis cs:70,220.455127769)
--(axis cs:71,212.510631413989)
--(axis cs:72,211.630852263337)
--(axis cs:73,206.774563883868)
--(axis cs:74,202.192273964219)
--(axis cs:75,197.022661692868)
--(axis cs:76,192.512279952631)
--(axis cs:77,187.798073451628)
--(axis cs:78,183.965992387863)
--(axis cs:79,180.062591843998)
--(axis cs:79,441.280265298859)
--(axis cs:79,441.280265298859)
--(axis cs:78,444.024483802614)
--(axis cs:77,445.754307500753)
--(axis cs:76,448.421053380703)
--(axis cs:75,450.424957354751)
--(axis cs:74,453.950583178639)
--(axis cs:73,456.606388497085)
--(axis cs:72,458.902481069996)
--(axis cs:71,461.765559062202)
--(axis cs:70,463.116300802428)
--(axis cs:69,467.511945752928)
--(axis cs:68,471.422965271215)
--(axis cs:67,475.205851705568)
--(axis cs:66,479.601129770576)
--(axis cs:65,482.429820350247)
--(axis cs:64,489.3781044882)
--(axis cs:63,491.745959365355)
--(axis cs:62,497.505348111728)
--(axis cs:61,502.271704087001)
--(axis cs:60,506.469810541648)
--(axis cs:59,511.484807698802)
--(axis cs:58,516.641228934961)
--(axis cs:57,522.828292264961)
--(axis cs:56,529.173684644994)
--(axis cs:55,535.617648909759)
--(axis cs:54,541.264960940102)
--(axis cs:53,546.807570146511)
--(axis cs:52,553.527041412048)
--(axis cs:51,559.318345705026)
--(axis cs:50,567.655099720661)
--(axis cs:49,575.891655323206)
--(axis cs:48,582.040869261953)
--(axis cs:47,587.804182885331)
--(axis cs:46,594.847092506844)
--(axis cs:45,599.578171353272)
--(axis cs:44,607.647886673221)
--(axis cs:43,615.210483757879)
--(axis cs:42,622.797037087625)
--(axis cs:41,631.679093926836)
--(axis cs:40,640.396201720766)
--(axis cs:39,649.986897400834)
--(axis cs:38,659.120906472076)
--(axis cs:37,669.391612316107)
--(axis cs:36,679.972768273827)
--(axis cs:35,690.444767938032)
--(axis cs:34,701.608876546516)
--(axis cs:33,712.484146926902)
--(axis cs:32,722.808010418582)
--(axis cs:31,733.414607546943)
--(axis cs:30,744.358556727248)
--(axis cs:29,755.089606990532)
--(axis cs:28,762.81769904158)
--(axis cs:27,770.920983217995)
--(axis cs:26,774.319786388439)
--(axis cs:25,773.801145974374)
--(axis cs:24,767.642600666062)
--(axis cs:23,755.988169322117)
--(axis cs:22,734.976726236066)
--(axis cs:21,705.760315092963)
--(axis cs:20,668.849570723693)
--(axis cs:19,624.979958298276)
--(axis cs:18,567.821825898892)
--(axis cs:17,502.294056366075)
--(axis cs:16,431.704444615797)
--(axis cs:15,360.19523580128)
--(axis cs:14,281.676263079984)
--(axis cs:13,222.171731526766)
--(axis cs:12,175.879990623534)
--(axis cs:11,97.9883324893069)
--(axis cs:10,28.5433796565064)
--(axis cs:9,4.45013892208515)
--(axis cs:8,0)
--(axis cs:7,0)
--(axis cs:6,0)
--(axis cs:5,0)
--(axis cs:4,0)
--(axis cs:3,0)
--(axis cs:2,0)
--(axis cs:1,0)
--(axis cs:0,0)
--cycle;

\addplot [semithick, color0]
table {%
0 17.0857142857143
1 29.0142857142857
2 42.9714285714286
3 61.0619047619048
4 86.4857142857143
5 122.490476190476
6 173.47619047619
7 246.952380952381
8 351.414285714286
9 496.833333333333
10 643.095238095238
11 703.680952380952
12 726.652380952381
13 747.961904761905
14 763.761904761905
15 768.428571428571
16 766
17 759.728571428571
18 748.371428571429
19 734.661904761905
20 719.97619047619
21 702.452380952381
22 685.847619047619
23 667.657142857143
24 650.385714285714
25 633.557142857143
26 618.47619047619
27 602.266666666667
28 586.085714285714
29 570.052380952381
30 555.638095238095
31 542.171428571429
32 529.052380952381
33 515.871428571429
34 502.595238095238
35 490.195238095238
36 478.785714285714
37 468.157142857143
38 457.633333333333
39 446.204761904762
40 436.304761904762
41 426.409523809524
42 416.161904761905
43 407.561904761905
44 397.857142857143
45 389.819047619048
46 381.514285714286
47 373.585714285714
48 365.019047619048
49 356.633333333333
50 348.833333333333
51 340.642857142857
52 332.833333333333
53 325.638095238095
54 318.528571428571
55 311.561904761905
56 304.257142857143
57 297.385714285714
58 292.233333333333
59 285.590476190476
60 279.62380952381
61 274.066666666667
62 268.857142857143
63 264.004761904762
64 258.142857142857
65 253.390476190476
66 248.490476190476
67 243.847619047619
68 239.009523809524
69 235.404761904762
70 231.247619047619
71 226.77619047619
72 222.714285714286
73 218.666666666667
74 214.461904761905
75 210.966666666667
76 206.547619047619
77 201.709523809524
78 197.019047619048
79 192.57619047619
};
\addplot [semithick, color1]
table {%
0 0
1 0
2 0
3 0
4 0
5 0
6 0
7 0
8 0
9 0.471428571428571
10 9.01428571428571
11 48.252380952381
12 119
13 177.561904761905
14 231.704761904762
15 295.942857142857
16 370.242857142857
17 439.509523809524
18 503.995238095238
19 561.566666666667
20 605.666666666667
21 641.371428571429
22 669.733333333333
23 689.147619047619
24 698.990476190476
25 703.595238095238
26 703.247619047619
27 698.842857142857
28 690.347619047619
29 681.9
30 671.02380952381
31 659.609523809524
32 647.352380952381
33 635.228571428571
34 622.714285714286
35 610.690476190476
36 598.795238095238
37 586.561904761905
38 574.076190476191
39 562.866666666667
40 551.504761904762
41 540.785714285714
42 530.271428571429
43 520.580952380952
44 511.395238095238
45 502.452380952381
46 494.504761904762
47 486.27619047619
48 478.32380952381
49 470.880952380952
50 462.042857142857
51 453.857142857143
52 446.738095238095
53 439.37619047619
54 433.133333333333
55 426.609523809524
56 419.657142857143
57 413.171428571429
58 406.390476190476
59 400.180952380952
60 394.138095238095
61 388.652380952381
62 382.928571428571
63 377.014285714286
64 371.171428571429
65 366.047619047619
66 360.338095238095
67 356.780952380952
68 351.619047619048
69 346.709523809524
70 341.785714285714
71 337.138095238095
72 335.266666666667
73 331.690476190476
74 328.071428571429
75 323.72380952381
76 320.466666666667
77 316.77619047619
78 313.995238095238
79 310.671428571429
};
\end{axis}

\end{tikzpicture}

%% file: pic/new22/22quct041_25k.tex
\begin{tikzpicture}

\definecolor{color0}{rgb}{0.886274509803922,0.290196078431373,0.2}
\definecolor{color1}{rgb}{0.203921568627451,0.541176470588235,0.741176470588235}
\pgfmathsetlengthmacro\MajorTickLength{
      \pgfkeysvalueof{/pgfplots/major tick length} * 0.5
    }
\begin{axis}[
axis line style={white},
legend cell align={left},
width = 0.37\linewidth,
height = 1in,
major tick length=\MajorTickLength,
legend style={inner xsep=1pt, inner ysep=-1pt, row sep=-3pt,at={(0.3,0.8)},anchor=north west},
font= \fontsize{7}{7.5}\selectfont,
tick align=inside,
tick pos=left,
x grid style={white},
xlabel={Days},
xmajorgrids,
xmin=0, xmax=80,
xtick style={color=white!33.3333333333333!black},
y grid style={white},
xtick={0,20,40,60,80},
xticklabels={0,20,40,60,80},
yticklabel style={rotate=90},
ytick={0,400,800},
yticklabels={0,400,800},
x label style={at={(0.5, 0.3)}},
ymajorgrids,
y label style={at={(0.55,0.5)}},
ymin=-48.6385516918637, ymax=865.852875125655,
ytick style={color=white!33.3333333333333!black}
]
\path [fill=color0, fill opacity=0.3, very thin]
(axis cs:0,20.1391511063455)
--(axis cs:0,14.5307518062759)
--(axis cs:1,24.2619824940097)
--(axis cs:2,35.0565940578158)
--(axis cs:3,48.2758893682555)
--(axis cs:4,66.4873767601975)
--(axis cs:5,93.2078049402034)
--(axis cs:6,132.285042462973)
--(axis cs:7,186.343572129352)
--(axis cs:8,264.587719094981)
--(axis cs:9,376.003452283108)
--(axis cs:10,533.424082860656)
--(axis cs:11,624.277243864777)
--(axis cs:12,643.955232017559)
--(axis cs:13,656.968123758137)
--(axis cs:14,658.520742245132)
--(axis cs:15,651.5581681529)
--(axis cs:16,637.76773324806)
--(axis cs:17,621.524054157744)
--(axis cs:18,602.343516217294)
--(axis cs:19,579.838375954488)
--(axis cs:20,557.09148945954)
--(axis cs:21,533.287352226559)
--(axis cs:22,511.290362446021)
--(axis cs:23,488.155061144802)
--(axis cs:24,465.572191251052)
--(axis cs:25,443.352234382619)
--(axis cs:26,421.723526130205)
--(axis cs:27,401.952176176523)
--(axis cs:28,382.563305847573)
--(axis cs:29,363.234010689152)
--(axis cs:30,345.520817872678)
--(axis cs:31,328.002800846073)
--(axis cs:32,311.908464198625)
--(axis cs:33,295.274100677019)
--(axis cs:34,279.279243974035)
--(axis cs:35,265.142348279077)
--(axis cs:36,250.696665082265)
--(axis cs:37,239.226521741235)
--(axis cs:38,226.810336047651)
--(axis cs:39,214.12224862987)
--(axis cs:40,203.974498516047)
--(axis cs:41,193.31029996434)
--(axis cs:42,183.366270092709)
--(axis cs:43,173.627696251827)
--(axis cs:44,164.490284459258)
--(axis cs:45,155.732827768962)
--(axis cs:46,147.703425250903)
--(axis cs:47,140.333651769378)
--(axis cs:48,132.477708463536)
--(axis cs:49,125.75382444524)
--(axis cs:50,119.961637748058)
--(axis cs:51,113.420293839085)
--(axis cs:52,107.706616976931)
--(axis cs:53,102.616637965086)
--(axis cs:54,96.7745815397972)
--(axis cs:55,92.2577087929091)
--(axis cs:56,87.9838378838093)
--(axis cs:57,83.271712229386)
--(axis cs:58,78.4386133975718)
--(axis cs:59,74.0211474329353)
--(axis cs:60,70.0799470089398)
--(axis cs:61,65.8570361089238)
--(axis cs:62,61.7310009889192)
--(axis cs:63,58.8477772127671)
--(axis cs:64,55.8702835982142)
--(axis cs:65,52.8816256010595)
--(axis cs:66,50.3270040255598)
--(axis cs:67,47.1270636132422)
--(axis cs:68,44.6832978505227)
--(axis cs:69,41.9826882934527)
--(axis cs:70,38.7574753914197)
--(axis cs:71,37.0674039295763)
--(axis cs:72,34.6785810873012)
--(axis cs:73,32.9896571631092)
--(axis cs:74,30.9530822638967)
--(axis cs:75,29.5127305055945)
--(axis cs:76,28.2532991448515)
--(axis cs:77,26.9667369574948)
--(axis cs:78,25.9656257872829)
--(axis cs:79,25.3491780558993)
--(axis cs:79,115.825579225654)
--(axis cs:79,115.825579225654)
--(axis cs:78,118.578063533105)
--(axis cs:77,120.596369838622)
--(axis cs:76,122.300098913401)
--(axis cs:75,125.23484230994)
--(axis cs:74,127.590607056492)
--(axis cs:73,130.505488467959)
--(axis cs:72,132.72918590299)
--(axis cs:71,135.854926167511)
--(axis cs:70,139.417281890134)
--(axis cs:69,142.114399085188)
--(axis cs:68,144.840973994137)
--(axis cs:67,148.5913829887)
--(axis cs:66,151.779792090945)
--(axis cs:65,155.28342294263)
--(axis cs:64,159.556900867805)
--(axis cs:63,164.171640262961)
--(axis cs:62,168.851523282925)
--(axis cs:61,173.89053670661)
--(axis cs:60,178.696752020186)
--(axis cs:59,183.755551596191)
--(axis cs:58,188.357503107283)
--(axis cs:57,192.883627576439)
--(axis cs:56,197.025870854055)
--(axis cs:55,201.742291207091)
--(axis cs:54,207.88561263496)
--(axis cs:53,212.907633879574)
--(axis cs:52,217.972994673555)
--(axis cs:51,225.133104219168)
--(axis cs:50,232.426711766505)
--(axis cs:49,239.372389146993)
--(axis cs:48,245.998019691804)
--(axis cs:47,252.94790162868)
--(axis cs:46,260.888807758806)
--(axis cs:45,269.063288735892)
--(axis cs:44,278.800977676664)
--(axis cs:43,288.362595010309)
--(axis cs:42,298.876448353893)
--(axis cs:41,309.010088385175)
--(axis cs:40,320.627443231526)
--(axis cs:39,333.110761078867)
--(axis cs:38,345.539178515456)
--(axis cs:37,358.336585054881)
--(axis cs:36,371.711101908027)
--(axis cs:35,386.430467254903)
--(axis cs:34,403.196484181305)
--(axis cs:33,418.31813233269)
--(axis cs:32,435.780856189724)
--(axis cs:31,454.33700497917)
--(axis cs:30,476.217046204992)
--(axis cs:29,497.892202903081)
--(axis cs:28,519.417276676699)
--(axis cs:27,542.319668483671)
--(axis cs:26,566.257056394066)
--(axis cs:25,591.540969500876)
--(axis cs:24,615.806449525647)
--(axis cs:23,643.077948563936)
--(axis cs:22,669.80672493262)
--(axis cs:21,695.440803113247)
--(axis cs:20,721.607539666674)
--(axis cs:19,747.297546375609)
--(axis cs:18,769.753571161347)
--(axis cs:17,791.262353609247)
--(axis cs:16,808.154596849027)
--(axis cs:15,820.169987186906)
--(axis cs:14,824.285082997586)
--(axis cs:13,819.575565562251)
--(axis cs:12,808.569039827102)
--(axis cs:11,798.363532834252)
--(axis cs:10,763.362324906334)
--(axis cs:9,636.093635095532)
--(axis cs:8,445.03364012832)
--(axis cs:7,310.724389035697)
--(axis cs:6,217.083889575862)
--(axis cs:5,151.22908826368)
--(axis cs:4,104.580584404851)
--(axis cs:3,73.3940135443658)
--(axis cs:2,52.166706913058)
--(axis cs:1,35.3205417778349)
--(axis cs:0,20.1391511063455)
--cycle;

\path [fill=color1, fill opacity=0.3, very thin]
(axis cs:0,0)
--(axis cs:0,0)
--(axis cs:1,0)
--(axis cs:2,0)
--(axis cs:3,0)
--(axis cs:4,0)
--(axis cs:5,0)
--(axis cs:6,0)
--(axis cs:7,0)
--(axis cs:8,0)
--(axis cs:9,0)
--(axis cs:10,-7.07075956379466)
--(axis cs:11,-2.38693198111125)
--(axis cs:12,60.6541858869486)
--(axis cs:13,135.868685282268)
--(axis cs:14,186.528410228841)
--(axis cs:15,244.852388783183)
--(axis cs:16,312.15492313929)
--(axis cs:17,381.013203352676)
--(axis cs:18,441.737814281888)
--(axis cs:19,492.98846466391)
--(axis cs:20,531.782371027633)
--(axis cs:21,560.067488760492)
--(axis cs:22,579.40179585295)
--(axis cs:23,588.06263373925)
--(axis cs:24,588.503127902814)
--(axis cs:25,583.2150522077)
--(axis cs:26,573.287929284653)
--(axis cs:27,560.019199087879)
--(axis cs:28,543.252841988828)
--(axis cs:29,527.301640980254)
--(axis cs:30,509.340840703965)
--(axis cs:31,491.141852477351)
--(axis cs:32,472.811055388043)
--(axis cs:33,455.331450607845)
--(axis cs:34,439.227095822313)
--(axis cs:35,420.643765172724)
--(axis cs:36,404.249740196931)
--(axis cs:37,388.430294263534)
--(axis cs:38,373.374189786395)
--(axis cs:39,358.496530596314)
--(axis cs:40,344.576122465596)
--(axis cs:41,331.772048377303)
--(axis cs:42,317.756719446445)
--(axis cs:43,305.071664394032)
--(axis cs:44,292.871694808193)
--(axis cs:45,280.77256967554)
--(axis cs:46,270.283403322349)
--(axis cs:47,260.010616633941)
--(axis cs:48,249.092224326047)
--(axis cs:49,240.45542457472)
--(axis cs:50,231.471123262456)
--(axis cs:51,222.634735896379)
--(axis cs:52,214.773789611998)
--(axis cs:53,207.374114662793)
--(axis cs:54,200.090002082048)
--(axis cs:55,192.826027594133)
--(axis cs:56,186.605681091077)
--(axis cs:57,176.709045927544)
--(axis cs:58,175.208974030903)
--(axis cs:59,169.570559252763)
--(axis cs:60,164.176915096123)
--(axis cs:61,158.981146174179)
--(axis cs:62,154.432359478059)
--(axis cs:63,149.562371668007)
--(axis cs:64,142.245453592785)
--(axis cs:65,140.925808058681)
--(axis cs:66,136.77748934777)
--(axis cs:67,132.927284843047)
--(axis cs:68,129.644662812068)
--(axis cs:69,126.370146062189)
--(axis cs:70,123.190630817155)
--(axis cs:71,119.986843777959)
--(axis cs:72,117.365773650744)
--(axis cs:73,114.362173791624)
--(axis cs:74,111.322625991977)
--(axis cs:75,109.250036272227)
--(axis cs:76,106.671684172989)
--(axis cs:77,104.563961481572)
--(axis cs:78,102.201363511387)
--(axis cs:79,98.2351725631036)
--(axis cs:79,210.541526466023)
--(axis cs:79,210.541526466023)
--(axis cs:78,211.837471440069)
--(axis cs:77,214.076815217457)
--(axis cs:76,216.949675050312)
--(axis cs:75,219.206274407385)
--(axis cs:74,222.065723522586)
--(axis cs:73,225.59899125692)
--(axis cs:72,229.284711786149)
--(axis cs:71,233.12966107641)
--(axis cs:70,236.6054856877)
--(axis cs:69,239.105582093151)
--(axis cs:68,243.423298352981)
--(axis cs:67,247.257181176371)
--(axis cs:66,251.154549487181)
--(axis cs:65,254.967395824814)
--(axis cs:64,259.414740581972)
--(axis cs:63,263.039570079566)
--(axis cs:62,267.68414537631)
--(axis cs:61,272.028562563685)
--(axis cs:60,277.881337331061)
--(axis cs:59,283.497401912285)
--(axis cs:58,288.858987134146)
--(axis cs:57,295.426876402553)
--(axis cs:56,300.539949976884)
--(axis cs:55,306.795331629168)
--(axis cs:54,314.308056170379)
--(axis cs:53,321.732681453712)
--(axis cs:52,329.206792912274)
--(axis cs:51,337.656526239543)
--(axis cs:50,345.858973824923)
--(axis cs:49,355.554284163144)
--(axis cs:48,365.781562081719)
--(axis cs:47,376.290354239846)
--(axis cs:46,387.677761726194)
--(axis cs:45,399.17888663514)
--(axis cs:44,411.982674123846)
--(axis cs:43,425.462316188492)
--(axis cs:42,437.990853369089)
--(axis cs:41,452.18911667124)
--(axis cs:40,467.851062000424)
--(axis cs:39,485.629682995919)
--(axis cs:38,502.771441281566)
--(axis cs:37,521.035725153942)
--(axis cs:36,539.556085045788)
--(axis cs:35,558.443613468053)
--(axis cs:34,578.569020682541)
--(axis cs:33,597.814180460116)
--(axis cs:32,619.363701893511)
--(axis cs:31,639.722225192552)
--(axis cs:30,659.71741172322)
--(axis cs:29,679.076999796445)
--(axis cs:28,696.184051215056)
--(axis cs:27,713.242936834451)
--(axis cs:26,726.692653239619)
--(axis cs:25,734.289802161232)
--(axis cs:24,738.700755592332)
--(axis cs:23,734.267463348129)
--(axis cs:22,721.306942011128)
--(axis cs:21,697.146103472517)
--(axis cs:20,664.790444506348)
--(axis cs:19,624.322214947741)
--(axis cs:18,572.349564358889)
--(axis cs:17,507.161553928878)
--(axis cs:16,437.709154530613)
--(axis cs:15,362.574795682836)
--(axis cs:14,286.287123751741)
--(axis cs:13,226.306071999285)
--(axis cs:12,179.93804712276)
--(axis cs:11,104.270427126742)
--(axis cs:10,24.5270702434063)
--(axis cs:9,0)
--(axis cs:8,0)
--(axis cs:7,0)
--(axis cs:6,0)
--(axis cs:5,0)
--(axis cs:4,0)
--(axis cs:3,0)
--(axis cs:2,0)
--(axis cs:1,0)
--(axis cs:0,0)
--cycle;

\addplot [semithick, color0]
table {%
0 17.3349514563107
1 29.7912621359223
2 43.6116504854369
3 60.8349514563107
4 85.5339805825243
5 122.218446601942
6 174.684466019417
7 248.533980582524
8 354.81067961165
9 506.04854368932
10 648.393203883495
11 711.320388349515
12 726.26213592233
13 738.271844660194
14 741.402912621359
15 735.864077669903
16 722.961165048544
17 706.393203883495
18 686.04854368932
19 663.567961165049
20 639.349514563107
21 614.364077669903
22 590.54854368932
23 565.616504854369
24 540.68932038835
25 517.446601941748
26 493.990291262136
27 472.135922330097
28 450.990291262136
29 430.563106796117
30 410.868932038835
31 391.169902912621
32 373.844660194175
33 356.796116504854
34 341.23786407767
35 325.78640776699
36 311.203883495146
37 298.781553398058
38 286.174757281553
39 273.616504854369
40 262.300970873786
41 251.160194174757
42 241.121359223301
43 230.995145631068
44 221.645631067961
45 212.398058252427
46 204.296116504854
47 196.640776699029
48 189.23786407767
49 182.563106796117
50 176.194174757282
51 169.276699029126
52 162.839805825243
53 157.76213592233
54 152.330097087379
55 147
56 142.504854368932
57 138.077669902913
58 133.398058252427
59 128.888349514563
60 124.388349514563
61 119.873786407767
62 115.291262135922
63 111.509708737864
64 107.71359223301
65 104.082524271845
66 101.053398058252
67 97.8592233009709
68 94.7621359223301
69 92.0485436893204
70 89.0873786407767
71 86.4611650485437
72 83.7038834951456
73 81.747572815534
74 79.2718446601942
75 77.373786407767
76 75.2766990291262
77 73.7815533980583
78 72.2718446601942
79 70.5873786407767
};
\addplot [semithick, color1]
table {%
0 0
1 0
2 0
3 0
4 0
5 0
6 0
7 0
8 0
9 0
10 8.72815533980583
11 50.9417475728155
12 120.296116504854
13 181.087378640777
14 236.407766990291
15 303.71359223301
16 374.932038834951
17 444.087378640777
18 507.043689320388
19 558.655339805825
20 598.28640776699
21 628.606796116505
22 650.354368932039
23 661.165048543689
24 663.601941747573
25 658.752427184466
26 649.990291262136
27 636.631067961165
28 619.718446601942
29 603.18932038835
30 584.529126213592
31 565.432038834951
32 546.087378640777
33 526.572815533981
34 508.898058252427
35 489.543689320388
36 471.902912621359
37 454.733009708738
38 438.072815533981
39 422.063106796117
40 406.21359223301
41 391.980582524272
42 377.873786407767
43 365.266990291262
44 352.427184466019
45 339.97572815534
46 328.980582524272
47 318.150485436893
48 307.436893203883
49 298.004854368932
50 288.665048543689
51 280.145631067961
52 271.990291262136
53 264.553398058252
54 257.199029126214
55 249.81067961165
56 243.572815533981
57 236.067961165049
58 232.033980582524
59 226.533980582524
60 221.029126213592
61 215.504854368932
62 211.058252427184
63 206.300970873786
64 200.830097087379
65 197.946601941748
66 193.966019417476
67 190.092233009709
68 186.533980582524
69 182.73786407767
70 179.898058252427
71 176.558252427184
72 173.325242718447
73 169.980582524272
74 166.694174757282
75 164.228155339806
76 161.81067961165
77 159.320388349515
78 157.019417475728
79 154.388349514563
};
\end{axis}

\end{tikzpicture}

%% file: ijcai21.bbl
\begin{thebibliography}{}

\bibitem[\protect\citeauthoryear{Aleta \bgroup \em et al.\egroup
  }{2020}]{2ndwave}
Alberto Aleta, David Martín-Corral, Ana~Pastore y~Piontti, Marco Ajelli, Maria
  Litvinova, Matteo Chinazzi, Natalie~E. Dean, M.~Elizabeth Halloran, Ira
  M.~Longini Jr, Stefano Merler, Alex Pentland, Alessandro Vespignani, Esteban
  Moro, and Yamir Moreno.
\newblock Modeling the impact of social distancing, testing, contact tracing
  and household quarantine on second-wave scenarios of the covid-19 epidemic.
\newblock {\em Nature Human Behaviour}, 4:964–971, 2020.

\bibitem[\protect\citeauthoryear{All}{2020}]{Alleghenynews}
Stay updated - allegheny county, 2020.
\newblock
  [https://www.alleghenycounty.us/Health-Department/Resources/COVID-19/Stay-Updated.aspx;
  accessed 27-October-2020].

\bibitem[\protect\citeauthoryear{ArcGIS}{2020}]{arcgis}
ArcGIS.
\newblock Arcgis business analyst, 2020.
\newblock [https://bao.arcgis.com/esriBAO/index.html; accessed 25-August-2020].

\bibitem[\protect\citeauthoryear{Bishop}{2020}]{shopnum}
David Bishop.
\newblock June 2020 online grocery scorecard: Growth in sales \& hh penetration
  continues, 2020.
\newblock
  [https://www.brickmeetsclick.com/june-2020-online-grocery-scorecard--growth-in-sales---hh-penetration-continues;
  accessed August-15-2020].

\bibitem[\protect\citeauthoryear{Bureau}{2020}]{population}
US~Census Bureau.
\newblock National population by characteristics: 2010-2019, 2020.
\newblock
  [https://www.census.gov/data/tables/time-series/demo/popest/2010s-national-detail.html;
  accessed 15-October-2020].

\bibitem[\protect\citeauthoryear{Chang \bgroup \em et al.\egroup
  }{2020}]{australia}
Sheryl~L. Chang, Nathan Harding, Cameron Zachreson, Oliver~M. Cliff, and
  Mikhail Prokopenko.
\newblock Modelling transmission and control of the covid-19 pandemic in
  australia.
\newblock {\em Nature Communications}, 11(5710), 2020.

\bibitem[\protect\citeauthoryear{Dietz}{1976}]{SEIR}
K.~Dietz.
\newblock The incidence of infectious diseases under the influence of seasonal
  fluctuations.
\newblock In J{\"u}rgen Berger, Wolfgang~J. B{\"u}hler, Rudolf Repges, and
  Petre Tautu, editors, {\em Mathematical Models in Medicine}, pages 1--15,
  Berlin, Heidelberg, 1976. Springer Berlin Heidelberg.

\bibitem[\protect\citeauthoryear{Doyle}{2020}]{firstcovid}
Patrick Doyle.
\newblock Allegheny county announces first two cases of covid-19, 2020.
\newblock
  [https://www.witf.org/2020/03/14/allegheny-county-announces-first-two-cases-of-covid-19/;
  accessed 15-October-2020].

\bibitem[\protect\citeauthoryear{Eilersen and Sneppen}{2020}]{costbenefit}
Andreas Eilersen and Kim Sneppen.
\newblock Cost–benefit of limited isolation and testing in covid-19
  mitigation.
\newblock {\em Nature Scientific Reports}, 10(18543), 2020.

\bibitem[\protect\citeauthoryear{for Disease~Control and
  Prevention}{2020}]{CDCinfectdeath}
Centers for Disease~Control and Prevention.
\newblock Cdc covid data tracker, 2020.
\newblock [https://covid.cdc.gov/covid-data-tracker/\#demographics; accessed
  27-October-2020].

\bibitem[\protect\citeauthoryear{Haarnoja \bgroup \em et al.\egroup
  }{2017}]{haarnoja2017reinforcement}
Tuomas Haarnoja, Haoran Tang, Pieter Abbeel, and Sergey Levine.
\newblock Reinforcement learning with deep energy-based policies.
\newblock In {\em ICML}, 2017.

\bibitem[\protect\citeauthoryear{Hamm}{2020}]{freqres}
Trent Hamm.
\newblock Don’t eat out as often, 2020.
\newblock [https://www.thesimpledollar.com/save-money/dont-eat-out-as-often;
  accessed 15-October-2020].

\bibitem[\protect\citeauthoryear{Hoertel \bgroup \em et al.\egroup
  }{2020}]{France}
Nicolas Hoertel, Martin Blachier, Carlos Blanco, Mark Olfson, Marc Massetti,
  Marina~Sánchez Rico, Frédéric Limosin, and Henri Leleu.
\newblock A stochastic agent-based model of the sars-cov-2 epidemic in france.
\newblock {\em Nature Medicine}, page 1417–1421, 2020.

\bibitem[\protect\citeauthoryear{Jones and Saad}{2019}]{freq}
Jeff Jones and Lydia Saad.
\newblock Gallup news service december wave one final topline, 2019.
\newblock [Timberline: 937008; JT: 335; Princeton Job number: 19-12-021;
  accessed 13-October-2020].

\bibitem[\protect\citeauthoryear{Kompella \bgroup \em et al.\egroup
  }{2020}]{kompella2020reinforcement}
Varun Kompella, Roberto Capobianco, Stacy Jong, Jonathan Browne, Spencer Fox,
  Lauren Meyers, Peter Wurman, and Peter Stone.
\newblock Reinforcement learning for optimization of covid-19 mitigation
  policies.
\newblock {\em arXiv preprint arXiv:2010.10560}, 2020.

\bibitem[\protect\citeauthoryear{Lauer \bgroup \em et al.\egroup
  }{2020}]{incubation}
Stephen~A. Lauer, Kyra~H. Grantz, Qifang Bi, Forrest~K. Jones, Qulu Zheng,
  Hannah~R. Meredith, Andrew~S. Azman, Nicholas~G. Reich, and Justin Lessler.
\newblock The incubation period of coronavirus disease 2019 (covid-19) from
  publicly reported confirmed cases: Estimation and application.
\newblock {\em Annals of Internal Medicine}, 2020.

\bibitem[\protect\citeauthoryear{Liu}{2020}]{Liu2020AME}
Changliu Liu.
\newblock A microscopic epidemic model and pandemic prediction using
  multi-agent reinforcement learning.
\newblock {\em ArXiv}, abs/2004.12959, 2020.

\bibitem[\protect\citeauthoryear{Long \bgroup \em et al.\egroup
  }{2020}]{immuneloss}
Quan-Xin Long, Xiao-Jun Tang, Qiu-Lin Shi, Qin Li, Hai-Jun Deng, Jun Yuan,
  Jie-Li Hu, Wei Xu, Yong Zhang, Fa-Jin Lv, Kun Su, Fan Zhang, Jiang Gong,
  Bo~Wu, Xia-Mao Liu, Jin-Jing Li, Jing-Fu Qiu, Juan Chen, and Ai-Long Huang.
\newblock Clinical and immunological assessment of asymptomatic sars-cov-2
  infections.
\newblock {\em Nature Medicine}, 26:1200--1204, 2020.

\bibitem[\protect\citeauthoryear{Meloni \bgroup \em et al.\egroup
  }{2011}]{meloni2011modeling}
Sandro Meloni, Nicola Perra, Alex Arenas, Sergio G{\'o}mez, Yamir Moreno, and
  Alessandro Vespignani.
\newblock Modeling human mobility responses to the large-scale spreading of
  infectious diseases.
\newblock {\em Scientific reports}, 1:62, 2011.

\bibitem[\protect\citeauthoryear{NCIRD}{2020}]{CDCchart}
NCIRD, 2020.
\newblock [National Center for Immunization and Respiratory Diseases (NCIRD),
  Division of Viral Diseases,
  https://www.cdc.gov/coronavirus/2019-ncov/covid-data/
  investigations-discovery/hospitalization-death-by-age.html; accessed
  15-October-2020].

\bibitem[\protect\citeauthoryear{of Health}{2019}]{hospital}
Pennsylvania~Department of~Health.
\newblock Hospital reports, 2019.
\newblock [https://www.health.pa.gov/topics/HealthStatistics/
  HealthFacilities/HospitalReports/Pages/hospital-reports.aspx; retrieved
  2020-08-31].

\bibitem[\protect\citeauthoryear{Qu and Li}{2019}]{qu2019exploiting}
Guannan Qu and Na~Li.
\newblock Exploiting fast decaying and locality in multi-agent mdp with tree
  dependence structure.
\newblock In {\em 2019 IEEE 58th Conference on Decision and Control (CDC)},
  pages 6479--6486. IEEE, 2019.

\bibitem[\protect\citeauthoryear{Qu \bgroup \em et al.\egroup
  }{2020}]{qu2020scalable}
Guannan Qu, Adam Wierman, and Na~Li.
\newblock Scalable reinforcement learning of localized policies for multi-agent
  networked systems.
\newblock In {\em Learning for Dynamics and Control}, pages 256--266. PMLR,
  2020.

\bibitem[\protect\citeauthoryear{Silva \bgroup \em et al.\egroup
  }{2020}]{covidabs}
Petr{\^o}nio~CL Silva, Paulo~VC Batista, H{\'e}lder~S Lima, Marcos~A Alves,
  Frederico~G Guimar{\~a}es, and Rodrigo~CP Silva.
\newblock Covid-abs: An agent-based model of covid-19 epidemic to simulate
  health and economic effects of social distancing interventions.
\newblock {\em arXiv preprint arXiv:2006.10532}, 2020.

\bibitem[\protect\citeauthoryear{Statista}{2016}]{statista}
Research~Department Statista.
\newblock How often people work out at the gym in us 2016, 2016.
\newblock [https://www.statista.com/statistics/638978; accessed
  15-October-2020].

\bibitem[\protect\citeauthoryear{Tambri~Housen and Sheel}{2020}]{howlong}
Amy Elizabeth~Parry Tambri~Housen and Meru Sheel.
\newblock How long are you infectious when you have coronavirus?, 2020.
\newblock
  [https://theconversation.com/how-long-are-you-infectious-when-you-have-coronavirus-135295;
  accessed 15-October-2020].

\bibitem[\protect\citeauthoryear{Wheaton}{2012}]{synpop}
WD~Wheaton.
\newblock 2009 us synthetic population ver. 2, 2012.

\bibitem[\protect\citeauthoryear{Yang \bgroup \em et al.\egroup
  }{2018}]{yang2018study}
Yaodong Yang, Lantao Yu, Yiwei Bai, Ying Wen, Weinan Zhang, and Jun Wang.
\newblock A study of ai population dynamics with million-agent reinforcement
  learning.
\newblock In {\em Proceedings of the 17th International Conference on
  Autonomous Agents and MultiAgent Systems}, pages 2133--2135, 2018.

\bibitem[\protect\citeauthoryear{Zheng \bgroup \em et al.\egroup
  }{2017}]{zheng2017magent}
Lianmin Zheng, Jiacheng Yang, Han Cai, Weinan Zhang, Jun Wang, and Yong Yu.
\newblock Magent: A many-agent reinforcement learning platform for artificial
  collective intelligence.
\newblock {\em arXiv preprint arXiv:1712.00600}, 2017.

\end{thebibliography}
